\documentclass[12pt]{article}
\usepackage{amsmath}
\usepackage{graphicx}
\usepackage{enumerate}
\usepackage{natbib}
\usepackage{url} 
\usepackage[flushleft]{threeparttable}
\usepackage{multirow}
\usepackage{verbatim}
\newcommand{\blind}{1}

\usepackage{varwidth}
\usepackage{amsmath}
\usepackage{numprint}
\usepackage{algorithm}
\usepackage{algpseudocode}
\usepackage{multirow}
\algrenewcommand\algorithmicrequire{\textbf{Input:}}
\algrenewcommand\algorithmicensure{\textbf{Output:}}
\usepackage{amsmath,amssymb}

\usepackage{mathrsfs}
\usepackage{bm, xcolor}
\usepackage{mathrsfs}
\usepackage{verbatim}
\usepackage{longtable}
\usepackage{threeparttable}
\usepackage{color}
\usepackage{amsmath}

\usepackage{enumerate}
\usepackage{amssymb,bbm}
\usepackage{amsbsy}
\usepackage{amsmath}
\usepackage{amsthm}
\usepackage{amsfonts}
\usepackage{latexsym}
\usepackage{xifthen} 
\usepackage{color}
\usepackage{manfnt}
\usepackage{ifthen}
\usepackage{bm}
\usepackage{caption}
\usepackage{subcaption}

\def\v#1{\mbi{#1}} 

\newcommand{\mR}{\mathbb{R}}
\newcommand{\R}{\mathbb{R}}

\newcommand{\A}{\mathcal{A}}

\newcommand{\E}{\mathbb{E}} 
\renewcommand{\P}{\mathbb{P}} 

\newcommand{\floor}[1]{\left\lfloor{#1} \right\rfloor}

\providecommand{\argmax}{\mathop{\rm argmax}} 
\providecommand{\argmin}{\mathop{\rm argmin}}

\makeatletter
\def\singlespace{\def\baselinestretch{1}\@normalsize}

\newtheorem{lemma}{Lemma}
\newtheorem{theorem}{Theorem}

\@addtoreset{equation}{section}

\renewcommand{\hat}{\widehat}

\def\singlespace{\def\baselinestretch{1}\@normalsize}

\makeatother

\def\newpage{\vfill\eject}

\newdimen\biblioindent    \biblioindent=30pt

 at 7truept

\def\beqr{\begin{eqnarray}}
	\def\eeqr{\end{eqnarray}}
\def\beqrs{\begin{eqnarray*}}
	\def\eeqrs{\end{eqnarray*}}

\def\beq{\begin{equation}}
\def\eeq{\end{equation}}
\def\beqn{\begin{eqnarray}}
\def\eeqn{\end{eqnarray}}
\def\beqnn{\begin{eqnarray*}}
\def\eeqnn{\end{eqnarray*}}

\def\A{{\bf A}}
\def\B{{\boldsymbol B}}

\def\M{{\boldsymbol M}}

\def\v{{\boldsymbol v}}

\def\M{{\bf M}}

\newtheorem*{assumption*}{\assumptionnumber}
\providecommand{\assumptionnumber}{}
\makeatletter
\newenvironment{assumption}[1]
{%
	\renewcommand{\assumptionnumber}{Assumption #1}%
	\begin{assumption*}%
		\protected@edef\@currentlabel{#1}%
	}
	{%
	\end{assumption*}
}
\makeatother

\newtheorem{lemm}{Lemma}[section]

\theoremstyle{definition}
\newtheorem{defi}{Definition}

\makeatletter

\newcommand{\Rmnum}[1]{\expandafter\@slowromancap\romannumeral #1@}
\makeatother

\usepackage{chngcntr}
\usepackage{apptools}
\AtAppendix{\counterwithin{lemma}{section}}

\usepackage{xr}
\begin{document}

	\def\spacingset#1{\renewcommand{\baselinestretch}%
		{#1}\small\normalsize} \spacingset{1}
	
	\if1\blind
	{
		\title{\bf Differentially Private Sliced Inverse Regression: Minimax Optimality and Algorithm}
		\author{Xintao Xia\\
			Department of Statistics, Iowa State University\\
			and \\
			Linjun Zhang\\
			Department of Statistics, Rutgers University\\
			and \\
			Zhanrui Cai\\
			Faculty of Business and Economics, The University of Hong Kong}
		\date{}
		\maketitle
	} \fi
	
	\if0\blind
	{
		\bigskip
		\bigskip
		\bigskip
		\begin{center}
			{\LARGE\bf Differentially Private Sliced Inverse Regression: Minimax Optimality and Algorithm}
		\end{center}
		\medskip
	} \fi
	
	\bigskip
	\begin{abstract}
		Privacy preservation has become a critical concern in high-dimensional data analysis due to the growing prevalence of data-driven applications. Since its proposal, sliced inverse regression has emerged as a widely utilized statistical technique to reduce the dimensionality of covariates while maintaining sufficient statistical information. In this paper, we propose optimally differentially private algorithms specifically designed to address privacy concerns in the context of sufficient dimension reduction. We establish lower bounds for differentially private sliced inverse regression in low and high dimensional settings. Moreover, we develop differentially private algorithms that achieve the minimax lower bounds up to logarithmic factors. Through a combination of simulations and real data analysis, we illustrate the efficacy of these differentially private algorithms in safeguarding privacy while preserving vital information within the reduced dimension space. As a natural extension, we can readily offer analogous lower and upper bounds for differentially private sparse principal component analysis, a topic that may also be of potential interest to the statistics and machine learning community.
	\end{abstract}
	
	\noindent%
	{\it Keywords:} sliced inverse regression, differential privacy, high dimension.
	\vfill
	
	\newpage
	\spacingset{1.9} 
	
	\section{Introduction}
	
	\allowdisplaybreaks
	
	\subsection{Sufficient Dimension Reduction}
	
	The collection and analysis of high-dimensional statistical data have become increasingly prevalent in light of technological advances \citep{fan2020statistical}. Dealing with high-dimensional data requires specialized methods that can effectively handle the curse of dimensionality. Suppose $Y\in\mR^1$ is a response variable and $\boldsymbol{x}=(x_1,\dots, x_p)^{\top}\in\mR^p$ are the associated covariates. A popular idea is to replace the high-dimensional $\boldsymbol{x}$ with a small number of linear combinations $\B^{\top}\boldsymbol{x}$ that are sufficient to predict the response, where $\B\in\mR^{p\times k}$ is a $p\times k$ matrix, with $k$ being much smaller than $p$. \textit{Sufficient dimension reduction} (SDR) combines the idea of linear dimension reduction with statistical sufficiency. Specifically, SDR seeks for a matrix $\B\in\mR^{p\times k}$ such that
	\begin{equation}
		\label{eq:condition_indep}
		Y\perp \!\!\! \perp\boldsymbol{x}\mid\B^{\top}\boldsymbol{x},
	\end{equation}
	where $\perp \!\!\! \perp$ stands for statistical independence. We call span$(\B)$ a sufficient dimension reduction subspace. One may notice that the subspace is not unique (Proposition 2.3 of \cite{li2018sufficient}). Following the literature, we define the intersection of all sufficient dimension reduction subspaces as the \textit{central subspace}, denoted by $\mathcal{S}_{Y\mid\boldsymbol{x}}$. The goal of interest for SDR is to identify the space $\mathcal{S}_{Y\mid\boldsymbol{x}}$.
	
	Since the pioneering work of sliced inverse regression \citep{li1991sliced}, numerous methods have been proposed for estimating the space $\mathcal{S}_{Y\mid\boldsymbol{x}}$ under different additional assumptions, such as the sliced average variance estimate \citep{cook1991sliced}, the contour regression \citep{li2005contour}, among others. Readers are directed to the seminal book by \cite{li2018sufficient} for a comprehensive review and examination of sufficient dimension reduction methodologies. Sliced inverse regression has gained significant popularity among these methods and is widely employed for dimensionality reduction in practical applications, especially in high dimensional data. 
	\cite{lin2018consistency} provides a framework to analyze the phase transition phenomenon of sliced inverse regression. \cite{lin2019sparse} proposed the Lasso-SIR method, which achieves optimality when $p=o(n^2)$, where $n$ is the sample size. When $p$ grows exponentially with $n$, \cite{tan2020sparse} proposed a refined sparse sliced inverse regression estimator with an optimal convergence rate. Sliced inverse regression was also studied for the online sequential data in \cite{cai2020online}. 
	
	\subsection{Differential Privacy}
	
	Privacy-preserving data analysis has gained increasing attention due to its critical role in safeguarding personal information. In healthcare, finance, and social science research, where sensitive personal data is frequently involved, privacy-preserving data analysis protects individual privacy by ensuring that personal information is not disclosed. Additionally, privacy-preserving data analysis builds trust between individuals and organizations that collect and store personal data. 
	
	Differential privacy (DP), introduced by \cite{dwork2006calibrating}, is a rigorous mathematical framework that quantifies the extent of privacy compromise that occurs when an individual's information is altered within a dataset. The fundamental objective of a DP algorithm is to safeguard the privacy of each individual against potential adversaries who have access to both the output of the algorithm and the information of other individuals. Recently, significant efforts have been dedicated to developing differentially private statistical algorithms to protect individual privacy in data analysis. These efforts encompass various areas, including estimation and inference on high-dimensional linear regression \citep{cai2021cost, cai2023private}, statistical inference using M-estimation \citep{avella2021differentially}, and adaptive false discovery rate control \citep{xia2023adaptive}, etc. It is worth noting that the mentioned works represent a subset of the extensive research conducted in this field.
	
	\subsection{Related Work and Our Contributions}
	
	The literature on differentially private sufficient dimension reduction is relatively limited. One related work on differentially private sliced inverse regression is \cite{he2023differentially}, which only proved the consistency of the proposed algorithm and omitted discussion on convergence rates of either the lower bound or the upper bound. For differentially private high-dimensional principal component analysis, \cite{hu2023privacy} considered the case where the reduced dimension equals 1 under concentrated DP.  However, the convergence rate in \cite{hu2023privacy} is not optimal in settings where the underlying data is sub-Gaussian. 
	
	In this paper, we aim to address these gaps in the literature by providing a comprehensive analysis of {\it Differentially Private Sliced Inverse Regression} (DP-SIR). Specifically, we establish a minimax lower bound that characterizes both the statistical error and privacy error for DP-SIR, considering both the low and high dimensional settings. Notably, the developed framework allows the dimension to increase exponentially compared with the sample size. The lower bound takes into account various factors that are related to the sliced inverse regression, such as the {\it sample size}, the {\it dimensionality}, the {\it eigen-gap} of the kernel matrix, the {\it sparsity}, and the {\it privacy cost}. We develop efficient algorithms for DP-SIR that match the lower bound up to a logarithmic rate. To evaluate the performance of these algorithms, we conduct simulations and real data applications to assess the statistical efficiency. Particularly, by considering a special case, the newly developed tools and methodology can be readily extended to DP {\it sparse} principal component analysis, which has also not been studied in the literature, while yielding similar lower and upper bounds. 
	
	In addition to the theoretical aspects, our work is also motivated by practical applications that necessitate the development of novel methodologies for dimension reduction with differential privacy (DP) guarantee. In genetic and genomic studies, safeguarding the privacy of individual subjects is of paramount importance while simultaneously identifying significant effects underlying the phenotype \citep{cai2022model, du2022robust}. Similarly, in business analytics, protecting the privacy of transactions and sales details is crucial, as potential competitors may exploit vulnerabilities to develop adversarial strategies. In this paper, we present two real data applications to demonstrate the efficacy of the proposed DP-SIR method in both dimension reduction and privacy protection.


	\subsection{Notation}
	
	We use bold lowercase letters (e.g., $\boldsymbol{\beta}$) to denote (random) vectors, and bold uppercase letters (e.g., $\boldsymbol{A}$) to denote matrices. For any vector $\boldsymbol{x}=(x_1,\dots,x_p)^{\top}$, we define the $\ell_q$ norm of $\boldsymbol{x}$ for $1\leq q<\infty$, as $\|\boldsymbol{x}\|_q:=(\sum_{i=1}^{p}|x_i|^q)^{1/q}$ with $|\cdot|$ representing the absolute value. The $\ell_\infty$ norm of $\boldsymbol{x}$ is defined as $\|\boldsymbol{x}\|_\infty:=\max_{i=1,\dots,p}|x_i|$. The support of $\boldsymbol{x}$ is a set with all indexes of nonzero elements in $\boldsymbol{x}$, denoted by $\text{supp}(\boldsymbol{x}):=\{i\in[p]\mid|x_i|>0\}$, where $[p]:=\{1,\dots,p\}$ is the index set. A random vector $\boldsymbol{x}\in\mR^p$ follows a sub-Gaussian distribution, if $\mathbb{P}(|\boldsymbol{x}^{\top}\boldsymbol{b}|\geq t)\leq 2\exp\{-t^2/(c^2\|\boldsymbol{b}\|_2^2)\}$ for a positive constant $c$ and all $t\geq 0$, $\boldsymbol{b}\in\mR^{p}$. For any $p\times q$-dimensional matrix $\boldsymbol{A}=(a_{ij})_{ij}\in\mR^{p\times q}$, the $k$th largest singular value is denoted by $\sigma_k(\boldsymbol{A})$. When $\boldsymbol{A}$ is positive semidefinite, $\sigma_k(\boldsymbol{A})$ equals the $k$th largest eigenvalue of $\boldsymbol{A}$, denoted by $\lambda_{k}(\boldsymbol{A})$. When $\boldsymbol{A}\in\mR^{p\times p}$ is a square matrix, the trace of $\boldsymbol{A}$ is defined as $\text{Tr}(\boldsymbol{A}):=\sum_{i=1}^{p}a_{ii}$. As for matrix norms: Frobenius norm $\|\boldsymbol{A}\|_F:=\sqrt{\text{Tr}(\boldsymbol{A}^{\top}\boldsymbol{A})}$, operator norm $\|\boldsymbol{A}\|_{op}=\sqrt{\sigma_1(\boldsymbol{A}^{\top}\boldsymbol{A})}$ and entry-wise maximum norm $\|\boldsymbol{A}\|_{\infty}:=\max_{1\leq i\leq p,1\leq j\leq q}|a_{ij}|$. The column space of $\boldsymbol{A}$, denoted by $\text{span}(\boldsymbol{A})$, is the linear subspace spanned by column vectors of $\boldsymbol{A}$. The $i$-th row of $\boldsymbol{A}$ is denoted by $\boldsymbol{A}_{i,*}$ and the $j$-th column of $\boldsymbol{A}$ is denoted by $\boldsymbol{A}_{*, j}$. Let $\text{supp}(\boldsymbol{A}):=\{i\in[q]\mid\|\boldsymbol{A}_{i,*}\|_2>0\}$ denote the row support of $\boldsymbol{A}$. For any subset $E\subset[p]$, let $|E|$ and $E^c:=[p]/E$ denote the cardinality and complement set of $E$, respectively. For two subsets $E$ and $F$, $\boldsymbol{A}_{E,F}$ represents the $|E|\times|F|$ submatrix formed by $(a_{ij})_{i\in E,j\in F}$. For any event $G$, the corresponding indicator function is denoted by $\mathbbm{1}\{G\}$. For a matrix $\B$, we define its projection matrix as $\boldsymbol{P}_{\B}=\B(\B^{\top}\B)^{-}\B^{\top}$, where $(\cdot)^{-}$ is the Moore–Penrose inverse. For two real-valued sequences $\{a_n\}$ and $\{b_n\}$, we write $a_n\lesssim b_n$ if $a_n\leq c b_n$ for a constant $c\in(0,\infty)$, and $a_n\gtrsim b_n$ if $a_n\geq c_0b_n$ for a constant $c_0\in(0,\infty)$. Finally, for a sequence $\{\boldsymbol{z}_i\}_{i=1}^{n}$ and a measurable set $I$, we write $\E_n[\boldsymbol{z}_i|\boldsymbol{z}_i\in I]=\sum_{i=1}^{n}\boldsymbol{z}_i\mathbbm{1}\{\boldsymbol{z}_i\in I\}/\sum_{i=1}^{n}\mathbbm{1}\{\boldsymbol{z}_i\in I\}$, that is, $\E_n[\cdot\mid\cdot]$ abbreviates the empirical conditional expectation.
	
	\section{Preliminaries}
	
	\subsection{Sliced Inverse Regression}
	
	Assume that the response $Y$ given the covariates $\boldsymbol{x}$ follows the multiple indexed model:
	\begin{equation}\label{eq:mim}
		Y=f(\boldsymbol{x}^{\top}\boldsymbol{\beta}_1,\dots,\boldsymbol{x}^{\top}\boldsymbol{\beta}_k,e),
	\end{equation}
	where $f(\cdot)$ is an unknown deterministic function and $e$ is the random error. To estimate $\B=(\boldsymbol{\beta}_1,\dots,\boldsymbol{\beta}_k)\in\mR^{p\times k}$, \cite{li1991sliced} proposed to divide the support of $Y$ into $H$ slices, where $H>k$. Denote the slices by $I_h=(q_{h-1},q_{h}]$ for $h=1,\dots,H$, where $-\infty= q_0\leq\dots\leq q_H=\infty$. The kernel matrix is defined by 
	\begin{equation}
		\label{eq:kernel}
		\M:= \sum_{l=1}^{H}\P(Y\in I_l)\text{Cov}\{\mathbb{E}(\boldsymbol{x}\mid Y\in I_l)\}.
	\end{equation}
	Let $\boldsymbol{I}_k$ be the $k\times k$ diagonal matrix. Then $\B$ can be estimated by:
	\begin{equation}
		\label{eq:tr_generalized_eigenvector}
		\B:=\argmax_{\B\in\mR^{p\times k}}\text{Tr}(\B^{\top}\M\B)\quad\text{such that}\quad\B^{\top}\boldsymbol{\Sigma}\B=\boldsymbol{I}_k,
	\end{equation}
	
	When the dimensionality $p$ is greater than sample size $n$, it is common to assume that the columns of $\B$ are sparse. Specifically, there exists a subset of the covariates $\boldsymbol{x}_{\mathcal{A}}\subset\boldsymbol{x}$ such that $Y\perp \!\!\! \perp\boldsymbol{x}\mid\boldsymbol{x}_{\mathcal{A}}$, where $\mathcal{A}\subset[p]$ and $|\mathcal{A}|< n$ \citep{yin2015sequential}. When $p$ grows exponentially with $n$, we follow  \cite{lin2019sparse, lin2021optimal}  and assume that the number of elements in $\text{supp}(\boldsymbol{\beta}_i)$ is less than $s$, where $s$ is a sparsity constraint parameter.

	\subsection{Differential Privacy}
	
	In this section, we introduce the necessary definitions and tools of differential privacy. The formal definition of differential privacy is as follows.
	
	\begin{defi}[Differential Privacy \citep{dwork2006calibrating}]
		\label{def:dp}
		A randomized algorithm $M(\cdot):\mathcal{D}\to\mathcal{R}$ is $(\varepsilon,\delta)$-DP for $\varepsilon,\delta>0$ if for every pair of neighboring data sets $D,D'\in\mathcal{D}$ that differ by one individual datum and every measurable set $\mathcal{S}\subset\mathcal{R}$ with respect to $M(\cdot)$,
		\begin{equation}
			\mathbb{P}\{M(D)\in\mathcal{S}\}\leq e^{\varepsilon}\mathbb{P}\{M(D')\in\mathcal{S}\}+\delta,
		\end{equation}
		where the probability measure $\mathbb{P}$ is induced by the randomness of $M(\cdot)$ only.
	\end{defi}
	In the definition, two data sets $D$ and $D'$ are treated as fixed, and the probability measures the randomness in the mechanism $M(\cdot)$. The level of privacy against an adversary is determined by the likelihood ratio of randomized outputs between two neighboring datasets, and this can be controlled by private parameters $(\varepsilon,\delta)$. When $\varepsilon$ and $\delta$ are small, the privacy constraint becomes more stringent, offering stronger privacy guarantees. Next, we introduce the definition of sensitivity, which characterizes the change in the algorithm output when only one single entry in the dataset is altered.
	
	\begin{defi}[Sensitivity]
		\label{def:sensitivity}
		For a vector-valued deterministic algorithm $\mathcal{T}(\cdot):D\in\mathcal{D}\to\mR^{p}$, the $\ell_q$ sensitivity of $\mathcal{T}(\cdot)$ is defined as
		\begin{equation}
			\Delta_q(\mathcal{T}):=\sup_{D,D'\in\mathcal{D}}\|\mathcal{T}(D)-\mathcal{T}(D')\|_q,
		\end{equation}
		where $D$ and $D'$ only differ in one single entry.
	\end{defi}
	
	In scenarios where the parameter is of high dimension, disclosing the entire estimation can result in a substantial private loss. However, exploiting the sparsity assumption allows us to disclose only the significant nonzero coordinates selectively. The {\it peeling} algorithm \citep{dwork2021differentially, xia2023adaptive, cai2023private} is a DP algorithm that identifies the top-$s$ most significant coordinates. We modify the {\it peeling} algorithm to accommodate matrix-valued parameters. The privacy guarantee is summarized in Lemma \ref{lem:peeling}.
	
	\begin{lemma}[Extended from Theorem 3 in \cite{dwork2021differentially}]
		\label{lem:peeling}
		For any matrix-valued function $\boldsymbol{A}(\cdot):\mathcal{D}\to\mR^{d_1\times d_2}$ satisfying $\|\boldsymbol{A}(D)-\boldsymbol{A}(D^{\prime})\|_{\infty}\leq\Delta_A$, where $D^{\prime}$ is a neighboring data set of $D$, 
		Algorithm \ref{alg:vector_noisy_ht} is $(\varepsilon,\delta)$-DP when $\sigma\geq\Delta_A$.
	\end{lemma}
	
	\begin{algorithm}
		\caption{Matrix-Noisy Hard Thresholding (Peeling)}
		\begin{algorithmic}[1]
			\Require  matrix-valued function $\boldsymbol{A}(\cdot)\in\mR^{d_1\times d_2}$, dataset $D$, sparsity $s^{\prime}$, privacy parameters $(\varepsilon,\delta)$, noisy scale $\sigma$.
			\State Initialization $S=\emptyset$;
			\For {$i=1,2,\dots,s^{\prime}$}
			\State Generate $\boldsymbol{w}_i\in\mR^{d_2}$ with $w_{1,i},\dots,w_{d_2,i}\stackrel{i.i.d.}{\sim}$Laplace$\big(\sigma\cdot 2\sqrt{3d_1s^{\prime}\log(2/\delta)}/\varepsilon\big)$;
			\State Select $j=\arg\max_{j\in[d_2]\backslash S}\|\boldsymbol{A}(D)_{*,j}\|_2$+$w_{j,i}$ and add $j$ to $S$, where $\boldsymbol{A}(D)_{*,j}$ is $j$-th column of matrix $\boldsymbol{A}(D)$;
			\EndFor
			\State Generate $\boldsymbol{\tilde{w}}\in\mR^{d_1\times s^{\prime}}$ with $\{w_{k,l}\}_{1\leq k\leq d_1,1\leq l\leq s^{\prime}}\stackrel{i.i.d.}{\sim}$ $N\big(\sigma\cdot 2\sqrt{2d_1s^{\prime}\log(2.5/\delta)}/\varepsilon\big)$;
			\Ensure $\boldsymbol{A}(D)_{*,S}+\boldsymbol{\tilde{w}}$.
		\end{algorithmic}
		\label{alg:vector_noisy_ht}
	\end{algorithm}
	
	\section{DP-SIR in Low Dimensions}\label{sec:method}
	
	\subsection{Minimax Lower Bound with Privacy Constraints}
	
	In this section, we analyze the minimax lower bound for the DP-SIR when the sample size $n$ is larger than the dimension $p$, and $p$ is allowed to increase with $n$. Specifically, $p=n^\alpha$ and $\alpha\in(0,1)$. We consider the following distribution space:
	\begin{align*}
		\mathcal P=&\bigg\{\{\bm x_i,Y_i\}_{i=1}^{n}\text{ are independently and identically distributed, such that:}\\
		&\bm x_i \in\mR^p\text{ follows a sub-Gaussian distribution with covariance matrix } \boldsymbol{\Sigma}; \\
		&\lambda_x^{-1}\le\lambda_{\min}(\boldsymbol{\Sigma})\le\lambda_{\max}(\boldsymbol{\Sigma})\le \lambda_x;Y_i\mid\bm x_i\text{ follows the model \eqref{eq:mim}};\\
		& \lambda\le\lambda_{k}(\boldsymbol{\Sigma}^{-1/2}\boldsymbol{M}\boldsymbol{\Sigma}^{-1/2})\leq\lambda_{1}(\boldsymbol{\Sigma}^{-1/2}\boldsymbol{M}\boldsymbol{\Sigma}^{-1/2})\leq \kappa\lambda\text{, where }\boldsymbol{M}\text{ is defined in \eqref{eq:kernel}}.\bigg\}.
	\end{align*}
	The parameter $\lambda$ is called the eigen-gap of the kernel matrix $\boldsymbol{M}$. The parameters $\lambda_x$ and $\kappa$ are positive constants. We allow $p$ and $\lambda$ to vary with the sample size $n$, and assume the dimension $k$ and slices $H$ to be bounded. Similar conditions were used in \cite{lin2019sparse,tan2020sparse} to study the minimax lower bound for non-private sparse SIR. 
	
	Let $\boldsymbol{\hat{B}}\in\mathbb{R}^{p\times k}$ be a $(\varepsilon,\delta)$-differentially private estimator for $\boldsymbol{B}$. Although $\B$ is not identifiable, the projection matrix of $\B$ is unique. Let $\boldsymbol{P}_{\B}$ be the projection matrix for $\B$, we define the loss function as
	$L(\boldsymbol{\hat{B}},\B)=\|\boldsymbol{P}_{\boldsymbol{\hat{B}}}-\boldsymbol{P}_{\B}\|_F^2$, where $\boldsymbol{P}_{\boldsymbol{\hat{B}}}$ is the projection matrix of $\boldsymbol{\hat{B}}(\boldsymbol{\hat{B}}\boldsymbol{\Sigma}\boldsymbol{\hat{B}})^{-1/2}$. The following theorem summarizes the minimax lower bound for DP-SIR in low dimensions.
	\begin{theorem}\label{thm:lower bound lowdim}
		Consider the parameter space defined above, and 
		let $\mathcal M_{\varepsilon,\delta}$ be the space of all $(\varepsilon,\delta)$-DP algorithms.  If $ p \lesssim n\varepsilon\lambda$, $0 < \varepsilon < 1$ and $\delta \lesssim n^{-(1+\gamma)}$ for some $\gamma > 0$, we have 
		\begin{equation}
			\label{eq:minimax-lower}
			\inf_{\boldsymbol{\hat{B}}\in\mathcal M_{\varepsilon,\delta}}\sup_{P\in\mathcal P}\E L(\boldsymbol{\hat{B}},\B)\gtrsim\frac{p}{\lambda^2n}+\frac{p^2}{\lambda^2n^2\varepsilon^2}.
		\end{equation}
	\end{theorem}
	
	Different from the DP linear model discussed in \cite{cai2021cost}, the results in (\ref{eq:minimax-lower}) involves $\lambda$, i.e., the eigen-gap of the kernel matrix $\boldsymbol{M}$. Note that $\lambda$ appears in both terms; the first term represents the statistical error, while the second term accounts for the additional requirement of differential privacy. Later we will introduce a DP estimator that achieves the lower bound \eqref{eq:minimax-lower} within logarithmic factors of $n$.
	
	\subsection{Private Estimation of Slices}
	\label{sec:dp-hist}
	A crucial problem in SIR is to divide the support of $Y$ into $H$ slices for a given integer $H$. Although the problem is well studied for non-private estimation, it may risk privacy loss under the $(\varepsilon,\delta)$-DP framework. When $Y$ is categorical, the slices can be naturally defined by unique values of $Y$. For example, if $Y$ is binary, we would select $H=2$ slices corresponding to the distinct values of $Y$ without any privacy issue. However, when $Y$ is continuous, traditional data-driven slicing methods, such as using the quantiles of $Y$ \citep{cai2020online}, will lead to a violation of the differential privacy principle. This section proposes a DP-histogram approach to cut the slices for $Y$. The approach applies not only to continuous or categorical responses but can also be extended to mixed response types.
	
	We begin by defining \textit{bins} as a partition of the domain of $Y$. When the response $Y$ is continuous, without loss of generality, we assume that the domain of $Y$ is bounded in $[-1,1]$. Because SIR remains invariant for monotonic transformation \citep{li1991sliced} and the kernel matrix $\M$ depends on $Y$ only through the indicator function $\mathbbm{1}\{Y\in I_h\}$, the kernel matrix $\M$ remains unchanged when $Y$ and $\{q_h\}_{h=0}^{H}$ are monotonically transformed. Without information about the range of $Y$, we propose simply using the transformation $(\pi/2)^{-1}\arctan(Y)$, which maps $\mathbb{R}$ to $(-1,1)$. We partition the domain of $Y$ into $m$ bins of equal size, denoted by $\{B_1,\dots,B_m\}$, each of which has width $2/m$. 
	
	Let $b_j$ represent the number of observations in bin $B_j$, and define private estimates as $d_j = \max\{b_j + \xi_j,0\}$, where $\xi_1,\dots,\xi_m \stackrel{i.i.d.}{\sim} \mbox{Laplace} (2/\varepsilon)$. When the response variable $Y$ is continuous, the estimated probability density function is
	\begin{equation}
		\label{eq:dp_hist}
		\hat{f}_Y(y)=\frac{m}{2}\sum_{j=1}^{m}\hat{f}_j\mathbbm{1}(y\in B_j),
	\end{equation}
	where $\hat{f}_j=d_j/\sum_{i=1}^{m}d_i$. Define $\hat F_Y(y)=\int_{-1}^{y}\widehat{f}_Y(t)dt$ as the estimated cumulative distribution function and $F_Y(y)$ as the true cumulative distribution function of $Y$. The true slices are defined by $I_h=(q_{h-1},q_h]$, where $q_0=-1$, $q_H=1$ and $q_h=\inf\{y\in\{-1+2/m,\dots,1-2/m\}: F_Y(y)\geq h/H\}$ for $h=1,\dots,H-1$. We propose estimating the cutoff points of the slices by $\hat{q}_0=-1$, $\hat{q}_H=1$, $\hat{q}_h=\inf\{y\in[-1,1]: \hat{F}_Y(y)\geq h/H\}$ for $h=1,\dots,H-1$. The corresponding estimated slices are defined by $\widehat{I}_h=(\hat{q}_{h-1},\hat{q}_h]$. The following lemma establishes the privacy guarantee and consistency of the proposed method.
	
	\begin{lemma}
		\label{lem:est_slices}
		The estimated slices $\{\widehat{I}_1,\dots,\widehat{I}_H\}$ are $(\varepsilon,0)$-differentially private. Assume that the density of the variable $Y$, $f_Y(\cdot)$, is Lipschitz continuous and bounded. Suppose that the values of density function evaluated at $q_h$ for $h=1,\dots,H-1,$ satisfy $f_Y(q_h)\geq C_*$ for a positive constant $C_*$, and $m\asymp n^{1/3}$, $\sqrt{\log(n)}/\varepsilon=o(n^{1/6})$. Then, the estimated cutoff points of the slices satisfy $|\hat{q}_h-q_h|\leq 2/n^{1/3}$ for $h=0,\dots,H,$ with probability at least $1-\exp\{-C^{\prime}\log(n)\}$ for a positive constant $C^{\prime}$.
	\end{lemma}
	
	The Lipschitz condition on the probability density function $f_Y(\cdot)$ is satisfied by many continuous distributions and is a common assumption in various studies; see, for example, \cite{wasserman2010statistical}. The second assumption, which imposes a lower bound on the density function at $q_h$ for $h=1,\dots,H-1$, ensures that the quantiles in the cutoff points are distinguishable. This assumption is crucial for a consistent estimation of the slice boundaries. However, it is not strictly necessary for the sliced inverse regression. When two points are indistinguishable, choosing between them is expected to have minimal impact. Lemma \ref{lem:est_slices} establishes that the estimated slices differ from the true slices by at least one bin with probability approaching $1$ as $n\to\infty$. We now briefly discuss the impact of the estimated slices on the kernel matrix with the formal proof provided in the appendix. Under regularity conditions, the eigen-gap of the kernel matrix constructed from the estimated slices converges to that of the kernel matrix with true slices as $n\to\infty$. Moreover, under certain smoothness and tail conditions on $\E(Y\mid\boldsymbol{x})$, the sliced stability condition \citep{lin2021optimal} guarantees that any selection of slices with adequate observations in each slice results in a comparable eigen-gap, when $H$ is a sufficiently large constant \citep{lin2019sparse}.
	
	\subsection{Algorithm in Low Dimensions}
	
	In this section, we introduce a practical algorithm that achieves the optimal rate while ensuring privacy. Consider the data $\{\boldsymbol{x}_i,Y_i\}_{i=1}^{n}$, which are independently and identically drawn from model (\ref{eq:mim}). Without loss of generality, we assume the samples are centered: $\bar{\boldsymbol{x}}_n:=\sum_{i=1}^{n}\boldsymbol{x}_i/n=\boldsymbol{0}$. The sample estimator of the covariance matrix for $\{\boldsymbol{x}_i\}_{i=1}^{n}$ is 
	\begin{equation*}
		\boldsymbol{\hat{\Sigma}}:=\frac{1}{n}\sum_{i=1}^{n}(\boldsymbol{x}_i-\bar{\boldsymbol{x}}_n)(\boldsymbol{x}_i-\bar{\boldsymbol{x}}_n)^{\top}=\frac{1}{n}\sum_{i=1}^{n}\boldsymbol{x}_i\boldsymbol{x}_i^{\top}.
	\end{equation*}
	Let $\{\hat{I}_h\}_{h=1}^{H}$ be the estimated slices obtained from the DP histogram as discussed in Section \ref{sec:dp-hist}. The corresponding sample estimator of the kernel matrix is given by
	\begin{equation*}
		\begin{split}
			\boldsymbol{\hat{M}}:&=\sum_{h=1}^{H}\hat{p}_h\{\mathbb{E}_n(\boldsymbol{x}_i\mid Y_i\in \hat{I}_h)-\bar{\boldsymbol{x}}_n\}\{\mathbb{E}_n(\boldsymbol{x}_i\mid Y_i\in \hat{I}_h)-\bar{\boldsymbol{x}}_n\}^{\top}\\
			&=\sum_{h=1}^{H}\hat{p}_h\mathbb{E}_n(\boldsymbol{x}_i\mid Y_i\in \hat{I}_h)\mathbb{E}_n(\boldsymbol{x}_i\mid Y_i\in \hat{I}_h)^{\top},
		\end{split}
	\end{equation*}
	where $\hat{p}_h=\sum_{i=1}^{n}\mathbbm{1}\{Y_i\in \hat{I}_h\}/n$ and $\mathbb{E}_n(\boldsymbol{x}_i\mid Y_i\in \hat{I}_h)=\sum_{i=1}^{n}\boldsymbol{x}_i\mathbbm{1}\{Y_i\in \hat{I}_h\}/(n\hat{p}_h)$. To estimate the unknown projection matrix $\boldsymbol{B}$, we consider the following relaxation of (\ref{eq:tr_generalized_eigenvector}):
	\begin{equation}
		\label{eq:gev}
		\boldsymbol{\hat{B}}=\argmin_{\B\in\mR^{p\times k}}-\text{Tr}(\B^{\top}\boldsymbol{\hat{M}}\B)+\lambda_{penalty}\|\B^{\top}\boldsymbol{\hat{\Sigma}}\B-\boldsymbol{I}_k\|_F^2.
	\end{equation}
	We use the stochastic gradient descent algorithm with one pass of the dataset to estimate $\boldsymbol{\hat{B}}$ in equation (\ref{eq:gev}). The gradient of (\ref{eq:gev}) is $-2\boldsymbol{\hat{M}}\boldsymbol{B}+2\lambda_{penalty}\boldsymbol{\hat{\Sigma}}\B(\B^{\top}\boldsymbol{\hat{\Sigma}}\B-\boldsymbol{I}_k)$. We first split the data into $T$ subsets of roughly equal sizes. Then, in each step, we evaluate the gradient using sub-data and perform truncation on the gradients to ensure the bounded sensitivity of the algorithm. The algorithm details are provided in Algorithm \ref{alg:ld}.
	\begin{algorithm}
		\caption{DP-SIR when $n>p$}
		\begin{algorithmic}[1]
			\Require dataset $\{\boldsymbol{x}_i,Y_i\}_{i=1}^{n}$, step size $\eta$, initial value $\boldsymbol{\hat{B}}^{(0)}$, iterations $T$, penalty $\lambda_{penalty}$, noise scale $\sigma$, privacy parameter $(\varepsilon,\delta)$, truncation level $C$ and $R$.
			\State Estimate sliced intervals $\{\hat{I}_h\}_{h=1}^{H}$ based on the DP histogram \eqref{eq:dp_hist} under $(\varepsilon,0)$-differential privacy;
			\State Random split data into $T$ parts of roughly equal sizes: $[n]=S_0\cup\dots\cup S_{T-1}$;
			\For {$t=0,1,2,\dots,T-1$}
			\State Generate $\boldsymbol{w}_t\in\mR^{p}$ with $w_{1,t},\dots,w_{p,t}\stackrel{i.i.d.}{\sim}N(0,2\sigma^2T^2\log(1.25T/\delta)/\varepsilon^2)$;
			\State Use sub-data $S_t$ to update:
			\begin{equation*}
				\begin{split}
					&\boldsymbol{\hat{B}}^{(t+0.5)}=\boldsymbol{\hat{B}}^{(t)}-2\eta\bigg[-\sum_{h=1}^{H}\mathbb{E}_n(\boldsymbol{x}_i\mid Y_i\in \hat{I}_h,i\in S_{t})\sum_{Y_i\in \hat{I}_h,i\in S_{t}}\Pi_{R}(\boldsymbol{x}_i^{\top}\boldsymbol{\hat{B}}^{(t)})/|S_t|\\
					&+\lambda_{penalty}\sum_{i\in S_{t}}\boldsymbol{x}_i\Pi_{R}(\boldsymbol{x}_i^{\top}\boldsymbol{\hat{B}}^{(t)})/|S_{t}|\big\{\sum_{i\in S_{t}}\Pi_{R}(\boldsymbol{\hat{B}}^{(t)\top}\boldsymbol{x}_i)\Pi_{R}(\boldsymbol{x}_i^{\top}\boldsymbol{\hat{B}}^{(t)})/|S_{t}|-\boldsymbol{I}_k\big\}\bigg],
				\end{split}
			\end{equation*}
			where $\Pi_{R}(\cdot)$ is the projection of each component of $k$-dimensional vector onto the ball $\mathbb{B}_{R}:=\{X\in\mR\mid |X|\leq R\}$ and $|S_t|$ is the number of elements in the set $S_t$;
			\State $\boldsymbol{\hat{B}}^{(t+1)}=\Pi_{C}(\boldsymbol{\hat{B}}^{(t+0.5)}+\boldsymbol{w}_t)$, where $\Pi_C(\boldsymbol{\hat{B}}^{(t+0.5)})$ is the projection of each column of $\boldsymbol{\hat{B}}^{(t+0.5)}$ onto the $l_2$ ball $\mathbb{B}_C^p=\{\boldsymbol{x}\in\mR^p\mid \|\boldsymbol{x}\|_2\leq C\}$;
			\EndFor
			\Ensure $\boldsymbol{\hat{B}}^{(T)}$.
		\end{algorithmic}
		\label{alg:ld}
	\end{algorithm}
	
	In the algorithm, we consider the scenario where slices are estimated in a differentially private manner. As detailed in Section \ref{sec:dp-hist}, if the response variable $Y$ is categorical or prior knowledge about the distribution of $Y$ is available, we recommend using natural slices without relying on the data. In Algorithm \ref{alg:ld}, we allocate the $(\varepsilon,0)$ privacy budget for slice estimation and the $(\varepsilon,\delta)$ for the gradient descent procedure. Since the performance of sliced inverse regression is not highly sensitive to the choice of slices in finite sample scenarios, we suggest assigning a smaller portion of the privacy budget to the histogram estimation. For example, one might allocate $(0.1\varepsilon,0)$ privacy budget for slice estimation and an additional $(\varepsilon,\delta)$ privacy budget for the stochastic gradient descent procedure.
	
	As will be discussed in Theorem \ref{thm:dpsir_errorbound}, the initial private value of Algorithm \ref{alg:ld}, denoted as $\boldsymbol{\hat{B}}^{(0)}$, should be both differentially private and within a constant error ball of the true parameter. This assumption is relatively mild. We propose a simple procedure that directly adds random noises to the kernel matrix and the covariance matrix to estimate the initial value. Due to the space limit, we summarize the procedure in Algorithm B.1 of the supplement. We prove in Lemma B.1 that the initial estimator is differentially private and prove in Theorem B.1 that the initial estimator satisfies the constant error bound requirement.
	
	\subsection{Privacy Guarantee and Upper Bound}
	
	In this section, we provide the theoretical guarantee on the upper bound of the proposed algorithm. To facilitate the analysis, we introduce the following technical assumptions, which will be discussed in detail shortly.
	\begin{assumption}A
		\label{assump:low_SIR}
		$\ $
		\begin{enumerate}
			\item Linearity condition: For $\B\in\mR^p$, the conditional expectation $\mathbb{E}(\boldsymbol{x}\mid\B^{\top}\boldsymbol{x})$ is linear in $\B^{\top}\boldsymbol{x}$; that is, there exists $c_0\in\mR$ and $\boldsymbol{c}\in\mR^{k}$, such that $$\mathbb{E}(\boldsymbol{x}\mid\B^{\top}\boldsymbol{x})=c_0+\boldsymbol{c}^{\top}\B^{\top}\boldsymbol{x}.$$
			\item Refined coverage condition:  The generalized eigendecomposition (GED) of $\M$ and $\boldsymbol{\Sigma}$ aims to find the eigenvalues $\lambda_1\geq\lambda_2\geq\dots\geq\lambda_p$ and eigenvectors $\v_i$ for $i=1,\dots, p$, that satisfy the equation: $\boldsymbol{M}\v_i = \lambda_i \boldsymbol{\Sigma} \v_i$ for $i=1,\dots, p$. The ordered generalized eigenvalues corresponding to $\M$ and $\boldsymbol{\Sigma}$ satisfy
			\begin{equation*}
				1\geq\kappa\lambda\geq\lambda_1\geq\dots\geq\lambda_k>\lambda>\lambda_{k+1}=\dots=\lambda_{p}=0,
			\end{equation*}
			where $\kappa>0$ is a fixed constant.
		\end{enumerate}
	\end{assumption}
	
	In Assumption \ref{assump:low_SIR}, the parameters $p$ and $\lambda$ are allowed to depend on the sample size $n$, with $p$ permitted to grow to infinity as $n$ increases. In the fixed slices framework, the number of slices $H$ is treated as a bounded integer, and the dimension of subspace $k$ is also an integer satisfying $k\leq H-1$. The linearity condition is commonly assumed in the sufficient dimension reduction literature, as demonstrated by \cite{cook2004testing}. This condition holds for covariates that follow an elliptically symmetric distribution, with the normal distribution being a special case. \cite{hall1993almost} showed that as the dimensionality $p$ increases, the linearity condition tends to hold reasonably well in many scenarios. \cite{li1991sliced} established that the linearity condition implies the identification of the central subspace $\mathcal{S}_{Y\mid\boldsymbol{x}}$ through the eigenspace of $\boldsymbol{\Sigma}^{-1}\text{Cov}\{\mathbb{E}(\boldsymbol{x}\mid Y)\}$. The refined coverage condition implies the convergage condition, $\text{span}[\boldsymbol{\Sigma}^{-1}\text{Cov}\{\mathbb{E}(\boldsymbol{x}\mid \tilde{Y})\}]=\mathcal{S}_{Y\mid \boldsymbol{x}}$, where $\tilde{Y}:=\sum_{h=1}^{H}\mathbbm{1}\{Y\in I_h\}$, suggesting that using sliced $Y$ is sufficient to identify the central subspace $\mathcal{S}_{Y\mid\boldsymbol{x}}$. In this paper, we assume that the refined coverage condition holds for the sliced matrix $\boldsymbol{\Sigma}^{-1}\text{Cov}\{\mathbb{E}(\boldsymbol{x}\mid \tilde{Y})\}$, which is equivalent to assuming that the coverage condition holds for the original matrix $\boldsymbol{\Sigma}^{-1}\text{Cov}\{\mathbb{E}(\boldsymbol{x}\mid Y)\}$ and that the \textit{sliced stable condition} holds for the \textit{central curve} $\mathbb{E}(\boldsymbol{x}\mid Y)$. We refer interested readers to \cite{lin2021optimal} for a more detailed discussion about the sliced stable condition. 
	
	\begin{assumption}B
		\label{assum:low_design}
		\begin{enumerate}
			\item The true parameter $\B$ satisfies $\|\B_{*,j}\|_2\leq c_0$ for some constant $c_0$ and $1\leq j\leq k$.
			\item Bounded design: there is a positive constant $c_x$ such that $\|\boldsymbol{x}_i\|_{\infty}\leq c_x$ for all $i=1,\dots,n$. 
			\item Bounded moments of the design: the random vector $\boldsymbol{x}_i$ follows the sub-Gaussian distribution, and the eigenvalues of covariance matrix $\boldsymbol{\Sigma}$ satisfy 
			\begin{equation*}
				\lambda_x^{-1}\leq \lambda_{min}(\boldsymbol{\Sigma})\leq \lambda_{max}(\boldsymbol{\Sigma})\leq\lambda_x
			\end{equation*}
			for a positive constant $\lambda_x$, where $\lambda_{min}(\boldsymbol{\Sigma})$ and $\lambda_{max}(\boldsymbol{\Sigma})$ denote the minimum and maximum eigenvalue of $\boldsymbol{\Sigma}$, respectively.
			\item Abundant observations: there exists a fixed positive constant $0<p_H<1$ such that the sample size within each slice $n_h=\sum_{i=1}^{n}\boldsymbol{1}\{Y_i\in I_h\}$ satisfies the condition $n_h\geq p_H n$ for all $h=1,\dots, H$.
		\end{enumerate}
	\end{assumption}
	
	
	The analysis of DP algorithms relies on the regularity condition imposed on the design, as outlined in Assumption \ref{assum:low_design}. The first condition assumes that the true parameter does not have irregular values. The second assumption, regarding a bounded design, is commonly used in the differential privacy literature \citep{cai2021cost} to ensure bounded sensitivity. The condition can be relaxed by using robust statistical loss functions such as the Huber loss \citep{avella2021differentially}. The condition also holds for real applications on gene expression and sales data considered in our paper. The third assumption, concerning the distribution of covariates and the covariance matrix, is frequently employed in high-dimensional statistics \citep{wainwright2019high}. It is important to note that the bounded design condition does not imply the sub-Gaussian condition for multivariate random variables. However, we can demonstrate that the second assumption holds with high probability under the third assumption for $c_x=O\{\log^{1/2}(n)\}$. The resulting error bounds would inflate to logarithm factors of $n$ under the low dimension setting and logarithm factors of $p$ under the high dimension setting. The fourth assumption requires a sufficient number of samples within each slice, which can be easily satisfied when slice boundaries are chosen based on DP histograms, as discussed in Section \ref{sec:dp-hist}.
	
	The following lemma demonstrates that the Algorithm \ref{alg:ld} is $(2\varepsilon,\delta)$-differentially private. Additionally, when $Y$ is categorical, the lemma indicates that Algorithm \ref{alg:ld} is $(\varepsilon,\delta)$-differentially private because we saved the privacy budget for slice estimation.
	
	\begin{lemma}[Privacy Guarantee of Algorithm \ref{alg:ld}]
		\label{lem:low_privacy}
		Under Assumption \ref{assum:low_design}, and truncation parameter $C\geq c_0$ and $\sigma$ satisfies 
		$$\sigma\geq 2\eta\{7Rc_x+\lambda_{penalty}(2Rc_x+4kR^3c_x)\}\sqrt{pk}T/n,$$
		then Algorithm \ref{alg:ld} is $(2\varepsilon,\delta)$-differentially private.
	\end{lemma}
	
	The convergence of Algorithm \ref{alg:ld} requires a regularity condition on the initial value $\boldsymbol{\hat{B}}^{(0)}$ and appropriate choice of step size and penalty parameter. When those conditions are satisfied, we prove the upper error bound of the proposed estimator in Theorem \ref{thm:dpsir_errorbound}. 
	
	\begin{theorem}[Convergence of Algorithm \ref{alg:ld}]
		\label{thm:dpsir_errorbound}
		Under Assumption \ref{assump:low_SIR} and \ref{assum:low_design}, further assume that $1/(n^{2/3}\lambda^2)\leq c$, $p\log n/(n\lambda^2)\leq c$ and $p^{2}\log(1/\delta)\log(n)^7/(n^2\lambda^2\varepsilon^2)\leq c$ for a small enough constant $c$. We choose the penalty parameter $\lambda_{penalty}=\lambda/c_1$, the step size $\eta=c_2/\lambda$ for small positive constants $c_1,c_2$, the number of iteration $T=c_3\log(n)$, the truncation level $R=2c_4C\sqrt{\log n}$ for $C\geq c_0$ and the constant $c_4$ only depending on the distribution of $\boldsymbol{x}$. Assume the initial value satisfies, $L(\boldsymbol{\hat{B}}^{(0)},\B)\leq C_1$, where $C_1$ is a small positive constant. Then, for any $C^{\prime}>0$, there exist a positive constant $C^{\prime\prime}>0$, such that the differentially private sliced inverse regression estimator $\boldsymbol{\hat{B}}^{(T)}$ in Algorithm \ref{alg:ld} satisfies
		\begin{equation}
			\label{eq:dpsir_error}
			L(\boldsymbol{\hat{B}}^{(T)},\B)=
			\|\boldsymbol{P}_{\boldsymbol{\hat{B}}^{(T)}}-\boldsymbol{P}_{\B}\|_F^2\leq C^{\prime\prime}\frac{p\log(n)}{\lambda^2n}+C^{\prime\prime}\frac{p^{2}\log(1/\delta)\log^7(n)}{\lambda^2n^2\varepsilon^2},
		\end{equation}
		with probability greater than $1-\exp\{-C^{\prime}\log(p)\}$. 
	\end{theorem}
	
	Besides Assumptions \ref{assump:low_SIR} and \ref{assum:low_design}, we also introduced assumptions concerning the parameter $\lambda$, which can be intuitively understood as conditions on the minimum strength of the underlying signal. The first assumption on $\lambda$ is an additional assumption due to the estimation of the slices. The second assumption on $\lambda$ is commonly employed in the statistical literature and has been observed in prior works on the sliced inverse regression, such as \cite{tan2020sparse}. The third assumption on $\lambda$ is an additional requirement accounting for the estimation costs of differential privacy. These assumptions on $\lambda$ are crucial in establishing the upper bound of the proposed algorithm.
	
	The error bound in (\ref{eq:dpsir_error}) consists of two main components. The first component corresponds to the statistical error, capturing the inherent uncertainty in estimating the target parameter. The second component represents an additional term stemming from the constraints imposed by differential privacy, accounting for the privacy-related noises introduced in the estimation process. Note that the upper bound matches the minimax lower bound in Theorem \ref{thm:lower bound lowdim} up to constant and logarithm terms.
	
	In practice, the true dimension $k$ is often unknown. Adopting the approach from \cite{zhu2006sliced}, we suggest initially using a large $H$, for example $H=\log(n)$, to estimate the kernel matrix $\boldsymbol{M}$ and the covariance matrix $\boldsymbol{\Sigma}$ by the Algorithm \ref{alg:ld_ini} of the appendix, followed by the application of the BIC criterion to select $k$. Formally, let $\tilde{\lambda}_1\geq\dots\geq\tilde{\lambda}_{H}$ represent the $H$ largest eigenvalues of the kernel matrix from Algorithm \ref{alg:ld_ini}, which are differentially private according to Theorem \ref{thm:dpsir_errorbound_ini}. Define the BIC for $l=1,\dots,H$ as:
	\begin{equation*}
		G(l)=n\sum_{i=1}^{l}\tilde{\lambda}_i^2/\sum_{i=1}^{H}\tilde{\lambda}_{i}^2-C_n\times{l(l+1)/2}.
	\end{equation*}
	The chosen dimension $\tilde{k}$ is determined by the largest BIC: $\tilde{k}=\arg\max_{1\leq l\leq H}G(l)$.
	\begin{theorem}
		\label{thm:sir_bic}
		Under the conditions in Theorem \ref{thm:dpsir_errorbound}, if the tuning parameter 
		$C_n$ satisfies $C_n/n\to 0$ and $C_n\lambda^2/\{p+p^3\log(1/\delta)/(n\varepsilon^2)\}\to\infty$ as $n\to\infty$,  then $\tilde{k}=k+o_p(1)$.
	\end{theorem}
	
	In practice, without any additional information about $\lambda$ and $k$, we suggest the following procedure. Firstly, the differentially private histogram method is applied to estimate the slices when necessary. Secondly, we utilize Algorithm B.1 to acquire differentially private initial values $\boldsymbol{\hat{B}}^{(0)}$, generalized eigenvalues $\{\tilde{\lambda}_l\}_{l=1}^{H}$, and apply the BIC criterion to select $\tilde{k}$. Lastly, we employ Algorithm 2 to obtain the estimator $\boldsymbol{\hat{B}}^{(T)}$. The upper bounds in Theorem 2 still hold when using $\tilde{k}$ as tuning parameters.
	
	\section{DP-SIR in High Dimensions}
	
	\subsection{Minimax Lower Bound with Privacy Constraints}
	
	In this section, we present the minimax lower bound result for the differentially private sliced inverse regression when $p$ grows exponentially with $n$, i.e., $p=o(e^n)$. We consider the following distribution space 
	\begin{align*}
		\mathcal P=&\bigg\{\{\bm x_i,Y_i\}_{i=1}^{n}\text{ are independently and identically distributed, such that:}\\
		&\bm x_i\in\mathbb{R}^p \text{ follows a sub-Gaussian distribution with covariance matrix } \boldsymbol{\Sigma}; \\
		&\lambda_x^{-1}\le\lambda_{\min}(\boldsymbol{\Sigma})\le\lambda_{\max}(\boldsymbol{\Sigma})\le \lambda_x;Y_i\mid\bm x_i\text{ follows model \eqref{eq:mim}};|\text{supp}(\boldsymbol{B})|\leq s;\\
		& \lambda\le\lambda_{k}(\boldsymbol{\Sigma}^{-1/2}\boldsymbol{M}\boldsymbol{\Sigma}^{-1/2})\leq\lambda_{1}(\boldsymbol{\Sigma}^{-1/2}\boldsymbol{M}\boldsymbol{\Sigma}^{-1/2})\leq \kappa\lambda\text{, where }\boldsymbol{M}\text{ is defined in \eqref{eq:kernel}}.\bigg\}.
	\end{align*}
	The conditions are identical to conditions for the low-dimensional case with an additional sparsity constraint on the support of $\B$, which is a common assumption for high-dimensional data analysis. Let $\boldsymbol{\hat{B}}\in\mathbb{R}^{p\times k}$ be a possible $(\varepsilon,\delta)$-differentially private estimator for $\boldsymbol{B}$. We have the following minimax lower bound result. 
	\begin{theorem}\label{thm:lower bound}
		Consider the parameter space defined above, and 
		let $\mathcal M_{\varepsilon,\delta}$ be the space of all $(\varepsilon,\delta)$-differentially private algorithms.  If $s\log p \lesssim n\varepsilon\lambda$, $0 < \varepsilon < 1$ and $\delta \lesssim n^{-(1+\gamma)}$ for some $\gamma > 0$, we have 
		\begin{equation}
			\label{eq:minimax-higher}
			\inf_{\widehat{\boldsymbol{B}}\in\mathcal M_{\varepsilon,\delta}}\sup_{P\in\mathcal P}\E L(\boldsymbol{\widehat{B}},\B)\gtrsim\frac{s\log p}{\lambda^2n}+\frac{s^2\log^2 (p) }{\lambda^2n^2\varepsilon^2}.
		\end{equation}
	\end{theorem}
	
	Analyzing DP-SIR in high dimensions is much more challenging because one needs to carefully balance the effect of noise added for privacy and the sparse signal underlying the data. The lower bound in (\ref{eq:minimax-higher}) involves the minimal signal $\lambda$ and the sparsity parameter $s$ in both terms. The first term in the lower bound corresponds to the statistical error, the same as in the existing literature \citep{tan2020sparse}. The second term corresponds to the additional differential privacy requirement. In the next section, we propose a differentially private estimator that attains the lower bound \eqref{eq:minimax-higher} within logarithm factors of $n$.
	
	\subsection{Algorithm in High Dimensions}
	
	In this section, we propose the practical algorithm that matches the optimal rate when $p$ grows exponentially with $n$. Under the sparsity assumption, the sparse SIR directions can be identified through the following optimization problem:
	\begin{equation}
		\label{eq:spare_sir_algo}
		\boldsymbol{\hat{B}}=\arg\max_{\B\in\mR^{p\times k}}\text{Tr}(\B^{\top}\widehat{\M}\B)\text{ s.t. }\B^{\top}\widehat{\boldsymbol{\Sigma}}\B=\boldsymbol{I}_k\text{ and }|\text{supp}(\B)|\leq s^{\prime},
	\end{equation}
	where $\text{supp}(\B)$ is the row support of matrix $\B$ and $s^{\prime}$ is the tuning parameter. Solving the optimization problem (\ref{eq:spare_sir_algo}) is highly challenging due to the non-convex constraints. These two constraints complicate the computation and the design of differentially private algorithms. Inspired by \cite{gao2021sparse}, we consider the following relaxation of (\ref{eq:spare_sir_algo}):
	\begin{equation}
		\label{eq:sparse_hard}
		\boldsymbol{\hat{B}}=\arg\min_{\B\in\mR^{p\times k}}-\text{Tr}(\B^{\top}\boldsymbol{\hat{M}}\B)+\lambda_{penalty}\|\B^{\top}\boldsymbol{\hat{\Sigma}}\B-\boldsymbol{I}_k\|_F^2\text{ s.t. }|\text{supp}(\B)|\leq s^{\prime}.
	\end{equation}
	Given a private initial value $\boldsymbol{\hat{B}}^{(0)}$ of equation (\ref{eq:sparse_hard}), we propose using the stochastic gradient algorithm to solve for $\boldsymbol{\hat{B}}$. The details are summarized in Algorithm \ref{alg:hd_dp_sir}. Specifically, at each step, we perform the gradient descent and the {\it peeling} algorithm to ensure that the updated parameters converge to the optimal value while satisfying the sparsity constraint. The algorithm also requires tuning parameters such as the sparsity $s^{\prime}$, the penalty $\lambda_{penalty}$, and the truncation parameter $C$. The choices and theoretical rates of these parameters will be carefully discussed in the next subsection.
	
	\begin{algorithm}
		\caption{DP-SSIR when $p=o(e^n)$}
		\begin{algorithmic}[1]
			\Require dataset $\{(\boldsymbol{x}_i,Y_i)\}_{i=1}^{n}$, step size $\eta$, initial value $\boldsymbol{\hat{B}}^{(0)}$, iterations $T$, sparsity $s^{\prime}$, penalty $\lambda_{penalty}$, noise scale $\sigma$, privacy parameter $(\varepsilon,\delta)$, truncation level $C$ and $R$.
			\State Estimate sliced intervals $\{\hat{I}_h\}_{h=1}^{H}$ based on the DP histogram \eqref{eq:dp_hist} under $(\varepsilon,0)$-differential privacy;
			\State Random split data into $T$ parts of roughly equal sizes: $[n]=S_0\cup\dots\cup S_{T-1}$;
			\For {$t=0,1,2,\dots,T-1$}
			\State Use sub-data $S_t$ to update:
			\begin{equation*}
				\begin{split}
					&\boldsymbol{\hat{B}}^{(t+0.3)}=\boldsymbol{\hat{B}}^{(t)}-2\eta\bigg[-\sum_{h=1}^{H}\mathbb{E}_n(\boldsymbol{x}_i\mid Y_i\in \hat{I}_h,i\in S_{t})\sum_{Y_i\in \hat{I}_h,i\in S_{t}}\Pi_{R}(\boldsymbol{x}_i^{\top}\boldsymbol{\hat{B}}^{(t)})/|S_t|\\
					&+\lambda_{penalty}\sum_{i\in S_{t}}\boldsymbol{x}_i\Pi_{R}(\boldsymbol{x}_i^{\top}\boldsymbol{\hat{B}}^{(t)})/|S_{t}|\big\{\sum_{i\in S_{t}}\Pi_{R}(\boldsymbol{\hat{B}}^{(t)\top}\boldsymbol{x}_i)\Pi_{R}(\boldsymbol{x}_i^{\top}\boldsymbol{\hat{B}}^{(t)})/|S_{t}|-\boldsymbol{I}_k\big\}\bigg],
				\end{split}
			\end{equation*}
			where $\Pi_{R}(\cdot)$ is the projection of each component of $k$-dimensional vector onto the ball $\mathbb{B}_{R}:=\{X\in\mR\mid |X|\leq R\}$ and $|S_t|$ is the number of elements in the set $S_t$;
			\State $\boldsymbol{\hat{B}}^{(t+0.6)}=\text{Matrix-NoisyHT}(\boldsymbol{\hat{B}}^{(t+0.3)},\sigma,s^{\prime},\varepsilon/T,\delta/T)$ 
			\State $\boldsymbol{\hat{B}}^{(t+1)}=\prod_{C}(\boldsymbol{\hat{B}}^{(t+0.6)})$, where $\prod_C(\boldsymbol{\hat{B}}^{(t+0.6)})$ is the projection of each column of $\boldsymbol{\hat{B}}^{(t+0.6)}$ onto the ball $\mathbb{B}_C^p=\{\boldsymbol{x}\in\mR^n\mid \|\boldsymbol{x}\|_2\leq C\}$.
			\EndFor
			\Ensure $\boldsymbol{\hat{B}}^{(T)}=\boldsymbol{\hat{B}}^{(T)}(\boldsymbol{\hat{B}}^{(T)\top}\boldsymbol{\hat{B}}^{(T)})^{-1/2}$
		\end{algorithmic}
		\label{alg:hd_dp_sir}
	\end{algorithm}
	
	To ensure that Algorithm \ref{alg:hd_dp_sir} is differentially private, it is crucial to ensure that the initial estimator $\boldsymbol{\hat{B}}^{(0)}$ is differentially private. However, achieving a consistent initial estimator in high-dimensional settings is significantly more challenging than in low-dimensional contexts. This difficulty arises because directly adding noise often fails to yield an accurate estimator, particularly when $p=O(n)$. Additionally, it is important to note that adding noise directly can obscure the underlying signals. To address this issue, we propose a consistent post-selection estimator following the idea in \cite{lin2018consistency}. Specifically, let the support set of the kernel matrix $\M$ be denoted by $\mathcal{T}:=\{i\in\{1,\dots,p\}\mid M_{i,i}\neq 0\}$. For $j\in\mathcal{T}$, there exists $\boldsymbol{v}\in\text{span}(\M)$ such that the $j$th component of $\boldsymbol{v}$ is nonzero. Due to the fact that $\boldsymbol{\Sigma}\cdot\text{span}(\B)=\text{span}(\M)$ by the conditions of sliced inverse regression, there exits $\boldsymbol{\beta}\in\text{span}(\B)$ such that $\boldsymbol{v}=\boldsymbol{\Sigma}\boldsymbol{\beta}$, which further implies $j\in\text{supp}(\boldsymbol{\Sigma}\boldsymbol{\beta})$. Under some sparsity conditions of $\B$ and $\boldsymbol{\Sigma}$, we have $|\mathcal{T}|\leq s\max_{1\leq i\leq p}|\text{supp}(\boldsymbol{\Sigma})_{*,i}|$. Thus, we can select the support of the diagonal of the kernel matrix and then run the low-dimensional DP-SIR algorithm. The details are summarized in the appendix's Algorithm \ref{alg:hd_ini}. Similar to the low dimensional setting, we require the initial estimator to be private and within a constant error bound of the true estimator. In the Lemma \ref{lem:hd_ini}, we prove that Algorithm \ref{alg:hd_ini} is private. In Theorem \ref{thm:hd_ini}, we prove that the output of Algorithm \ref{alg:hd_ini} is consistent, which is a stronger result than the constant error requirement in Theorem \ref{thm:dp_sparse_sir_conv}. The detailed discussions on the initial estimator can be found in the supplement.

	\subsection{Privacy Guarantee and Upper Bound}
	
	When the dimension $p$ is larger than $n^2$, directly adding noise to the matrix pair $(\boldsymbol{\hat{M}},\boldsymbol{\hat{\Sigma}})$ does not yield an accurate estimation of the eigenspace. Similar to Assumption \ref{assum:low_design} for low-dimensional DP-SIR, we assume the following regularity conditions to analyze the upper bound of Algorithm \ref{alg:hd_dp_sir}. It requires that the true parameter $\B$ has sparse rows.
	
	\begin{assumption}C
		\label{assum:high_design}
		The true parameter $\B$ satisfies $\|\B_{*,i}\|_2\leq c_0$ for some constant $c_0$ and $|\text{supp}(\B_{*,i})|\leq s$ for $i=1,\dots,k$.
	\end{assumption}
	
	\begin{lemma}[Privacy Guarantee of Algorithm \ref{alg:hd_dp_sir}]
		\label{lem:high_privacy}
		Under Assumption \ref{assum:low_design} and Assumption \ref{assum:high_design}, if the noisy scale satisfies
		$$\sigma\geq2\eta\{7c_xR+\lambda_{penalty}(2c_xR+4kc_xR^3)\}\frac{T}{n},$$
		then Algorithm \ref{alg:hd_dp_sir} is $(2\varepsilon,\delta)$-differentially private.
	\end{lemma}
	
	\begin{theorem}[Convergence of Algorithm \ref{alg:hd_dp_sir}]
		\label{thm:dp_sparse_sir_conv}
		Under Assumption \ref{assump:low_SIR}, \ref{assum:low_design} and \ref{assum:high_design}, further assume $1/(n^{2/3}\lambda^2)\leq c$, $s\log(p)/(n\lambda^2)\leq c$ and $s^2\log(p)^2\log(n)^2\log(1/\delta)/(\lambda^2n^2\varepsilon^2)<c$ for a sufficiently small constant $c>0$. We choose the penalty parameter $\lambda_{penalty}=\lambda/c_1$, the step size $\eta=c_2/\lambda$ for small positive constants $c_1,c_2$, the number of iterations $T=c_3\log(n)$, the truncation level $R=2c_4C\sqrt{\log n}$ for $C\geq c_0$ and the constant $c_4$ only depending on the distribution of $\boldsymbol{x}$.  Assume that the initial value satisfies $L(\boldsymbol{\hat{B}}^{(0)},\B)\leq c_5$, where $c_5$ is a small positive constant depending only on $c$ and $\lambda_x$. The peeling parameter satisfies $s\leq s^{\prime}\leq c^{\prime}s$ for a large constant $c^{\prime}$. Then, for any $C^{\prime}>0$, there exist a positive contact $C^{\prime\prime}>0$, such that the differentially private sliced inverse regression estimator $\boldsymbol{\hat{B}}^{(T+1)}$ in Algorithm \ref{alg:hd_dp_sir} satisfies $\text{supp}(\boldsymbol{\hat{B}}^{(T+1)})\leq s^{\prime}$ and
		\begin{equation}
			\label{eq:dpspsir_error}
			L(\boldsymbol{\hat{B}}^{(T+1)},\B)\leq C^{\prime\prime}\frac{s\log(p)\log(n)}{\lambda^2n}+C^{\prime\prime}\frac{s^2\log^2(p)\log^7(n)\log(1/\delta)}{\lambda^2n^2\varepsilon^2}
		\end{equation}
		with probability greater than $1-\exp(-C^{\prime}\log(n))$.
	\end{theorem}
	
	The error bound (\ref{eq:dpspsir_error}) consists of two components: statistical errors and differentially private errors. This error bound matches the optimal lower bound derived in Theorem \ref{thm:lower bound}. Similarly to the scenario where $p<n$, we introduce three assumptions on the parameter $\lambda$ in the theorem. The second condition on $\lambda$ is commonly used in the statistical literature and has been observed in previous studies \citep{tan2020sparse}. This condition serves as a standard requirement that ensures reliable statistical estimation. The third condition on $\lambda$ serves as an additional criterion, specifically tailored to address the impact of differentially private errors. It is introduced to guarantee estimation consistency considering privacy constraints and is relatively mild. 
	

	\section{Numerical Study}

	\subsection{Simulation}
	
	In this section, we evaluate the performance of the proposed algorithms through simulations. We consider the following models that are common to the sliced inverse regression in the literature:
	\begin{center}
		\begin{varwidth}{\textwidth}
			\begin{enumerate}
				\item[(M1):] $Y=\boldsymbol{\beta}_1^{\top}\boldsymbol{x}+e$,
				\item[(M2):] $Y=\text{exp}(\boldsymbol{\beta}_2^{\top}\boldsymbol{x})+e$,
				\item[(M3):] $Y=25(\boldsymbol{\beta}_3^{\top}\boldsymbol{x})/\{1+(\boldsymbol{\beta}^{\top}_4\boldsymbol{x}+1)^2\}+0.1e$,
				\item[(M4):] $Y=\text{sin}(\boldsymbol{\beta}_3^{\top}\boldsymbol{x})\text{exp}(\boldsymbol{\beta}^{\top}_4\boldsymbol{x}+e)$.
			\end{enumerate}
		\end{varwidth}
	\end{center}
	In the above models, the parameters are $\boldsymbol{\beta}_1=(\mu_1,\mu_2,0,\dots,0)^{\top}$, $\boldsymbol{\beta}_2=(\mu_3,\mu_4,0\dots,0)^{\top}$, $\boldsymbol{\beta}_3=(\mu_5,\mu_6,0,\dots,0)^{\top}$ and $\boldsymbol{\beta}_4=(\mu_7,\mu_8,0,\dots,0)^{\top}$. In low-dimensional settings, the coordinates of $(\mu_1, \dots, \mu_8)$ are independently sampled from a uniform distribution with support $(-10,10)$. The coordinates are independently sampled from a uniform distribution with support $(-10,-5)$ for high-dimensional settings. The covariate vectors $\boldsymbol{x}=(X_1,\dots,X_p)$ are generated from a multivariate normal distribution with means $0$, variances $0.25$ and correlations $\text{Cor}(X_i,X_j)=0.5^{|i-j|}$. Then we truncate the $\boldsymbol{x}$ entries in the interval $[-1.5,1.5]$. The error $e$ follows a normal distribution with mean $0$ and variance $1$. The number of slices is set to $H=20$ in low-dimensional settings and $H=10$ in high-dimensional settings. Slice estimation uses the DP-histogram method with a privacy parameter of $\varepsilon=0.1$. The number of bins is fixed at $m=100$ in low-dimensional settings and $m=50$ in high-dimensional settings. The estimated space dimension, $\hat{k}$, is selected by DP-BIC. The true dimension for models (M1) and (M2) is $k=1$, while for models (M3) and (M4), the true dimension is $k=2$. The DP budget $(\varepsilon,\delta)$ is set to $(1,1/n^{1.1})$. We compare the following methods in the simulation:
	
	\begin{center}
		\begin{varwidth}{\textwidth}
			\begin{enumerate}
				\item[(SIR):] Sliced Inverse Regression proposed by \cite{li1991sliced}.
				\item[(DP-Ini):] Initial estimator Algorithm B.1.
				\item[(DP-RF):] DP Rayleigh flow proposed by \cite{hu2023privacy}.
				\item[(DP-SIR):] Proposed DP sliced inverse regerssion in Algorithm \ref{alg:ld}.
				\item[(DP-SIni):] Initial estimator Algorithm B.2 for sparse SIR.
				\item[(DP-TRF):] DP truncated Rayleigh flow proposed by \cite{hu2023privacy}.
				\item[(DP-SSIR):] Proposed DP spares sliced inverse regression in Algorithm \ref{alg:hd_dp_sir}.
			\end{enumerate}
		\end{varwidth}
	\end{center}
	\vspace{0.35cm}
	
	We briefly introduce those methods. SIR is the oracle estimator without additional noise, which is statistically optimal in the non-private setting. DP-Ini is the initial estimator that introduces noise directly to the covariance matrix, resembling the method discussed by \cite{dwork2014analyze} for DP-PCA. DP-RF and DP-TRF are proposed by \cite{hu2023privacy}, and use DP-Ini as the private initial value.
	These two methods are applicable only when the target spaces are one-dimensional, and they do not attain optimal performance in terms of error rate. 
	
	To evaluate performance, we measure the loss $|\boldsymbol{P}{\B} - \boldsymbol{P}{\boldsymbol{\hat{B}}}|_F$, where $\boldsymbol{\hat{B}}$ denotes the estimator from the algorithms. We perform $1000$ replications of the simulation. Table \ref{tab:1} summarizes the results for models (M1) and (M2) with sample sizes of $n=(20000, 40000)$ and dimensions $p=(15, 30)$, as well as for models (M3) and (M4) with $n=(30000, 50000)$ and $p=(10, 15)$. For models (M3) and (M4), the DP-RF method consistently employs $k=1$, while the other methods utilize $\hat{k}$ determined by DP-BIC. As an optimal benchmark, the classical sliced inverse regression consistently outperforms all DP algorithms across all scenarios. The proposed method shows improved performance compared to the DP initial estimator, although it does not reach the efficacy of classical SIR due to additional privacy costs. Furthermore, the DP Rayleigh flow method introduced by \cite{hu2023privacy} performs even worse than the initial estimator. This discrepancy is due to the noise scale of the DP Rayleigh flow method being $\log(n)$ times greater than that of the initial estimator, significantly affecting the accuracy of the algorithm's estimations.

	\begin{table}[ht]
		\centering
		\begin{tabular}{ccccccc}
			\hline\hline
			Models &$(n,p)$&SIR&DP-Ini&DP-RF&DP-SIR&$\hat{k}$ \\
			\hline
			\multirow{4}{5em}{M1($k=1$)} 
			& (20000,15)& 0.018 & 0.237 & 0.381 & 0.222 & 1.0 \\ 
			& (20000,30)&0.026 & 0.764 & 1.017 & 0.731 & 1.0 \\ 
			& (40000,15)&0.013 & 0.123 & 0.194 & 0.115 & 1.0 \\ 
			& (40000,30)&0.018 & 0.364 & 0.562 & 0.340 & 1.0 \\ 
			\hline
			\multirow{4}{5em}{M2($k=1$)} 
			& (20000,15)&0.029 & 0.272 & 0.459 & 0.257 & 1.0 \\ 
			& (20000,30)&0.043 & 0.950 & 1.143 & 0.926 & 1.0 \\ 
			& (40000,15)&0.021 & 0.144 & 0.222 & 0.135 & 1.0 \\ 
			& (40000,30)&0.029 & 0.416 & 0.643 & 0.391 & 1.0 \\ 
			\hline
			\multirow{4}{5em}{M3($k=2$)} 
			& (30000,10)&0.200 & 0.409 & 1.005 & 0.400 & 1.8 \\ 
			& (30000,15)&0.316 & 0.813 & 1.016 & 0.800 & 2.2 \\ 
			& (50000,10)&0.195 & 0.317 & 1.002 & 0.312 & 1.8 \\ 
			& (50000,15)&0.191 & 0.473 & 1.006 & 0.463 & 1.8 \\ 
			\hline
			\multirow{4}{5em}{M4($k=2$)} 
			& (30000,10)& 0.195 & 0.340 & 1.004 & 0.333 & 1.8 \\ 
			& (30000,15)&0.202 & 0.623 & 1.015 & 0.612 & 2.1 \\ 
			& (50000,10)&0.200 & 0.276 & 1.002 & 0.271 & 1.8 \\ 
			& (50000,15)&0.194 & 0.373 & 1.006 & 0.361 & 1.8 \\ 
			\hline
		\end{tabular}
		\caption{The means of the distance based on $1000$ replications, where SIR stands for the classic sliced inverse regression, DP-Ini corresponds to the DP-initial estimator in Algorithm \ref{alg:ld_ini}, DP-RF stands for the DP-Rayleigh Flow \citep{hu2023privacy}, DP-SIR corresponds to the proposed Algorithm \ref{alg:ld}.   }
		\label{tab:1}
	\end{table}
	
	Next, we compare the performance of SIR, DP-SIni, DP-TRF, and DP-SSIR in high-dimensional settings. The peeling number $s^{\prime}$ in step 5 of Algorithm \ref{alg:hd_dp_sir} is set to $6$. The classical sliced inverse regression considers only the first $6$ features, which represent the true active set, and serves as the benchmark oracle method, referred to as Oracle-SIR in the table. DP-TRF and DP-SSIR use the estimator obtained from Algorithm \ref{alg:hd_ini} as the private initial value. As in the previous comparison, we evaluate the loss based on the projection matrix $\|\boldsymbol{P}_{\B} - \boldsymbol{P}_{\boldsymbol{\hat{B}}}\|_F$, repeating each simulation $1000$ times. Table \ref{tab:2} presents the results for $n=(1000, 2000)$ and $p=(1000, 2000)$ for models M1 and M2, as well as for $n=(2000, 4000)$ and $p=(2000, 4000)$ for models M3 and M4. We observe that DP-SSIR performs well across all settings. While it does not match Oracle-SIR due to challenges in estimating the support and the additional privacy costs, it consistently outperforms the truncated DP-TRF method proposed by \cite{hu2023privacy}. Furthermore, similar to the low-dimensional setting, our initial estimator also demonstrates superior performance compared to the method introduced by \cite{hu2023privacy}.
	
	\begin{table}[ht]
		\centering
		\begin{tabular}{ccccccc}
			\hline\hline
			Models &$(n,p)$&Oracle-SIR&DP-SIni&DP-TRF&DP-SSIR &$\hat{k}$\\
			\hline
			\multirow{4}{5em}{M1($k=1$)} 
			& (1000,1000)&0.037 & 0.455 & 0.522 & 0.385 & 1.0 \\ 
			& (1000,2000)&0.037 & 0.471 & 0.537 & 0.405 & 1.0 \\ 
			& (2000,1000)&0.025 & 0.218 & 0.246 & 0.173 & 1.0 \\ 
			& (2000,2000)&0.025 & 0.218 & 0.239 & 0.175 & 1.0 \\ 
			\hline
			\multirow{4}{5em}{M2($k=1$)} 
			& (2000,2000)&0.082 & 0.627 & 0.709 & 0.522 & 1.0\\
			& (2000,4000)&0.082 & 0.754 & 0.817 & 0.681 & 1.0\\ 
			& (4000,2000)&0.059 & 0.258 & 0.307 & 0.188 & 1.0 \\ 
			& (4000,4000)&0.059 & 0.257 & 0.319 & 0.185 & 1.0 \\ 
			\hline
			\multirow{4}{5em}{M3($k=2$)} 
			& (2000,2000)&0.260 & 0.747 & 1.237 & 0.624 & 1.9 \\ 
			& (2000,4000)&0.261 & 0.738 & 1.245 & 0.640 & 1.9 \\  
			& (4000,2000)&0.205 & 0.584 & 1.210 & 0.465 & 1.8 \\ 
			& (4000,4000)&0.205 & 0.556 & 1.213 & 0.435 & 1.8 \\ 
			\hline
			\multirow{4}{5em}{M4($k=2$)} 
			& (2000,2000)&0.443 & 0.929 & 1.129 & 0.612 & 1.9\\ 
			& (2000,4000)&0.434 & 0.935 & 1.128 & 0.673 & 1.9 \\ 
			& (4000,2000)&0.357 & 0.846 & 1.062 & 0.516 & 1.9 \\ 
			& (4000,4000)&0.355 & 0.833 & 1.071 & 0.524 & 1.9\\ 
			\hline\hline
		\end{tabular}
		\caption{The means of the distance based on $1000$ replications, where oracle SIR stands for the classic sliced inverse regression with true support, DP-SIni corresponds to the initial estimator in Algorithm \ref{alg:hd_ini}, DP-TRF stands for the truncated DP-Rayleigh Flow \citep{hu2023privacy}, DP-SSIR corresponds to the proposed Algorithm \ref{alg:hd_dp_sir}.}
		\label{tab:2}
	\end{table}
	
	\subsection{Real Data Application: Supermarket Dataset}
	
	We apply the proposed methods on a supermarket dataset, with the goal of identifying a concise group of products that substantially impact daily customer visits. The dataset consists of $n = 464$ entries from a supermarket \citep{liu2022model}. The response variable $Y$ is the number of customers per day. The covariates $\boldsymbol{x}$ are the sales volumes of $p = 6398$ products on the same day. Due to privacy concerns, specific product names have been removed. All variables have been standardized to have zero mean and unit variance. 
	
\begin{figure}[ht]
		\centering
		\includegraphics[width=0.5\textwidth]{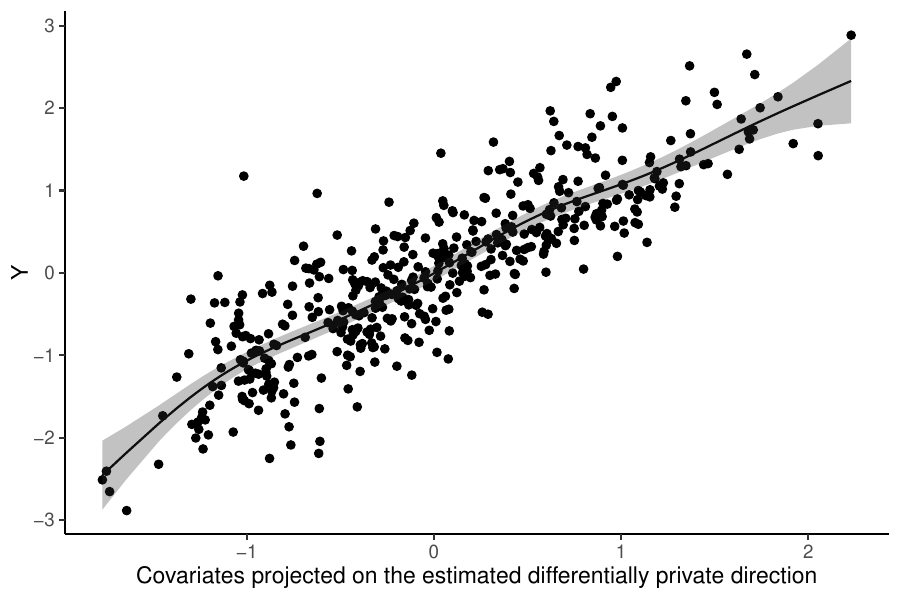}
		\caption{The scatter plot between the response $Y$ and $\hat{X}$ which is the projection of $\boldsymbol{x}$. The black solid curve is the fitted spline regression curve, and the gray shaded areas are corresponding confidence regions.}
		\label{fig:2}
	\end{figure}

	We implemented our proposed Algorithm \ref{alg:hd_dp_sir} with the private initial value obtained from Algorithm \ref{alg:hd_ini}. The number of slices was set to $H=7$ and the privacy parameter was set to $(\varepsilon,\delta)=(2,n^{-1.1})$. The peeling size was set to $s^{\prime}=7$. The number of directions, $\hat{k}$, was selected by DP-BIC and set to $1$. To evaluate our proposed method, we first projected the covariates $\boldsymbol{x}$ onto the estimated direction, denoted as $\hat{x}$. We then used the R package \textit{mgcv} with default settings to fit a spline regression of $Y$ based on $\hat{x}$. The results are presented in Figure \ref{fig:2}. We observed that the relationship between $Y$ and $\hat{x}$ was very close to linear, indicating that our proposed method effectively identified the relevant direction.
	
\begin{figure}[ht]
		\centering
		\includegraphics[width=0.7\linewidth]{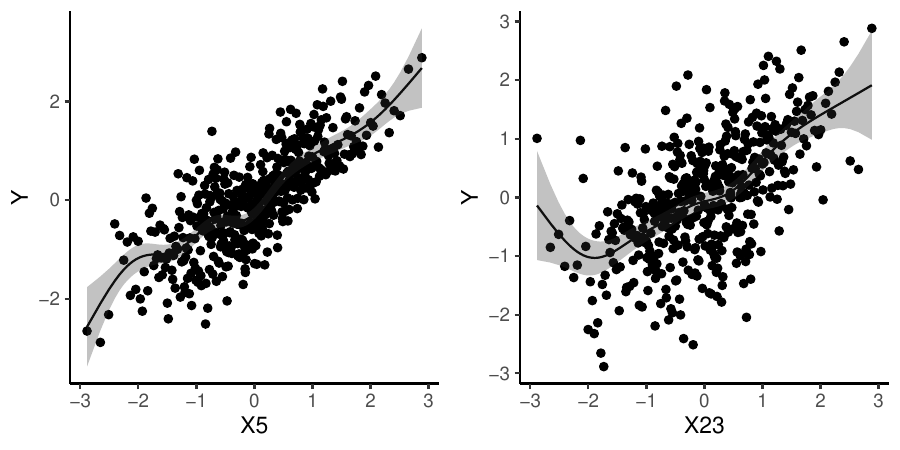}
		\caption{The scatter plots between the response $Y$ and $X_5$ and $X_{23}$. The black solid curves are the fitted spline regression curves for three variables, respectively, and the gray shaded areas are corresponding confidence regions.}
		\label{fig:3}
	\end{figure}

	This supermarket dataset was also analyzed by \cite{liu2022model} and \cite{chen2018error} for feature screening. Our proposed DP-SSIR method selected $X_3$, $X_5$, $X_{10}$, $X_{11}$, $X_{23}$, $X_{39}$ and $X_{139}$. All variables, except $X_5$ and $X_{23}$, are also selected by \cite{liu2022model} using PC-Knockoff. The scatter plots between the response $Y$ and covariates $X_5$, $X_{23}$ are presented in Figure \ref{fig:3}. The plots show that the variables are indeed related to the response $Y$. 
	
	In the end, we compare our proposed DP-SSIR with other methods. We project covariates $\boldsymbol{x}$ on the estimated directions by Lasso-SIR, DP-SIni, DP-TRF, and DP-SSIR, respectively. We then use the R package \textit{ mgcv} with default settings to fit the spline regression of $Y$ given the projected covariates and report the adjusted $R^2$ in Table \ref{tab:3}. The proposed DP-SSIR method has the largest adjusted $R^2$ among all private methods. It has a smaller $R^2$ compared to the non-private Lasso-SIR method.
	
	\begin{table}[ht]
		\centering
		\begin{tabular}{c|c|c|c|c}
			\hline
			Method  & Lasso-SIR&DP-SIni&DP-TRF&DP-SSIR \\\hline
			Adjusted $R^2$  &0.868&0.715&0.71&0.747 \\
			\hline
		\end{tabular}
		\caption{Adjusted $R^2$ for spline fittings using the projection directions selected by Lasso-SIR, DP-SIni, DP-TRF, and DP-SSIR in the supermarket dataset.}
		\label{tab:3}
	\end{table}
	
	\section{Conclusion and Discussion}
	
	In this paper, we proposed the DP sliced inverse regression and comprehensively analyzed both the lower and upper bounds in various dimensional settings. Our theoretical analysis highlights the trade-off between statistical accuracy and privacy constraints when privacy protection is a concern. Furthermore, extensive numerical experiments demonstrate that the proposed algorithms perform well in practice, with only minimal loss in accuracy when $(\varepsilon,\delta)$ -DP is satisfied.
	
	There are several promising avenues for future research. Although we focused on sliced inverse regression as a dimension reduction technique, there are other methods worth exploring, such as the Sliced Average Variance Estimate (SAVE), the Minimum Average Variance Estimate (MAVE), Contour Regression, and Directional Regression, among others \citep{li2018sufficient}. Investigating the privacy costs associated with these alternative dimension reduction tools would be of interest. While the inverse-type dimension reduction tools are intrinsically related to the generalized eigen-decomposition, the forward-type dimension reduction methods (such as MAVE)  are more closely related to general nonparametric regression methods. It would also be useful to investigate the privacy preserving methodologies in other high dimensional statistical learning problems, such as causal discovery \citep{cai2022causal} and independence testing \citep{cai2022distribution, cai2024asymptotic}. 
	Further research in this field will contribute to a deeper understanding of the privacy trade-offs inherent in statistical methods and facilitate the development of more comprehensive and effective privacy-preserving techniques.

\appendix
\renewcommand\thetheorem{\Alph{section}.\arabic{theorem}}
\renewcommand\thealgorithm{\Alph{section}.\arabic{algorithm}}
\numberwithin{figure}{section}

\renewcommand{\thesection}{\Alph{section}}

\newpage

\section{Extension to DP Sparse PCA}\label{sec: dp_pca}

As one of the most popular statistical analysis tools, \textit{Principal Component Analysis} (PCA) can be understood as an unsupervised dimension reduction method. The goal of PCA is to retain as much variance as possible in the transformed variables. While classical textbooks focus on the case when the data's original dimension $p$ is smaller than the sample size $n$, recent literature has witnessed growing studies on the sparse PCA when $p>n$, see for example, \cite{zou2006sparse,ma2013sparse, qiu2023gradient}. It is easy to observe that PCA can be formulated as a special case of sliced inverse regression. Precisely, at the population level, the first $k$ principle components or projection directions can be estimated by
\begin{equation}
	\label{eq:pca}
	\B:=\argmax_{\B\in\mR^{p\times k}}\text{Tr}(\B^{\top}\boldsymbol{\Sigma}\B)\quad\text{such that}\quad\B^{\top}\B=\boldsymbol{I}_k.
\end{equation}
The readers may observe similarities between (\ref{eq:pca}) and \eqref{eq:tr_generalized_eigenvector}.

Creating DP algorithms to identify significant loadings has emerged as a important problem, as highlighted in the work of \cite{chaudhuri2012near}. For the low-dimension settings where $p<n$, \cite{liu2022dp} developed optimal \textit{DP Principal Component Analysis} (DP-PCA) under $(\varepsilon,\delta)$-DP constraint. However, the development of the optimal \textit{DP Sparse Principal Component Analysis} (DP-SPCA) in cases where $p=o\{\exp(n)\}$ has not appeared in the literature before. We can extend our proposed algorithm and develop a similar rate based on the DP sparse SIR results. Specifically, the DP-SPCA directions can be identified through the following optimization problem:
\begin{equation*}
	\boldsymbol{\hat{B}}=\argmax_{\B\in\mR^{p\times k}}\text{Tr}(\B^{\top}\boldsymbol{\Sigma}\B)\text{ s.t. }\B^{\top}\B=\boldsymbol{I}_k\text{ and }|\text{supp}(\B)|\leq s,
\end{equation*}
where $\boldsymbol{\Sigma}=\text{Cov}(\boldsymbol{X})$. The above formulation is a special case of \eqref{eq:spare_sir_algo} and has the following relaxation
\begin{equation}
	\label{eq:sparse_pca}
	\boldsymbol{\hat{B}}=\argmin_{\B\in\mR^{p\times k}}-\text{Tr}(\B^{\top}\boldsymbol{\Sigma}\B)+\lambda_{penalty}\|\B^{\top}\B-\boldsymbol{I}_k\|_F^2\text{ s.t. }|\text{supp}(\B)|\leq s.
\end{equation}
The proposed Algorithm \ref{alg:hd_dp_sir} can be applied to estimate \eqref{eq:sparse_pca}. Using similar techniques developed in this paper, one can obtain a similar lower bound result for DP-SPCA as in Theorem \ref{thm:lower bound}, by substituting the corresponding quantities analogously. We can also prove that solving \eqref{eq:sparse_pca} can achieve the optimal upper bound up to logarithm factors, which are similar to the upper bound in Theorem \ref{thm:dp_sparse_sir_conv}.

\section{Algorithm Initialization}\label{sec: ini_alg}

\subsection{Initialization for Algorithm \ref{alg:ld}}
In this section, we present the initialization algorithms for the differentially private low-dimensional SIR. We propose Algorithm \ref{alg:ld_ini} to estimate the initial value for Algorithm \ref{alg:ld}.

\begin{algorithm}
	\caption{Initial Estimator for Algorithm \ref{alg:ld}}
	\begin{algorithmic}[1]
		\Require $\{\boldsymbol{x}_{i},Y_{i}\}_{i=1,\dots,n}$, privacy parameter $(\varepsilon,\delta)$, number of slices $H$, dimension $k$, noise scale $\sigma^2_1$ and $\sigma^2_2$.
		\State Estimate sliced intervals $\{\hat{I}_h\}_{h=1}^{H}$ based on the DP histogram under $(\varepsilon, 0)$-DP;
		\State Estimate $\boldsymbol{\hat{\Sigma}}$ and $\boldsymbol{\hat{M}}$ using data $\{\boldsymbol{x}_{i},Y_{i}\}_{i=1,\dots,n}$;
		\State Let $\boldsymbol{E}_1\in\mR^{p\times p}$ and $\boldsymbol{E}_2\in\mR^{p\times p}$ be two symmetric matrices where the upper triangle (including the diagonal) elements are i.i.d. samples generated from $N(0,\sigma^2_1\cdot8\log(2.5/\delta)/\varepsilon^2)$ and $N(0,\sigma^2_2\cdot8\log(2.5/\delta)/\varepsilon^2)$, respectively;
		\State Let $\boldsymbol{\tilde{\Sigma}}:=\boldsymbol{\hat{\Sigma}}+\boldsymbol{E}_1$ and $\boldsymbol{\tilde{M}}:=\boldsymbol{\hat{M}}+\boldsymbol{E}_2$ be private estimates of $\boldsymbol{\Sigma}$ and $\boldsymbol{M}$;
		\State Calculate:
		\begin{equation*}
			\boldsymbol{\tilde{B}}=\argmax_{\B\in\mR^{p\times k}}\text{Tr}(\B^{\top}\boldsymbol{\tilde{M}}\B)\quad\text{such that}\quad\B^{\top}\boldsymbol{\tilde{\Sigma}}\B=\boldsymbol{I}_k;
		\end{equation*}
		\Ensure $\boldsymbol{\hat{B}}^{(0)}:=\boldsymbol{\tilde{B}}$ and ordered estimated generalized eigenvalues $\{\tilde{\lambda}_1,\dots,\tilde{\lambda}_H\}$.
	\end{algorithmic}
	\label{alg:ld_ini}
\end{algorithm}

\begin{lemm}[Privacy Guarantee of Algorithm \ref{alg:ld_ini}]
	\label{lem:low_privacy_ini}
	Under Assumption \ref{assum:low_design} in the main document, if 
	$$\sigma_1\geq\frac{2pc_x^2}{n}\text{ and }\sigma_2\geq\frac{7pc_x^2}{n},$$
	then Algorithm \ref{alg:ld_ini} is $(2\varepsilon,\delta)$-DP.
\end{lemm}

\begin{theorem}[Convergence of Algorithm B.1]
	\label{thm:dpsir_errorbound_ini}
	Suppose Assumption \ref{assump:low_SIR} and \ref{assum:low_design} in the main document, and conditions in Lemma \ref{lem:est_slices} hold. Assume that parameters satisfy $1/(n^{2/3}\lambda^2)\leq c$, $p/(n\lambda^2)\leq c$, and $p^{3}\log(1/\delta)/(n^2\lambda^2\varepsilon^2)\leq c$ for a small positive constant $c$. Then, for any $C^{\prime}>0$, there exist a constant $C>0$, such that the DP initial estimator $\boldsymbol{\hat{B}}^{(0)}$ in Algorithm \ref{alg:ld_ini} satisfies
	\begin{equation}
		\label{eq:dpsir_ini_error}
		L(\boldsymbol{\hat{B}}^{(0)},\B)\leq C\frac{p}{\lambda^2n}+C\frac{p^{3}\log(1/\delta)}{\lambda^2n^2\varepsilon^2},
	\end{equation}
	with probability greater than $1-\exp(-C^{\prime}\log(p))$.
\end{theorem}

Note that Theorem \ref{thm:dpsir_errorbound} requires that the initial value $\boldsymbol{\hat{B}}^{(0)}$ be within a constant error ball of the true parameter. Based on Theorem \ref{thm:dpsir_errorbound_ini}, our proposed initial value clearly satisfies the requirement. In practice, we can either use $\boldsymbol{\hat{B}}^{(0)}$ or $\boldsymbol{\hat{B}}^{(0)}\{\text{diag}(\tilde{\lambda}_1,\dots,\tilde{\lambda}_k)/\lambda_{penalty}+\boldsymbol{I}_k\}^{1/2}$ as the initial estimator, where the latter is the solution of the penalized loss function \eqref{eq:gev} as shown in Lemma \ref{lem:grad_desc}.

\subsection{Initialization for Algorithm \ref{alg:hd_dp_sir}}

Next, we propose the initial estimator for the differentially private high-dimensional sparse SIR. The proposed Algorithm \ref{alg:hd_ini} shares a similar idea with the Diagonal Thresholding SIR (DT-SIR) proposed in \cite{lin2018consistency}. Here we use the peeling algorithm to select the large diagonal elements of the kernel matrix, then perform the low-dimensional Algorithm \ref{alg:ld_ini} to the selected subset. We first introduce Assumption \ref{assum:sparse_cov}, which was also assumed by \cite{lin2018consistency} for non-private SIR in the ultra-high dimension case.

\begin{algorithm}
	\caption{Initial Estimator for Algorithm 3}
	\begin{algorithmic}[1]
		\Require  $\{\boldsymbol{x}_{i},Y_{i}\}_{i=1,\dots,n}$, number of slices $H$,  dimension $k$, peeling size $s^{\prime}$, privacy parameter $(\varepsilon,\delta)$, and noisy level $\sigma,\sigma_1,\sigma_2$.
		\State Estimate sliced intervals $\{\hat{I}_h\}_{h=1}^{H}$ based on the DP histogram under $(\varepsilon, 0)$-DP;
		\State Initialization $\mathcal{\hat{P}}=\emptyset$.
		\For {$i=1,2,\dots,s^{\prime}$}
		\State Generate $\boldsymbol{w}\in\mR^{p}$ with $w_{i,1},\dots,w_{i,p}\stackrel{i.i.d.}{\sim}$Laplace$(\sigma\sqrt{3s^{\prime}\log(2/\delta)}/(\varepsilon/2))$;
		\State Select $j=\arg\max_{j\in[p]\backslash\mathcal{\hat{P}}}|\hat{M}_{j,j}|$+$w_{i,j}$ and add $j$ to $\mathcal{\hat{P}}$;
		\EndFor
		\State Denote the selected subset by $\widehat{\mathcal{P}}$;
		\State Apply Algorithm \ref{alg:ld_ini} without Line 1 (private slice estimation) to $\boldsymbol{\hat{M}}_{\mathcal{\hat{P}}\mathcal{\hat{P}}},\boldsymbol{\hat{\Sigma}}_{\mathcal{\hat{P}}\mathcal{\hat{P}}}$ and noise scale $\sigma_1,\sigma_2$ and privacy parameter $(\varepsilon/2,\delta/2)$.
		\Ensure $\boldsymbol{\hat{B}}^{(0)}$ and $\{\tilde{\lambda}_1,\dots,\tilde{\lambda}_H\}$.
	\end{algorithmic}
	\label{alg:hd_ini}
\end{algorithm}

\begin{assumption}D
	\label{assum:sparse_cov}
	Let $r_i$ be the number of nonzero elements in the $i$th row of the covariance matrix $\boldsymbol{\Sigma}$. The row sparsity of $\boldsymbol{\Sigma}$, $\max_{1\leq i\leq p}r_i\leq L$, is bounded by a constant $L$. Let the support set of the diagonal elements of $\M$ be denoted by $\mathcal{P}:=\{i\in\{1,\dots,p\}\mid M_{i,i}\neq 0\}$. The nonzero diagonal elements of $\M$ are uniformly bounded away from zero: $M_{i,i}\geq c$ for a positive constant $c$ and all $i\in\mathcal{P}$.
\end{assumption}

\begin{lemm}[Privacy Guarantee of Algorithm \ref{alg:hd_ini}]
	\label{lem:hd_ini}
	Under Assumption A, B, C, for
	$$\sigma\geq 7\frac{c_x^2}{n},\sigma_1\geq 2\frac{s^{\prime}c_x^2}{n}\text{ and }\sigma_2\geq 7\frac{s^{\prime}c_x^2}{n},$$
	the outpout of Algorithm \ref{alg:hd_ini} is $(2\varepsilon,\delta)$-DP
\end{lemm}

\begin{theorem}[Convergence of Algorithm \ref{alg:hd_ini}]
	\label{thm:hd_ini}
	Under Assumption A, B, C and \ref{assum:sparse_cov}, assume that $\sqrt{s\log(1/\delta)}\log(p)/(n\varepsilon)\to 0$, $s/(n\lambda^2)\to 0$, $s^{3}\log(1/\delta)/(n^2\lambda^2\varepsilon^2)\to 0$ and $Ls<s^{\prime}<Cs$ for a constant $C$, then the selected set satisfies $\mathcal{P}\subset\mathcal{\hat{P}}$ and the estimator $\boldsymbol{\hat{B}}^{(0)}$, $\{\tilde{\lambda}_1,\dots,\tilde{\lambda}_H\}$ satisfies
	\begin{equation*}
		L(\boldsymbol{\hat{B}}^{(0)},\B)\to 0\text{ and }|\tilde{\lambda}_j-\lambda_j|\to 0\text{ for }j=1,\dots,H,
	\end{equation*}
	with probability converging to one as $n,p\to\infty$ .
\end{theorem}

\section{Additional Numerical Studies}\label{sec: add_real_data}

\subsection{Additional Real Data Application: Cancer Classification}

We utilized our proposed methods on a real dataset to distinguish between patients with cancer and those without cancer, using \textit{mass spectrometry data}. This data was produced by the National Cancer Institute (NCI) and Eastern Virginia Medical School (EVMS), where the mass spectrometry information was gathered using the SELDI technique. The training dataset contains $44$ cancer patients (suffering from either ovarian or prostate cancer) and $56$ healthy controls. Each individual in this dataset is associated with $10,000$ features, of which $7,000$ are real features, and the remaining $3,000$ are regarded as \textit{random} features. The test dataset contains $800$ patients without labels. A synthetic dataset is generated by applying Lasso-logistic regression to predict the labels in the test dataset. The total number of observations is $n=800$. The dataset has been previously analyzed for the NeurIPS 2003 feature selection challenge by \cite{guyon2003design} and for evaluating the Lasso-SIR model \citep{lin2019sparse}, which is closely related to our study. The data is publicly available from the UCI machine learning repository \citep{Dua:2019}.

\begin{figure}[ht]
	\centering
	\includegraphics{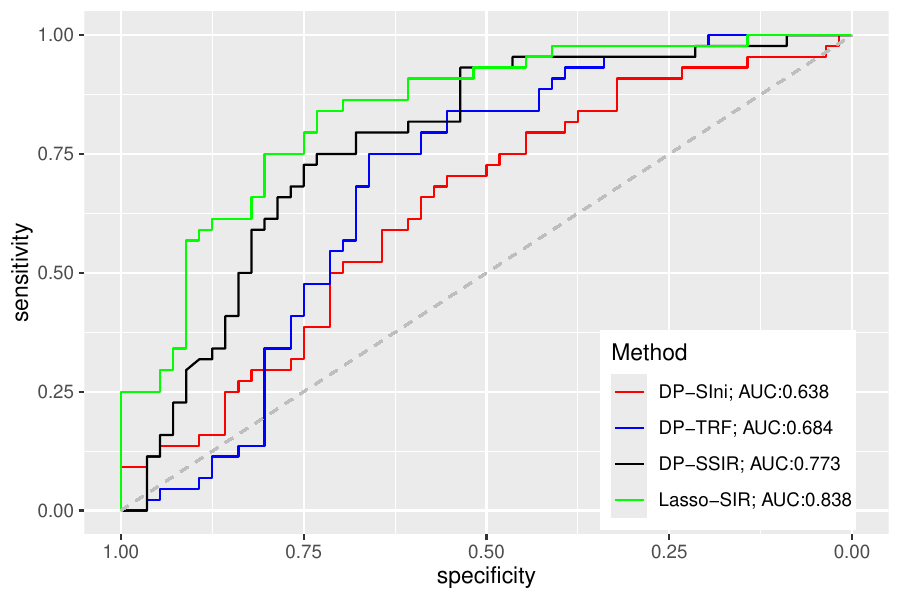}
	\caption{The ROC curves of the DP-SIni, DP-TRF, DP-SSIR, and Lasso-SIR.}
	\label{fig:1}
\end{figure}

We first normalized the covariates $\boldsymbol{x}$, ensuring that each feature has mean zero and variance one. As $Y$ is a binary random variable, we set the number of slices to $H=2$ and the number of directions to $k=1$. We implemented non-private optimal Lasso-SIR \citep{lin2019sparse} to estimate the dimension reduction direction, followed by the logistic regression. The parameters for differential privacy, $(\varepsilon,\delta)$, were set to $(2,1/n^{1.1})$. We first applied Algorithm \ref{alg:hd_ini} with the peeling size $s=10$ for the initial value for DP estimators. We then applied the DP truncate Rayleigh flow method and our proposed Algorithm \ref{alg:hd_dp_sir} to estimate the dimension reduction direction and the corresponding components. The number of iterations in the gradient step is set to $T=\floor{\log(n)}$, where $\floor{x}$ denotes the biggest integer less than or equal to $x$. Subsequently, we employed the logistic regression given the reduced direction. The fitted models were applied to the validation set, and the receiver operating characteristic (ROC) curves were computed. The results are depicted in Figure \ref{fig:1}. Our proposed method performed slightly worse than non-private Lasso-SIR and significantly outperformed the DP truncate Rayleigh flow method \citep{hu2023privacy}.

\subsection{Finite-Sample Sensitivity Analysis of DP Histogram}

In this section, we examine the impact of tuning parameters on the privately estimated slices introduced in Section \ref{sec:dp-hist}. As demonstrated in the proofs of Theorems \ref{thm:dpsir_errorbound} and \ref{thm:dp_sparse_sir_conv}, the influence of DP histogram-based slices on SIR arises primarily through the refined coverage condition (Assumption \ref{assump:low_SIR}). We demonstrated in the proof of Lemma \ref{lem:est_slices} that when the sample size $n$ becomes large, and the privacy parameter $\varepsilon$ and the eigen-gap $\lambda$ remain sufficiently bounded, the kernel matrix constructed with estimated slices satisfies the refined coverage condition with eigengap close to $\lambda$. We will illustrate the finite-sample performance of the privately estimated slices through numerical studies, examining various choices of tuning parameters, including sample size $n$, the privacy parameter $\varepsilon$, the number of bins $m$, and the number of slices $H$. We use the following settings: the sample size is $n=1000$, and the dimension is $p=10$ for covariate $\boldsymbol{x}$. For illustration, we consider two-generation models, (M1) and (M3), as defined in Section 5.1 of the main document. The results are presented as averages over $5000$ replications.

\begin{figure}[ht]
	\centering
	\includegraphics[width=0.8\linewidth]{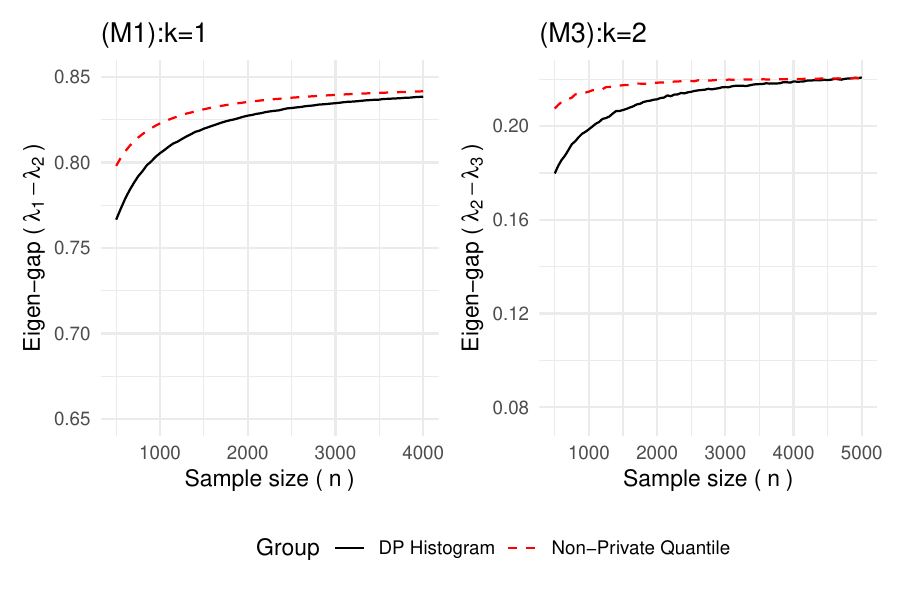}
	\caption{Comparison of the eigen-gaps between the kernel matrix based on sample quantiles and that based on DP estimated slices for privacy parameter $\varepsilon=0.1$ under two generation models and varying sample sizes. The plot uses $m=\lceil 8n^{1/3}\rceil$ bins and $H=10$ slices.}
	\label{fig:n.eg}
\end{figure}

\begin{figure}[ht]
	\centering
	\includegraphics[width=0.8\linewidth]{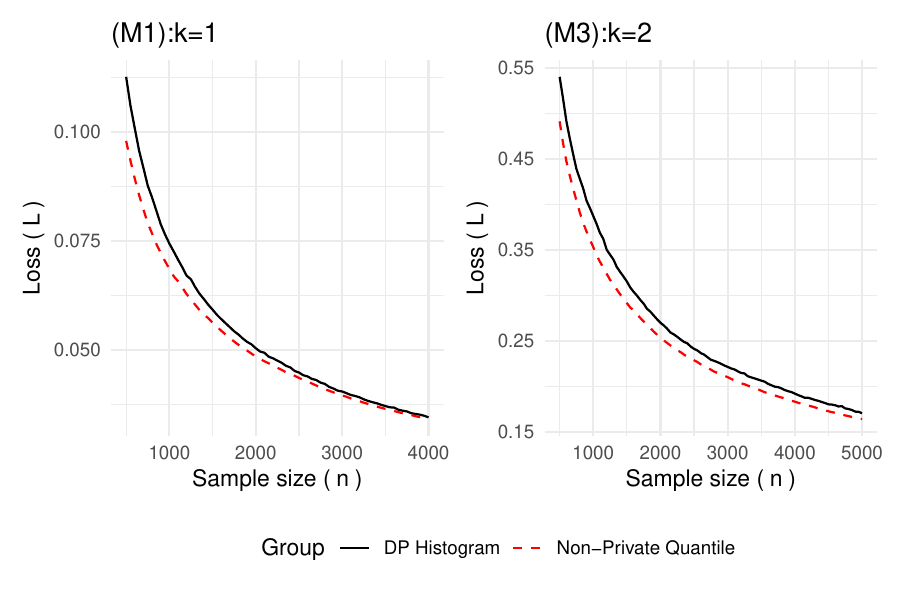}
	\caption{Comparison of the loss in the estimated matrix $\boldsymbol{B}$ derived from the kernel matrix based on sample quantiles versus that based on DP estimated slices for privacy parameter $\varepsilon=0.1$ under two generation models and varying sample size. The plot uses $m=\lceil 8n^{1/3}\rceil$ bins and $H=10$ slices.}
	\label{fig:n.loss}
\end{figure}

We examine the theoretical results in Lemma \ref{lem:est_slices}. Specifically, we examine the eigen-gap of the kernel matrix constructed using DP estimated slices and compare it with the eigen-gap of the kernel matrix based on sample quantiles for various sample sizes $n$. Additionally, we assess the loss $L$ in the estimated matrix $\boldsymbol{B}$ when using the kernel matrix derived from DP estimated slices versus the one derived from sample quantiles. The results are shown in Figure \ref{fig:n.eg} and \ref{fig:n.loss}. As the number of bins $m=\lceil 8n^{1/3}\rceil$ increases with the sample size $n$, Figure \ref{fig:n.eg} demonstrates that the difference between the eigen-gaps of the kernel matrices based on DP estimated slices and sample quantiles converges to zero. Similarly, Figure \ref{fig:n.loss} shows that the difference in loss between the two methods also converges to zero. The numerical results are consistent with the findings in Lemma \ref{lem:est_slices}. With a sample size of $n=2000$, the relative difference remains approximately $4\%$ for model (M1) and $6\%$ for model (M3), respectively.

\begin{figure}[ht]
	\centering
	\includegraphics[width=0.8\linewidth]{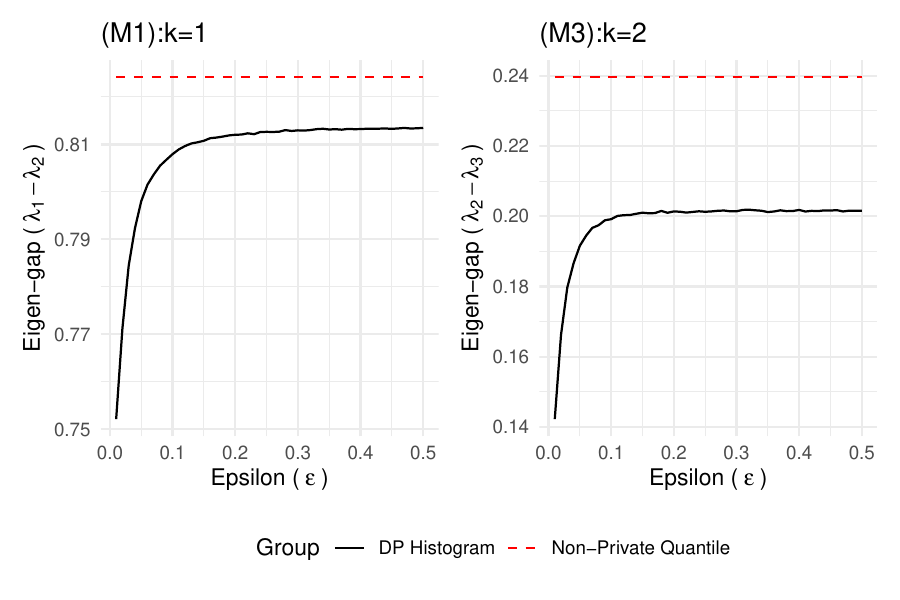}
	\caption{Comparison of the eigen-gaps between the kernel matrix based on DP estimated slices and that based on sample quantiles for two generation mechanisms and varying privacy parameter $\varepsilon$. The plot uses $m=20$ bins and $H=10$ slices.}
	\label{fig:ep.eg}
\end{figure}

\begin{figure}[ht]
	\centering
	\includegraphics[width=0.8\linewidth]{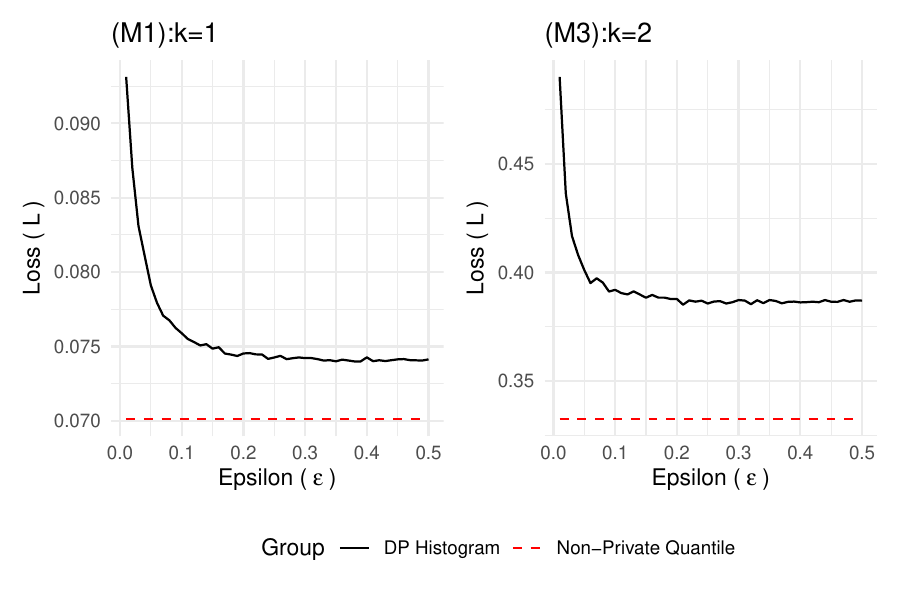}
	\caption{Comparison of the loss in the estimated matrix $\boldsymbol{B}$ derived from the kernel matrix based on DP estimated slices versus that based on sample quantiles for two generation mechanisms and varying privacy parameter $\varepsilon$. The plot uses $m=20$ bins and $H=10$ slices.}
	\label{fig:ep.loss}
\end{figure}

Next, we analyze the impact of various parameters, starting with examining the effect of the privacy parameter $\varepsilon$. Specifically, we compare the eigen-gap and the loss in the estimated matrix $\boldsymbol{B}$ between the kernel matrix based on DP estimated slices and that based on sample quantiles across two generation mechanisms and varying values of $\varepsilon$. The results, presented in Figure \ref{fig:ep.eg} and Figure \ref{fig:ep.loss}, demonstrate that, as the privacy budget $\varepsilon$ increases, the eigen-gap becomes larger, and the loss in the estimated matrix correspondingly decreases. The results also match our theoretical results (Theorem \ref{thm:dpsir_errorbound} and Theorem \ref{thm:dp_sparse_sir_conv}) that the loss in the final estimator is closely related to the eigen-gap of the kernel matrix. When the eigen-gap remains relatively stable, the loss in the estimated matrix similarly remains consistent. For privacy parameters, $\varepsilon\geq 0.1$, the eigen-gap remains nearly constant but is smaller than results based on non-private quantiles. This difference can be attributed to the approximation error introduced by using a finite number of bins, as shown later.

\begin{figure}[ht]
	\centering
	\includegraphics[width=0.8\linewidth]{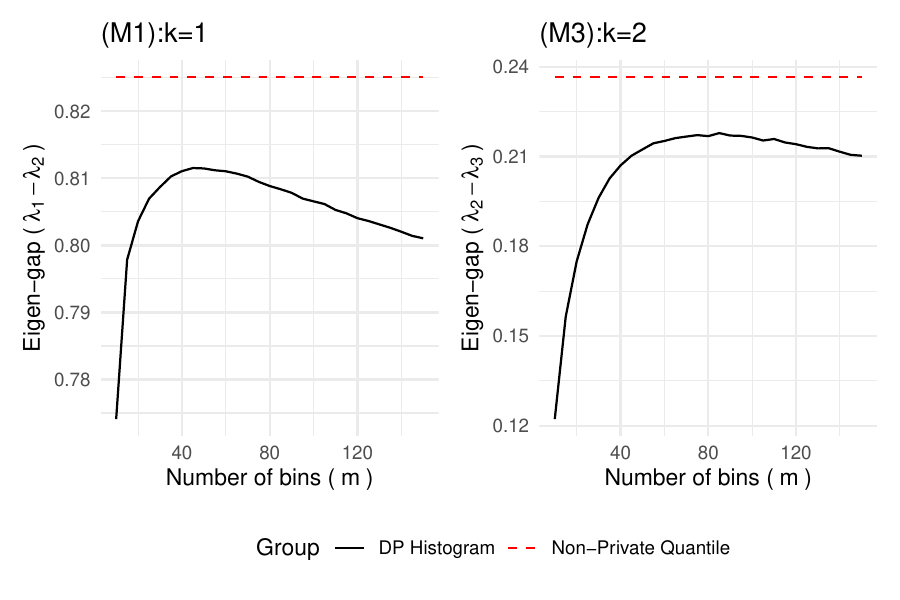}
	\caption{Comparison of the eigen-gaps between the kernel matrix based on DP estimated slices and that based on sample quantiles for two generation mechanisms and varying number of bins $m$. The plot uses privacy parameter $\varepsilon=0.1$ bins and $H=10$ slices.}
	\label{fig:m.eg.ep0.1}
\end{figure}

\begin{figure}[ht]
	\centering
	\includegraphics[width=0.8\linewidth]{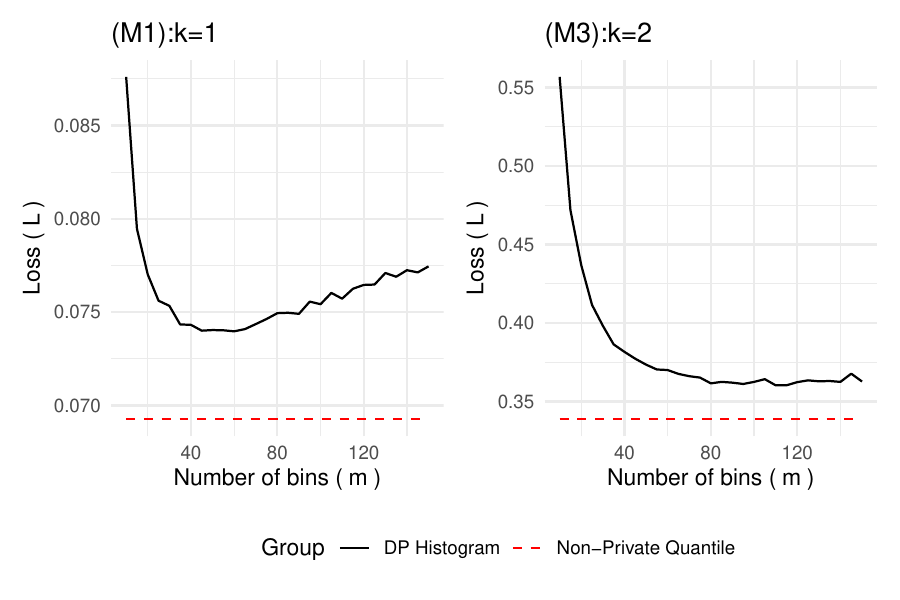}
	\caption{Comparison of the loss in the estimated matrix $\boldsymbol{B}$ derived from the kernel matrix based on DP estimated slices versus that based on sample quantiles for two generation mechanisms and varying number of bins $m$. The plot uses privacy parameter $\varepsilon=0.1$ and $H=10$ slices.}
	\label{fig:m.loss.ep0.1}
\end{figure}

\begin{figure}[ht]
	\centering
	\includegraphics[width=0.8\linewidth]{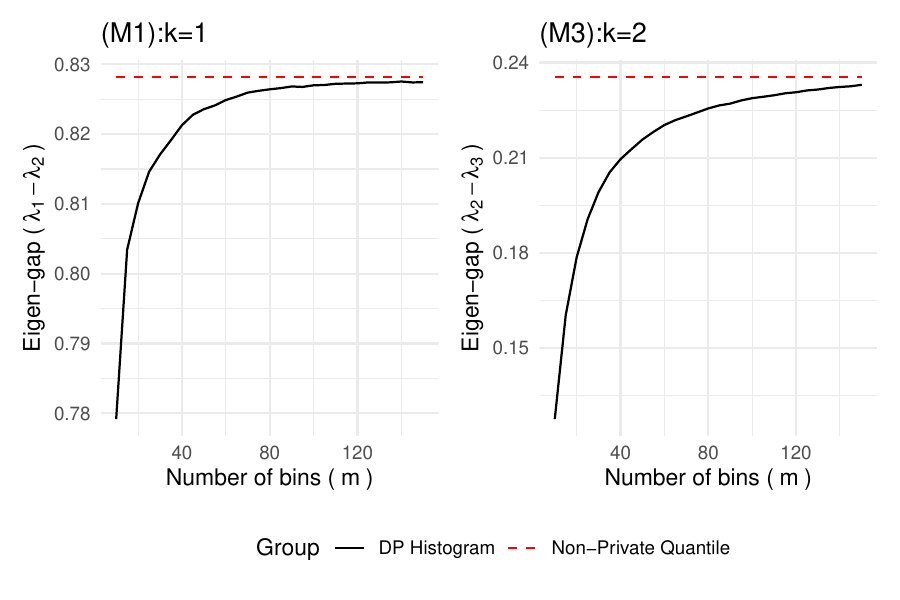}
	\caption{Comparison of the eigen-gaps between the kernel matrix based on DP estimated slices and that based on sample quantiles for two generation mechanisms and varying number of bins $m$. The plot uses privacy parameter $\varepsilon=1$ and $H=10$ slices.}
	\label{fig:m.eg.ep1}
\end{figure}

\begin{figure}[ht]
	\centering
	\includegraphics[width=0.8\linewidth]{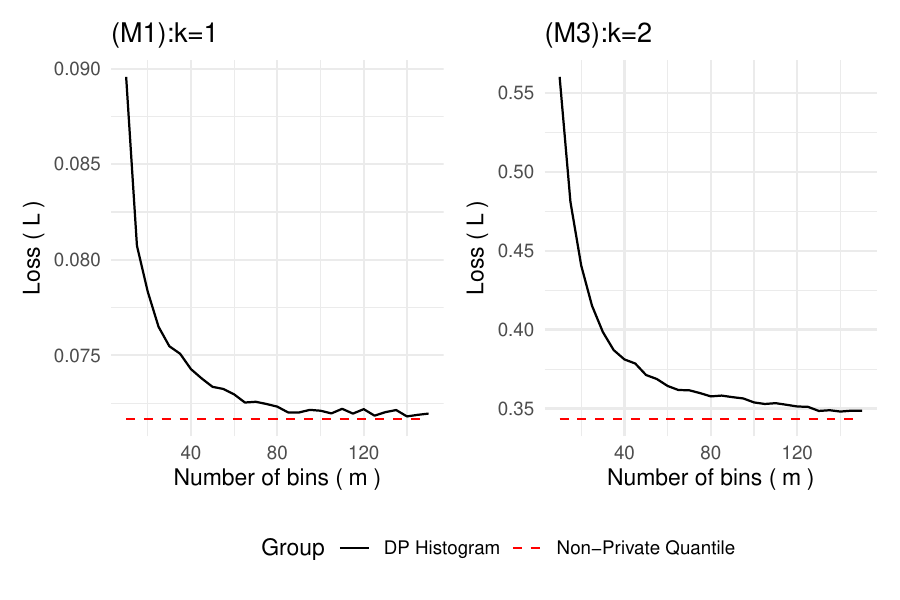}
	\caption{Comparison of the loss in the estimated matrix $\boldsymbol{B}$ derived from the kernel matrix based on DP estimated slices versus that based on sample quantiles for two generation mechanisms and varying number of bins $m$. The plot uses privacy parameter $\varepsilon=1$ and $H=10$ slices.}
	\label{fig:m.loss.ep1}
\end{figure}

Next, we examine the impact of the number of bins, a fundamental consideration in kernel-based density estimation. Selecting the appropriate number of bins significantly affects estimation accuracy in DP histograms, as discussed in \cite{wasserman2010statistical}. Our numerical analysis considers two privacy regimes: a high-privacy setting with  $\varepsilon=0.1$ and a low-privacy setting with $\varepsilon=1$. In both settings, we compare the eigen-gap and the loss in the estimated matrix between the kernel matrix based on DP estimated slices and that based on sample quantiles across two generation mechanisms and various values of $m$. The results for $\varepsilon=0.1$ are presented in Figures \ref{fig:m.eg.ep0.1} and \ref{fig:m.loss.ep0.1}. Since the loss in the estimated matrix is closely associated with the corresponding eigen-gap, we focus our discussion on the eigen-gap, noting that similar patterns hold for the loss. As the number of bins $m$ increases, the eigen-gap rises sharply and decreases gradually. In the first phase, where the eigen-gap increases significantly, each bin contains sufficient observations, so increasing the number of bins minimally impacts estimation accuracy within each bin while significantly reducing approximation errors associated with finite binning. In the second phase, where the eigen-gap decreases slowly, insufficient observations per bin cause estimation accuracy within bins to dominate the effect. The results for $\varepsilon=1$ are shown in Figures \ref{fig:m.eg.ep1} and \ref{fig:m.loss.ep1}. As the number of bins $m$ increases, the eigen-gap initially sharply rises and remains approximately stable. In the first phase, where the eigen-gap increases significantly, each bin contains a sufficient number of observations; therefore, increasing the number of bins minimally affects estimation accuracy within each bin while substantially reducing approximation errors due to finite binning. In the second phase, where the approximation error is sufficiently small, further increases in the number of bins show little effect. These results align with our theoretical findings in Lemma \ref{lem:est_slices}.

\begin{figure}[ht]
	\centering
	\includegraphics[width=0.8\linewidth]{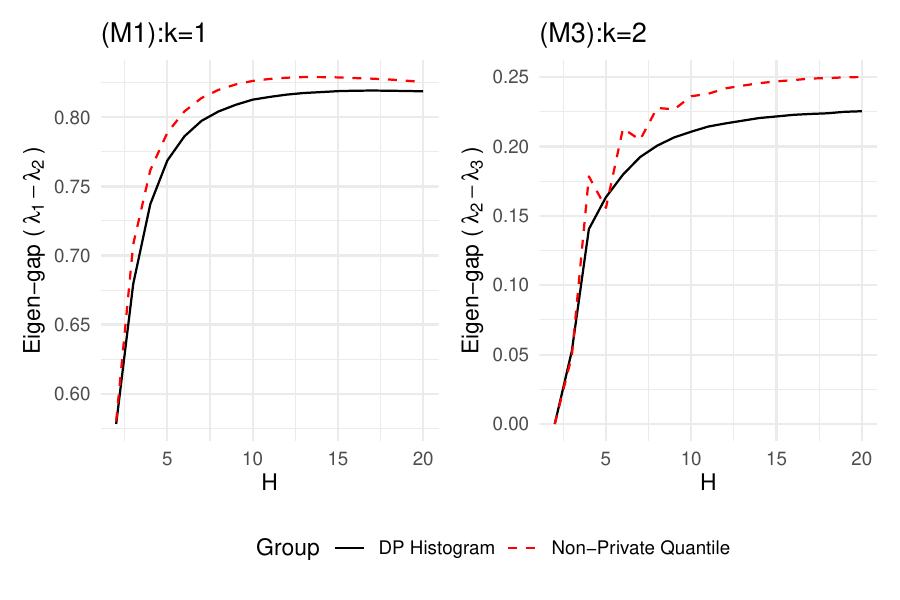}
	\caption{Comparison of the eigen-gaps between the kernel matrix based on DP estimated slices and that based on sample quantiles for two generation mechanisms and varying number of slices $H$. The plot uses privacy parameter $\varepsilon=0.1$ and the number of bins $m=50$.}
	\label{fig:H.eg}
\end{figure}

\begin{figure}[ht]
	\centering
	\includegraphics[width=0.8\linewidth]{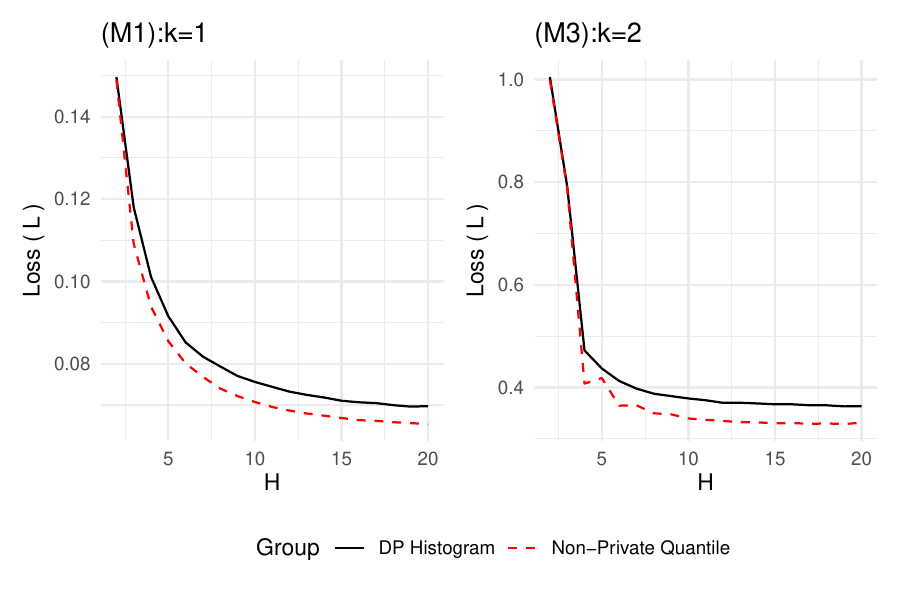}
	\caption{Comparison of the loss in the estimated matrix $\boldsymbol{B}$ derived from the kernel matrix based on DP estimated slices versus that based on sample quantiles for two generation mechanisms and varying number of slices $H$. The plot uses privacy parameter $\varepsilon=0.1$ and the number of bins $m=50$.}
	\label{fig:H.loss}
\end{figure}

Lastly, we examine the impact of the number of slices $H$. As demonstrated by \cite{lin2021optimal}, as the number of slices $H\to\infty$, the kernel matrix based on slices from sample quantiles, $\text{Cov}\{\mathbb{E}(\boldsymbol{x}\mid \widetilde{Y})\}$, converges to the matrix $\text{Cov}\{\mathbb{E}(\boldsymbol{x}\mid Y)\}$, where $\widetilde{Y}=\sum_{l=1}^{H}l\cdot\mathbbm{1}\{Y\in I_l\}$ takes different values among all the $H$ slices $\{I_h\}_{h=1}^{H}$. The result can be extended to where slices are derived from DP histograms due to Lemma \ref{lem:est_slices}. We conduct a numerical analysis to evaluate the finite-sample performance of slices obtained via DP histograms. The results, shown in Figures \ref{fig:H.eg} and \ref{fig:H.loss}, indicate that with sufficiently large privacy budget $\varepsilon$ and eigen-gap $\lambda$, increasing the number of slices $H$ results in an increase in the eigen-gap. Furthermore, the difference between the eigen-gap of the kernel matrix based on private estimates and that derived from non-private quantiles remains relatively small, with a slight increase as the number of slices $H$ grows. This slight increase can be attributed to the random noise introduced to satisfy privacy requirements. Furthermore, for $H\geq 10$,  the rate of increase in the eigen-gap concerning $H$ becomes notably slower, suggesting that selecting a reasonably large $H$ is sufficient.

\subsection{Tuning of the Sparsity Parameter $s^{\prime}$ in Sparse-SIR}

In the high dimensional sparse SIR problem, the proposed Algorithm requires a sparsity tuning parameter $s^{\prime}$ in Step 5. The sparsity tuning parameter is also necessary for the natural sparse SIR estimator and the refined natural sparse SIR estimator proposed by \cite{tan2020sparse}. In practice, we propose using the DP validation to select the sparsity tuning parameter. We begin by splitting the data into a training set (indexed by $\mathcal{T}$) and a validation set (indexed by $\mathcal{V}$). For the training set, we obtain private estimates under different sparsity tuning parameters. Then, using the validation set, we privately select and return the sparsity tuning parameter with the best performance in the validation data. The procedure is described as follows.

Let $\mathcal{S}=\{s_1,\dots,s_r\}$ be a set of $r$ candidate tuning parameters, and we use $s_{\max}=\max_{i=1,\dots,r}s_i$ to denote the upper bound of the tuning parameters. The upper bound on the sparsity is a prerequisite for DP algorithms, such as the peeling mechanism \citep{dwork2021differentially} and the sparse vector technique \citep{dwork2014algorithmic}. We apply Algorithm \ref{alg:hd_ini} to the training set with the peeling size $s_{\max}$ to obtain an ordered subset $\widehat{\mathcal{P}}$ by the peeling mechanism, and private estimates of the kernel matrix and the covariance matrix $\boldsymbol{M}$ and $\boldsymbol{\Sigma}$, denoted by $\boldsymbol{\tilde{M}}_{\widehat{\mathcal{P}}\widehat{\mathcal{P}}}(\mathcal{T})$ and $\boldsymbol{\tilde{\Sigma}}_{\widehat{\mathcal{P}}\widehat{\mathcal{P}}}(\mathcal{T})$, where $(\mathcal{T})$ indicates the dependence on the training set. For any $s\in\mathcal{S}$, let $\widehat{\mathcal{P}}_{s}$ be the first $s$ elements of $\widehat{\mathcal{P}}$. It is important to note that the peeling algorithm iteratively returns the noisy maximum value; therefore, the set $\widehat{\mathcal{P}}_{s}$ contains the noisy top $s$ elements of diagonal of $\boldsymbol{M}$. The private estimate with sparsity parameter $s$ is defined as
\begin{equation}
	\label{eq:candidate}
	\boldsymbol{\hat{B}}_{s}(\mathcal{T})=\argmax_{\B\in\mR^{s\times k}}\text{Tr}\{\B^{\top}\boldsymbol{\tilde{M}}_{\widehat{\mathcal{P}}_{s}\widehat{\mathcal{P}}_{s}}(\mathcal{T})\B\}\text{ s.t. }\B^{\top}\boldsymbol{\tilde{\Sigma}}_{\widehat{\mathcal{P}}_{s}\widehat{\mathcal{P}}_{s}}(\mathcal{T})\B=\boldsymbol{I}_k,
\end{equation}
where $\boldsymbol{\tilde{M}}_{\widehat{\mathcal{P}}_{s}\widehat{\mathcal{P}}_{s}}(\mathcal{T}),\boldsymbol{\tilde{\Sigma}}_{\widehat{\mathcal{P}}_{s}\widehat{\mathcal{P}}_{s}}(\mathcal{T})$ are sub-matrices of $\boldsymbol{\tilde{M}}_{\widehat{\mathcal{P}}\widehat{\mathcal{P}}}(\mathcal{T}),\boldsymbol{\tilde{\Sigma}}_{\widehat{\mathcal{P}}\widehat{\mathcal{P}}}(\mathcal{T})$ with the index set $\widehat{\mathcal{P}}_{s}$, respectively. Then, we obtain a set of candidate private estimates corresponding to the sparsity candidate set $\mathcal{S}$, denoted by $\mathcal{B}=\{\boldsymbol{\hat{B}}_{s_1}(\mathcal{T}),\dots,\boldsymbol{\hat{B}}_{s_r}(\mathcal{T})\}$.

For the validation set, we evaluate the performance of $\mathcal{B}$ by applying the loss function using the validation data. For any $s\in\mathcal{S}$, we define the validation loss by
\begin{equation}
	\label{eq:candidate_loss}
	\text{loss}(s):=\text{loss}(\widehat{\boldsymbol{B}}_{s}(\mathcal{T})):=-\text{Tr}(\widehat{\boldsymbol{B}}_{s}(\mathcal{T})^{\top}\widehat{\boldsymbol{M}}_{\widehat{\mathcal{P}}_{s}\widehat{\mathcal{P}}_{s}}(\mathcal{V})\widehat{\boldsymbol{B}}_{s}(\mathcal{T}))
\end{equation}
where $\widehat{\boldsymbol{M}}_{\widehat{\mathcal{P}}_{s}\widehat{\mathcal{P}}_{s}}(\mathcal{V})$ is a sub-matrix of $\boldsymbol{\hat{M}}(\mathcal{V})$ using the validation dataset, where $(\mathcal{V})$ indicates the dependence on the validation set. When the loss function $ \text{loss}(\cdot)$ has bounded sensitivity, the exponential mechanism \citep{dwork2014algorithmic} can be applied to the validation errors to select a value of $s$ in a DP manner. Let the sensitivity of $\text{loss}(\cdot)$ be bounded by $\Delta_l>0$. Then, we proposed to use an exponential mechanism that outputs $s\in\mathcal{S}$ with probability proportional to
\begin{equation*}
	\exp\bigg(\frac{\varepsilon\times \text{loss}(s)}{2\Delta_l}\bigg),
\end{equation*}
where $\varepsilon$ is the privacy parameter. The whole algorithm is formally outlined in Algorithm \ref{alg:dp_cv}.

\begin{algorithm}
	\caption{DP Validation}
	\begin{algorithmic}[1]
		\Require  $\{\boldsymbol{x}_{i},Y_{i}\}_{i=1,\dots,n}$, candidate set $\mathcal{S}=\{s_1,\dots,s_r\}$, number of slices $H$, dimension $k$, noisy level $\sigma$, truncation level $C_B$, privacy parameter $(\varepsilon,\delta)$;
		\State Estimate sliced intervals $\{\hat{I}_h\}_{h=1}^{H}$ based on the DP histogram under $(\varepsilon, 0)$-DP;
		\State Randomly partition the dataset into a training set (indexed by $\mathcal{T}$) and a validation set (indexed by $\mathcal{V}$), such that $\mathcal{T}\cup\mathcal{V}=[n]$ and $\mathcal{T}\cap\mathcal{V}=\emptyset$;
		\State Apply Algorithm \ref{alg:hd_ini} with the training set $\{\boldsymbol{x}_{i},Y_{i}\}_{i\in\mathcal{T}}$, privacy parameter $(\varepsilon,\delta)$ and sparsity parameter $s_{\max}$ to obtain an ordered index set $\widehat{\mathcal{P}}$ and a set of candidate private estimates $\mathcal{B}$ as outlined in \eqref{eq:candidate};
		\For {$s=s_1,s_2,\dots,s_r$}
		\State Truncation: $\widehat{\boldsymbol{B}}_{s}=\Pi_{C_B}(\widehat{\boldsymbol{B}}_{s})$, where $\Pi_{C_B}\{\cdot\}$ is the projection of each column of a matrix onto the $l_2$ ball $\mathbb{B}_{C_B}^s=\{\boldsymbol{x}\in\mR^s\mid \|\boldsymbol{x}\|_2\leq C_B\}$;
		\State Calculate $l_{s}:=\text{loss}(s)$ as described in \eqref{eq:candidate_loss}, where $\widehat{\boldsymbol{M}}(\mathcal{V})$ is the kernel matrix calculated with estimate sliced intervals $\{\hat{I}_h\}_{h=1}^{H}$ and the testing set $\{\boldsymbol{x}_{i},Y_{i}\}_{i\in\mathcal{V}}$;
		\EndFor
		\State Calculate normalized probability $p_{s}=\exp\{\varepsilon l_{s}/(2\sigma)\}/\sum_{s=s_1}^{s_r}\exp\{\varepsilon l_{s}/(2\sigma)\}$ for $s=s_1,\dots,s_r$;
		\State Sample $s^{\prime}$ from the multinomial distribution with support $\{s_1,\dots,s_r\}$ and corresponding probabilities $\{p_{s_1},\dots,p_{s_r}\}$;
		\Ensure $s^{\prime}$.
	\end{algorithmic}
	\label{alg:dp_cv}
\end{algorithm}

\begin{lemm}
	\label{lem:dp_cv}
	Suppose that Assumption \ref{assump:low_SIR}, \ref{assum:low_design}, \ref{assum:high_design} and \ref{assum:sparse_cov} hold, and let the parameters for Algorithm \ref{alg:hd_ini} be as specified in Lemma \ref{lem:hd_ini}. When 
	\begin{equation*}
		\sigma\geq C_B^2k\frac{7s_{\max}^{1/2}c_x^2}{|\mathcal{V}|},
	\end{equation*}
	the output of Algorithm \ref{alg:dp_cv} is $(3\varepsilon,\delta)$-DP.
\end{lemm}

\begin{proof}[Proof of Lemma \ref{lem:dp_cv}]
	The proof involves verifying the definition of differential privacy. We begin by introducing some notation. Let $\mathcal{D}:=\mathcal{T}\cup\mathcal{V}$ to denote the original data set and $\mathcal{D}^{\prime}:=\mathcal{T}^{\prime}\cup\mathcal{V}^{\prime}$ to denote a neighboring data set. Let the Algorithm \ref{alg:dp_cv} be written as $\mathcal{A}(\mathcal{D})=\mathcal{A}(\mathcal{T},\mathcal{V})$ to separate the dependence on the training set and the validation set. When the difference between $\mathcal{D}$ and $\mathcal{D}^{\prime}$ occurs in the training set, we use the notation $\mathcal{T}^{\prime}\neq\mathcal{T}$. We consider two distinct cases: one in which $\mathcal{T}^{\prime}$ and $\mathcal{T}$ differe in the value of a single individual, and another in which $\mathcal{V}^{\prime}$ and $\mathcal{V}$ differe in the value of a single individual. When $\mathcal{V}^{\prime}\neq\mathcal{V}$, for any possible outcome $s^{\prime}\in\mathcal{S}$, 
	\begin{equation*}
		\begin{split}
			\frac{\Pr(\mathcal{\mathcal{A}(\mathcal{T},\mathcal{V})}=s^{\prime})}{\Pr(\mathcal{\mathcal{A}(\mathcal{T},\mathcal{V}^{\prime})}=s^{\prime})}=\frac{\E\{\Pr(\mathcal{\mathcal{A}(\mathcal{T},\mathcal{V})}=s^{\prime}\mid\boldsymbol{Z}_{\mathcal{T}})\}}{\E\{\Pr(\mathcal{\mathcal{A}(\mathcal{T},\mathcal{V}^{\prime})}=s^{\prime}\mid\boldsymbol{Z}_{\mathcal{T}}^{\prime})\}},
		\end{split}
	\end{equation*}
	where $\boldsymbol{Z}_{\mathcal{T}},\boldsymbol{Z}_{\mathcal{T}}^{\prime}$ represent the randomness in the training procedure. Then, we have
	\begin{equation*}
		\begin{split}
			&\frac{\Pr(\mathcal{\mathcal{A}(\mathcal{T},\mathcal{V})}=s^{\prime})}{\Pr(\mathcal{\mathcal{A}(\mathcal{T},\mathcal{V}^{\prime})}=s^{\prime})}=\frac{\E\big\{\frac{\exp(\varepsilon l_{s^{\prime}}(\mathcal{T},\mathcal{V})/(2\sigma))}{\sum_i\exp(\varepsilon l_{s_i}(\mathcal{T},\mathcal{V})/(2\sigma))}\big\}}{\E\big\{\frac{\exp(\varepsilon l_{s^{\prime}}(\mathcal{T},\mathcal{V}^{\prime})/(2\sigma))}{\sum_i\exp(\varepsilon l_{s_i}(\mathcal{T},\mathcal{V}^{\prime})/(2\sigma))}\big\}}\\
			\leq&\E\bigg\{\frac{\frac{\exp(\varepsilon l_{s^{\prime}}(\mathcal{T},\mathcal{V})/(2\sigma))}{\sum_i\exp(\varepsilon l_{s_i}(\mathcal{T},\mathcal{V})/(2\sigma))}}{\frac{\exp(\varepsilon l_{s^{\prime}}(\mathcal{T},\mathcal{V}^{\prime})/(2\sigma))}{\sum_i\exp(\varepsilon l_{s_i}(\mathcal{T},\mathcal{V}^{\prime})/(2\sigma))}}\bigg\}=\E\bigg\{\frac{\exp(\varepsilon l_{s^{\prime}}(\mathcal{T},\mathcal{V})/(2\sigma))}{\exp(\varepsilon l_{s^{\prime}}(\mathcal{T},\mathcal{V}^{\prime})/(2\sigma))}\times\frac{\sum_i\exp(\varepsilon l_{s_i}(\mathcal{T},\mathcal{V}^{\prime})/(2\sigma))}{\sum_i\exp(\varepsilon l_{s_i}(\mathcal{T},\mathcal{V})/(2\sigma))}\bigg\}\\
			\leq&\E\bigg\{\exp\frac{\varepsilon(l_{s^{\prime}}(\mathcal{T},\mathcal{V})-l_{s^{\prime}}(\mathcal{T},\mathcal{V}^{\prime}))}{2\sigma}\times\max_i\exp\frac{\varepsilon(l_{s_i}(\mathcal{T},\mathcal{V})-l_{s_i}(\mathcal{T},\mathcal{V}^{\prime}))}{2\sigma}\bigg\},
		\end{split}
	\end{equation*}
	where we use the notation $l_{s}(\mathcal{T},\mathcal{V}^{\prime})$ to indicate the dependence of the loss function on the training set and the validation set, we use the Jensen's inequality and the independence between $\boldsymbol{Z}_{\mathcal{T}},\boldsymbol{Z}_{\mathcal{T}}^{\prime}$ in the second inequality. When $\sigma\geq\sup_{i,\mathcal{V},\mathcal{V}^{\prime}}|l_{s_i}(\mathcal{T},\mathcal{V})-l_{s_i}(\mathcal{T},\mathcal{V}^{\prime})|$, we have
	\begin{equation*}
		\frac{\Pr(\mathcal{\mathcal{A}(\mathcal{T},\mathcal{V})}=s^{\prime})}{\Pr(\mathcal{\mathcal{A}(\mathcal{T},\mathcal{V}^{\prime})}=s^{\prime})}\leq\E\bigg\{\exp\bigg(\frac{\varepsilon}{2}\bigg)\times\exp\bigg(\frac{\varepsilon}{2}\bigg)\bigg\}=\exp(\varepsilon).
	\end{equation*}    
	Note that 
	\begin{equation*}
		\begin{split}
			|l_{s_i}(\mathcal{T},\mathcal{V})-l_{s_i}(\mathcal{T},\mathcal{V}^{\prime})|&\leq |\text{Tr}(\widehat{\boldsymbol{B}}_{s_i}^{\top}(\widehat{\boldsymbol{M}}(\mathcal{V})-\widehat{\boldsymbol{M}}(\mathcal{V}^{\prime}))\widehat{\boldsymbol{B}}_{s_i})|\\
			&\leq C_B^2k\frac{7c_x^2s_{\max}^{1/2}}{|\mathcal{V}|},
		\end{split}
	\end{equation*}
	where we use the fact that $\|\widehat{\boldsymbol{M}}(\mathcal{V})-\widehat{\boldsymbol{M}}(\mathcal{V}^{\prime})\|_{\infty}\leq7c_x^2/|\mathcal{V}|$ as in the proof of Lemma \ref{lem:high_privacy}. By the composition theorem, the output $s^{\prime}$ is $(3\varepsilon,\delta)$-DP.

	When $\mathcal{T}^{\prime}\neq\mathcal{T}$, by the Lemma \ref{lem:hd_ini}, the candidate sets $\mathcal{B}$ is $(2\varepsilon,\delta)$-DP. By the post-processing property of DP, the output $s^{\prime}$ is $(2\varepsilon,\delta)$-DP.
	
	\hfill $\square$
\end{proof}

\begin{figure}[ht]
	\centering
	\includegraphics[width=0.6\linewidth]{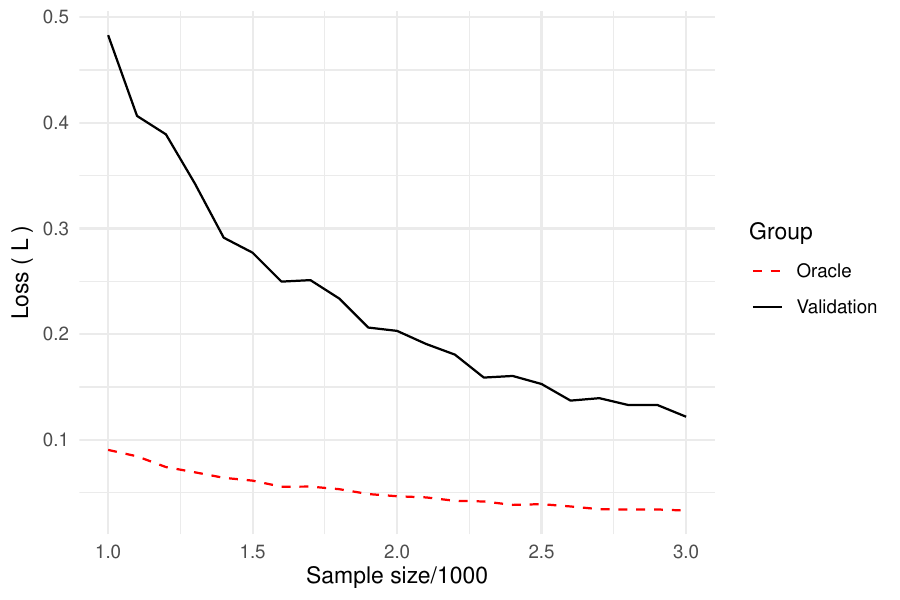}
	\caption{Average loss over $2000$ repetitions plotted against sample size $n$ with privacy parameters $(\varepsilon,\delta)=(1,n^{-1.1})$.}
	\label{fig:cv}
\end{figure}

We conduct a numerical study to demonstrate the performance of the validation procedure. Specifically, we consider the following scenario: a sample of size $n$ with dimension $p=1000$ is generated according to model (M1), as described in Section 5.1 of the main document. We compare the performance of the proposed estimator (Algorithm \ref{alg:hd_dp_sir}) using the sparsity parameter selected by validation (Algorithm \ref{alg:dp_cv}) over a uniform grid of values ranging from $1$ to $4s$, with that of the oracle estimator, which assumes knowledge of the true sparsity parameter $s$. The results are reported as averages over $2000$ replications. As shown in Figure \ref{fig:cv}, selecting the sparsity parameter by validation leads to errors comparable with the oracle estimator.

\section{Proof of the DP Slice Estimation}

\subsection{Proof of Lemma \ref{lem:est_slices}}
First, we consider the theoretical performance of the proposed slice selection method using the estimated DP histogram. We will also demonstrate how the slice estimation impacts the performance of the sliced inverse regression. The following lemma establishes that, under regularity conditions for the continuous response $Y$, the estimated cumulative distribution function is $(\varepsilon,0)$-DP and is a consistent estimator of the true cumulative distribution function of $Y$.

\begin{lemm}[Modified from Theorem 4.4 in \cite{wasserman2010statistical}]
	\label{lem:dp_his}
	The estimated density $\hat{f}_Y(\cdot)$ function, defined in \eqref{eq:dp_hist}, is $(\varepsilon,0)$-DP. Assume the variable $Y$ has a bounded domain and a density function $f_Y(\cdot)$, which is bounded by a constant, $f_Y(\cdot)\leq C$, and Lipschitz continuous. Then,
	\begin{equation*}
		\E\sup_y|F_Y(y)-\hat{F}_Y(y)|=O\bigg(\sqrt{\frac{\log(n)}{n}}+m^{-2}+\frac{m\log(m)}{n\varepsilon}\bigg),
	\end{equation*}
	where $F_Y(\cdot)$ is the cumulative distribution function of $Y$. Especially, when $m=n^{1/3}$ and $\sqrt{\log(n)}/\varepsilon=o(n^{1/6})$, we have
	\begin{equation*}
		\E\sup_y|F_Y(y)-\hat{F}_Y(y)|=O\bigg(\sqrt{\frac{\log(n)}{n}}\bigg),
	\end{equation*}
	where the error rate matches the statistical minimax rate up to $\sqrt{\log(n)}$.
\end{lemm}

Recall that when $Y$ is continuous, we consider the true slices defined by: $q_0=-1$, $q_H=1$ and $q_h=\inf\{x\in\{-1+2/m,\dots,1-2/m\}: F_Y(x)\geq h/H\}$ for $h=1,\dots,H-1$, where the choice of slices is defined based on the boundary of bins $B_j$. Thus, the above lemma further implies that the slices satisfy $|q_h-\hat{q}_h|\leq 2/m$ for all $h=0,\dots,H$ with probability approaching $1$ as $n\to\infty$. The theoretical proof is shown below.

\begin{proof}[Proof of Lemma \ref{lem:est_slices}]

	We reformulate some results in Lemma \ref{lem:dp_his}. By the proof of Theorem 4.2 in \cite{wasserman2010statistical}, the Vapnik-Chervonenkis dimension of the class of sets of the form $(-\infty,q_h]$ is $1$, and by the Vapnik-Chervonenkis bound, we have 
	\begin{equation}
		\label{eq:pf_lemma_4_3}
		\begin{split}
			&\mathbb{P}\bigg(\max_{1\leq j\leq m}|\sum_{i=1}^{j}b_j/n-\mathbb{P}(Y\leq -1+2j/m)|\geq 8\sqrt{\frac{\log(n)}{n}}\bigg)\\
			\leq&8n\exp\{-2\log(n)\}\leq 8\exp(-\log(n)).
		\end{split}
	\end{equation}
	By the fact that the Laplace distribution has a sub-exponential tail, we have the following result by 2.8.1 in \cite{vershynin2010introduction},
	\begin{equation*}
		\mathbb{P}(|\xi_j|\geq C_1\log(m)/\varepsilon)\leq 2\exp\{-2\log(m)\},
	\end{equation*}
	where $C_1$ is a positive constant. Furthermore, by the union bound, we have
	\begin{equation}
		\label{eq:pf_lemma_4_4}
		\mathbb{P}(\max_{1\leq j\leq m}|\xi_j|\geq C_1\log(m)/\varepsilon)\leq 2m\exp\{-2\log(m)\}=2\exp\{-\log(m)\}.
	\end{equation}
	Let $F_m(y)$ be the cumulative function defined on the bins:
	\begin{equation*}
		F_m(y):=\sum_{1\leq j\leq m}\mathbb{P}(Y\leq -1+2j/m)\mathbbm{1}\{-1+2(j-1)/m<y\leq-1+2j/m\}.
	\end{equation*}
	Under the events defined in \eqref{eq:pf_lemma_4_3} and \eqref{eq:pf_lemma_4_4}, we have
	\begin{equation*}
		\begin{split}
			\sup_y|F_Y(y)-\hat{F}_Y(y)|&\leq\sup_y|F_m(y)-\hat{F}_Y(y)|+\sup_y|F_Y(y)-F_m(y)|\\
			&\leq C^{\prime}\bigg(\sqrt{\frac{\log(n)}{n}}+\frac{m\log(m)}{n\varepsilon}+m^{-2}\bigg),
		\end{split}
	\end{equation*}
	where $C^{\prime}$ is a positive constant. We provide some details about the inequality: the first term in the inequality corresponds to the result \eqref{eq:pf_lemma_4_3}, the second term in the inequality corresponds to the result \eqref{eq:pf_lemma_4_4} and the proof of Theorem 4.4(2) in \cite{wasserman2010statistical}, and the third term in the inequality corresponds to $\sup_y|F_Y(y)-F_m(y)|$ by (21) in \cite{wasserman2010statistical}. Especially, when $m=n^{1/3}$ and $\sqrt{\log(n)}/\varepsilon=o(n^{1/6})$, we have
	\begin{equation*}
		\sup_y|F_Y(y)-\hat{F}_Y(y)|\leq 2C^{\prime}\bigg(\sqrt{\frac{\log(n)}{n}}+\frac{\log(m)}{n^{2/3}\varepsilon}\bigg)\leq 3C^{\prime}\sqrt{\frac{\log(n)}{n}},
	\end{equation*}
	with probability at least $1-8\exp\{-\log(n)\}-2\exp\{-\log(n)/3\}$.
	
	Then, we consider the estimation of the slices. By the conditions in Lemma \ref{lem:est_slices} that the density function $f_Y(y)$ is continuous in a compact domain, the density function is bounded by a constant $C^*$ from above. For $h\in\{1,\dots,H-1\}$, we have
	\begin{equation*}
		\begin{split}
			F_Y(q_{h}-4/m)&=F_Y(q_h-2/m)-\int_{q_h-4/m}^{q_h-2/m}f_Y(t)dt\leq F_Y(q_h-2/m)-C_*\int_{q_h-4/m}^{q_h-2/m}dt\\
			&\leq h/H-2C_*/m,
		\end{split}
	\end{equation*}
	where we use the condition that the density function satisfies $f_Y(q_h)\geq C_*$ in the second inequality, and we use the definition of $q_h$ in the third inequality. Moreover,
	\begin{equation*}
		\begin{split}
			F_Y(q_{h}+2/m)&=F_Y(q_h)+\int_{q_h}^{q_h+2/m}f_Y(t)dt\geq F_Y(q_h)+C_*\int_{q_h}^{q_h+2/m}dt\\
			&\geq h/H+2C_*/m,
		\end{split}
	\end{equation*}
	where we use the condition that the density function is bounded by $C_*$ from below in the second inequality, and we use the definition of $q_h$ in the third inequality. Thus, we have
	\begin{equation*}
		\hat{F}(q_{h}+2/m)\geq F_Y(q_{h}+2/m)-3C^{\prime}\sqrt{\frac{\log(n)}{n}}\geq h/H+\frac{2C_*}{n^{1/3}}-3C^{\prime}\sqrt{\frac{\log(n)}{n}}>h/H,
	\end{equation*}
	which further implies that $\hat{q}_h\leq q_{h}+2/m$. On the other hand, our aim is to show that $\hat{F}(q_{h}-4/m)\geq h/H$ is impossible. Similarly, we have
	\begin{equation*}
		\hat{F}(q_{h}-4/m)\leq F_Y(q_{h}-4/m)+3C^{\prime}\sqrt{\frac{\log(n)}{n}}\leq h/H-\frac{2C_*}{n^{1/3}}+3C^{\prime}\sqrt{\frac{\log(n)}{n}}<h/H,
	\end{equation*}
	which results in a contradiction. The event $\hat{F}(q_{h}-4/m)<h/H$ further implies that $\hat{q}_h\geq q_{h}-2/m$. Thus, we have $|\hat{q}_h-q_h|\leq 2/m$, for all $h=1,\dots,H$.
	\hfill $\square$
\end{proof}
So far, we have considered the case where the response variable $Y$ is continuous. The proposed method can also be applied to categorical responses. When the response variable $Y$ is categorical with $m$ unique values, we define $m$ bins, $\{B_1,\dots,B_m\}$, corresponding to these unique values of $Y$. The estimated probability mass function (pmf) for categorical $Y$ is $\widehat{p}_Y(j)=d_j/\sum_{i=1}^{m}d_i$. We define the true partitions by $m_0=0,m_h=\inf\{l\in\{1,\dots,m\}: \sum_{j=1}^{l}\mathbb{P}(Y\in B_j)\geq h/H\}$ and the corresponding true slices by $I_h=\bigcup_{l=m_{h-1}+1}^{m_h}B_l$. The slices are estimated based on the estimated pmf by $\widehat{I}_h=\bigcup_{l=\hat{m}_{h-1}+1}^{\hat{m}_h}B_l$, where $\hat{m}_0=0,\hat{m}_h=\inf\{l\in\{1,\dots,m\}: \sum_{j=1}^{l}\widehat{p}_Y(j)\geq h/H\}$. Then, the consistency of $\widehat{I}_h$ can be shown similarly by the proof of Lemma \ref{lem:est_slices}.

\subsection{Impact on the Kernel Matrix}

Before proving the desired properties of the proposed estimator, we examine the impact of the estimated slices on the kernel matrix. As will be shown later, the eigengap of the kernel matrix is important for the convergence of the proposed method. When the response $Y$ is continuous, for any vector in the central subspace, $\boldsymbol{v}\in\mathcal{S}_{Y\mid\boldsymbol{x}}$, we have
\begin{equation*}
	\lambda\boldsymbol{v}^{\top}\boldsymbol{\Sigma}^{-1}\boldsymbol{v}<\boldsymbol{v}^{\top}\boldsymbol{\Sigma}^{-1/2}\boldsymbol{M}\boldsymbol{\Sigma}^{-1/2}\boldsymbol{v}<\kappa\lambda\boldsymbol{v}^{\top}\boldsymbol{\Sigma}^{-1}\boldsymbol{v},
\end{equation*}
by Assumption \ref{assump:low_SIR}, where the matrix is defined as
\begin{equation*}
	\boldsymbol{M}:=\sum_{h=1}^{H}p_h\{\mathbb{E}(\boldsymbol{x}_i\mid Y_i\in I_h)-\mathbb{E}(\boldsymbol{x})\}\{\mathbb{E}(\boldsymbol{x}_i\mid Y_i\in I_h)-\mathbb{E}(\boldsymbol{x})\}^{\top},p_h=\mathbb{P}(Y_i\in I_h).
\end{equation*}

The eigen-gap in the kernel matrix is critical for SIR. As demonstrated in Lemma \ref{lem:Est.gap}, the kernel matrix with estimated slices meets the same conditions, up to a constant factor.

\begin{lemm}
	\label{lem:Est.gap}
	Suppose the conditions in Lemma \ref{lem:est_slices} and Assumption \ref{assump:low_SIR} hold. Assume that $1/(\lambda n^{1/3})<c$ for a small constant $c$. The ordered generalized eigenvalues $\{\lambda_1\geq\lambda_2\geq\dots\geq\lambda_p\}$ corresponding to the kernel matrix with estimated slices and $\boldsymbol{\Sigma}$ satisfy
	\begin{equation*}
		1\geq\kappa\lambda/2\geq\lambda_1\geq\dots\geq\lambda_k>\lambda/2>\lambda_{k+1}=\dots=\lambda_{p}=0,
	\end{equation*}
	with probability at least
	$1-\exp\{-C^{\prime}\log(n)\}$ for a positive constant $C^{\prime}$.
\end{lemm}
\begin{proof}[Proof of Lemma \ref{lem:Est.gap}]
	
	Without loss of generality, we assume $\mathbb{E}(\boldsymbol{x})=\boldsymbol{0}$. By Lemma \ref{lem:est_slices}, we have $|\widehat{q}_h-q_h|\leq 2/n^{1/3}$ for $h=1,\dots,H$ with probability at least $1-\exp\{-C^{\prime}\log(n)\}\stackrel{n\to\infty}{\longrightarrow}1$. Let $\{q_h^e\}_{h=1}^{H}$ be a set of cutoff points satisfying $|q^e_h-q_h|\leq 2/n^{1/3}$ and we use notations $\{I_h^e\}_{h=1}^{H}$ to denote the corresponding slices. The kernel matrix with estimated slices is defined as
	\begin{equation*}   \boldsymbol{M}_e:=\sum_{h=1}^{H}p_h^e\{\mathbb{E}(\boldsymbol{x}_i\mid Y_i\in I_h^e)\}\{\mathbb{E}(\boldsymbol{x}_i\mid Y_i\in I_h^e)\}^{\top},
	\end{equation*}
	where $p_h^e=\mathbb{P}(Y_i\in I_h^e)$. Thus, we consider the difference:
	\begin{equation*}
		\begin{split}
			&|\boldsymbol{v}^{\top}\boldsymbol{\Sigma}^{-1/2}(\boldsymbol{M}-\boldsymbol{M}^e)\boldsymbol{\Sigma}^{-1/2}\boldsymbol{v}|\\
			\leq&\bigg|\boldsymbol{v}^{\top}\boldsymbol{\Sigma}^{-1/2}\sum_{h=1}^{H}\bigg[\mathbb{E}\{\boldsymbol{x}_i\mathbbm{1}(Y_i\in I_h)-\boldsymbol{x}_i\mathbbm{1}(Y_i\in I_h^e)\}\mathbb{E}(\boldsymbol{x}_i\mid Y_i\in I_h)^{\top}\bigg]\boldsymbol{\Sigma}^{-1/2}\boldsymbol{v}\bigg|\\
			&+\bigg|\boldsymbol{v}^{\top}\boldsymbol{\Sigma}^{-1/2}\sum_{h=1}^{H}\bigg[\mathbb{E}\{\boldsymbol{x}_i\mathbbm{1}(Y_i\in I^e_h)\}\{\mathbb{E}(\boldsymbol{x}_i\mid Y_i\in I_h)-\mathbb{E}(\boldsymbol{x}_i\mid Y_i\in I^e_h)\}^{\top}\bigg]\boldsymbol{\Sigma}^{-1/2}\boldsymbol{v}\bigg|.
		\end{split}
	\end{equation*}
	where we use the triangle inequality. We first consider the first term. Note that
	\begin{equation*}
		\begin{split}
			&\mathbb{E}\{\boldsymbol{x}_i\mathbbm{1}(Y_i\in I_h)-\boldsymbol{x}_i\mathbbm{1}(Y_i\in I^e_h)\}\\
			=&\mathbb{E}\{\boldsymbol{x}_i\mathbbm{1}(Y_i\in I_h/I^e_h)-\boldsymbol{x}_i\mathbbm{1}(Y_i\in I^e_h/I_h)\}\\
			=&\mathbb{E}\{\boldsymbol{x}_i|Y_i\in I_h/I^e_h\}\mathbb{P}(Y_i\in I_h/I^e_h)-\mathbb{E}\{\boldsymbol{x}_i|Y_i\in I^e_h/I_h\}\mathbb{P}(Y_i\in I^e_h/I_h).
		\end{split}
	\end{equation*}
	By the assumption \ref{assum:low_design}, we know $\boldsymbol{x}$ is sub-Gaussian distributed, and thus, $\E(\boldsymbol{v}^{\top}\boldsymbol{\Sigma}^{-1/2}\boldsymbol{x}_i)\leq C_X\|\boldsymbol{v}^{\top}\boldsymbol{\Sigma}^{-1/2}\|_2$ for a positive constant $C_X$. The first term satisfies,
	\begin{equation*}
		\begin{split}
			&\bigg|\boldsymbol{v}^{\top}\boldsymbol{\Sigma}^{-1/2}\sum_{h=1}^{H}\bigg[\mathbb{E}\{\boldsymbol{x}_i\mathbbm{1}(Y_i\in I_h)-\boldsymbol{x}_i\mathbbm{1}(Y_i\in I^e_h)\}\mathbb{E}(\boldsymbol{x}_i\mid Y_i\in I_h)^{\top}\bigg]\boldsymbol{\Sigma}^{-1/2}\boldsymbol{v}\bigg|\\
			\leq&C_X^2(\boldsymbol{v}^{\top}\boldsymbol{\Sigma}^{-1}\boldsymbol{v})\sum_{h=1}^{H}\{\mathbb{P}(Y_i\in I_h/I^e_h)+\mathbb{P}(Y_i\in I_h/I^e_h)\}\leq 4HC_X^2C^*(\boldsymbol{v}^{\top}\boldsymbol{\Sigma}^{-1}\boldsymbol{v})/n^{1/3},
		\end{split}
	\end{equation*}
	where we use the fact that the density $f_Y(\cdot)$ is bounded by $C^*$. The second term satisfies,
	\begin{equation*}
		\begin{split}
			&\mathbb{E}(\boldsymbol{x}_i\mid Y_i\in I_h)-\mathbb{E}(\boldsymbol{x}_i\mid Y_i\in I^e_h)\\
			=&\mathbb{E}\{\boldsymbol{x}_i\mathbbm{1}(Y_i\in I_h)\}/p_h-\mathbb{E}\{\boldsymbol{x}_i\mathbbm{1}(Y_i\in I^e_h)\}/p_h-\mathbb{E}\{\boldsymbol{x}_i\mathbbm{1}(Y_i\in I^e_h)\}(1/p_h^e-1/p_h).
		\end{split}
	\end{equation*}
	By the fact $1/p_h,1/p^e_h\geq 1/(2H)$, we have the second term satisfies
	\begin{equation*}
		\begin{split}
			&\bigg|\boldsymbol{v}^{\top}\boldsymbol{\Sigma}^{-1/2}\sum_{h=1}^{H}\bigg[\mathbb{E}\{\boldsymbol{x}_i\mathbbm{1}(Y_i\in I^e_h)\}\{\mathbb{E}(\boldsymbol{x}_i\mid Y_i\in I_h)-\mathbb{E}(\boldsymbol{x}_i\mid Y_i\in I^e_h)\}^{\top}\bigg]\boldsymbol{\Sigma}^{-1/2}\boldsymbol{v}\bigg|\\
			\leq&C_X^2C^*H(\boldsymbol{v}^{\top}\boldsymbol{\Sigma}^{-1}\boldsymbol{v})(4H+8H^2)/n^{1/3}.
		\end{split}
	\end{equation*}
	Thus 
	\begin{equation*}
		\begin{split}
			\boldsymbol{v}^{\top}\boldsymbol{\Sigma}^{-1/2}\boldsymbol{M}^e\boldsymbol{\Sigma}^{-1/2}\boldsymbol{v}&\leq\boldsymbol{v}^{\top}\boldsymbol{\Sigma}^{-1/2}\boldsymbol{M}\boldsymbol{\Sigma}^{-1/2}\boldsymbol{v}+|\boldsymbol{v}^{\top}\boldsymbol{\Sigma}^{-1/2}(\boldsymbol{M}-\boldsymbol{M}^e)\boldsymbol{\Sigma}^{-1/2}\boldsymbol{v}|\\
			&\leq\{\kappa\lambda+O(n^{-1/3})\}\boldsymbol{v}^{\top}\boldsymbol{\Sigma}^{-1}\boldsymbol{v}
		\end{split}
	\end{equation*}
	and
	\begin{equation*}
		\begin{split}
			\boldsymbol{v}^{\top}\boldsymbol{\Sigma}^{-1/2}\boldsymbol{M}^e\boldsymbol{\Sigma}^{-1/2}\boldsymbol{v}&\geq\boldsymbol{v}^{\top}\boldsymbol{\Sigma}^{-1/2}\boldsymbol{M}\boldsymbol{\Sigma}^{-1/2}\boldsymbol{v}-|\boldsymbol{v}^{\top}\boldsymbol{\Sigma}^{-1/2}(\boldsymbol{M}-\boldsymbol{M}^e)\boldsymbol{\Sigma}^{-1/2}\boldsymbol{v}|\\
			&\geq\{\lambda-O(n^{-1/3})\}\boldsymbol{v}^{\top}\boldsymbol{\Sigma}^{-1}\boldsymbol{v}.
		\end{split}
	\end{equation*}
	When $1/(\lambda n^{1/3})$ is small enough, the previous inequality imples that the ordered generalized eigenvalues $\{\lambda_1\geq\lambda_2\geq\dots\geq\lambda_p\}$ corresponding to $\boldsymbol{M}^e$ and $\boldsymbol{\Sigma}$ satisfy
	\begin{equation*}
		1\geq\kappa\lambda/2\geq\lambda_1\geq\dots\geq\lambda_k>\lambda/2>\lambda_{k+1}=\dots=\lambda_{p}=0,
	\end{equation*}
	where $\kappa>0$ is a fixed constant. 
	\hfill $\square$
\end{proof}

\section{Proof of the Minimax Lower Bounds}\label{sec: minimax}

\subsection{Proof of Theorem 1}
It's enough to consider the case where $k=1$ and $f$ is the identity function, in which case we have $$
Y=\bm x^\top\bm \beta+e.
$$
We first establish the relation between the estimation of the regression coefficients $\boldsymbol{\beta}$ and the estimation of the eigenvector $\boldsymbol{B}$. Assume $\boldsymbol{x}\sim N(\boldsymbol{0},\boldsymbol{\Sigma})$ and $e\sim N(0,\sigma_e^2)$. We consider the sliced intervals $I_1=(-\infty,a)$ and $I_2=[a,\infty)$. Notice that
\begin{equation*}
	\begin{split}
		\mathbb{E}(\boldsymbol{x}|Y\in I_1)&=\mathbb{E}(\boldsymbol{x}|\boldsymbol{x}^{\top}\boldsymbol{\beta}\leq a-e)=\frac{-\boldsymbol{\Sigma}\boldsymbol{\beta}}{\sqrt{\boldsymbol{\beta}^{\top}\boldsymbol{\Sigma}\boldsymbol{\beta}}}\mathbb{E}\bigg[ \frac{\phi((a-e)/\sqrt{\boldsymbol{\beta}^{\top}\boldsymbol{\Sigma}\boldsymbol{\beta}})}{\Phi((a-e)/\sqrt{\boldsymbol{\beta}^{\top}\boldsymbol{\Sigma}\boldsymbol{\beta}})}\bigg],
	\end{split}
\end{equation*}
where $\phi(\cdot)$ and $\Phi(\cdot)$ are the pdf and cdf of the standard normal distribution, respectively, and the expectation $\mathbb{E}$ is with respect to $e$. Similarly, we have
\begin{equation*}
	\begin{split}
		\mathbb{E}(\boldsymbol{x}|Y\in I_2)=\mathbb{E}(\boldsymbol{x}|\boldsymbol{x}^{\top}\boldsymbol{\beta}\geq a-e)=\frac{\boldsymbol{\Sigma}\boldsymbol{\beta}}{\sqrt{\boldsymbol{\beta}^{\top}\boldsymbol{\Sigma}\boldsymbol{\beta}}}\mathbb{E}\bigg[\frac{\phi((a-e)/\sqrt{\boldsymbol{\beta}^{\top}\boldsymbol{\Sigma}\boldsymbol{\beta}})}{1-\Phi((a-e)/\sqrt{\boldsymbol{\beta}^{\top}\boldsymbol{\Sigma}\boldsymbol{\beta}})}\bigg].
	\end{split}
\end{equation*}
Furthermore, we have the probabilities,
\begin{equation*}
	\mathbb{P}(Y\in I_1)=\mathbb{P}(\bm x^{\top}\bm\beta+e\leq a)=\Phi(a/\sqrt{\boldsymbol{\beta}^{\top}\boldsymbol{\Sigma}\boldsymbol{\beta}+\sigma_e^2}),
\end{equation*}
\begin{equation*}
	\mathbb{P}(Y\in I_2)=\mathbb{P}(\bm x^{\top}\bm\beta+e\geq a)=1-\Phi(a/\sqrt{\boldsymbol{\beta}^{\top}\boldsymbol{\Sigma}\boldsymbol{\beta}+\sigma_e^2}).
\end{equation*}
Notice that we have
\begin{equation*}
	\begin{split}
		\mathbb{E}(\boldsymbol{x})=&\frac{\boldsymbol{\Sigma}\boldsymbol{\beta}}{\sqrt{\boldsymbol{\beta}^{\top}\boldsymbol{\Sigma}\boldsymbol{\beta}}}\bigg\{-\mathbb{E}\bigg[ \frac{\phi((a-e)/\sqrt{\boldsymbol{\beta}^{\top}\boldsymbol{\Sigma}\boldsymbol{\beta}})}{\Phi((a-e)/\sqrt{\boldsymbol{\beta}^{\top}\boldsymbol{\Sigma}\boldsymbol{\beta}})}\bigg]\times\Phi(a/\sqrt{\boldsymbol{\beta}^{\top}\boldsymbol{\Sigma}\boldsymbol{\beta}+\sigma_e^2})\\
		&+\mathbb{E}\bigg[\frac{\phi((a-e)/\sqrt{\boldsymbol{\beta}^{\top}\boldsymbol{\Sigma}\boldsymbol{\beta}})}{1-\Phi((a-e)/\sqrt{\boldsymbol{\beta}^{\top}\boldsymbol{\Sigma}\boldsymbol{\beta}})}\bigg]\times(1-\Phi(a/\sqrt{\boldsymbol{\beta}^{\top}\boldsymbol{\Sigma}\boldsymbol{\beta}+\sigma_e^2}))\bigg\}\\
		=&\boldsymbol{0}.
	\end{split}
\end{equation*}
Then, the kernel matrix is
\begin{equation*}
	\begin{split}
		\boldsymbol{M}=&\mathbb{P}(Y\in I_1)\mathbb{E}(\boldsymbol{x}|Y\in I_1)\mathbb{E}(\boldsymbol{x}|Y\in I_1)^{\top}+\mathbb{P}(Y\in I_2)\mathbb{E}(\boldsymbol{x}|Y\in I_2)\mathbb{E}(\boldsymbol{x}|Y\in I_2)^{\top}\\
		=&\frac{\bm\Sigma\bm\beta\bm\beta^{\top}\bm\Sigma}{\boldsymbol{\beta}^{\top}\boldsymbol{\Sigma}\boldsymbol{\beta}}\cdot\bigg\{\Phi\bigg(\frac{a}{\sqrt{\boldsymbol{\beta}^{\top}\boldsymbol{\Sigma}\boldsymbol{\beta}+\sigma_e^2}}\bigg)\mathbb{E}\bigg[ \frac{\phi((a-e)/\sqrt{\boldsymbol{\beta}^{\top}\boldsymbol{\Sigma}\boldsymbol{\beta}})}{\Phi((a-e)/\sqrt{\boldsymbol{\beta}^{\top}\boldsymbol{\Sigma}\boldsymbol{\beta}})}\bigg]^2\\
		&+\Phi\bigg(\frac{-a}{\sqrt{\boldsymbol{\beta}^{\top}\boldsymbol{\Sigma}\boldsymbol{\beta}+\sigma_e^2}}\bigg)\mathbb{E}\bigg[ \frac{\phi((a-e)/\sqrt{\boldsymbol{\beta}^{\top}\boldsymbol{\Sigma}\boldsymbol{\beta}})}{1-\Phi((a-e)/\sqrt{\boldsymbol{\beta}^{\top}\boldsymbol{\Sigma}\boldsymbol{\beta}})}\bigg]^2\bigg\}.
	\end{split}
\end{equation*}
By the fact that $\boldsymbol{M}\boldsymbol{\beta}=\lambda\boldsymbol{\Sigma}\boldsymbol{\beta}$, we have
\begin{equation*}
	\begin{split}
		\lambda=&\Phi\bigg(\frac{a}{\sqrt{\boldsymbol{\beta}^{\top}\boldsymbol{\Sigma}\boldsymbol{\beta}+\sigma_e^2}}\bigg)\mathbb{E}\bigg[ \frac{\phi((a-e)/\sqrt{\boldsymbol{\beta}^{\top}\boldsymbol{\Sigma}\boldsymbol{\beta}})}{\Phi((a-e)/\sqrt{\boldsymbol{\beta}^{\top}\boldsymbol{\Sigma}\boldsymbol{\beta}})}\bigg]^2\\
		&+\Phi\bigg(\frac{-a}{\sqrt{\boldsymbol{\beta}^{\top}\boldsymbol{\Sigma}\boldsymbol{\beta}+\sigma_e^2}}\bigg)\mathbb{E}\bigg[ \frac{\phi((a-e)/\sqrt{\boldsymbol{\beta}^{\top}\boldsymbol{\Sigma}\boldsymbol{\beta}})}{1-\Phi((a-e)/\sqrt{\boldsymbol{\beta}^{\top}\boldsymbol{\Sigma}\boldsymbol{\beta}})}\bigg]^2.
	\end{split}
\end{equation*}

When the error variance $\sigma_e^2$ is bounded by a constant, then the eigen-gap $\lambda$ is also bound by a constant. Then, it remains to consider the case where $\sigma_e^2$ converges to $\infty$ as $n\to\infty$.

We pick $a\asymp\sigma_e\sqrt{\log(\sigma_e)}$ such that $\Phi((-a)/\sqrt{\boldsymbol{\beta}^{\top}\boldsymbol{\Sigma}\boldsymbol{\beta}+\sigma_e^2}))\asymp 1/\sigma_e$. Thus, the eigen-gap satisfies $\lambda\gtrsim 1/\sigma_e$.

Let $M(\bm Y,\bm X)$ be an estimator of $\bm B$ using data $\{(Y_i,\bm x_i)\}_{i=1}^{n}$. Then, we link the estimation of $\bm B$ and the estimation of $\bm\beta$. We know that when the dimension of $\bm B$ equals to $1$, we have the relation $\|\bm P_{M(\bm Y,\bm X)}-\bm P_{\bm B}\|_F^2=2\sin^2(\theta)$, where $\theta$ is the angle between $\bm B$ and $M(\bm Y,\bm X)$, which is defined as
\begin{equation*}
	\theta=\arccos\bigg(\frac{\bm B^{\top}M(\bm Y,\bm X)}{\|\bm B\|_2\cdot\| M(\bm Y,\bm x)\|_2}\bigg).
\end{equation*}
Under the assumption $\|\bm P_{M(\bm Y,\bm X)}-\bm P_{\bm B}\|^2$, we have $\theta\to 0$. Furthermore, we have 
\begin{equation*}
	\| M(\bm Y,\bm X)/\|M(\bm Y,\bm X)\|_2- \bm B/\|\bm B\|_2\|_2^2=2-2\cos(\theta)=4\sin^2(\theta/2)\asymp\theta^2\asymp\|\bm P_{M(\bm Y,\bm X)}-\bm P_{\bm B}\|_F^2.
\end{equation*}
Thus, it remains to consider the estimation of the coefficient $\bm \beta$.

We use the score attack developed in \cite{cai2023score} to prove the lower bound. For any estimator $M(\bm Y, \bm X)$ of $\boldsymbol{\beta}$, we consider the score attack  $$	\A_{{\bm \beta}}((\widetilde Y, \widetilde{\bm x}), M(\bm Y, \bm X)) = \big\langle M(\bm Y, \bm X ), \{\widetilde Y - (\widetilde{\bm x}^\top {\bm \beta})\}\widetilde {\bm x} \big\rangle.
$$ 
For each $i \in [n]$, we denote $\A_{{\bm \beta}}((Y_i, \bm x_i), M(\bm y, \bm X))$ by $A_i$. We first present the following two lemmas.
\begin{lemm}\label{prop: low-dim attack soundness}
	Suppose $\E(\bm x \bm x^\top)$ is diagonal and $\lambda_{\max}(\E(\bm x \bm x^\top)) < C < \infty$ and $\text{var}(e)=\sigma_e^2<\infty$. If the estimator $M$ is $(\varepsilon, \delta)$-DP with $0 < \varepsilon < 1$ and satisfies $\|M(\bm Y, \bm X) - {\bm \beta}\|^2_2 \lesssim p$, then
	\begin{align}\label{eq: low-dim glm attack soundness}
		\sum_{i \in [n]} \E A_i \leq 2n\varepsilon\sqrt{\E\|M(\bm Y, \bm X) - {\bm \beta}\|^2_2\sigma_e^2}C + 4\sqrt{2}\delta n\sqrt{p\log(1/\delta)\sigma_e^2}C.
	\end{align}
\end{lemm}
The proof is in the proof of Theorem 4.1 in \cite{cai2021cost}.
\begin{lemm}\label{prop: low-dim attack completeness}
	Suppose $d = o\left(p^{1-\gamma}\right)$ for some $\gamma > 0$. For every $M$ satisfying $\E_{\bm Y, \bm X|{\bm \beta}}\|M(\bm Y, \bm X) - {\bm \beta}\|^2_2 \lesssim 1$ at every ${\bm \beta}$, we have 
	\begin{align}\label{eq: low-dim glm attack completeness}
		\sum_{i \in [n]} \E_{\bm \pi}\E_{\bm Y, \bm X|{\bm \beta}} A_i \gtrsim p\sigma_e^2, 
	\end{align}
	where $\bm \pi$ refers to the $Beta(3,3)$ prior for ${\bm\beta}$.
\end{lemm}

We now prove the minimax lower bound over $\mathcal P$, we may consider only those $M$ satisfying $\|M(\bm Y, \bm X) - {\bm \beta}\|^2_2 \lesssim p$, for any $M$ violating this bound lies outside $\Theta$ and cannot be optimal. For now we also assume that $M$ is such that $\E_{\bm Y, \bm X|{\bm \beta}}\|M(\bm Y, \bm X) - {\bm \beta}\|^2_2 \lesssim 1$ at every ${\bm \beta}$. We have
\begin{align*}
	p\sqrt{\sigma_e^2} \lesssim \sum_{i \in [n]}\E_{\bm \pi} \E_{\bm Y, \bm X|{\bm \beta}} A_i \leq 2n\varepsilon\sqrt{\E\|M(\bm Y, \bm X) - {\bm \beta}\|^2_2}C + 4\sqrt{2}\delta n \sqrt{p\log(1/\delta)}C.
\end{align*}
It follows that
\begin{align*}
	2n\varepsilon\E_{\bm \pi} \sqrt{\E_{\bm Y, \bm X|{\bm \beta}} \|M(\bm Y, \bm X) - {\bm \beta}\|^2_2} C \gtrsim p \sqrt{\sigma_e^2}- 4\sqrt{2}n\delta \sqrt{p\log(1/\delta)}C.
\end{align*}
The assumption that $\delta < n^{-(1+\gamma)}$ for some $\gamma > 0$ implies that for $n$ sufficiently large, $p - 4\sqrt{2}n\delta  \sqrt{p\log(1/\delta)}C \gtrsim p$. We then conclude that
\begin{align*}
	\E_{\bm \pi}\E_{\bm Y, \bm X|{\bm \beta}} \|M(\bm Y, \bm X) - {\bm \beta}\|^2_2 \gtrsim\sigma_e^2 \frac{p^2}{n^2\varepsilon^2}.
\end{align*}
Because the sup-risk is always greater than the Bayes risk, we have 
\begin{align*}
	\sup_{{\bm \beta} \in \Theta}\E_{\bm Y, \bm X|{\bm \beta}} \|M(\bm Y, \bm X) - {\bm \beta}\|^2_2 \gtrsim \sigma_e^2\frac{p^2}{n^2\varepsilon^2}.
\end{align*}
The bound is true for any $M$ satisfying $\E_{\bm Y, \bm X|{\bm \beta}}\|M(\bm Y, \bm X) - {\bm \beta}\|^2_2 \lesssim 1$; it extends to all $M \in \mathcal M_{\varepsilon, \delta}$ as we assumed $p \lesssim n\varepsilon$ and therefore $p^2/(n\varepsilon)^2 \lesssim 1$. The proof is complete by combining with the non-private minimax lower bound $\inf_M\sup_{{\bm \beta} \in \R^p} \E\|M(\bm Y, \bm X) - {\bm \beta}\|_2^2 \gtrsim \sigma_e^2 p/n$.

As a result, we have $$
\inf_{\hat B\in\mathcal M_{\epsilon,\delta}}\sup_{P\in\mathcal P}\E L(\boldsymbol{\hat{B}},\B)\gtrsim\frac{p}{\lambda^2n}+\frac{p^2}{\lambda^2n^2\varepsilon^2}.
$$
\subsubsection{Proof of Lemma~\ref{prop: low-dim attack completeness}}

By the completeness part of Theorem 2.1 in \cite{cai2023score}, we know $$\sum_{i \in [n]} \E_{\bm Y, \bm X|\bm \beta} A_i = \sum_{j \in [p]} \frac{\partial}{\partial \beta_j} \E_{\bm Y, \bm X|\bm \beta} M(\bm Y, \bm X)_j.$$ By Proposition 2.2 in \cite{cai2023score} and the assumption that $\E\|M(\bm Y, \bm X) - \bm \beta\|^2_2 \lesssim 1$ at every $\bm \beta$, the proof is complete by plugging the choice of $\bm \pi(\bm \beta)$, the product of $d$ copies of the Beta$(3,3)$ density, and evaluating the integrals.

\subsection{Proof of Theorem 3}
We use the score attack developed in \cite{cai2023score} to prove the lower bound. For any estimator $M(\bm Y, \bm X))$, we consider the sparse score attack  $$	\A_{{\bm \beta},s}((\widetilde Y, \widetilde{\bm x}), M(\bm Y, \bm X)) = \big\langle (M(\bm Y, \bm X) )_{S(M(\bm Y, \bm X) )}, \{\widetilde Y - (\widetilde{\bm x}^\top {\bm \beta})\}( \widetilde {\bm x})_{S(\bm \beta)} \big\rangle,
$$ where $M(\bm Y, \bm X)_{S(M(\bm Y, \bm X) )}$ denotes the trimmed vector which keeps top-$s$ indices of $M(\bm Y, \bm X) $ and sets remaining to zero. For each $i \in [n]$, we denote $\A_{{\bm \beta},s}((Y_i, \bm x_i), M(\bm Y, \bm X))$ by $A_i$. We first present the following two lemmas.
\begin{lemm}\label{prop: high-dim attack soundness}
	Suppose $\E(\bm x \bm x^\top)$ is diagonal and $\lambda_{\max}(\E(\bm x \bm x^\top)) < C < \infty$, and $\text{var}(e)=\sigma^2\leq\infty$. If the estimator $M$ is $(\varepsilon, \delta)$-DP with $0 < \varepsilon < 1$ and satisfies $\|M(\bm Y, \bm X) - {\bm \beta}\|^2_2 \lesssim o(1)$, then
	\begin{align}\label{eq: high-dim glm attack soundness}
		\sum_{i \in [n]} \E A_i \leq 2n\varepsilon\sqrt{\E\|M(\bm Y, \bm X) - {\bm \beta}\|^2_2\sigma_e^2} C + 4\sqrt{2}\delta n \sqrt{s\log(1/\delta)\sigma_e^2}C.
	\end{align}
\end{lemm}
The proof is in the proof of Theorem 4.3 in \cite{cai2021cost}.
\begin{lemm}\label{prop: high-dim glm attack completeness}
	Suppose $s = o(p^{1-\gamma})$ for some $\gamma > 0$. For every $M$ satisfying $\E_{\bm Y, \bm X|{\bm \beta}}\|M(\bm Y, \bm X) - {\bm \beta}\|^2_2 \lesssim 1$ at every ${\bm \beta}$, we have 
	\begin{align}\label{eq: high-dim glm attack completeness}
		\sum_{i \in [n]} \E_{\bm \pi}\E_{\bm Y, \bm X|{\bm \beta}} A_i \gtrsim \sigma_e^2s\log(p/s), 
	\end{align}
	where $\bm \pi$ refers to the sparse truncated normal prior for ${\bm \beta}$.
\end{lemm}

We now prove the minimax lower bound over $\mathcal P$, we may consider only those $M$ satisfying $\|M(\bm Y, \bm X) - {\bm \beta}\|^2_2 \lesssim 1$, for any $M$ violating this bound lies outside $\Theta$ and cannot be optimal. For now we also assume that $M$ is such that $\E_{\bm Y, \bm X|{\bm \beta}}\|M(\bm Y, \bm X) - {\bm \beta}\|^2_2 \lesssim 1$ at every ${\bm \beta}$. We have
\begin{align*}
	\sigma_e s\log p \lesssim \sum_{i \in [n]}\E_{\bm \pi} \E_{\bm Y, \bm X|{\bm\beta}} A_i \leq 2n\varepsilon\sqrt{\E\|M(\bm Y, \bm X) - {\bm \beta}\|^2_2} C+ 4\sqrt{2}\delta n \sqrt{s\log(1/\delta)}C.
\end{align*}
It follows that
\begin{align*}
	2n\varepsilon\E_{\bm \pi} \sqrt{\E_{\bm Y, \bm X|{\bm \beta}} \|M(\bm Y, \bm X) - {\bm \beta}\|^2_2} C\gtrsim s\log(p/s) \sigma_e- 4\sqrt{2}n\delta  \sqrt{s\log(1/\delta)}C.
\end{align*}
The assumption that $\delta < n^{-(1+\gamma)}$ for some $\gamma > 0$ implies that for $n$ sufficiently large, $s\log(p/s) - 4\sqrt{2}n\delta \sqrt{s\log(1/\delta)}C \gtrsim s\log(p/s)$. We then conclude that
\begin{align*}
	\E_{\bm \pi}\E_{\bm Y, \bm X|{\bm \beta}} \|M(\bm Y, \bm X) - {\bm \beta}\|^2_2 \gtrsim \sigma_e^2\frac{(s\log(p/s))^2}{n^2\varepsilon^2}.
\end{align*}
Because the sup-risk is always greater than the Bayes risk, we have 
\begin{align*}
	\sup_{{\bm B} \in \Theta}\E_{\bm Y, \bm X|{\bm \beta}} \|M(\bm Y, \bm X) - {\bm \beta}\|^2_2 \gtrsim \sigma_e^2\frac{(s\log(p/s))^2}{n^2\varepsilon^2}.
\end{align*}
The bound is true for any $M$ satisfying $\E_{\bm Y, \bm X|{\bm \beta}}\|M(\bm Y, \bm X) - {\bm \beta}\|^2_2 \lesssim 1$; it extends to all $(\varepsilon, \delta)$-DP estimator $M\in M_{\varepsilon,\delta}$ as we assumed $s\log(p/s) \lesssim n\varepsilon$ and therefore $(s\log(p/s))^2/(n\varepsilon)^2 \lesssim 1$. The proof is complete by combining with the non-private minimax lower bound 
$$\inf_M\sup_{{\bm \beta} \in \R^d, \|{\bm \beta}\|_0 \leq s} \E\|M(\bm Y, \bm X) - {\bm \beta}\|_2^2 \gtrsim  \sigma_e^2s\log(p/s)/n.$$

As a result, we have $$
\inf_{\hat B\in\mathcal M_{\varepsilon,\delta}}\sup_{P\in\mathcal P}\E L(\boldsymbol{\hat{B}},\B)\gtrsim  \frac{s\log p}{\lambda^2n}+\frac{s^2(\log p)^2 }{\lambda^2n^2\epsilon^2}.
$$
\subsubsection{Proof of Lemma~\ref{prop: high-dim glm attack completeness}}

For the linear model, we have the density function
\begin{equation*}
	f_{\bm \beta}(Y\mid\bm x)=\frac{1}{\sqrt{2\pi\sigma^2}}\exp\bigg(-\frac{(Y-\bm x^{\top}\bm \beta)^2}{2\sigma^2}\bigg);\bm x\sim f_{\bm x},
\end{equation*}
where $f_{\bm x}$ is multivariate normal distribution with mean $\bm 0$ and covariance matrix $\bm \Sigma$. It remains to bound $\mathbb{E}_{\bm \pi}\{\sum_{i=1}^{n}\mathbb{E}_{\bm Y,\bm X|\bm \beta}(A_i)\}$ from below. By the fact that
\begin{equation*}
	\mathbb{E}_{\bm Y,\bm X\mid\bm \beta}\langle\bm \beta,(Y_i-\bm x_i^{\top}\bm \beta)\bm x_{i}\rangle=0,
\end{equation*}
we have
\begin{equation*}
	\mathbb{E}_{\bm \pi}\sum_{i=1}^{n}\mathbb{E}_{\bm Y,\bm X\mid\bm \beta}A_i=\mathbb{E}_{\bm \pi}\sum_{i=1}^{n}\mathbb{E}_{\bm Y,\bm X\mid\bm \beta}\langle M(\bm y,\bm X)_{S(M(\bm Y,\bm X))},(Y_i-\bm x_i^{\top}\bm \beta)(\bm x_{i})_{S(\bm \beta)}\rangle.
\end{equation*}
Because the likelihood function $f_{\bm \beta}(y\mid\bm x)$ satisfies
\begin{equation*}
	\frac{\partial}{\partial\beta_j}\log(f_{\bm \beta}(Y\mid\bm x)):=\frac{1}{\sigma^2}(Y-\bm x^{\top}\bm \beta)x_j.
\end{equation*}
We have
\begin{equation*}
	\begin{split}
		\mathbb{E}_{\bm \pi}\sum_{i=1}^{n}\mathbb{E}_{\bm Y,\bm X\mid\bm \beta}A_i&=\mathbb{E}_{\bm \pi}\sum_{i=1}^{n}\mathbb{E}_{\bm Y,\bm X\mid\bm \beta}\sum_{j\in S(\bm \beta)}M_j(\bm Y,\bm X)_{S(M(\bm Y,\bm X))}\times\frac{\partial}{\partial\beta_j}\log f_{\bm \beta}(Y_i\mid\bm x_i)\sigma^2\\
		&=\mathbb{E}_{\pi}\mathbb{E}_{\bm Y,\boldsymbol{X}\mid\boldsymbol{\beta}}\sum_{j\in S(\bm \beta)}M_j(\bm Y,\bm X)_{S(M(\bm Y,\bm X))}\times\sum_{i=1}^{n}\frac{\partial}{\partial\beta_j}\log f_{\bm \beta}(Y_i\mid\bm x_i)\sigma^2,
	\end{split}
\end{equation*}
where $M_j(\bm Y,\bm X)$ is the $j$-th coefficient of $M(\bm Y,\bm X)$. Then, we have
\begin{equation*}
	\begin{split}
		&\mathbb{E}_{\bm Y,\bm X\mid\bm \beta}\sum_{j\in S(\bm \beta)}M_j(\bm Y,\bm X)_{S(M(\bm Y,\bm X))}\times\sum_{i=1}^{n}\frac{\partial}{\partial\beta_j}\log f_{\bm \beta}(Y_i\mid\bm x_i)\\
		=&\mathbb{E}_{\bm Y,\bm X\mid\bm \beta}\sum_{j\in S(\bm beta)}M_j(\bm Y,\bm X)_{S(M(\bm Y,\bm X))}\times\frac{\partial}{\partial\beta_j}(\log \Pi_{i=1}^{n}f_{\bm \beta}(Y_i\mid\bm x_i))\\
		=&\int M_j(\bm Y,\bm X)_{S(M(\bm Y,\bm X))}\cdot\frac{\partial}{\partial\beta_j}(\log \Pi_{i=1}^{n}f_{\bm \beta}(Y_i\mid\bm x_i))\cdot\Pi_{i=1}^{n}\{f_{\bm \beta}(Y_i\mid\bm x_i)\cdot f(\bm x_i)d\bm x_idY_i\}\\
		=&\int M_j(\bm Y,\bm X)_{S(M(\bm Y,\bm X))}\cdot\frac{\partial}{\partial\beta_j}(\Pi_{i=1}^{n}f_{\bm \beta}(Y_i\mid\bm x_i))\cdot\Pi_{i=1}^{n}\{f(\bm x_i)d\bm x_idY_i\}\\
		=&\frac{\partial}{\partial\beta_j}\int M_j(\bm Y,\bm X)_{S(M(\bm Y,\bm X))}\cdot\Pi_{i=1}^{n}f_{\bm \beta}(Y_i\mid\bm x_i)\cdot\Pi_{i=1}^{n}\{f(\bm x_i)d\bm x_idY_i\}\\
		=&\frac{\partial}{\partial\beta_j}\mathbb{E}_{\bm Y,\bm X\mid\bm \beta}M_j(\bm Y,\bm X)_{S(M(\bm Y,\bm X))}
	\end{split}
\end{equation*}
where we use the fact that the function $M_j(\bm Y,\bm X)_{S(M(\bm Y,\bm X))}\cdot\Pi_{i=1}^{n}f(\bm x_i)$ doesn't contain $\beta_j$ explicitly. We define the functions $\bm g(\bm \beta):=\mathbb{E}_{\bm Y,\bm X\mid\bm \beta}\{M(\bm Y,\bm X)_{S(M(\bm Y,\bm X))}\}$. Thus,
\begin{equation*}
	\mathbb{E}_{\bm \pi}\sum_{i=1}^{n}\mathbb{E}_{\bm y,\bm X\mid\bm \beta}A_i=\mathbb{E}_{\bm \pi}\sum_{j\in S(\bm \beta)}\frac{\partial}{\partial\beta_j}g(\bm \beta)_j.
\end{equation*}

Then, we consider the prior distribution $\bm  \pi$. The prior distribution of $\bm \beta$ is defined as follows: let $\{\beta_1^0,\dots,\beta_p^0\}$ follow i.i.d. normal distribution with mean $0$ and variance $\gamma^2$; let $S$ be the index set of top $s$ of $\{\beta_1^0,\dots,\beta_p^0\}$; let $\bm \beta=(\beta_1^0\times 1\{1\in S\},\dots,\beta_p^0\times 1\{p\in S\})$. Notice that the prior distribution $\bm \pi$ is not absolutely continuous with respect to the Lebesgue measure on $\mathbb{R}^p$, thus the Stein Lemma can not be directly used. We use the following decomposition of the prior distribution. Consider all possible combinations of the active sets of $\bm \beta$, and let $S_k$ for $k=\{1,\dots,\binom{p}{s}\}$ be the mutually different index sets of choosing $s$ different elements from $\{1,\dots,p\}$. We define the distribution $p_{S_k}(\beta_1,\dots,\beta_p)=p_{p,(s)}(\bm \beta_{S_k})\times \mathbbm{1}(\bm \beta_{[p]/S_k}=\boldsymbol{0})$, where $p_{p,(s)}$ is the p.d.f. of top $s$ elements of $p$ normal distributions and all other variables are zero. Thus, the prior distribution can be written as $p_{\pi}:=\sum_{k=1}^{\binom{d}{s}}p_{S_k}/\binom{d}{s}$. Then, we have
\begin{equation*}
	\mathbb{E}_{\bm \pi}\sum_{i=1}^{n}\mathbb{E}_{\bm Y,\bm X\mid\bm \beta}A_i=\sum_{k=1}^{\binom{p}{s}}\mathbb{E}_{p_{S_k}}\sum_{j\in S_k}\partial_j g_j(\bm \beta)/\binom{p}{s}.
\end{equation*}
Because of the permutation, it is enough to consider $\mathbb{E}_{p_{S_k}}\sum_{j\in S_k}\partial_j g_j(\bm \beta)$. Notice that the support of $\bm g(\bm \beta)$ is the same as $\bm \beta$ and thus the distribution of $\bm g(\bm \beta)$ given $S_k$ is absolutely continuous with respect to $p_{S_k}$. By Lemma 2.1 in \cite{cai2023score}, we have
\begin{equation}
	\label{eq:lower}
	\begin{split}
		\mathbb{E}_{\bm \pi}\sum_{i=1}^{n}\mathbb{E}_{\bm Y,\bm X\mid\bm \beta}A_i\geq&\mathbb{E}_{p_{S_k}}\bigg[\sum_{j\in S_k}-\beta_j\frac{p_{S_k,j}^{\prime}(\bm \beta)}{p_{S_k,j}(\bm \beta)}\bigg]-\mathbb{E}_{p_{S_k}}\bigg[\sum_{j\in S_k}|g(\boldsymbol{\beta})_j-\beta_j||\frac{p_{S_k,j}^{\prime}(\bm \beta)}{p_{S_k,j}(\bm \beta)}|\bigg].
	\end{split}
\end{equation}
It remains to lower bound the expectation terms. We use $\pi_{p,k}(x)$ to denote the density function of the $k$-th largest entry of $p$ $i.i.d.$ normal variables with mean $0$ and variance $1$. The $k$-th largest entry has density $$
\pi_{p,k}(x)=\frac{p!}{(p-k)!(k-1)!}\phi(x)\Phi(x)^{p-k}(1-\Phi(x))^{k-1}.
$$
As a result 
\begin{align*}
	\pi^{\prime}_{p,k}(x)=&-\frac{p!}{(p-k)!(k-1)!}x\phi(x)\Phi(x)^{p-k}(1-\Phi(x))^{k-1}\\
	&+(p-k)\frac{p!}{(p-k)!(k-1)!}\phi^2(x)\Phi(x)^{p-k-1}(1-\Phi(x))^{k-1}\\
	&-(k-1)\frac{p!}{(p-k)!(k-1)!}\phi^2(x)\Phi(x)^{p-k}(1-\Phi(x))^{k-2}.
\end{align*}
$$
\frac{\pi^{\prime}_{p,k}(x)}{\pi_{p,k}(x)}=-x+(p-k)\frac{\phi(x)}{\Phi(x)}-(k-1)\frac{\phi(x)}{1-\Phi(x)}
$$
We have 
\begin{equation*}
	\begin{split}
		&\mathbb{E}_{p_{S_k}}\bigg[\sum_{j\in S_k}-\beta_j\frac{p_{S_k,j}^{\prime}(\bm\beta)}{p_{S_k,j}(\bm\beta)}\bigg]\\
		\geq&\sum_{k=1}^s\E_{\pi_{p,k}}[\beta_k^2]-\bigg|\sum_{k=1}^s\E_{\pi_{p,k}}[x\cdot\{(p-k)\frac{\phi(x)}{\Phi(x)}-(k-1)\frac{\phi(x)}{1-\Phi(x)}\}]\bigg|
	\end{split}
\end{equation*}
By the Proof of Proposition 5.2 in \cite{cai2023score}, we have $\sum_{k=1}^s\E_{\pi_{p,k}}[\beta_k^2]\gtrsim s\log(p)$. Then, it remains to show
\begin{equation*}
	\mathbb{E}_{p_{S_k}}\bigg[\sum_{j\in S_k}-\beta_j\frac{p_{S_k,j}^{\prime}(\bm\beta)}{p_{S_k,j}(\bm\beta)}\bigg]\gtrsim s\log(p)-\sqrt{s}\log(p),
\end{equation*}
where the second term is based on the results below. It remains to show
$$
\sum_{k=1}^s\E_{\pi_{p,k}}\big[x\cdot\{(d-k)\frac{\phi(x)}{\Phi(x)}-(k-1)\frac{\phi(x)}{1-\Phi(x)}\}\big]\lesssim\sqrt{s}\log(p).
$$
As the ordered density satisfies
$$
(p-k)\frac{\phi(x)}{\Phi(x)}\pi_{p,k}(x)=p\cdot\frac{(p-1)!}{(p-k-1)!(k-1)!}\phi^2(x)\Phi(x)^{p-k-1}(1-\Phi(x))^{k-1}=p\cdot\phi(x)\cdot\pi_{p-1,k}(x).
$$
	We have $$
	\E_{\pi_{p,k}}\left[x\cdot[(p-k)\frac{\phi(x)}{\Phi(x)}]\right]=p\cdot\E_{\pi_{p-1,k}}[x\cdot\phi(x)]
	$$
	Similarly, we have 
	\begin{align*}
		(k-1)\frac{\phi(x)}{1-\Phi(x)}\pi_{p,k}(x)=p\cdot\phi(x)\cdot\pi_{p-1,k-1}(x),
	\end{align*}
	and $$
	\E_{\pi_{p,k}}\left[x\cdot[(k-1)\frac{\phi(x)}{1-\Phi(x)}]\right]=p\cdot\E_{\pi_{p-1,k-1}}[x\cdot\phi(x)]
	$$
	Then 
	\begin{align*}
		&\sum_{k=1}^s\E_{\pi_{p,k}}\left[x\cdot[(p-k)\frac{\phi(x)}{\Phi(x)}-(k-1)\frac{\phi(x)}{1-\Phi(x)}]\right]\\
		=&p\cdot\sum_{k=1}^s \left[\E_{\pi_{p-1,k}}[x\cdot \phi(x)]-\E_{\pi_{p-1,k-1}}[x\cdot \phi(x)]\right]\\
		&=p\cdot\E_{\pi_{p-1,s}}[x\cdot\phi(x)].
	\end{align*}
	We consider the expectation in more detail. 
	Notice that
	\begin{equation*}
		\begin{split}
			p\cdot\E_{\pi_{p-1,s}^{s}}[x\cdot\phi(x)]&=\int_{-\infty}^{\infty}\frac{p!}{(p-1-s)!(s-1)!}\phi(x)\Phi(x)^{p-1-s}(1-\Phi(x))^{s-1}x\phi(x)dx\\
			&=\int_{-\infty}^{0}\frac{p!}{(p-1-s)!(s-1)!}\phi(x)\Phi(x)^{p-1-s}(1-\Phi(x))^{s-1}x\phi(x)dx\\
			&+\int_{0}^{\infty}\frac{p!}{(p-1-s)!(s-1)!}\phi(x)\Phi(x)^{p-1-s}(1-\Phi(x))^{s-1}x\phi(x)dx.
		\end{split}
	\end{equation*}
	The first part satisfies
	\begin{equation*}
		\begin{split}
			&|\int_{-\infty}^{0}\frac{p!}{(p-1-s)!(s-1)!}\phi(x)\Phi(x)^{p-1-s}(1-\Phi(x))^{s-1}x\phi(x)dx|\\
			\leq&\max_x|x\phi(x)|\int_{-\infty}^{0}p^{s+1}(1/2)^{p-1-s}\phi(x)dx\stackrel{p\to\infty,s=o(p/\log(p))}{\longrightarrow}0.
		\end{split}
	\end{equation*}
	The second part satisfies
	\begin{equation*}
		\begin{split}
			&|\int_{0}^{\infty}\frac{p!}{(p-1-s)!(s-1)!}\phi(x)\Phi(x)^{p-1-s}(1-\Phi(x))^{s-1}x\phi(x)dx|\\
			\leq&\int_{0}^{\infty}p^{s+1}\frac{1}{\sqrt{2\pi(s-1)}(\frac{s-1}{e})^{s-1}}\phi(x)\Phi(x)^{p-1-s}(1-\Phi(x))^{s-1}x\phi(x)dx\\
			\lesssim&s^{1.5}\int_{0}^{\infty}(\frac{pe}{s})^{s+1}\phi(x)\Phi(x)^{p-1-s}(1-\Phi(x))^{s-1}x\phi(x)dx,
		\end{split}
	\end{equation*}
	where we use Stirling's approximation in the first inequality. Let $t\asymp\sqrt{\log(ep/s)}\to\infty$ such that $\phi(t)/t\asymp(ep/s)^{-1}$. By the Mill's ratio, we have 
	$$(1-\Phi(t))\asymp\phi(t)/t\asymp(ep/s)^{-1}.$$
	Furthermore,
	\begin{equation*}
		\begin{split}
			&s^{1.5}\int_{t}^{\infty}(\frac{pe}{s})^{s+1}\phi(x)\Phi(x)^{p-1-s}(1-\Phi(x))^{s-1}x\phi(x)dx\\
			\leq&s^{1.5}\int_{t}^{\infty}(\frac{pe}{s})^{s+1}\phi(t)^2(1-\Phi(x))^{s-2}d\Phi(x)\\
			\lesssim& \sqrt{s}\log(ep/s)\int_{\infty}^{t}(\frac{pe}{s})^{s-1}d(1-\Phi(x))^{s-1}\\
			=&\sqrt{s}\log(ep/s)(\frac{pe}{s})^{s-1}(1-\Phi(t))^{s-1}\asymp\sqrt{s}\log(ep/s),
		\end{split}
	\end{equation*}
	where we use Mills's ratio and the fact that $\phi(x)$ is monotone decreasing at $[t,\infty)$ in the first inequality. Then, by the additive property of the integral, we have
	\begin{equation*}
		\begin{split}
			&s^{1.5}\int_{0}^{\infty}(\frac{pe}{s})^{s+1}\phi(x)\Phi(x)^{p-1-s}(1-\Phi(x))^{s-1}x\phi(x)dx\\
			=&s^{1.5}\int_{0}^{t}(\frac{pe}{s})^{s+1}\phi(x)\Phi(x)^{p-1-s}(1-\Phi(x))^{s-1}x\phi(x)dx\\
			&+s^{1.5}\int_{t}^{\infty}(\frac{de}{s})^{s+1}\phi(x)\Phi(x)^{p-1-s}(1-\Phi(x))^{s-1}x\phi(x)dx
		\end{split}
	\end{equation*}
	By Mills's ratio, $x\phi(x)\leq(x^2+1)(1-\Phi(x))$, the remaining part satisfies
	\begin{equation*}
		\begin{split}
			&s^{1.5}\int_{0}^{t}(\frac{pe}{s})^{s+1}\phi(x)\Phi(x)^{p-1-s}(1-\Phi(x))^{s-1}x\phi(x)dx\\
			\leq&s^{1.5}\int_{0}^{t}x^2(\frac{pe}{s})^{s+1}\Phi(x)^{p-1-s}(1-\Phi(x))^{s}\phi(x)dx\\
			\lesssim&s^{1.5}\log(ep/s)(\frac{pe}{s})^{s+1}\int_{0}^{t}\Phi(x)^{p-1-s}(1-\Phi(x))^{s}d\Phi(x).
		\end{split}
	\end{equation*}
	By integral by parts, we have
	\begin{equation*}
		\begin{split}
			&\int_{0}^{t}\Phi(x)^{p-1-s}(1-\Phi(x))^{s}d\Phi(x)=\frac{1}{p-s}\int_{0}^{t}(1-\Phi(x))^{s}d\Phi(x)^{p-s}\\
			=&\frac{1}{p-s}\big\{\Phi(t)^{p-s}(1-\Phi(t))^{s}-\Phi(0)^{p-s}(1-\Phi(0))^{s}\big\}+\frac{s-1}{p-s}\int_{0}^{t}(1-\Phi(x))^{s-1}\Phi(x)^{p-s}d\Phi(x)\\
			\lesssim &\frac{1}{p-s}\Phi(t)^{p-s}(1-\Phi(t))^{s}+\frac{s-1}{p-s}\int_{0}^{t}(1-\Phi(x))^{s-1}\Phi(x)^{p-s}d\Phi(x).
		\end{split}
	\end{equation*}
	By mathematical induction, we have
	\begin{equation*}
		\begin{split}
			&\int_{0}^{t}\Phi(x)^{p-1-s}(1-\Phi(x))^{s}d\Phi(x)\\
			=&\sum_{k=0}^{s-1}\frac{\Pi_{l=0}^{k-1}(s+1-l)}{\Pi_{l=0}^{k}(p-s+l)}\Phi(t)^{p-s+k}(1-\Phi(t))^{s-k}+\frac{s!}{p\times(p-1)\times(p-s)}\int_{0}^{t}\Phi(x)^{p-1}d\Phi(x).
		\end{split}
	\end{equation*}
	The first term satisfies
	\begin{equation*}
		\begin{split}
			&s^{1.5}\log(ep/s)(\frac{pe}{s})^{s+1}\sum_{k=0}^{s-1}\frac{\Pi_{l=0}^{k-1}(s+1-l)}{\Pi_{l=0}^{k}(p-s+l)}\Phi(t)^{p-s+k}(1-\Phi(t))^{s-k}\\
			\lesssim&\sqrt{s}\log(ep/s)\sum_{k=0}^{s-1}\Phi(t)^{p-s}\lesssim\sqrt{s}\log(ep/s)s\Phi(t)^{p-s}.
		\end{split}
	\end{equation*}
	Because
	\begin{equation*}
		s\Phi(t)^{p-s}=s\{1-(1-\Phi(t))\}^{p-s}\asymp s\exp\{-(p-s)(1-\Phi(t))=s\exp(-s/e)\lesssim 1,
	\end{equation*}
	we have the first term is $O(\sqrt{s}\log(ep/s))$. It remains to consider the second term. Note that
	\begin{equation*}
		\begin{split}
			&s^{1.5}\log(ep/s)(\frac{pe}{s})^{s+1}\frac{s!}{(p-1)\times(p-s)}\int_{0}^{t}\Phi(x)^{p-1}d\Phi(x)\\
			\lesssim&s^{1.5}\log(ep/s)(\frac{pe}{s})\int_{0}^{t}\Phi(x)^{p-1}d\Phi(x)\lesssim\sqrt{s}\log(ep/s)\int_0^{t}d\Phi(x)^p.
		\end{split}
	\end{equation*}
	Further notice that
	\begin{equation*}
		\Phi(t)^{p}=\{1-(1-\Phi(t))\}^{p}\asymp\exp\{-p(1-\Phi(t))=\exp(-s/e)\lesssim 1.
	\end{equation*}
	Combining all the results, we have
	$$
	\sum_{k=1}^s\E_{\pi_{p,k}^{s}}\left[x\cdot[(p-k)\frac{\phi(x)}{\Phi(x)}-(k-1)\frac{\phi(x)}{1-\Phi(x)}]\right]\lesssim\sqrt{s}\log(p/s).
	$$
	
	Then, it remains to consider the second term of \eqref{eq:lower}:
	\begin{equation*}
		\begin{split}
			&\mathbb{E}_{p_{S_k}}\bigg[\sum_{j\in S_k}|g(\boldsymbol{\beta})_j-\beta_j||\frac{p_{S_k,j}^{\prime}(\bm\beta)}{p_{S_k,j}(\bm\beta)}|\bigg]\\
			\leq&\mathbb{E}_{p_{S_k}}\bigg[\sum_{j\in S_k}|g(\boldsymbol{\beta})_j-\beta_j|\cdot|\beta_j|\bigg]+\mathbb{E}_{p_{S_k}}\bigg[\sum_{j\in S_k}|g(\boldsymbol{\beta})_j-\beta_j|\cdot|\beta_j-\frac{p_{S_k,j}^{\prime}(\bm\beta)}{p_{S_k,j}(\bm\beta)}|\bigg]\\
			\leq&\sqrt{\mathbb{E}_{\boldsymbol{X}\mid\boldsymbol{\beta}}\|M(\bm Y,\bm X)-\boldsymbol{\beta}\|_2^2}\cdot\sqrt{\sum_{k=1}^s\E_{\pi_{p,k}}[\beta_k^2]}\\
			&+\mathbb{E}_{p_{S_k}}\bigg[\sum_{j\in S_k}|g(\boldsymbol{\beta})_j-\beta_j|\cdot|\beta_j-\frac{p_{S_k,j}^{\prime}(\bm\beta)}{p_{S_k,j}(\bm\beta)}|\bigg],
		\end{split}
	\end{equation*}
	where we use the triangle inequality in the first inequlity, and we use the Cauchy–Schwarz inequality in the second inequality. The first term is $o(\sqrt{s\log(p)})$ by the assumption $\mathbb{E}_{\bm Y,\boldsymbol{X}\mid\boldsymbol{\beta}}\|M(\bm Y,\bm X)-\boldsymbol{\beta}\|_2^2=o(1)$. It remains to consider the second term. Then, we have 
	\begin{equation*}
		\begin{split}
			&\E_{\pi_{p,j}}\bigg\{\beta_j/\gamma^2-\frac{p_{S_k,j}^{\prime}(\bm\beta)}{p_{S_k,j}(\bm\beta)}\bigg\}^2=\E_{\pi_{p,j}}\bigg\{(p-j)\frac{\phi(x)}{\Phi(x)}-(j-1)\frac{\phi(x)}{1-\Phi(x)}\bigg\}^2\\
			\leq&\E_{\pi_{p,j}}2\bigg\{(p-j)\frac{\phi(x)}{\Phi(x)}\bigg\}+\E_{\pi_{p,j}}\bigg\{(j-1)\frac{\phi(x)}{1-\Phi(x)}\bigg\}\\
			=&2p(p-j)\E_{\pi_{p-1,j}}\bigg\{\frac{\phi(x)^2}{\Phi(x)}\bigg\}+2p(j-1)\E_{\pi_{p-1,j-1}}\bigg\{\frac{\phi(x)^2}{1-\Phi(x)}\bigg\},
		\end{split}
	\end{equation*}
	where we use the inequality $(a+b)^2\leq 2a^2+2b^2$ in the second inequality. We consider the two parts separately. The first term
	\begin{equation*}
		\begin{split}
			p(p-j)\E_{\pi_{p-1,j}}\bigg\{\frac{\phi(x)^2}{\Phi(x)}\bigg\}&=(p-j)\int_{-\infty}^{\infty}\frac{p!}{(p-1-j)!(j-1)!}\phi(x)\Phi(x)^{p-2-j}(1-\Phi(x))^{j-1}\phi(x)^2dx\\
			&=(p-j)\int_{-\infty}^{0}\frac{p!}{(p-1-j)!(j-1)!}\phi(x)\Phi(x)^{p-2-j}(1-\Phi(x))^{j-1}\phi(x)^2dx\\
			&+(p-j)\int_{0}^{\infty}\frac{p!}{(p-1-j)!(j-1)!}\phi(x)\Phi(x)^{p-2-j}(1-\Phi(x))^{j-1}\phi(x)^2dx.
		\end{split}
	\end{equation*}
	The first part satisfies
	\begin{equation*}
		\begin{split}
			&(p-j)|\int_{-\infty}^{0}\frac{p!}{(p-1-j)!(j-1)!}\phi(x)\Phi(x)^{p-2-j}(1-\Phi(x))^{j-1}\phi(x)^2dx|\\
			\lesssim&\int_{-\infty}^{0}p^{j+2}(1/2)^{p-2-j}\phi(x)dx\stackrel{p\to\infty,j=o(p/\log(p))}{\longrightarrow}0.
		\end{split}
	\end{equation*}
	The second part satisfies
	\begin{equation*}
		\begin{split}
			&(p-j)|\int_{0}^{\infty}\frac{p!}{(p-1-j)!(j-1)!}\phi(x)\Phi(x)^{p-2-j}(1-\Phi(x))^{j-1}\phi(x)^2dx|\\
			\leq&\int_{0}^{\infty}p^{j+2}\frac{1}{\sqrt{2\pi(j-1)}(\frac{j-1}{e})^{j-1}}\phi(x)\Phi(x)^{p-2-j}(1-\Phi(x))^{j-1}\phi(x)^2dx\\
			\lesssim&j^{2.5}\int_{0}^{\infty}(\frac{pe}{j})^{j+2}\phi(x)\Phi(x)^{p-2-j}(1-\Phi(x))^{j-1}\phi(x)^2dx,
		\end{split}
	\end{equation*}
	where we use Stirling's approximation in the first inequality. Let $t\asymp\sqrt{\log(ep/j^{m})}\to\infty$, where $m=(j+1)/(j+2)$, such that $\phi(t)/t\asymp(ep/j^m)^{-1}$. By Mill's ratio, we have 
	$$(1-\Phi(t))\asymp\phi(t)/t\asymp(ep/j^m)^{-1}.$$
	Furthermore,
	\begin{equation*}
		\begin{split}
			&j^{2.5}\int_{t}^{\infty}(\frac{pe}{j})^{j+2}\phi(x)\Phi(x)^{p-2-j}(1-\Phi(x))^{j-1}\phi(x)^2dx\\
			\leq&j^{2.5}\int_{t}^{\infty}(\frac{pe}{j})^{j+2}\phi(t)^2(1-\Phi(x))^{j-1}d\Phi(x)\\
			\lesssim& j^{2.5}\log(ep/j^m)\int_{t}^{\infty}(\frac{pe}{j})^{j+2}(1-\Phi(x))^{j-1}(1-\Phi(t))^{2}d\Phi(x),
		\end{split}
	\end{equation*}
	where we use the fact that $\phi(x)$ is monotone decreasing at $[t,\infty)$ and $\Phi(x)\leq 1$ in the first inequality, and we use Mill's ratio in the second inequality. Then, we have
	\begin{equation*}
		\begin{split}
			&j^{1.5}\log(ep/j^m)\int_{\infty}^{t}(\frac{pe}{j})^{j+2}(1-\Phi(t))^{2}d(1-\Phi(x))^{j}\\
			\lesssim& j^{1.5}\log(ep/j^m)(\frac{pe}{j})^{j+2}(1-\Phi(t))^{j+2}\asymp j^{1.5}\log(ep/j^m)\frac{j^{m(j+2)}}{j^{j+2}}\\
			=&j^{0.5}\log(ep/j^m).
		\end{split}
	\end{equation*}
	
	Then, by the additive property of the integral, we have
	\begin{equation*}
		\begin{split}
			&j^{2.5}\int_{0}^{\infty}(\frac{pe}{j})^{j+2}\phi(x)\Phi(x)^{p-j-2}(1-\Phi(x))^{j-1}\phi(x)^2dx\\
			=&j^{2.5}\int_{0}^{t}(\frac{pe}{j})^{j+2}\phi(x)\Phi(x)^{p-j-2}(1-\Phi(x))^{j-1}\phi(x)^2dx\\
			&+j^{2.5}\int_{t}^{\infty}(\frac{pe}{j})^{j+2}\phi(x)\Phi(x)^{p-j-2}(1-\Phi(x))^{j-1}\phi(x)^2dx
		\end{split}
	\end{equation*}
	By Mills's ratio, $x\phi(x)\leq(x^2+1)(1-\Phi(x))$, the remaining part satisfies
	\begin{equation*}
		\begin{split}
			&j^{2.5}\int_{0}^{t}(\frac{pe}{j})^{j+2}\phi(x)\Phi(x)^{p-2-j}(1-\Phi(x))^{j-1}\phi(x)^2dx\\
			\leq&j^{2.5}\int_{0}^{t}x^4(\frac{pe}{j})^{j+2}\Phi(x)^{p-2-s}(1-\Phi(x))^{j+1}\phi(x)dx\\
			\lesssim&j^{2.5}\log^2(ep/j^m)(\frac{pe}{j})^{j+2}\int_{0}^{t}\Phi(x)^{p-2-j}(1-\Phi(x))^{j+1}d\Phi(x).
		\end{split}
	\end{equation*}
	Using integral by parts, we have
	\begin{equation*}
		\begin{split}
			&\int_{0}^{t}\Phi(x)^{p-2-j}(1-\Phi(x))^{j+1}d\Phi(x)=\frac{1}{p-j-1}\int_{0}^{t}(1-\Phi(x))^{j+1}d\Phi(x)^{p-j-1}\\
			=&\frac{1}{p-j-1}\big\{\Phi(t)^{p-j-1}(1-\Phi(t))^{j+1}-\Phi(0)^{p-j-1}(1-\Phi(0))^{j+1}\big\}\\
			&+\frac{j+1}{p-j-1}\int_{0}^{t}(1-\Phi(x))^{j}\Phi(x)^{p-j-1}d\Phi(x)\\
			\lesssim &\frac{1}{p-j-1}\Phi(t)^{p-j-1}(1-\Phi(t))^{j+1}+\frac{j+1}{p-j-1}\int_{0}^{t}(1-\Phi(x))^{j}\Phi(x)^{p-j-1}d\Phi(x).
		\end{split}
	\end{equation*}
	By mathematical induction, we have
	\begin{equation*}
		\begin{split}
			&\int_{0}^{t}\Phi(x)^{p-2-j}(1-\Phi(x))^{j+1}d\Phi(x)\\
			=&\sum_{k=1}^{j+1}\frac{\Pi_{l=1}^{k-1}(j+2-l)}{\Pi_{l=1}^{k}(p-j-2+l)}\Phi(t)^{p-j-2+k}(1-\Phi(t))^{j-k+2}\\
			&+\frac{(j+1)!}{(p-1)\times\cdot\times(p-j-1)}\int_{0}^{t}\Phi(x)^{p-1}d\Phi(x).
		\end{split}
	\end{equation*}
	The first term satisfies
	\begin{equation*}
		\begin{split}
			&j^{2.5}\log^2(ep/j)(\frac{pe}{j})^{j+2}\sum_{k=1}^{j+1}\frac{\Pi_{l=1}^{k-1}(j+2-l)}{\Pi_{l=1}^{k}(p-j-2+l)}\Phi(t)^{p-j-2+k}(1-\Phi(t))^{j-k+2}\\
			\lesssim&j^{1.5}\log^2(ep/j^m)\sum_{k=1}^{j+1}\Phi(t)^{p-j-2}\lesssim\log^2(ep/j^m)j^{2.5}\Phi(t)^{p-j-2}.
		\end{split}
	\end{equation*}
	Notice that
	\begin{equation*}
		j^{1.5}\Phi(t)^{p-j-2}=j^{1.5}\{1-(1-\Phi(t))\}^{p-j-2}\asymp j^{1.5}\exp\{-(p-j-2)(1-\Phi(t))=j^{1.5}\exp(-j^m/e).
	\end{equation*}
	As $j\to\infty$, $j^{1.5}\exp(-j^m/e)\asymp j^{1.5}\exp(-j/e)\lesssim 1$, we have that the first term is $O(\sqrt{j}\log^2(ep/j^{m}))$. It remains to consider the second term. Note that
	\begin{equation*}
		\begin{split}
			&j^{2.5}\log^2(ep/j^m)(\frac{pe}{j})^{j+2}\frac{(j+1)!}{(p-1)\times(p-j-1)}\int_{0}^{t}\Phi(x)^{p-1}d\Phi(x)\\
			\lesssim&j^{2.5}\log^2(ep/j^m)(\frac{pe}{j})\int_{0}^{t}\Phi(x)^{p-1}d\Phi(x)\lesssim j^{1.5}\log^2(ep/j^{m})\int_0^{t}d\Phi(x)^p.
		\end{split}
	\end{equation*}
	Moreover,
	\begin{equation*}
		j\Phi(t)^{p}=j\{1-(1-\Phi(t))\}^{p}\asymp j\exp(-j^m/e)\lesssim 1.
	\end{equation*}
	Combining all the results, we have
	$$
	p(p-j)\E_{\pi_{p-1,j}^{s}}\bigg\{\frac{\phi(x)^2}{\Phi(x)}\bigg\}/\gamma^2 \lesssim\sqrt{j}\log^2(p/j^m)\asymp\sqrt{j}\log^2(p/j).
	$$
	
	For the second term, to simplify the notation, we consider $j$ instead of $j-1$:
	\begin{equation*}
		\begin{split}
			&2pj\E_{\pi_{p-1,j}^{s}}\bigg\{\frac{\phi(x)^2}{1-\Phi(x)}\bigg\}\\
			&=j\int_{-\infty}^{\infty}\frac{p!}{(p-1-j)!(j-1)!}\phi(x)\Phi(x)^{p-1-j}(1-\Phi(x))^{j-2}\phi(x)^2dx\\
			&=j\int_{-\infty}^{0}\frac{p!}{(p-1-j)!(j-1)!}\phi(x)\Phi(x)^{p-1-j}(1-\Phi(x))^{j-2}\phi(x)^2dx\\
			&+j\int_{0}^{\infty}\frac{p!}{(p-1-j)!(j-1)!}\phi(x)\Phi(x)^{p-1-j}(1-\Phi(x))^{j-2}\phi(x)^2dx.
		\end{split}
	\end{equation*}
	The first part satisfies
	\begin{equation*}
		\begin{split}
			&j\cdot\int_{-\infty}^{0}\frac{p!}{(p-1-j)!(j-1)!}\phi(x)\Phi(x)^{p-1-j}(1-\Phi(x))^{j-2}\phi(x)^2dx\\
			\lesssim& j\cdot\int_{-\infty}^{0}p^{j+2}(1/2)^{p-3}\phi(x)dx\stackrel{p\to\infty,j=o(p/\log(p))}{\longrightarrow}0.
		\end{split}
	\end{equation*}
	The second part satisfies
	\begin{equation*}
		\begin{split}
			&j\cdot \int_{0}^{\infty}\frac{p!}{(p-1-j)!(j-1)!}\phi(x)\Phi(x)^{p-1-j}(1-\Phi(x))^{j-2}\phi(x)^2dx\\
			\leq&j\cdot \int_{0}^{\infty}p^{j+1}\frac{1}{\sqrt{2\pi(j-1)}(\frac{j-1}{e})^{j-1}}\phi(x)\Phi(x)^{p-1-j}(1-\Phi(x))^{j-2}\phi(x)^2dx\\
			\lesssim&j^{2.5}\int_{0}^{\infty}(\frac{pe}{j})^{j+1}\phi(x)\Phi(x)^{p-1-j}(1-\Phi(x))^{j-2}\phi(x)^2dx,
		\end{split}
	\end{equation*}
	where we use Stirling's approximation in the first inequality. Let $t\asymp\sqrt{\log(ep/j^{m})}\to\infty$, where $m=j/(j+1)$, such that $\phi(t)/t\asymp(ep/j^m)^{-1}$. By Mill's ratio, we have 
	$$(1-\Phi(t))\asymp\phi(t)/t\asymp(ep/j^m)^{-1}.$$
	Furthermore,
	\begin{equation*}
		\begin{split}
			&j^{2.5}\int_{t}^{\infty}(\frac{pe}{j})^{j+1}\phi(x)\Phi(x)^{p-1-j}(1-\Phi(x))^{j-2}\phi(x)^2dx\\
			\leq&j^{2.5}\int_{t}^{\infty}(\frac{pe}{j})^{j+1}\phi(t)^2(1-\Phi(x))^{j-2}d\Phi(x)\\
			\lesssim& j^{2.5}\log(ep/j^m)\int_{t}^{\infty}(\frac{pe}{j})^{j+1}(1-\Phi(x))^{j-2}(1-\Phi(t))^{2}d\Phi(x),
		\end{split}
	\end{equation*}
	where we use the fact that $\phi(x)$ is monotone decreasing at $[t,\infty)$ and $\Phi(x)\leq 1$ in the first inequality, and we use Mill's ratio in the second inequality. Then, we have
	\begin{equation*}
		\begin{split}
			&j^{1.5}\log(ep/j^m)\int_{\infty}^{t}(\frac{pe}{j})^{j+1}(1-\Phi(t))^{2}d(1-\Phi(x))^{j-1}\\
			\lesssim& j^{1.5}\log(ep/j^m)(\frac{pe}{j})^{j+1}(1-\Phi(t))^{j+1}\asymp j^{1.5}\log(ep/j^m)\frac{j^{m(j+1)}}{j^{j+1}}\\
			=&j^{0.5}\log(ep/j^m).
		\end{split}
	\end{equation*}
	
	Then, by the additive property of the integral, we have
	\begin{equation*}
		\begin{split}
			&j^{2.5}\int_{0}^{\infty}(\frac{pe}{j})^{j+1}\phi(x)\Phi(x)^{p-j-1}(1-\Phi(x))^{j-2}\phi(x)^2dx\\
			=&j^{2.5}\int_{0}^{t}(\frac{pe}{j})^{j+1}\phi(x)\Phi(x)^{p-j-1}(1-\Phi(x))^{j-2}\phi(x)^2dx\\
			&+j^{2.5}\int_{t}^{\infty}(\frac{pe}{j})^{j+1}\phi(x)\Phi(x)^{p-j-1}(1-\Phi(x))^{j-2}\phi(x)^2dx
		\end{split}
	\end{equation*}
	By Mills's ratio, $x\phi(x)\leq(x^2+1)(1-\Phi(x))$, the remaining part satisfies
	\begin{equation*}
		\begin{split}
			&j^{2.5}\int_{0}^{t}(\frac{pe}{j})^{j+1}\phi(x)\Phi(x)^{p-1-j}(1-\Phi(x))^{j-2}\phi(x)^2dx\\
			\leq&j^{2.5}\int_{0}^{t}x^4(\frac{pe}{j})^{j+1}\Phi(x)^{p-1-s}(1-\Phi(x))^{j}\phi(x)dx\\
			\lesssim&j^{2.5}\log^2(ep/j^m)(\frac{pe}{j})^{j+1}\int_{0}^{t}\Phi(x)^{p-1-j}(1-\Phi(x))^{j}d\Phi(x).
		\end{split}
	\end{equation*}
	By integral by parts, we have
	\begin{equation*}
		\begin{split}
			&\int_{0}^{t}\Phi(x)^{p-1-j}(1-\Phi(x))^{j}d\Phi(x)=\frac{1}{p-1}\int_{0}^{t}(1-\Phi(x))^{j}d\Phi(x)^{p-j}\\
			=&\frac{1}{p-j}\big\{\Phi(t)^{p-j}(1-\Phi(t))^{j}-\Phi(0)^{p-j}(1-\Phi(0))^{j}\big\}\\
			&+\frac{j}{p-j}\int_{0}^{t}(1-\Phi(x))^{j-1}\Phi(x)^{p-j}d\Phi(x)\\
			\lesssim &\frac{1}{p-j}\Phi(t)^{p-j}(1-\Phi(t))^{j}+\frac{j}{p-j}\int_{0}^{t}(1-\Phi(x))^{j-1}\Phi(x)^{p-j}d\Phi(x).
		\end{split}
	\end{equation*}
	By mathematical induction, we have
	\begin{equation*}
		\begin{split}
			&\int_{0}^{t}\Phi(x)^{p-1-j}(1-\Phi(x))^{j-1}d\Phi(x)\\
			=&\sum_{k=1}^{j}\frac{\Pi_{l=1}^{k-1}(j+1-l)}{\Pi_{l=1}^{k}(p-j-1+l)}\Phi(t)^{p-j-1+k}(1-\Phi(t))^{j-k+1}\\
			&+\frac{j!}{(p-1)\times\cdots\times(p-j)}\int_{0}^{t}\Phi(x)^{p-1}d\Phi(x).
		\end{split}
	\end{equation*}
	The first term satisfies
	\begin{equation*}
		\begin{split}
			&j^{2.5}\log^2(ep/j^m)(\frac{pe}{j})^{j+1}\sum_{k=1}^{j}\frac{\Pi_{l=1}^{k-1}(j+1-l)}{\Pi_{l=1}^{k}(p-j-1+l)}\Phi(t)^{p-j-1+k}(1-\Phi(t))^{j-k+1}\\
			\lesssim&j^{1.5}\log^2(ep/j^m)\sum_{k=1}^{j}\Phi(t)^{p-j}\lesssim\log^2(ep/j^m)j^{2.5}\Phi(t)^{p-j}.
		\end{split}
	\end{equation*}
	Note that
	\begin{equation*}
		j^{1.5}\Phi(t)^{p-j}=j^{1.5}\{1-(1-\Phi(t))\}^{p-j}\asymp j^{1.5}\exp\{-(p-j)(1-\Phi(t))=j^{1.5}\exp(-j^m/e).
	\end{equation*}
	By the fact as $j\to\infty$, $j^{1.5}\exp(-j^m/e)\asymp j^{1.5}\exp(-j/e)\lesssim 1$, we have that the first term is $O(\sqrt{j}\log^2(ep/j^{m}))$. It remains to consider the second term. Note that
	\begin{equation*}
		\begin{split}
			&j^{2.5}\log^2(ep/j^m)(\frac{pe}{j})^{j+1}\frac{(j+1)!}{(p-1)\times(p-j-1)}\int_{0}^{t}\Phi(x)^{p-1}d\Phi(x)\\
			\lesssim&j^{2.5}\log^2(ep/j^m)(\frac{pe}{j})\int_{0}^{t}\Phi(x)^{p-1}d\Phi(x)\lesssim j^{1.5}\log^2(ep/j^{m})\int_0^{t}d\Phi(x)^p.
		\end{split}
	\end{equation*}
	Besides,
	\begin{equation*}
		j\Phi(t)^{p}=j\{1-(1-\Phi(t))\}^{p}\asymp j\exp(-j^m/e)\lesssim 1.
	\end{equation*}
	Combining all the results, we have
	$$
	p(p-j)\E_{\pi_{p-1,j}^{s}}\bigg\{\frac{\phi(x)^2}{\Phi(x)}\bigg\}/\gamma^2 \lesssim\sqrt{j}\log^2(p/j^m)\asymp\sqrt{j}\log^2(p/j).
	$$
	Using the Cauchy–Schwarz inequality, we have
	\begin{equation*}
		\begin{split}
			&\mathbb{E}_{p_{S_k}}\bigg[\sum_{j\in S_k}|g(\boldsymbol{\beta})_j-\beta_j|\cdot|\beta_j/\gamma^2-\frac{p_{S_k,j}^{\prime}(\bm\beta)}{p_{S_k,j}(\bm\beta)}|\bigg]\\
			\lesssim&\sqrt{\mathbb{E}_{\bm Y,\boldsymbol{X}\mid\boldsymbol{\beta}}\|M(\bm Y,\bm X)-\boldsymbol{\beta}\|_2^2}\\
			&\times\sqrt{\sum_{k=1}^s\bigg[2p(p-k)\E_{\pi_{p-1,k}^{s}}\bigg\{\frac{\phi(x)^2}{\Phi(x)}\bigg\}/\gamma^2+2p(k-1)\E_{\pi_{p-1,k-1}^{s}}\bigg\{\frac{\phi(x)^2}{1-\Phi(x)}\bigg\}/\gamma^2\bigg]}\\
			\lesssim&\sqrt{\mathbb{E}_{\bm Y,\boldsymbol{X}\mid\boldsymbol{\beta}}\|M(\bm Y,\bm X)-\boldsymbol{\beta}\|_2^2}\cdot \sqrt{\sum_{k=1}^{s}k\log^2(p/k)}\\
			=&o(s\log(p/s)).
		\end{split}
	\end{equation*}
	
	Combine all the results together, we have
	\begin{equation*}
		\sum_{i \in [n]} \E_{\bm \pi}\E_{\bm y, \bm X|{\bm \beta}} A_i \gtrsim s\log(p/s).
	\end{equation*}
	
	\section{Proof of the Upper Bound for DP-SIR when $n>p$}\label{sec: low_dim_proof}
	
	\subsection{Proof of the Estimator in Algorithm \ref{alg:ld} }
	We first show that the proposed estimator in the Algorithm \ref{alg:ld} is $(2\varepsilon,\delta)$-DP.
	
	\begin{proof}[Proof of Lemma \ref{lem:low_privacy}]
		
		For convenience, we define the variable $\widetilde{Y}=\sum_{l=1}^{H}l\cdot\mathbbm{1}\{Y\in \widehat{I}_l\}$, which takes different values among all the $H$ estimated slices $\{\widehat{I}_h\}_{h=1}^{H}$. The estimated slices are defined by the first step of Algorithm \ref{alg:ld}. The sample version $\{\widetilde{Y}_i\}_{i=1}^{n}$ can be defined analogously. The gradient, utilized in the Line 5 of Algorithm \ref{alg:ld} , at the $t$-th step can be written as
		\begin{equation}
			\label{eq:gradient}
			2\eta\{-\boldsymbol{\hat{M}}^{(t)}\boldsymbol{B}+\lambda_{penalty}\boldsymbol{\hat{\Sigma}}^{(t)}\boldsymbol{B}(\boldsymbol{B}^{\top}\boldsymbol{\hat{\Sigma}}^{(t)}\boldsymbol{B}-\boldsymbol{I}_k)\},
		\end{equation}
		where we omit the truncation operators and introduce the notations
		\begin{equation*}
			\boldsymbol{\hat{M}}^{(t)}:=\sum_{h=1}^{H}\mathbb{E}_n\{\mathbbm{1}(\widetilde{Y}_i=h)\mid i\in\mathcal{S}_t\}\mathbb{E}_n(\boldsymbol{x}_i\mid \widetilde{Y}_i=h,i\in\mathcal{S}_t)\mathbb{E}_n(\boldsymbol{x}_i\mid \widetilde{Y}_i=h,i\in\mathcal{S}_t)^{\top},
		\end{equation*}
		\begin{equation*}
			\boldsymbol{\hat{\Sigma}}^{(t)}:=\frac{1}{|\mathcal{S}_t|}\sum_{i\in\mathcal{S}_t}\boldsymbol{x}_i\boldsymbol{x}_i^{\top}.
		\end{equation*}
		The superscript $(t)$ indicates the dependence on the sub-data $\mathcal{S}_t$. Line 6 in Algorithm \ref{alg:ld} is the truncation operator. Thus, it remains to consider the $\ell_2$ sensitivity of the gradient defined in \eqref{eq:gradient}. To simplify the notation, we omit the notation of the sub-data set. We use $\mathcal{S}^{\prime}$ to denote a neighboring data set of $\mathcal{S}=\{\boldsymbol{x}_i,Y_i\}_{i=1}^{n}$. Without loss of generality, we assume the difference occurs in the first sample, i.e., $\mathcal{S}^{\prime}=\{\{\boldsymbol{x}_1^{\prime},Y_1^{\prime}\},\{\boldsymbol{x}_i,Y_i\}_{i=2}^{n}\}$. We consider the two parts of the gradient \eqref{eq:gradient} separately. 
		
		The first part related to the sensitivity of $\boldsymbol{\hat{M}}$ matrix can be written as,
		\begin{equation*}
			\begin{split}
				&\boldsymbol{\hat{M}}(\mathcal{S})-\boldsymbol{\hat{M}}(\mathcal{S}^{\prime})\\
				=&\sum_{h=1}^{H}\{\hat{p}_h\mathbb{E}_n(\boldsymbol{x}_i\mid\tilde{Y}_i=h)\mathbb{E}_n(\boldsymbol{x}_i\mid\tilde{Y}_i=h)^{\top}-\hat{p}_h^{\prime}\mathbb{E}_n(\boldsymbol{x}^{\prime}_i\mid\tilde{Y}_i^{\prime}=h)\mathbb{E}_n(\boldsymbol{x}_i^{\prime}\mid\tilde{Y}_i^{\prime}=h)^{\top}\}.
			\end{split}
		\end{equation*}
		The above term has two possible forms, depending on whether $Y_1$ and $Y_1^{\prime}$ belong to the same slice. If $\widetilde{Y}^{\prime}=1$ and thus $\{\boldsymbol{x}_1^{\prime},Y_1^{\prime}\}$ belongs to the same slice as $\{\boldsymbol{x}_1,Y_1\}$, then we have
		\begin{equation*}
			\begin{split}
				&(\boldsymbol{\hat{M}}(\mathcal{S})-\boldsymbol{\hat{M}}(\mathcal{S}^{\prime}))\boldsymbol{B}\\
				=&\hat{p}_1\bigg[\{\mathbb{E}_n(\boldsymbol{x}^{\prime}\mid\widetilde{Y}^{\prime}=1)\}\{\mathbb{E}_n(\boldsymbol{x}\mid\widetilde{Y}=1)-\mathbb{E}_n(\boldsymbol{x}^{\prime}\mid\widetilde{Y}^{\prime}=1)\}^{\top}\boldsymbol{B}\\
				&+\{\mathbb{E}_n(\boldsymbol{x}\mid\widetilde{Y}=1)-\mathbb{E}_n(\boldsymbol{x}^{\prime}\mid\widetilde{Y}^{\prime}=1)\}\{\mathbb{E}_n(\boldsymbol{x}\mid\widetilde{Y}=1)\}^{\top}\boldsymbol{B}\bigg].
			\end{split}
		\end{equation*}
		Furthermore, we have
		\begin{equation*}
			\begin{split}
				&\|(\boldsymbol{\hat{M}}(\mathcal{S})-\boldsymbol{\hat{M}}(\mathcal{S}^{\prime}))\boldsymbol{B}\|_{\infty}\\
				\leq&\hat{p}_1\bigg[\|\mathbb{E}_n(\boldsymbol{x}^{\prime}\mid\widetilde{Y}^{\prime}=1)\|_{\infty}\|\{\mathbb{E}_n(\boldsymbol{x}\mid\widetilde{Y}=1)-\mathbb{E}_n(\boldsymbol{x}^{\prime}\mid\widetilde{Y}^{\prime}=1)\}\boldsymbol{B}\|_{\infty}\\
				&+\|\mathbb{E}_n(\boldsymbol{x}\mid\widetilde{Y}=1)-\mathbb{E}_n(\boldsymbol{x}^{\prime}\mid\widetilde{Y}^{\prime}=1)\|_{\infty}\|\{\mathbb{E}_n(\boldsymbol{x}\mid\widetilde{Y}=1)\}\boldsymbol{B}\|_{\infty}\bigg]\\
				\leq &\hat{p}_1\big(c_x\frac{2R}{n\widehat{p}_1}+\frac{2c_x}{n\widehat{p}_1}R\big)\leq\frac{4c_xR}{n},
			\end{split}
		\end{equation*}
		where we use the inequality $\|\boldsymbol{a}\boldsymbol{a}^{\top}-\boldsymbol{b}\boldsymbol{b}^{\top}\|_{\infty}\leq\|\boldsymbol{a}(\boldsymbol{a}-\boldsymbol{b})^{\top}\|_{\infty}+\|(\boldsymbol{b}-\boldsymbol{a})\boldsymbol{b}^{\top}\|_{\infty}$ in the first inequality and we use Assumption \ref{assum:low_design} and the property of the truncation operator in the second inequality. Then, we consider the case where $\widetilde{Y}_1^{\prime}\neq\widetilde{Y}_1$. Without loss of generality, we assume $\widetilde{Y}_1^{\prime}=2$. Then, by algebra, we have the decomposition
		\begin{equation*}
			\begin{split}
				&(\boldsymbol{\hat{M}}(\mathcal{S})-\boldsymbol{\hat{M}}(\mathcal{S}^{\prime}))\boldsymbol{B}\\
				=&-\frac{(\sum_{i=1}^{n}\boldsymbol{x}_i\mathbbm{1}\{\widetilde{Y}_i=1\})(\sum_{i=1}^{n}\boldsymbol{x}_i\mathbbm{1}\{\widetilde{Y}_i=1\})^{\top}\boldsymbol{B}}{n^2}\frac{1/n}{\hat{p}_1(\hat{p}_1-1/n)}+\frac{\boldsymbol{x}_1\boldsymbol{x}_1^{\top}\boldsymbol{B}}{n^2\hat{p}_1}\\
				&+\frac{(\sum_{i=1}^{n}\boldsymbol{x}_i\mathbbm{1}\{\widetilde{Y}_i=2\}\boldsymbol{B})(\sum_{i=1}^{n}\boldsymbol{x}_i\mathbbm{1}\{\widetilde{Y}_i=2\})^{\top}}{n^2}\frac{1/n}{\hat{p}_2(\hat{p}_2+1/n)}-\frac{\boldsymbol{x}_1^{\prime}\boldsymbol{x}_1^{\prime \top}\boldsymbol{B}}{n^2(\hat{p}_2+1/n)}\\
				&+2\frac{\boldsymbol{x}_1n(\hat{p}_1-1/n)}{n^2\hat{p}_1}\mathbb{E}_n(\boldsymbol{x}^{\prime}\mid\tilde{Y}^{\prime}=1)^{\top}\boldsymbol{B}-2\frac{\boldsymbol{x}_1^{\prime}n\hat{p}_2}{n^2(\hat{p}_2+1/n)}\mathbb{E}_n(\boldsymbol{x}\mid\tilde{Y}=2)^{\top}\boldsymbol{B}
			\end{split}
		\end{equation*}
		Thus, we have
		\begin{equation*}
			\begin{split}
				&\|(\boldsymbol{\hat{M}}(\mathcal{S})-\boldsymbol{\hat{M}}(\mathcal{S}^{\prime}))\boldsymbol{B}\|_{\infty}\\
				\leq&\frac{1}{n}\max_{h=1,2}\|\mathbb{E}_n(\boldsymbol{x}\mid\widetilde{Y}=h)\|_{\infty}\|\mathbb{E}_n(\boldsymbol{x}\mid\widetilde{Y}=h)\boldsymbol{B}\|_{\infty}(\frac{\hat{p}_1-1/n}{\hat{p}_1}+\frac{\hat{p}_2}{\hat{p}_2+1/n})\\
				&+\frac{1}{n^2}\max\{\|\boldsymbol{x}_1^{\prime}\|_{\infty}\|\boldsymbol{x}_1^{\prime}\boldsymbol{B}\|_{\infty},\|\boldsymbol{x}_1\|_{\infty}\|\boldsymbol{x}_1\boldsymbol{B}\|_{\infty}\}(\frac{1}{\hat{p}_1}+\frac{1}{\hat{p}_2+1/n})\\
				&+\frac{2}{n}\max\{\|\boldsymbol{x}_1^{\prime}\|_{\infty}\|\mathbb{E}_n(\boldsymbol{x}\mid\widetilde{Y}=2)\boldsymbol{B}\|_{\infty},\|\boldsymbol{x}_1\|_{\infty}\|\mathbb{E}_n(\boldsymbol{x}^{\prime}\mid\widetilde{Y}^{\prime}=1)\boldsymbol{B}\|_{\infty}\}(\frac{\hat{p}_1-1/n}{\hat{p}_1}+\frac{\hat{p}_2}{\hat{p}_2+1/n})\\
				\leq&\frac{c_xR}{n}\bigg(2+\frac{2}{np_H}+4\bigg)\leq\frac{7c_xR}{n}
			\end{split}
		\end{equation*}
		for $n$ large enough, where we use the Assumption \ref{assum:low_design}. In summary, we have
		\begin{equation*}
			\|(\boldsymbol{\hat{M}}(\mathcal{S})-\boldsymbol{\hat{M}}(\mathcal{S}^{\prime}))\boldsymbol{B}\|_{\infty}\leq\frac{7c_xR}{n}.
		\end{equation*}
		
		The sensitivity of the second term $\boldsymbol{\hat{\Sigma}}\boldsymbol{B}(\boldsymbol{B}^{\top}\boldsymbol{\hat{\Sigma}}\boldsymbol{B}-\boldsymbol{I}_k)$ has the decomposition,
		\begin{equation*}
			\begin{split}
				&\|\boldsymbol{\hat{\Sigma}}(\mathcal{S})\boldsymbol{B}(\boldsymbol{B}^{\top}\boldsymbol{\hat{\Sigma}}(\mathcal{S})\boldsymbol{B}-\boldsymbol{I}_k)-\boldsymbol{\hat{\Sigma}}(\mathcal{S}^{\prime})\boldsymbol{B}(\boldsymbol{B}^{\top}\boldsymbol{\hat{\Sigma}}(\mathcal{S}^{\prime})\boldsymbol{B}-\boldsymbol{I}_k)\|_{\infty}\\
				\leq & \|(\boldsymbol{\hat{\Sigma}}(\mathcal{S})-\boldsymbol{\hat{\Sigma}}(\mathcal{S}^{\prime}))\boldsymbol{B}\|_{\infty}+\|(\boldsymbol{\hat{\Sigma}}(\mathcal{S})\boldsymbol{B}\boldsymbol{B}^{\top}\boldsymbol{\hat{\Sigma}}(\mathcal{S})-\boldsymbol{\hat{\Sigma}}(\mathcal{S}^{\prime})\boldsymbol{B}\boldsymbol{B}^{\top}\boldsymbol{\hat{\Sigma}}(\mathcal{S}^{\prime}))\boldsymbol{B}\|_{\infty},
			\end{split}
		\end{equation*}
		where we use the triangle inequality. It remains to consider the sensitivity of the two parts separately. For the first part, we have
		\begin{equation*}
			\begin{split}
				&(\boldsymbol{\hat{\Sigma}}(\mathcal{S})-\boldsymbol{\hat{\Sigma}}(\mathcal{S}^{\prime}))\boldsymbol{B}=\frac{1}{n}\sum_{i=1}^{n}(\boldsymbol{x}_i\boldsymbol{x}_i^{\top}-\boldsymbol{x}_i^{\prime}\boldsymbol{x}_i^{\prime \top})\boldsymbol{B}=\frac{1}{n}(\boldsymbol{x}_1\boldsymbol{x}_1^{\top}-\boldsymbol{x}_1^{\prime}\boldsymbol{x}_1^{\prime \top})\boldsymbol{B},
			\end{split}
		\end{equation*}
		Thus, we have
		\begin{equation*}
			\|(\boldsymbol{\hat{\Sigma}}(\mathcal{S})-\boldsymbol{\hat{\Sigma}}(\mathcal{S}^{\prime}))\boldsymbol{B}\|_{\infty}\leq\frac{1}{n}(\|\boldsymbol{x}_1\|_{\infty}\|\boldsymbol{x}_1^{\top}\boldsymbol{B}\|_{\infty}+\|\boldsymbol{x}_1^{\prime}\|_{\infty}\|\boldsymbol{x}_1^{\prime \top}\boldsymbol{B}\|_{\infty})\leq\frac{2Rc_x}{n},
		\end{equation*}
		where we sue Assumption \ref{assum:low_design} and the property of truncation operator. For the second part, note that
		\begin{equation*}
			\begin{split}
				\|\boldsymbol{B}^{\top}\boldsymbol{\hat{\Sigma}}\boldsymbol{B}\|_{\infty}\leq\|\frac{1}{n}\sum_{i=1}^{n}\boldsymbol{B}^{\top}\boldsymbol{x}_i\boldsymbol{x}_i^{\top}\boldsymbol{B}\|_{\infty}\leq \frac{1}{n}\sum_{i=1}^{n}\|\boldsymbol{B}^{\top}\boldsymbol{x}_i\|_{\infty}\|\boldsymbol{x}_i^{\top}\boldsymbol{B}\|_{\infty}\leq R^2,
			\end{split}
		\end{equation*}
		\begin{equation*}
			\begin{split}
				\|\boldsymbol{\hat{\Sigma}}\boldsymbol{B}\|_{\infty}\leq\|\frac{1}{n}\sum_{i=1}^{n}\boldsymbol{x}_i\boldsymbol{x}_i^{\top}\boldsymbol{B}\|_{\infty}\leq \frac{1}{n}\sum_{i=1}^{n}\|\boldsymbol{x}_i\|_{\infty}\|\boldsymbol{x}_i^{\top}\boldsymbol{B}\|_{\infty}\leq c_xR,
			\end{split}
		\end{equation*}
		\begin{equation*}
			\begin{split}
				\|\boldsymbol{B}^{\top}\boldsymbol{\Sigma}(\mathcal{S})\boldsymbol{B}-\boldsymbol{B}^{\top}\boldsymbol{\Sigma}(\mathcal{S}^{\prime})\boldsymbol{B}\|_{\infty}&=\bigg\|\frac{1}{n}\sum_{i=1}^{n}\boldsymbol{B}^{\top}(\boldsymbol{x}_i\boldsymbol{x}_i^{\top}-\boldsymbol{x}_i^{\prime}\boldsymbol{x}_i^{\prime \top})\boldsymbol{B}\bigg\|_{\infty}\\
				&=\bigg\|\frac{1}{n}\boldsymbol{B}^{\top}(\boldsymbol{x}_1\boldsymbol{x}_1^{\top}-\boldsymbol{x}_1^{\prime}\boldsymbol{x}_1^{\prime \top})\boldsymbol{B}\bigg\|_{\infty}\leq\frac{2R^2}{n}.
			\end{split}
		\end{equation*}
		Thus, we have
		\begin{equation*}
			\begin{split}
				&\|(\boldsymbol{\hat{\Sigma}}(\mathcal{S})\boldsymbol{B}\boldsymbol{B}^{\top}\boldsymbol{\hat{\Sigma}}(\mathcal{S})-\boldsymbol{\hat{\Sigma}}(\mathcal{S}^{\prime})\boldsymbol{B}\boldsymbol{B}^{\top}\boldsymbol{\hat{\Sigma}}(\mathcal{S}^{\prime}))\boldsymbol{B}\|_{\infty}\\
				\leq& \|(\boldsymbol{\hat{\Sigma}}(\mathcal{S})\boldsymbol{B}-\boldsymbol{\hat{\Sigma}}(\mathcal{S}^{\prime})\boldsymbol{B})\boldsymbol{B}^{\top}\boldsymbol{\Sigma}\boldsymbol{B}\|_{\infty}+\|\boldsymbol{\hat{\Sigma}}(\mathcal{S}^{\prime})\boldsymbol{B}(\boldsymbol{B}^{\top}\boldsymbol{\Sigma}(\mathcal{S})\boldsymbol{B}-\boldsymbol{B}^{\top}\boldsymbol{\Sigma}(\mathcal{S}^{\prime})\boldsymbol{B})\|_{\infty}\\
				\leq& k\|\boldsymbol{\hat{\Sigma}}(\mathcal{S})\boldsymbol{B}-\boldsymbol{\hat{\Sigma}}(\mathcal{S}^{\prime})\boldsymbol{B}\|_{\infty}\|\boldsymbol{B}^{\top}\boldsymbol{\Sigma}(\mathcal{S})\boldsymbol{B}\|_{\infty}+k\|\boldsymbol{\hat{\Sigma}}(\mathcal{S}^{\prime})\boldsymbol{B}\|_{\infty}\|\boldsymbol{B}^{\top}\boldsymbol{\Sigma}(\mathcal{S})\boldsymbol{B}-\boldsymbol{B}^{\top}\boldsymbol{\Sigma}(\mathcal{S}^{\prime})\boldsymbol{B}\|_{\infty}\\
				\leq&  k\bigg\{\frac{2Rc_x}{n}\cdot R^2+c_xR\cdot \frac{2R^2}{n}\bigg\}\leq 4k\frac{c_xR^3}{n},
			\end{split}
		\end{equation*}
		where we use the triangle inequality in the first inequality, we use the relationship $\|\cdot\|_1\leq\|\cdot\|_0\times\|\cdot\|_{\infty}$ in the second inequality. Thus, the $l_{\infty}$ sensitivity of the gradient \eqref{eq:gradient} is bounded by 
		\begin{equation*}
			2\eta\{7Rc_x+\lambda_{penalty}(2Rc_x+4kR^3c_x)\}\frac{1}{n/T}.
		\end{equation*}
		By the relationship $\|\cdot\|_2\leq\sqrt{\|\cdot\|_0}\times\|\cdot\|_{\infty}$, we know that the $l_2$ sensitivity is bounded by $\sqrt{pk}\cdot 2\eta\{7Rc_x+\lambda_{penalty}(2Rc_x+4kR^3c_x)\}T/n$. By the Gaussian mechanism, each iteration is $(\varepsilon/T,\delta/T)$-DP. Thus, by the composition theorem, the gradient descent in Algorithm \ref{alg:ld} is $(\varepsilon,\delta)$-DP. Note that the slices estimation is $(\varepsilon,\delta)$-DP, we have that Algorithm \ref{alg:ld} is $(2\varepsilon,\delta)$-DP.
		
		When the initial value satisfies $\|\boldsymbol{B}^{\top}\boldsymbol{\hat{\Sigma}}\boldsymbol{B}-\boldsymbol{I}_k\|_{\infty}\leq\omega$, then the sensitivity of the second term $\boldsymbol{\hat{\Sigma}}\boldsymbol{B}(\boldsymbol{B}^{\top}\boldsymbol{\hat{\Sigma}}\boldsymbol{B}-\boldsymbol{I}_k)$ has the decomposition,
		\begin{equation*}
			\begin{split}
				&\|\boldsymbol{\hat{\Sigma}}(\mathcal{S})\boldsymbol{B}(\boldsymbol{B}^{\top}\boldsymbol{\hat{\Sigma}}(\mathcal{S})\boldsymbol{B}-\boldsymbol{I}_k)-\boldsymbol{\hat{\Sigma}}(\mathcal{S}^{\prime})\boldsymbol{B}(\boldsymbol{B}^{\top}\boldsymbol{\hat{\Sigma}}(\mathcal{S}^{\prime})\boldsymbol{B}-\boldsymbol{I}_k)\|_{\infty}\\
				\leq & \|(\boldsymbol{\hat{\Sigma}}(\mathcal{S})-\boldsymbol{\hat{\Sigma}}(\mathcal{S}^{\prime}))\boldsymbol{B}(\boldsymbol{B}^{\top}\boldsymbol{\hat{\Sigma}}(\mathcal{S})\boldsymbol{B}-\boldsymbol{I}_k)\|_{\infty}+\|\boldsymbol{\hat{\Sigma}}(\mathcal{S}^{\prime})\boldsymbol{B}(\boldsymbol{B}^{\top}\boldsymbol{\hat{\Sigma}}(\mathcal{S})\boldsymbol{B}-\boldsymbol{B}^{\top}\boldsymbol{\hat{\Sigma}}(\mathcal{S}^{\prime})\boldsymbol{B})\|_{\infty}\\
				\leq&\frac{2\omega Rc_xk}{n}+\frac{2c_xR^3k}{n},
			\end{split}
		\end{equation*}
		where we use the triangle inequality. Thus, the $l_{\infty}$ sensitivity of the gradient \eqref{eq:gradient} is bounded by 
		\begin{equation*}
			2\eta\{7Rc_x+\lambda_{penalty}(2\omega Rc_x+2c_xR^3)k\}\frac{1}{n/T}.
		\end{equation*}
		Thus, the noise scale can be further reduced by introducing an element-wide truncation to the part 
		\begin{equation*}
			\big\{\sum_{i\in S_{t}}\Pi_{R}(\boldsymbol{\hat{B}}^{(t)\top}\boldsymbol{x}_i)\Pi_{R}(\boldsymbol{x}_i^{\top}\boldsymbol{\hat{B}}^{(t)})/|S_{t}|-\boldsymbol{I}_k\big\}
		\end{equation*}
		in Line 5 of Algorithm \ref{alg:ld} and Line 4 of Algorithm \ref{alg:hd_dp_sir}.
		
		\hfill $\square$
	\end{proof}
	
	Next, we prove the upper bound of the estimator in Algorithm \ref{alg:ld}.
	
	\begin{proof}[Proof of Theorem \ref{thm:dpsir_errorbound}]
		
		We first consider the effect of truncation. Define events where the truncation operator in estimation didn't happen by
		\begin{equation*}
			\begin{split}
				E:=&\big\{\max_{t=0,\dots,T-1}|\boldsymbol{x}_i^{\top}\boldsymbol{\hat{B}}^{(t)}|\leq R\text{ for all }i\in S_t\big\}.
			\end{split}
		\end{equation*}
		We use $\|\cdot\|_{\psi_2}$ to denote the sub-Gaussian norm and $\|\cdot\|_{\psi_1}$ to denote the sub-exponential norm, respectively. Under condition \ref{assum:low_design}, and the independence between $\boldsymbol{x}_i$ and $\boldsymbol{\hat{B}}^{(t)}$ due to data splitting, applying the Chernoff bound, we have the large deviation result 
		$$\mathbb{P}(|\boldsymbol{x}_i^{\top}\boldsymbol{\hat{B}}^{(t)}|\geq R)\leq 2\exp\{-cR^2/(C^2\|\boldsymbol{x}_i\|_{\psi_2}^2)\},$$
		where we use the fact that $\boldsymbol{x}_i^{\top}\boldsymbol{\hat{B}}^{(t)}$ is sub-Gaussian distributed with the sub-Gaussian norm bounded by $C\|\boldsymbol{x}_i\|_{\psi_2}$ and $c$ is an absolute constant. Note that the absolute constant $c$ is frequently used in high-dimensional literature and textbooks; for example, Theorem 2.6.2 in \cite{vershynin2010introduction}. Note that the sub-Gaussian norm of $\boldsymbol{x}_i^{\top}\boldsymbol{\hat{B}}^{(t)}$ by definition is less than or equals to $\|\boldsymbol{\hat{B}}^{(t)}\|_2\|\boldsymbol{x}_i\|_{\psi_2}\leq C\|\boldsymbol{x}_i\|_{\psi_2}$, where we use the assumption that the $\boldsymbol{x}_i$ is a random vector following a sub-Gaussian distribution and $\|\boldsymbol{\hat{B}}^{(t)}\|_2\leq C$ by the truncation. Note that the set $E$ is the intersection of $n$ subsets, the union bound for set $E$ can be derived from the inequality: $(1-p_1)\times(1-p_2)\times\dots\times(1-p_n)\geq 1-p_1-\dots-p_n$. Thus, we have
		\begin{equation*}
			\begin{split}
				\mathbb{P}(E)&\geq 1-\sum_{i=1}^{n}\mathbb{P}(|\boldsymbol{x}_i^{\top}\boldsymbol{\hat{B}}^{(t)}|\geq R)\geq 1-2n\exp\{-cR^2/(C^2\|\boldsymbol{x}_i\|_{\psi_2}^2)\},
			\end{split}
		\end{equation*}
		where the inequality holds by applying the Chernoff bound $n$ times. By choosing the truncation level $R=2\sqrt{C^2\Psi_x^2\log(n)/c}$, we have $\mathbb{P}(E_1)\geq 1-2n\exp(-4\log(n))=1-2\exp(-3\log(n))\stackrel{n\to\infty}{\to}1$. Under the event $E$, it remains to consider the case where the truncations do not happen.
		
		We first consider the statistical estimation error, which is a function of the estimation errors of $\boldsymbol{\Sigma}$ and $\M$. Note that the sample estimators of $\M$ and $\boldsymbol{\Sigma}$ are defined by,
		\begin{equation*}
			\boldsymbol{\hat{M}}=\sum_{h=1}^{H}\hat{p}_{h}[\mathbb{E}_n(\boldsymbol{x}\mid\widetilde{Y}=h)-\mathbb{E}_n\boldsymbol{x}][\mathbb{E}_n(\boldsymbol{x}\mid\widetilde{Y}=h)-\mathbb{E}_n\boldsymbol{x}]^{\top},
		\end{equation*}
		where $\hat{p}_h=\mathbb{E}(\mathbbm{1}\{\widetilde{Y}=h\})$, $\mathbb{E}_n(\boldsymbol{x}\mid\widetilde{Y}=h)=\sum_{i=1}^{n}\boldsymbol{x}_i\mathbbm{1}\{\widetilde{Y}=h\}/\sum_{i=1}^{n}\mathbbm{1}\{\widetilde{Y}=h\}$, and
		\begin{equation*}
			\boldsymbol{\hat{\Sigma}}=\frac{1}{n}\sum_{i=1}^{n}\{\boldsymbol{x}_i-\mathbb{E}_n(\boldsymbol{x}_i)\}\{\boldsymbol{x}_i-\mathbb{E}_n(\boldsymbol{x}_i)\}^{\top},
		\end{equation*}
		respectively. We first consider the term $\|\boldsymbol{\hat{\Sigma}}-\boldsymbol{\Sigma}\|_{op}$. Note that
		\begin{equation*}
			\begin{split}
				\|\boldsymbol{\hat{\Sigma}}-\boldsymbol{\Sigma}\|_{op}&=\|\boldsymbol{\Sigma}^{1/2}(\boldsymbol{\Sigma}^{-1/2}\boldsymbol{\hat{\Sigma}}\boldsymbol{\Sigma}^{-1/2}-\boldsymbol{I}_p)\boldsymbol{\Sigma}^{1/2}\|_{op}\\
				&\leq\lambda_x\|\boldsymbol{\Sigma}^{-1/2}\boldsymbol{\hat{\Sigma}}\boldsymbol{\Sigma}^{-1/2}-\boldsymbol{I}_p\|_{op}=\lambda_x\big\|\mathbb{E}_n(\boldsymbol{\hat{z}}_i\boldsymbol{\hat{z}}_i^{\top})-\boldsymbol{I}_p\big\|_{op}\\
				&=\lambda_x\big\|\mathbb{E}_n(\boldsymbol{\hat{z}}_i\boldsymbol{\hat{z}}_i^{\top})-\mathbb{E}_n(\boldsymbol{z}_i\boldsymbol{z}_i^{\top})\big\|_{op}+\lambda_x\big\|\mathbb{E}_n(\boldsymbol{z}_i\boldsymbol{z}_i^{\top})-\boldsymbol{I}_p\big\|_{op}
			\end{split}
		\end{equation*}
		where we use the notation $\boldsymbol{\hat{z}}_i:=\boldsymbol{\Sigma}^{-1/2}\{\boldsymbol{x}_{i}-\mathbb{E}_n(\boldsymbol{X}_{i})\}$ to represent sample normalized vectors and $\boldsymbol{z}_i:=\boldsymbol{\Sigma}^{-1/2}\{\boldsymbol{x}_{i}-\mathbb{E}(\boldsymbol{x}_{i})\}\sim(\boldsymbol{0},\boldsymbol{I}_p)$ to represent normalized vectors. The first term corresponding the the effect of estimating $\mathbb{E}(\boldsymbol{x})$ satisfies
		\begin{equation*}
			\begin{split}
				&\|\mathbb{E}_n(\boldsymbol{\hat{z}}_i\boldsymbol{\hat{z}}_i^{\top})-\mathbb{E}_n(\boldsymbol{z}_i\boldsymbol{z}_i^{\top})\|_{op}\\
				=&\|\mathbb{E}_n\{(\boldsymbol{\hat{z}}_i-\boldsymbol{z}_i)(\boldsymbol{\hat{z}}_i-\boldsymbol{z}_i)^{\top}\}-\mathbb{E}_n\{\boldsymbol{\hat{z}}_i(\boldsymbol{\hat{z}}_i-\boldsymbol{z}_i)^{\top}\}-\mathbb{E}_n\{(\boldsymbol{\hat{z}}_i-\boldsymbol{z}_i)\boldsymbol{\hat{z}}_i^{\top}\}\|_{op}\\
				\leq&\|\mathbb{E}_n\{(\boldsymbol{\hat{z}}_i-\boldsymbol{z}_i)(\boldsymbol{\hat{z}}_i-\boldsymbol{z}_i)^{\top}\}\|_{op}+\|\mathbb{E}_n\{\boldsymbol{\hat{z}}_i(\boldsymbol{\hat{z}}_i-\boldsymbol{z}_i)^{\top}\}\|_{op}+\|\mathbb{E}_n\{(\boldsymbol{\hat{z}}_i-\boldsymbol{z}_i)\boldsymbol{\hat{z}}_i^{\top}\}\|_{op}\\
				=&\|\mathbb{E}_n\{(\boldsymbol{\hat{z}}_i-\boldsymbol{z}_i)(\boldsymbol{\hat{z}}_i-\boldsymbol{z}_i)^{\top}\}\|_{op}=\|\mathbb{E}_n(\boldsymbol{\hat{z}}_i-\boldsymbol{z}_i)\|_2^2\leq\{4\sqrt{p}+2\sqrt{\log(1/\delta)}\}^2/n,
			\end{split}
		\end{equation*}
		with a probability of at least $1-\delta$, where we use the triangle inequality in the second inequality, we use the fact that $\mathbb{E}_n(\boldsymbol{\hat{z}}_i)=\boldsymbol{0}$ in the third equality, and we use the Lemma \ref{lem:con} in the last inequality. As we will show later, the term due to estimating mean is relatively small. Furthermore, the remaining term satisfies
		\begin{equation*}
			\begin{split}
				\mathbb{P}\bigg(\bigg\|\frac{1}{n}\sum_{i=1}^{n}\boldsymbol{z}_i\boldsymbol{z}_i^{\top}-\boldsymbol{I}_p\bigg\|_{op}\leq t\bigg)\geq 1-\exp(-Cp-Cnt^2)
			\end{split}
		\end{equation*}
		where we use  Lemma \ref{lem:random_matrix_eigen}. For $t=C\sqrt{p/n}$, we conclude
		\begin{equation*}
			\|\boldsymbol{\hat{\Sigma}}-\boldsymbol{\Sigma}\|_{op}\leq 2C\lambda_x\sqrt{\frac{p}{n}},
		\end{equation*}
		with probability at least $1-\exp(-2Cp)\stackrel{p\to\infty}{\longrightarrow}1$. Then we consider the term $\|\M-\boldsymbol{\hat{M}}\|_{op}$. Similar to the arguments for $\|\boldsymbol{\Sigma}-\boldsymbol{\hat{\Sigma}}\|_{op}$, without loss of generality, we assume $\mathbb{E}_n(\boldsymbol{x})=0$ and use the following notations $\M_h:=\mathbb{E}(\boldsymbol{x}\mathbbm{1}\{\tilde{Y}=h\}),\boldsymbol{\hat{m}}_h:=\sum_{i=1}^{n}\boldsymbol{x}_i\mathbbm{1}\{\tilde{Y}=h\}/n$ and $\boldsymbol{\Sigma}_h=\M_h\M_h^{\top},\boldsymbol{\hat{\Sigma}}_h=\boldsymbol{\hat{m}}_h\boldsymbol{\hat{m}}_h^{\top}$ Then, we can write $\boldsymbol{M}$ and $\boldsymbol{\hat{M}}$ as
		\begin{equation*}
			\M=\sum_{i=1}^{H}p_h^{-1}\boldsymbol{\Sigma}_h,\text{ and }\boldsymbol{\hat{M}}=\sum_{i=1}^{H}\hat{p}_h^{-1}\boldsymbol{\hat{\Sigma}}_h.
		\end{equation*}
		Notice that $\hat{p}_h$ is the sample average of $n$ i.i.d. Binomial random variables. For $h=1,\dots,H$, by Bernstein's inequality, we have
		\begin{equation*}
			\mathbb{P}\{|\hat{p}_h-p_h|\geq\sqrt{p_h(1-p_h)}\sqrt{\log(p)/n}\}\leq 2\exp\bigg\{-\frac{\log(p)}{2(1+\sqrt{\log(p)/n}/3)}\bigg\}.
		\end{equation*}
		By $p_h(1-p_h)\leq 1/4$ and $\log(p)/n\to 0$, then
		\begin{equation*}
			\mathbb{P}\{|\hat{p}_h-p_h|\geq \sqrt{\log(p)/n}/2\}\leq 2\exp\bigg\{-\frac{\log(p)}{2(1+\sqrt{\log(p)/n}/3)}\bigg\}\leq 2p^{-1/3}.
		\end{equation*}
		Furthermore,
		\begin{equation*}
			\begin{split}
				&\mathbb{P}\{|\hat{p}_h^{-1}-p_h^{-1}|\geq\sqrt{\log(p)/n}/2/p_h^2\}\\
				=&\mathbb{P}\{|\hat{p}_h-p_h|\geq\sqrt{\log(p)/n}/2\hat{p}_h/p_h\}\\
				\leq &2p^{-1/3}+\mathbb{P}\{|\hat{p}_h-p_h|\geq\sqrt{\log(p)/n}/2(1+\sqrt{\log(p)/n}/2/p_h)\}\\
				\leq& 2p^{-1/3}+\mathbb{P}\{|\hat{p}_h-p_h|\geq\sqrt{\log(p)/n}\}\leq 2(p^{-1/3}+p^{-4/3}),
			\end{split}
		\end{equation*}
		where we use Bernstein's inequality in the last inequality. Furthermore, we have
		\begin{align*}
			\|\boldsymbol{\hat{M}}-\M\|_{op}\leq&\sum_{h=1}^{H}|\hat{p}_{h}^{-1}-p_h^{-1}|\|\boldsymbol{\Sigma}_{h}\|_{op}+\sum_{h=1}^{H}\hat{p}_{h}^{-1}\|\boldsymbol{\Sigma}_{h}-\boldsymbol{\hat{\Sigma}}_{h}\|_{op}\\
			\leq& C\sqrt{\frac{\log(p)}{n}}\sum_{h=1}^{H}\|\boldsymbol{\Sigma}_{h}\|_{op}+\sum_{h=1}^{H}\hat{p}_{h}^{-1}\|(\M_{h}-\boldsymbol{\hat{m}}_{h,\mathcal{I}\mathcal{I}})(\M_{h}-\boldsymbol{\hat{m}}_{h})^{\top}\\
			&+\M_{h}(\boldsymbol{\hat{m}}_{h}-\M_{h})^{\top}+(\boldsymbol{\hat{m}}_{h}-\M_{h})\M_{h}^{\top}\|_{op}.
		\end{align*}
		The first term is sufficiently small compared to the second term, so it is enough to consider the second term. In the second term, the dominant term is $\sum_{h=1}^{H}\|(\boldsymbol{\hat{m}}_{h}-\M_{h})\M_{h}^{\top}\|_{op}$. Notice that we have
		\begin{equation*}
			\begin{split}
				&\lambda_{min}(\boldsymbol{\Sigma}^{-1/2})\|\M_h\|_2\leq\|\boldsymbol{\Sigma}^{-1/2}\M_h\|_2\leq\sum_{h=1}^{H}\frac{1}{p_h}\|\boldsymbol{\Sigma}^{-1/2}\M_h\|_2\\
				=&\text{Tr}(\boldsymbol{\Sigma}^{-1/2}\M\boldsymbol{\Sigma}^{-1/2})\leq k\lambda/\kappa\leq C,
			\end{split}
		\end{equation*}
		for some constant $C$, where we use the fact that
		\begin{equation*}
			\sum_{h=1}^{H}\frac{\boldsymbol{\Sigma}^{-1/2}\M_h(\boldsymbol{\Sigma}^{-1/2}\M_h)^{\top}}{p_h}=\boldsymbol{\Sigma}^{-1/2}\M\boldsymbol{\Sigma}^{-1/2}.
		\end{equation*}
		Thus,
		\begin{equation*}
			\begin{split}
				\|(\boldsymbol{\hat{m}}-\M_{h})\M_{h}^{\top}\|_{op}=\|\boldsymbol{\hat{m}}_{h}-\M_{h}\|_2\|\M_{h}\|_2\leq C\sqrt{\frac{p}{n}}\lambda_x,
			\end{split}
		\end{equation*}
		with probability at least $1-\exp\{-C\log(p)\}$. Then, we conclude
		\begin{equation*}
			\|\boldsymbol{\hat{M}}-\M\|_{op}\leq C\sqrt{\frac{p}{n}},
		\end{equation*}
		for a universal constant $C$ with probability at least $1-\exp\{-C\log(p)\}$. Furthermore, the estimated eigenvalues satisfy
		\begin{equation*}
			\begin{split}
				|\lambda_{k+1}-\hat{\lambda}_{k+1}|&\leq\|\boldsymbol{\Sigma}^{-1/2}\M\boldsymbol{\Sigma}^{-1/2}-\boldsymbol{\hat{\Sigma}}^{-1/2}\boldsymbol{\hat{M}}\boldsymbol{\hat{\Sigma}}^{-1/2}\|_{op}\\
				&\leq\|\boldsymbol{\hat{\Sigma}}^{-1/2}(\boldsymbol{\Sigma}^{1/2}-\boldsymbol{\hat{\Sigma}}^{1/2})\boldsymbol{\Sigma}^{-1/2}\M\boldsymbol{\Sigma}^{-1/2}\|_{op}+\|\boldsymbol{\hat{\Sigma}}^{-1/2}(\M-\boldsymbol{\hat{M}})\boldsymbol{\Sigma}^{-1/2}\|_{op}\\
				&+\|\boldsymbol{\hat{\Sigma}}^{-1/2}\boldsymbol{\hat{M}}\boldsymbol{\hat{\Sigma}}^{-1/2}(\boldsymbol{\Sigma}^{-1/2}-\boldsymbol{\hat{\Sigma}}^{-1/2})\boldsymbol{\Sigma}^{1/2}\|_{op}\\
				&\leq\|\boldsymbol{\hat{\Sigma}}^{-1/2}\|_{op}\|\boldsymbol{\Sigma}^{1/2}-\boldsymbol{\hat{\Sigma}}^{1/2}\|_{op}\|\boldsymbol{\Sigma}^{-1/2}\M\boldsymbol{\Sigma}^{-1/2}\|_{op}\\
				&+\|\boldsymbol{\hat{\Sigma}}^{-1/2}\|_{op}\|\M-\boldsymbol{\hat{M}}\|_{op}\|\boldsymbol{\Sigma}^{-1/2}\|_{op}\\
				&+\|\boldsymbol{\hat{\Sigma}}^{-1/2}\boldsymbol{\hat{M}}\boldsymbol{\hat{\Sigma}}^{-1/2}\|_{op}\|\boldsymbol{\Sigma}^{1/2}-\boldsymbol{\hat{\Sigma}}^{1/2}\|_{op}\|\boldsymbol{\Sigma}^{-1/2}\|_{op}\\
				&=O_p\bigg(\sqrt{\frac{p}{n}}\bigg),
			\end{split}
		\end{equation*}
		where we use Lemma 2.1 in \cite{sibson1979studies} in the first inequality, we use the triangle inequality in the second inequality, and we use the sub-multiplicativity of matrix norm in the third inequality. By the proof of Lemma \ref{lem:est_slices}, the generalized eigenvalues corresponding to the estimated slices satisfy $\lambda_k\geq\lambda/2$. Then, the following event holds with a probability of at least $1-\exp\{-C\log(p)\}$,
		\begin{equation*}
			\hat{\lambda}_{k+1}\leq \lambda_k/2,
		\end{equation*}
		where we use the assumption $\lambda_{k+1}=0$ and $p/(n\lambda^2)$ is small enough. We define
		\begin{equation*}
			\Delta=\frac{\lambda_k-\hat{\lambda}_{k+1}}{\sqrt{(1+\lambda_1^2)(1+\hat{\lambda}_{k+1}^2)}}\geq\frac{\lambda_k}{2\sqrt{(1+\lambda_1^2)(1+\lambda_k^2/4)}}.
		\end{equation*}
		We define the notion,
		\begin{equation*}
			c(\boldsymbol{A},\B;\mathcal{X})=\inf_{\|\boldsymbol{x}\|_2=1,\boldsymbol{x}\in\mathcal{X}}\sqrt{(\boldsymbol{x}^{\top}\boldsymbol{A}\boldsymbol{x})^2+(\boldsymbol{x}^{\top}\B\boldsymbol{x})^2}.
		\end{equation*}
		Then, we have
		\begin{equation*}
			c(\M,\boldsymbol{\Sigma})=\inf_{\|\boldsymbol{x}\|_2=1}\sqrt{(\boldsymbol{x}^{\top}\M\boldsymbol{x})^2+(\boldsymbol{x}^{\top}\boldsymbol{\Sigma}\boldsymbol{x})^2}\geq\lambda_{min}(\boldsymbol{\Sigma}),
		\end{equation*}
		\begin{equation*}
			c(\boldsymbol{\hat{M}},\boldsymbol{\hat{\Sigma}})=\inf_{\|\boldsymbol{x}\|_2=1}\sqrt{(\boldsymbol{x}^{\top}\boldsymbol{\hat{M}}\boldsymbol{x})^2+(\boldsymbol{x}^{\top}\boldsymbol{\hat{\Sigma}}\boldsymbol{x})^2}\geq\lambda_{min}(\boldsymbol{\Sigma})/2,
		\end{equation*}
		where we use the fact $\|\boldsymbol{\hat{\Sigma}}-\boldsymbol{\Sigma}\|_{op}\to 0$. Let $\epsilon_{noisy}=\sqrt{\|\M-\boldsymbol{\hat{M}}\|_{op}^2+\|\boldsymbol{\hat{\Sigma}}-\boldsymbol{\Sigma}\|_{op}^2}$, then by Theorem 3.1 in \cite{sun1983perturbation} and Corollary 4.5 in \cite{stewart1979pertubation},
		\begin{align*}
			\|\mathbf{P}-\mathbf{\hat{P}}\|_F^2&\leq\bigg\{ \frac{\epsilon_{noisy}\sqrt{k}}{\Delta c(\M,\boldsymbol{\Sigma})c(\boldsymbol{\hat{M}},\boldsymbol{\hat{\Sigma}})}\sqrt{\|\M\|_{op}^2+\|\boldsymbol{\Sigma}\|_{op}^2}\bigg\}^2\\
			&\leq C\frac{p}{\lambda^2n},
		\end{align*}
		where $\mathbf{P}$ and $\hat{\mathbf{P}}$ are the projection matrix on the eigen-subspace of matrix pair $(\M,\boldsymbol{\Sigma})$ and $(\boldsymbol{\hat{M}},\boldsymbol{\hat{\Sigma}})$. Thus, we derived the statistical error bound.
		
		Then, we consider the error due to differential privacy. Let $\boldsymbol{\hat{B}}$ be the non-private estimator of $\B$ via the generalized eigen decomposition of $(\boldsymbol{\hat{M}},\boldsymbol{\hat{\Sigma}})$. We use the notation $\boldsymbol{\hat{A}}:=\boldsymbol{\hat{B}}(\boldsymbol{I}_k+\boldsymbol{\hat{\Lambda}}_k/\lambda_{penalty})^{1/2}$ where $\boldsymbol{\hat{\Lambda}}_k=\text{diag}\{\hat{\lambda}_1,\dots,\hat{\lambda}_k\}$ are the generalized eigenvalues of $\boldsymbol{\hat{B}}$. For two matrices $\boldsymbol{X},\boldsymbol{Y}\in\mR^{p\times r}$, the distance measure $\text{dist}(\cdot,\cdot)$ is defined by
		\begin{equation*}
			\text{dist}(\boldsymbol{X},\boldsymbol{Y})=\min_{\boldsymbol{H}\in\mathbb{O}^{r\times r}}\|\boldsymbol{X}\boldsymbol{H}-\boldsymbol{Y}\|_F,
		\end{equation*}
		where $\mathbb{O}^{r\times r}$ is space containing all $r\times r$ orthogonal matrix. By Lemma \ref{lem:grad_desc}, for the penalty satisfying $\lambda_{penalty}=\lambda_1/c$, for the step size satisfying $\eta\leq C_1/\{\lambda_1(\lambda_x+2)\}$, and for the current value satisfying, $$\text{dist}(\boldsymbol{\hat{B}}^{(t)},\boldsymbol{\hat{A}})\leq\frac{1}{\sqrt{\lambda_x}}\min\bigg\{\frac{\lambda_k}{C_2\lambda_1\lambda_x^2},\frac{\sqrt{1+2c}}{4}\bigg\},$$
		where $c,C_1,C_2$ are some universal constants, then after one iteration, the distance satisfies
		$$\text{dist}(\boldsymbol{\hat{B}}^{(t+0.5)},\boldsymbol{\hat{A}})^2\leq\bigg(1-\frac{\eta\lambda_k}{4\lambda_x}\bigg)\text{dist}(\boldsymbol{\hat{B}}^{(t)},\boldsymbol{\hat{A}})^2.$$
		The error induced by additional noise is
		\begin{align*}
			\text{dist}(\boldsymbol{\hat{B}}^{(t+1)},\boldsymbol{\hat{A}})&\leq \text{dist}(\boldsymbol{\hat{B}}^{(t+0.5)},\boldsymbol{\hat{A}})+\|\boldsymbol{w}_t\|_F,
		\end{align*}
		where $\boldsymbol{w}_t$ is the added noise in Line 5 of Algorithm \ref{alg:ld}, we use the triangle inequality and the fact that the truncation didn't happen. By applying Lemma \ref{lem:con}, we have
		\begin{equation*}
			\|\boldsymbol{w}_t\|_F\leq 6\sqrt{kp}\sqrt{\sigma^2\frac{2T^2\log(1.25T/\delta)}{\varepsilon^2}},
		\end{equation*}
		with probability at least $1-\exp(-p)$. Thus, by applying the previous result $T$ times, we have that, for a contraction parameter $0<c_{con}=\eta\lambda_k/4\lambda_x<1$,
		\begin{equation*}
			\text{dist}(\boldsymbol{\hat{B}}^{(T)},\boldsymbol{\hat{A}})^2\leq 2c_{con}^{T}\text{dist}(\boldsymbol{\hat{B}}^{(0)},\boldsymbol{\hat{A}})^2+2\times 36kp\sigma^2\frac{2T^2\log(1.25T/\delta)}{\varepsilon^2c_{con}},
		\end{equation*}
		where we use Lemma A.2 in \cite{cai2021cost} and the inequality $(a+b)^2\leq 2a^2+2b^2$. Combining the statistical error rate, we conclude
		\begin{equation*}
			\begin{split}
				\text{dist}(\boldsymbol{\hat{B}}^{(T)},\boldsymbol{A})^2&\leq2\text{dist}^2(\boldsymbol{\hat{B}}^{(T)},\boldsymbol{\hat{A}})+2\text{dist}(\boldsymbol{A},\boldsymbol{\hat{A}})^2\\
				&\leq 2c_{con}^{T}\text{dist}(\boldsymbol{\hat{B}}^{(0)},\boldsymbol{\hat{A}})^2+2\times 36kp\sigma^2\frac{2T^2\log(1.25T/\delta)}{\varepsilon^2c_{con}}+C\frac{p}{\lambda^2n/T},
			\end{split}
		\end{equation*}
		where $\boldsymbol{A}:=\B(\boldsymbol{I}_k+\boldsymbol{\Lambda}_k/\lambda_{penalty})^{1/2}$. For $T=O\{\log(n)\}$, the first term is smaller than the third term. Thus, by the choice of $\sigma$ in Lemma \ref{lem:low_privacy}, we have
		\begin{equation*}
			\text{dist}(\boldsymbol{\hat{B}}^{(T)},\boldsymbol{A})^2\leq C^{\prime}\frac{p\log(n)}{\lambda^2n}+C^{\prime}\frac{p^{2}\log(1/\delta)\log^7(n)}{\lambda^2n^2\varepsilon^2},
		\end{equation*}
		for a universal constant $C^{\prime}$. Furthermore, by the proof of Corollary 4.2 in \cite{gao2021sparse}, for a positive constant $C^{\prime\prime}$, the following holds with probability greater than $1-\exp\{-C^{\prime}\log(p)\}\to 1$ as $p\to\infty$:
		\begin{equation*}
			\text{dist}(\boldsymbol{\hat{B}}^{(T)}(\boldsymbol{\hat{B}}^{(T)\top}\boldsymbol{\Sigma}\boldsymbol{\hat{B}}^{(T)})^{-1/2},\boldsymbol{B})^2\leq \frac{C^{\prime\prime}}{2\lambda_x}\frac{p\log(n)}{\lambda^2n}+\frac{C^{\prime\prime}}{2\lambda_x}\frac{p^{2}\log(1/\delta)\log^7(n)}{\lambda^2n^2\varepsilon^2}.
		\end{equation*}
		We use the notation $\boldsymbol{\hat{B}}^{(n)}:=\boldsymbol{\hat{B}}^{(T)}(\boldsymbol{\hat{B}}^{(T)\top}\boldsymbol{\Sigma}\boldsymbol{\hat{B}}^{(T)})^{-1/2}$ to denote the normalized matrix. Applying the Hölder's inequality, we have
		\begin{equation*}
			\text{dist}(\boldsymbol{\Sigma}^{1/2}\boldsymbol{\hat{B}}^{(n)},\boldsymbol{\Sigma}^{1/2}\boldsymbol{B})^2\leq\|\boldsymbol{\Sigma}^{1/2}\|^2_{op}\times\text{dist}(\boldsymbol{\hat{B}}^{(n)},\boldsymbol{B})^2=\lambda_x\text{dist}(\boldsymbol{\hat{B}}^{(n)},\boldsymbol{B})^2.
		\end{equation*}
		By the fact $\boldsymbol{\hat{B}}^{(n)\top}\boldsymbol{\Sigma}\boldsymbol{\hat{B}}^{(n)}=\boldsymbol{I}_k$ and $\boldsymbol{B}^{\top}\boldsymbol{\Sigma}\boldsymbol{B}=\boldsymbol{I}_k$, and apply Lemma 2 in \cite{tan2020sparse}, we have the upper bound
		\begin{equation*}
			\begin{split}
				L(\boldsymbol{\hat{B}}^{(T)},\B)=&
				\|\boldsymbol{P}_{\boldsymbol{\hat{B}}^{(T)}}-\boldsymbol{P}_{\B}\|_F^2=2\times	\text{dist}(\boldsymbol{\Sigma}^{1/2}\boldsymbol{\hat{B}}^{(n)},\boldsymbol{\Sigma}^{1/2}\boldsymbol{B})^2\\
				=&C^{\prime\prime}\frac{p\log(n)}{\lambda^2n}+C^{\prime\prime}\frac{p^{2}\log(1/\delta)\log^7(n)}{\lambda^2n^2\varepsilon^2}.
			\end{split}
		\end{equation*}
		\hfill $\square$
	\end{proof}
	
	\subsection{Proof of the Initial Estimator}
	
	We first show that the Algorithm \ref{alg:ld_ini}, the initial estimate for the low-dimensional SIR,  is $(2\varepsilon,\delta)$-DP.
	
	\begin{proof}[Proof of Lemma \ref{lem:low_privacy_ini}]
		
		We begin the proof with the sensitivities of sample estimators $\boldsymbol{\hat{\Sigma}}$ and $\boldsymbol{\hat{M}}$. We use $\mathcal{S}^{\prime}$ to denote a neighboring data set of the original dataset $\mathcal{S}=\{(\boldsymbol{x}_i,Y_i)\}_{i=1}^{n}$. Without loss of generality, we assume the difference occurs in the first sample, i.e., $\mathcal{S}^{\prime}=\{(\boldsymbol{x}_1^{\prime},Y_1^{\prime}),(\boldsymbol{x}_i,Y_i)_{i=2}^{n}\}$. Then we have
		\begin{equation*}
			\boldsymbol{\hat{\Sigma}}(\mathcal{S})-\boldsymbol{\hat{\Sigma}}(\mathcal{S}^{\prime})
			=\frac{1}{n}\sum_{i=1}^{n}(\boldsymbol{x}_i\boldsymbol{x}_i^{\top}-\boldsymbol{x}_i^{\prime}\boldsymbol{x}_i^{\prime \top})=\frac{1}{n}(\boldsymbol{x}_1\boldsymbol{x}_1^{\top}-\boldsymbol{x}_1^{\prime}\boldsymbol{x}_1^{\prime \top}),
		\end{equation*}
		where we use the notation $\boldsymbol{\hat{\Sigma}}(\mathcal{S})$ to indicate the dependence on the data set $\mathcal{S}$ and $\boldsymbol{\hat{\Sigma}}(\mathcal{S}^{\prime})$ to indicate the dependence on the data set $\mathcal{S}^{\prime}$. The difference satisfies
		\begin{equation*}
			\begin{split}
				\|\boldsymbol{\hat{\Sigma}}(\mathcal{S})-\boldsymbol{\hat{\Sigma}}(\mathcal{S}^{\prime})\|_F&=\frac{1}{n}\|\boldsymbol{x}_1\boldsymbol{x}_1^{\top}-\boldsymbol{x}_1^{\prime}\boldsymbol{x}_1^{\prime \top}\|_F\leq\frac{1}{n}\|\boldsymbol{x}_1\boldsymbol{x}_1^{\top}\|_F+\frac{1}{n}\|\boldsymbol{x}_1^{\prime}\boldsymbol{x}_1^{\prime \top}\|_F\\
				&=\frac{1}{n}(\|\boldsymbol{x}_1\|_2^2+\|\boldsymbol{x}_1^{\prime}\|_2^2)\leq \frac{p}{n}(\|\boldsymbol{x}_1\|^2_{\infty}+\|\boldsymbol{x}_1^{\prime}\|^2_{\infty})\leq\frac{2p}{n}c_x^2,
			\end{split}
		\end{equation*}
		where we use the triangle inequality in the second inequality, and we use the inequality $\|\cdot\|_2^2\leq\|\cdot\|_0\times\|\cdot\|^2_{\infty}$ in the fourth inequality, and we use the bounded design condition in the last inequality.
		
		Then, we consider the estimator of $\boldsymbol{\hat{M}}$. Note that
		\begin{equation*}
			\begin{split}
				&\boldsymbol{\hat{M}}(\mathcal{S})-\boldsymbol{\hat{M}}(\mathcal{S}^{\prime})\\
				&=\sum_{h=1}^{H}\bigg\{\widehat{p}_h\mathbb{E}_n(\boldsymbol{x}\mid\widetilde{Y}=h)\mathbb{E}_n(\boldsymbol{x}\mid\widetilde{Y}=h)^{\top}-\widehat{p}_h^{\prime}\mathbb{E}_n(\boldsymbol{x}^{\prime}\mid\widetilde{Y}^{\prime}=h)\mathbb{E}_n(\boldsymbol{x}^{\prime}\mid\widetilde{Y}^{\prime}=h)^{\top}\bigg\},
			\end{split}
		\end{equation*}
		where we use the notation $\boldsymbol{\hat{M}}(\mathcal{S})$ to indicate the dependence on the data set $\mathcal{S}$ and $\boldsymbol{\hat{M}}(\mathcal{S}^{\prime})$ to indicate the dependence on the data set $\mathcal{S}^{\prime}$; we use the notation $\widehat{p}_h^{\prime}=\sum_{i=1}^{n}\mathbbm{1}\{Y_i^{\prime}\in\widehat{I}_h\}/n$ and $\mathbb{E}_n(\boldsymbol{x}^{\prime}\mid\widetilde{Y}^{\prime}=h)=\widehat{p}_h^{\prime -1}\sum_{i=1}^{n}\boldsymbol{x}_i^{\prime}\mathbbm{1}\{Y_i^{\prime}\in\widehat{I}_h\}/n$. The above term has two possible forms, depending on whether $Y_1$ and $Y_1^{\prime}$ belong to the same slice. Without loss of generality, we assume $\widetilde{Y}_1=1$. If $\widetilde{Y}_1^{\prime}=1$ and thus $\{\boldsymbol{x}_1^{\prime},Y_1^{\prime}\}$ belongs to the same slice as $\{\boldsymbol{x}_1,Y_1\}$, then we have
		\begin{equation*}
			\begin{split}
				&\|\boldsymbol{\hat{M}}(\mathcal{S})-\boldsymbol{\hat{M}}(\mathcal{S}^{\prime})\|_F\\
				=&\|\hat{p}_1\big\{\mathbb{E}_n(\boldsymbol{x}\mid\widetilde{Y}=1)\mathbb{E}_n(\boldsymbol{x}\mid\widetilde{Y}=1)^{\top}-\mathbb{E}_n(\boldsymbol{x}^{\prime}\mid\widetilde{Y}^{\prime}=1)\mathbb{E}_n(\boldsymbol{x}^{\prime}\mid\widetilde{Y}^{\prime}=1)^{\top}\big\}\|_F\\
				\leq&\hat{p}_1\bigg\{\|\mathbb{E}_n(\boldsymbol{x}\mid\widetilde{Y}=1)-\mathbb{E}_n(\boldsymbol{x}^{\prime}\mid\widetilde{Y}^{\prime}=1)\|_2\|\mathbb{E}_n(\boldsymbol{x}^{\prime}\mid\widetilde{Y}^{\prime}=1)\|_2\\
				&+\|\mathbb{E}_n(\boldsymbol{x}\mid\widetilde{Y}=1)-\mathbb{E}_n(\boldsymbol{x}^{\prime}\mid\widetilde{Y}^{\prime}=1)\|_2\|\mathbb{E}_n(\boldsymbol{x}\mid\widetilde{Y}=1)\|_2\bigg\}\\
				\leq&\widehat{p}_1\bigg\{\frac{1}{n\widehat{p}_1}\|\boldsymbol{x}_1-\boldsymbol{x}_1^{\prime}\|_2\|\mathbb{E}_n(\boldsymbol{x}\mid\widetilde{Y}=1)\|_2+\frac{1}{n\widehat{p}_1}\|\boldsymbol{x}_1-\boldsymbol{x}_1^{\prime}\|_2\|\mathbb{E}_n(\boldsymbol{x}^{\prime}\mid\widetilde{Y}^{\prime}=1)\|_2\bigg\}\\
				\leq&\frac{4c_x^2p}{n},
			\end{split}
		\end{equation*}
		where we use the inequality $\|\boldsymbol{a}\boldsymbol{a}^{T}-\boldsymbol{b}\boldsymbol{b}^{T}\|_F\leq\|\boldsymbol{a}(\boldsymbol{a}-\boldsymbol{b})^{T}\|_F+\|(\boldsymbol{b}-\boldsymbol{a})\boldsymbol{b}^{T}\|_F$ in the second inequality and we use $\|\cdot\|_2\leq\sqrt{\|\cdot\|_0}\cdot\|\cdot\|_{\infty}$ in the last inequality.
		
		Then, we consider the case where $\{\boldsymbol{x}_1^{\prime},Y_1^{\prime}\}$ belongs to the different slice from $\{\boldsymbol{x}_1,Y_1\}$. Without loss of generality, we assume that $\widetilde{Y}_1^{\prime}=2$. Then we have the decomposition,
		\begin{equation*}
			\begin{split}
				&\boldsymbol{\hat{M}}(\mathcal{S})-\boldsymbol{\hat{M}}(\mathcal{S}^{\prime})\\
				=&-\frac{(\sum_{i=1}^{n}\boldsymbol{x}_i\mathbbm{1}\{\widetilde{Y}_i=1\})(\sum_{i=1}^{n}\boldsymbol{x}_i\mathbbm{1}\{\widetilde{Y}_i=1\})^{\top}}{n^2}\frac{1/n}{\hat{p}_1(\hat{p}_1-1/n)}+\frac{\boldsymbol{x}_1\boldsymbol{x}_1^{\top}}{n^2\hat{p}_1}\\
				&+\frac{(\sum_{i=1}^{n}\boldsymbol{x}_i\mathbbm{1}\{\widetilde{Y}_i=2\})(\sum_{i=1}^{n}\boldsymbol{x}_i\mathbbm{1}\{\widetilde{Y}_i=2\})^{\top}}{n^2}\frac{1/n}{\hat{p}_2(\hat{p}_2+1/n)}-\frac{\boldsymbol{x}_1^{\prime}\boldsymbol{x}_1^{\prime \top}}{n^2(\hat{p}_2+1/n)}\\
				&+2\frac{\boldsymbol{x}_1n(\hat{p}_1-1/n)}{n^2\hat{p}_1}\mathbb{E}_n(\boldsymbol{x}^{\prime}\mid\widetilde{Y}^{\prime}=1)^{\top}-2\frac{\boldsymbol{x}_1^{\prime}n\hat{p}_2}{n^2(\hat{p}_2+1/n)}\mathbb{E}_n(\boldsymbol{x}\mid\widetilde{Y}=2)^{\top},
			\end{split}
		\end{equation*}
		by similar arguments in the proof of Lemma \ref{lem:low_privacy}. Thus, we have
		\begin{equation*}
			\begin{split}
				&\|\boldsymbol{\hat{M}}(\mathcal{S})-\boldsymbol{\hat{M}}(\mathcal{S}^{\prime})\|_F\\
				\leq&\frac{1}{n}\max_{h=1,2}\|\mathbb{E}_n(\boldsymbol{x}\mid\widetilde{Y}=h)\|_{2}\|\mathbb{E}_n(\boldsymbol{x}\mid\widetilde{Y}=h)\|_{2}(\frac{\hat{p}_1-1/n}{\hat{p}_1}+\frac{\hat{p}_2}{\hat{p}_2+1/n})\\
				&+\frac{1}{n^2}\max\{\|\boldsymbol{x}_1^{\prime}\|_{2}\|\boldsymbol{x}_1^{\prime}\|_{2},\|\boldsymbol{x}_1\|_{2}\|\boldsymbol{x}_1\|_{2}\}(\frac{1}{\hat{p}_1}+\frac{1}{\hat{p}_2+1/n})\\
				&+\frac{2}{n}\max\{\|\boldsymbol{x}_1^{\prime}\|_{2}\|\mathbb{E}_n(\boldsymbol{x}\mid\widetilde{Y}=2)\|_{2},\|\boldsymbol{x}_1\|_{2}\|\mathbb{E}_n(\boldsymbol{x}^{\prime}\mid\widetilde{Y}^{\prime}=1)\|_{2}\}(\frac{\hat{p}_1-1/n}{\hat{p}_1}+\frac{\hat{p}_2}{\hat{p}_2+1/n})\\
				\leq&\frac{c_x^2p}{n}\bigg(2+\frac{2}{np_H}+4\bigg)\leq\frac{7c_x^2p}{n}
			\end{split}
		\end{equation*}
		for $n$ large enough. In summary,
		\begin{equation*}
			\|\boldsymbol{\hat{M}}(\mathcal{S})-\boldsymbol{\hat{M}}(\mathcal{S}^{\prime})\|_F\leq\frac{7c_x^2p}{n}.
		\end{equation*}
		
		By Lemma \ref{lem:est_slices}, the estimated slices $\{\widehat{I}_h\}_{h=1}^{H}$ is $(\varepsilon,0)$-DP. Given the estimated slices $\{\widehat{I}_h\}_{h=1}^{H}$, by the Gaussian mechanism in Lemma \ref{lemma:laplace_gaussian_mechanism}, we conclude that reporting $\boldsymbol{\tilde{M}}$ is $(\varepsilon/2,\delta/2)$-DP and reporting $\boldsymbol{\tilde{\Sigma}}$ is $(\varepsilon/2,\delta/2)$-DP. By the composition theorem in Lemma \ref{lem:post_combin_dp}, we conclude that $(\boldsymbol{\tilde{\Sigma}},\boldsymbol{\tilde{M}})$ is $(2\varepsilon,\delta)$-DP. As a result, the estimates from the generalized eigen decomposition problem is also $(2\varepsilon,\delta)$-DP. 
		\hfill $\square$
	\end{proof}
	
	We then consider the error bound of the proposed estimator.
	
	\begin{proof}[Proof of Theorem \ref{thm:dpsir_errorbound_ini}]
		
		We first consider the error bounds of the private estimators $\boldsymbol{\tilde{\Sigma}}$ and $\boldsymbol{\tilde{M}}$. Recall that the non-private sample estimators of $\M$ and $\boldsymbol{\Sigma}$ are,
		\begin{equation*}
			\boldsymbol{\hat{M}}=\sum_{h=1}^{H}\hat{p}_{h}\{\mathbb{E}_n(\boldsymbol{x}\mid\tilde{Y}=h)-\mathbb{E}_n(\boldsymbol{x})][\mathbb{E}_n(\boldsymbol{X}\mid\tilde{Y}=h)-\mathbb{E}_n(\boldsymbol{x})\}^{\top},
		\end{equation*}
		and
		\begin{equation*}
			\boldsymbol{\hat{\Sigma}}=\frac{1}{n}\sum_{i=1}^{n}\{\boldsymbol{x}_i-\mathbb{E}_n(\boldsymbol{x}_i)\}\{\boldsymbol{x}_i-\mathbb{E}_n(\boldsymbol{x}_i)\}^{\top},
		\end{equation*}
		respectively. We first consider the term corresponding to the statistical error of $\boldsymbol{\Sigma}$: $\|\boldsymbol{\hat{\Sigma}}-\boldsymbol{\Sigma}\|_{op}$. By the proof of Theorem \ref{thm:dpsir_errorbound}, we conclude
		\begin{equation*}
			\|\boldsymbol{\hat{\Sigma}}-\boldsymbol{\Sigma}\|_{op}\leq C_1\sqrt{\frac{p}{n}},
		\end{equation*}
		with probability at least $1-\exp\{-C\log(\max\{p,n\})\}$ for two positive constant $C,C_1$. By the strong local semicircle law (Theorem 2.1 (iv) in \cite{erdHos2012rigidity}), the operator norm of additional noise $\boldsymbol{E}_1$ satisfies $$\|\boldsymbol{E}_1\|_{op}\leq C_1\sqrt{p}\sigma_1\frac{2\sqrt{2\log(2.5/\delta)}}{\varepsilon},$$
		with probability at least $1-\exp\{-C\log(p)\}$. Furthermore,
		\begin{equation*}
			\|\boldsymbol{\tilde{\Sigma}}-\boldsymbol{\hat{\Sigma}}\|_{op}\leq C_1\sqrt{\frac{p}{n}}+C_1\sigma_1\sqrt{p}\frac{2\sqrt{2\log(2.5/\delta)}}{\varepsilon},
		\end{equation*}
		with probability approaching $1$ as $p\to\infty$. Then we consider the term $\|\M-\boldsymbol{\hat{M}}\|_{op}$. By the result of Lemma \ref{lem:est_slices}, the expected number of observations in each estimated slice is lower bound by a constant. By the proof of Theorem \ref{thm:dpsir_errorbound}, we have
		\begin{equation*}
			\begin{split}
				\|\M-\boldsymbol{\hat{M}}\|_{op}\leq C_1\sqrt{\frac{p}{n}},
			\end{split}
		\end{equation*}
		with probability at least $1-\exp\{-C\log(\max\{p,n\})\}$. By the strong local semicircle law (Theorem 2.1 (iv) in \cite{erdHos2012rigidity}), the operator norm of additional noise $\boldsymbol{E}_2$ satisfies 
		$$\|\boldsymbol{E}_2\|_{op}\leq C_1\sqrt{p}\sigma_2\frac{2\sqrt{2\log(2.5/\delta)}}{\varepsilon},$$
		with probability at least $1-\exp\{-C\log(p)\}$. Furthermore, we conclude
		\begin{equation*}
			\|\boldsymbol{\tilde{M}}-\M\|_{op}\leq\|\boldsymbol{\hat{M}}-\M\|_{op}+\|\boldsymbol{E}_2\|_{op}\leq C_1\sqrt{\frac{p}{n}}+C_1\sigma_2\sqrt{p}\frac{2\sqrt{2\log(2.5/\delta)}}{\varepsilon}.
		\end{equation*}
		Furthermore, the estimated eigenvalues satisfy
		\begin{equation*}
			\begin{split}
				|\lambda_{K+1}-\tilde{\lambda}_{K+1}|&\leq\|\boldsymbol{\Sigma}^{-1/2}\M\boldsymbol{\Sigma}^{-1/2}-\boldsymbol{\tilde{\Sigma}}^{-1/2}\boldsymbol{\tilde{M}}\boldsymbol{\tilde{\Sigma}}^{-1/2}\|_{op}\\
				&\leq\|\boldsymbol{\tilde{\Sigma}}^{-1/2}(\boldsymbol{\Sigma}^{1/2}-\boldsymbol{\tilde{\Sigma}}^{1/2})\boldsymbol{\Sigma}^{-1/2}\M\boldsymbol{\Sigma}^{-1/2}\|_{op}+\|\boldsymbol{\tilde{\Sigma}}^{-1/2}(\M-\boldsymbol{\tilde{M}})\boldsymbol{\Sigma}^{-1/2}\|_{op}\\
				&+\|\boldsymbol{\tilde{\Sigma}}^{-1/2}\boldsymbol{\tilde{M}}\boldsymbol{\tilde{\Sigma}}^{-1/2}(\boldsymbol{\Sigma}^{-1/2}-\boldsymbol{\tilde{\Sigma}}^{-1/2})\boldsymbol{\Sigma}^{1/2}\|_{op}\\
				&\leq\|\boldsymbol{\tilde{\Sigma}}^{-1/2}\|_{op}\|\boldsymbol{\Sigma}^{1/2}-\boldsymbol{\tilde{\Sigma}}^{1/2}\|_{op}\|\boldsymbol{\Sigma}^{-1/2}\M\boldsymbol{\Sigma}^{-1/2}\|_{op}\\
				&+\|\boldsymbol{\tilde{\Sigma}}^{-1/2}\|_{op}\|\M-\boldsymbol{\tilde{M}}\|_{op}\|\boldsymbol{\Sigma}^{-1/2}\|_{op}\\
				&+\|\boldsymbol{\tilde{\Sigma}}^{-1/2}\boldsymbol{\tilde{M}}\boldsymbol{\tilde{\Sigma}}^{-1/2}\|_{op}\|\boldsymbol{\Sigma}^{1/2}-\boldsymbol{\tilde{\Sigma}}^{1/2}\|_{op}\|\boldsymbol{\Sigma}^{-1/2}\|_{op}\\
				&\leq C_1\bigg\{L\sqrt{\frac{p}{n}}+(\sigma_1+\sigma_2)\sqrt{p}\frac{2\sqrt{2\log(2.5/\delta)}}{\varepsilon}\bigg\},
			\end{split}
		\end{equation*}
		where we use Lemma 2.1 in \cite{sibson1979studies} in the first inequality; we use the triangle inequality in the second inequality; we use the sub-multiplicativity of matrix norm in the third inequality. By the proof of Lemma \ref{lem:est_slices}, the generalized eigenvalues corresponding to the estimated slices satisfies $\lambda_k\geq\lambda/2$. Then, $\tilde{\lambda}_{k+1}\leq \lambda_k/2$ holds with a probability approaching $1$,	where we use the assumption $\lambda_{k+1}=0$, and $p/(n\lambda^2),(\sigma_1+\sigma_2)\sqrt{p}\sqrt{\log(2.5/\delta)}/(\varepsilon\lambda)$ are small enough.
		
		Then, we have
		\begin{equation*}
			\Delta:=\frac{\lambda_k-\tilde{\lambda}_{k+1}}{\sqrt{(1+\lambda_1^2)(1+\tilde{\lambda}_{k+1}^2)}}\geq\frac{\lambda_k}{2\sqrt{(1+\lambda_1^2)(1+\lambda_k^2/4)}}.
		\end{equation*}
		Then,
		\begin{equation*}
			c(\M,\boldsymbol{\Sigma}):=\inf_{\|\boldsymbol{x}\|_2=1}\sqrt{(\boldsymbol{x}^{\top}\M\boldsymbol{x})^2+(\boldsymbol{x}^{\top}\boldsymbol{\Sigma}\boldsymbol{x})^2}\geq\lambda_{min}(\boldsymbol{\Sigma}),
		\end{equation*}
		\begin{equation*}
			c(\boldsymbol{\tilde{M}},\boldsymbol{\tilde{\Sigma}}):=\inf_{\|\boldsymbol{x}\|_2=1}\sqrt{(\boldsymbol{x}^{\top}\boldsymbol{\tilde{M}}\boldsymbol{x})^2+(\boldsymbol{x}^{\top}\boldsymbol{\tilde{\Sigma}}\boldsymbol{x})^2}\geq\lambda_{min}(\boldsymbol{\Sigma})/2,
		\end{equation*}
		where we use the fact $\|\boldsymbol{\tilde{\Sigma}}-\boldsymbol{\Sigma}\|_{op}\to 0$. Let $\epsilon_{noisy}=\sqrt{\|\M-\boldsymbol{\tilde{M}}\|_{op}^2+\|\boldsymbol{\tilde{\Sigma}}-\boldsymbol{\Sigma}\|_{op}^2}$. Then by Theorem 3.1 in \cite{sun1983perturbation} and Corollary 4.5 in \cite{stewart1979pertubation},
		\begin{align*}
			\|\mathbf{P}-\mathbf{\hat{P}}\|_F^2&\leq\bigg\{ \frac{\epsilon_{noisy}\sqrt{k}}{\Delta c(\M,\boldsymbol{\Sigma})c(\boldsymbol{\hat{M}},\boldsymbol{\hat{\Sigma}})}\sqrt{\|\M\|_{op}^2+\|\boldsymbol{\Sigma}\|_{op}^2}\bigg\}^2\\
			&\leq C_2\frac{p}{\lambda^2 n}+C_2\frac{p^{3}\log(1/\delta)}{n^2\varepsilon^2\lambda^2},
		\end{align*}
		for a positive constant $C_2$ where $\mathbf{P}$ and $\hat{\mathbf{P}}$ are the projection matrix on the eigen-subspace of matrix pair $(\M,\boldsymbol{\Sigma})$ and $(\boldsymbol{\tilde{M}},\boldsymbol{\tilde{\Sigma}})$.
		\hfill $\square$
	\end{proof}

	Then, we show the consistency of the selection based on DP-BIC.
	
	\begin{proof}[Proof of Theorem \ref{thm:sir_bic}]
		
		By the proof Theorem \ref{thm:dpsir_errorbound_ini}, we have
		\begin{equation*}
			|\lambda_i-\tilde{\lambda}_i|=O_p\bigg(\frac{1}{\lambda}\sqrt{\frac{p}{n}}+\frac{p^{3/2}\sqrt{\log(1/\delta)}}{n\varepsilon\lambda}\bigg)
		\end{equation*}
		for $i=1,\dots,H.$ Because the number of nonzero eigenvalues of $\M$ is $k$, the eigenvalues $\lambda_{l}=0$ for $l>k$. Then,
		\begin{equation*}
			\tilde{\lambda}_l^2=|\tilde{\lambda}_l-\lambda_l|^2=O_p\bigg(\frac{1}{\lambda}\sqrt{\frac{p}{n}}+\frac{p^{3/2}\sqrt{\log(1/\delta)}}{n\varepsilon\lambda}\bigg)^2,
		\end{equation*}
		for $l>k$. Furthermore, by the convergence of $\tilde{\lambda}_l$ to $\lambda_l$, we have $\sum_{i=1}^{H}\tilde{\lambda}_{i}^{2}\to\sum_{i=1}^{H}\lambda_{i}^{2}>C_1\lambda^2$ for a positive constant $C_1$ with probability approaching $1$ as $n,p\to\infty$. Similarly, we have $\sum_{i=1}^{H}\tilde{\lambda}_i^2\leq C_2\lambda^2$ for a positive constant $C_2$ with probability approaching $1$ as $n,p\to\infty$.
		
		For $l<k$, we consider the difference between $G(k)-G(l)$,
		\begin{align*}
			G(k)-G(l)&=n\sum_{i=l+1}^{k}\tilde{\lambda}_{i}^2/\sum_{i=1}^{H}\tilde{\lambda}_i^2-C_n\{k(k+1)-l(l+1)\}/2.
		\end{align*}
		Note that we have
		$$\sum_{i=l+1}^{k}\tilde{\lambda}_i^2\geq \sum_{i=l+1}^{k}\lambda_i^2/2>\lambda^2/2,$$
		with probability approaching $1$, where we use the convergence of $\tilde{\lambda}$ and Assumption \ref{assump:low_SIR}. Then, we have 
		$$\sum_{i=l+1}^{k}\tilde{\lambda}_{i}^2/\sum_{i=1}^{p}\tilde{\lambda}_i^2\geq C_3,$$
		for a constant $C_3>0$. By the condtion $C_n/n\to 0$, we have $G(k)-G(l)>0$. On the other side, For $l>k$,
		\begin{align*}
			G(k)-G(l)&=-n\sum_{i=k+1}^{l}\tilde{\lambda}_{i}^2/\sum_{i=1}^{H}\tilde{\lambda}_i^2+C_n\{l(l+1)-k(k+1)\}/2.
		\end{align*}
		By the fact 
		$$n\sum_{i=k+1}^{l}\tilde{\lambda}_{i}^2=O_p\bigg(\frac{p}{\lambda^2}+\frac{p^3\log(1/\delta)}{n\varepsilon^2\lambda^2}\bigg),$$
		and the condition $C_n\lambda^2/\{p+p^3\log(1/\delta)/(n\varepsilon^2)\}\to\infty$, we have $G(k)-G(l)>0$ with probability approaching $1$ as $n,p\to\infty$. The desired result follows from the two inequalities.
		\hfill $\square$
	\end{proof}
	
	\section{Proof of the Upper Bound for DP-SSIR when $p=o(e^n)$}\label{sec: high_dim_proof}
	\subsection{Proof of the Estimator in Algorithm \ref{alg:hd_dp_sir}}
	We begin the proof of Lemma \ref{lem:peeling} which serves as the basis of Algorithm \ref{alg:hd_dp_sir}.
	
	\begin{proof}[Proof of Lemma \ref{lem:peeling}]
		
		Algorithm \ref{alg:vector_noisy_ht} consists of two steps: first, it selects $s^{\prime}$ indices from a total of $d_2$ indices; second, it returns a $d_1\times s^{\prime}$ matrix. We consider the two terms separately. By the assumption, the $\ell_{\infty}$ sensitivity of $\boldsymbol{A}(\cdot)$ is bounded by $\sigma$. Then, the sensitivity of $\|\boldsymbol{A}(D)_{*,j}\|_2$ satisfies
		\begin{equation*}
			\bigg|\sqrt{\sum_{i=1}^{d_1}A(D)_{i,j}^2}-\sqrt{\sum_{i=1}^{d_1}A(D^{\prime})_{i,j}^2}\bigg|\leq\sqrt{\sum_{i=1}^{d_1}\{A(D)_{i,j}-A(D^{\prime})_{i,j}\}^2}\leq\sqrt{d_1}\sigma,
		\end{equation*}
		where $D^{\prime}$ is a neighboring data set of $D$, and we use the triangle inequality. By the Theorem 3 in \cite{dwork2021differentially}, output the index set $S$ is $(\varepsilon/2,\delta/2)$-DP. Similarly, the $\ell_2$ sensitivity of the $d_1\times s^{\prime}$ matrix $\boldsymbol{A}(D)_{*,S}$ is bounded by $\sqrt{d_1s^{\prime}}\sigma$. By the Gaussian mechanism, output the randomly perturbed matrix $\boldsymbol{A}(D)_{*,S}+\boldsymbol{\tilde{w}}$ is $(\varepsilon/2,\delta/2)$-DP. By the DP composition theorem, the output of Algorithm \ref{alg:vector_noisy_ht} is $(\varepsilon,\delta)$-DP.
		\hfill $\square$
	\end{proof}
	
	Then, we show that the estimator in Algorithm \ref{alg:hd_dp_sir} is $(2\varepsilon,\delta)$-DP.
	
	\begin{proof}[Proof of Lemma \ref{lem:high_privacy}]
		
		The gradient in the proposed Algorithm \ref{alg:hd_dp_sir} is defined on Line 4:
		\begin{equation*}
			2\eta\{-\boldsymbol{\hat{M}}\boldsymbol{B}+\lambda_{penalty}\boldsymbol{\hat{\Sigma}}\boldsymbol{B}(\boldsymbol{B}^{\top}\boldsymbol{\hat{\Sigma}}\boldsymbol{B}-\boldsymbol{I}_k)\}.
		\end{equation*}
		By the proof of Lemma \ref{lem:low_privacy}, the $l_{\infty}$ sensitivity is bounded by 
		\begin{equation*}
			2\eta\{7c_xR+\lambda_{penalty}(2c_xR+4kc_xR^3)\}\frac{T}{n}.
		\end{equation*}
		By applying Lemma \ref{lem:peeling}, each iteration is $(\varepsilon/T,\delta/T)$-differential private. By the DP composition theorem, the gradient descent of the algorithm (Line 3 to Line 6)is $(\varepsilon,\delta)$-DP. Finally, by Lemma \ref{lem:est_slices} and the DP composition theorem, the proposed estimator is $(2\varepsilon,\delta)$-DP.

		\hfill $\square$
	\end{proof}
	
	Next, we are ready to prove the main theorem.
	
	\begin{proof}[Proof of Theorem \ref{thm:dp_sparse_sir_conv}]
		
		The proof consists of several parts. Firstly, we analyze the convergence of sample estimators of $\M$ and $\boldsymbol{\Sigma}$. Secondly, we demonstrate that the gradient descent step leads to a substantial reduction in loss, and the noisy hard thresholding incurs a relatively small loss.
		
		\textbf{Step 1: Estimation of $\M$ and $\boldsymbol{\Sigma}$}
		
		We begin with the lemma about convergence properties of two sample estimators $\boldsymbol{\hat{M}}$ and $\boldsymbol{\hat{\Sigma}}$. Let $\mathcal{I}\subset\{1,\dots,p\}$ be a subset of covariates. Let $\boldsymbol{\hat{B}}(\mathcal{I})$ be the optimal solution of the following optimization problem:
		\begin{equation}
			\label{eq:fixI}
			\max\langle\boldsymbol{\hat{M}},\B\B^{\top}\rangle\text{ s.t. }\B^{\top}\boldsymbol{\hat{\Sigma}}\B=\boldsymbol{I}_k\text{ and }\text{supp}(\B)\subset\mathcal{I},
		\end{equation}
		where we use $\mathcal{I}$ to indicate the dependence on the subset $\mathcal{I}$. The first lemma proves the convergence rate of estimators $\boldsymbol{\hat{M}}_{\mathcal{I}\mathcal{I}}$ and $\boldsymbol{\hat{\Sigma}}_{\mathcal{I}\mathcal{I}}$, where $\boldsymbol{\hat{M}}_{\mathcal{I}\mathcal{I}}$ and $\boldsymbol{\hat{\Sigma}}_{\mathcal{I}\mathcal{I}}$ are $\mathcal{I}\times\mathcal{I}$ submatrices of $\boldsymbol{\hat{M}}$ and $\boldsymbol{\hat{\Sigma}}$, respectively.
		\begin{lemm}
			\label{lemma:estS}
			For any constant $C^{\prime}>0$, there exists some constant $C>0$, such that
			\begin{equation}
				\label{eq:estm}
				\|\boldsymbol{\hat{M}}_{\mathcal{I}\mathcal{I}}-\M_{\mathcal{I}\mathcal{I}}\|_{op}^2\leq\frac{C}{n}s^{\prime}\log(ep/s^{\prime}),
			\end{equation}
			\begin{equation}
				\label{eq:estsigma}
				\|\boldsymbol{\hat{\Sigma}}_{\mathcal{I}\mathcal{I}}-\boldsymbol{\Sigma}_{\mathcal{I}\mathcal{I}}\|_{op}^2\leq\frac{C}{n}s^{\prime}\log(ep/s^{\prime}),
			\end{equation}
			with probability at least $1-\exp\{-C^{\prime}s^{\prime}\log(ep/s^{\prime})\}$, for any $\mathcal{I}\subset \{1,\dots,p\}$, $|\mathcal{I}|\leq 2s^{\prime}+s$ and $s\leq s^{\prime}$, where $|\mathcal{I}|$ is the cardinality of $\mathcal{I}$ and $s$ is the true sparsity. For simplicity, we denote the event where (\ref{eq:estm}) and (\ref{eq:estsigma}) hold simultaneously by $\mathcal{E}_1$.
		\end{lemm}
		
		Under the conclusion of Lemma \ref{lemma:estS}, we study the property of a near oracle estimator. The name ``near oracle" comes from the fact $\mathcal{S}\subset\mathcal{I}$ and $|\mathcal{I}|/n\to 0$ as $n\to\infty$, where $\mathcal{S}$ is the true support of $\B$.
		\begin{lemm}
			\label{lemma:oracle}
			Under the conditions in Lemma \ref{lemma:estS}, we assume $\sqrt{s^{\prime}\log(ep/s^{\prime})/n}\leq c$ and $s^{\prime}\log(ep/s^{\prime})/(n\lambda^2)\leq c$ for $c$ small enough, where $C^{\prime\prime}$ only depends on $C,C^{\prime},\lambda_x$. For any constants $C>0$, there exists some constant $C^{\prime}>0$,
			\begin{equation}
				\label{eq:oracle}
				\|\B(\B^{\top}\B)^{-1}\B^{\top}-\boldsymbol{\hat{B}}(\mathcal{I})(\boldsymbol{\hat{B}}(\mathcal{I})^{\top}\boldsymbol{\hat{B}}(\mathcal{I}))^{-1}\boldsymbol{\hat{B}}(\mathcal{I})^{\top}\|_F\leq C\frac{\sqrt{s^{\prime}\log(ep/s^{\prime})/n}}{\lambda},
			\end{equation}
			holds with probability at least $1-\exp\{-C^{\prime}s^{\prime}\log(ep/s^{\prime})\}$, where $\B$ is the true parameter and the index set $\mathcal{I}$ satisfies $\mathcal{S}\subset\mathcal{I}$, $|\mathcal{I}|\leq 2s^{\prime}+s$, and $s\leq s^{\prime}$. For simplicity, we denote the event (\ref{eq:oracle}) by $\mathcal{E}_2$.
		\end{lemm}
		
		We make a few comments before the next step. The lemma has two conditions on the parameters. The first assumption on $\sqrt{s^{\prime}\log(ep/s^{\prime})/n}$ is frequently assumed in the high-dimensional literature. The second order assumption on $s^{\prime}\log(ep/s^{\prime})/(n\lambda^2)$ is the minimal signal condition on SIR, which is also assumed by \cite{tan2020sparse}. The results in Lemma \ref{lemma:estS} and \ref{lemma:oracle} hold uniformly for all $\mathcal{I}\subset\{1,\dots,p\}.$ Without specification, we will assume the event $\mathcal{E}_1\cap\mathcal{E}_2$ holds in the following proof.
		
		\textbf{Step 2: Convergence of Gradient Descent}
		
		In this step, we analyze the convergence of the proposed algorithm. The algorithm contains two parts, gradient descent and hard thresholding. We provide two lemmas to illustrate the two parts respectively.
		
		First, we consider the gradient descent step. Notice that in the $t$-th iteration, we only need to consider the support of $\boldsymbol{\hat{B}}^{(t)}$, $\boldsymbol{\hat{B}}^{(t+1)}$ and $\B$. For any index set $\mathcal{S}_t$, we use subscript $ \B_{\mathcal{S}_t}$ to denote the sub-matrix to simplify the notation. For two matrices $\boldsymbol{X},\boldsymbol{Y}\in\mR^{p\times r}$, the distance measure $\text{dist}(\cdot,\cdot)$ is defined by
		\begin{equation*}
			\text{dist}(\boldsymbol{X},\boldsymbol{Y})=\min_{\boldsymbol{H}\in\mathbb{O}^{r\times r}}\|\boldsymbol{X}\boldsymbol{H}-\boldsymbol{Y}\|_F,
		\end{equation*}
		where $\mathbb{O}^{r\times r}$ is space containing all $r\times r$ orthogonal matrix. The next lemma shows that the distance between the oracle estimator and the current estimator decreases linearly.
		
		\begin{lemm}
			\label{lem:grad_desc}
			Assume event $\mathcal{E}_1\cap\mathcal{E}_2$ holds for $\mathcal{I}=\text{supp}(\boldsymbol{\hat{B}}^{(t)})\cup\text{supp}(\boldsymbol{\hat{B}}^{(t+1)})\cup\mathcal{S}$. For clarity, let $\mathcal{S}_t=\text{supp}(\boldsymbol{\hat{B}}^{(t)})\cup\text{supp}(\boldsymbol{\hat{B}}^{(t+1)})\cup\mathcal{S}$ be the index set which covers the support of the parameter used in the $t$-th iteration. Define
			\begin{equation}
				\boldsymbol{\hat{A}}(\mathcal{S}_t):=\boldsymbol{\hat{B}}(\mathcal{S}_t)\bigg\{\boldsymbol{I}+\frac{\boldsymbol{\hat{\Lambda}}_k(\mathcal{S}_t)}{\lambda_{penalty}}\bigg\}^{1/2},
			\end{equation}
			where $\lambda_{penalty}=\lambda_1/c$ and $c $ is a fixed constant, $\boldsymbol{\hat{B}}(\mathcal{S}_t)$ is near oracle estimator defined in (\ref{eq:fixI}) and $\boldsymbol{\hat{\Lambda}}_k(\mathcal{S}_t)$ is the diagonal matrix with top $k$ eigenvalues corresponding to $\boldsymbol{\hat{B}}(\mathcal{S}_t)$. Let the step size
			\begin{equation*}
				\eta\leq\frac{C_1}{\lambda_1(\lambda_x+2)},
			\end{equation*}
			and
			\begin{equation}
				\text{dist}(\boldsymbol{\hat{B}}^{(t)},\boldsymbol{\hat{A}}(\mathcal{S}_t))\leq\frac{1}{\sqrt{\lambda_x}}\min\bigg\{\frac{\lambda_k}{C_2\lambda_1\lambda_x^2},\frac{\sqrt{1+2c}}{4}\bigg\},
			\end{equation}
			then after the iteration,
			\begin{equation}
				\text{dist}(\boldsymbol{\hat{B}}^{(t+0.3)}_{\mathcal{S}_t},\boldsymbol{\hat{A}}(\mathcal{S}_t))^2\leq\bigg(1-\frac{\eta\lambda_k}{4\lambda_x}\bigg)\text{dist}(\boldsymbol{\hat{B}}^{(t)},\boldsymbol{\hat{A}}(\mathcal{S}_t))^2,
			\end{equation}
			where $\boldsymbol{\hat{B}}^{(t+0.3)}_{\mathcal{S}_t}\in\mR^{p\times r}$ is a matrix that has the same entries as those in $\boldsymbol{\hat{B}}^{(t+0.3)}$ on $\mathcal{S}_t\times r$ before thresholding and zeros elsewhere.
		\end{lemm}
		Lemma \ref{lem:grad_desc} illustrates the convergence property of the gradient descent step. Notice that Lemma \ref{lem:grad_desc} only considers the distance concerning the distance measure $\text{dist}(\cdot,\cdot)$, which is rotation invariant. Lemma \ref{lem:grad_desc} is a special case of Theorem 7 in \cite{gao2021sparse}.
		
		\textbf{Step 3: Peeling Mechanism}
		
		Then, we consider the effect of the peeling algorithm.
		\begin{lemm}
			\label{lem:peeling_effect}
			Let $\boldsymbol{A}_0=\B(\boldsymbol{I}_k+\boldsymbol{\Lambda}_k/\lambda_{penalty})^{1/2}$, where $\B$ is the true parameter and $\boldsymbol{\Lambda}_k$ is a diagonal matrix with top $k$ eigenvalues and . we have
			\begin{equation*}
				\text{dist}(\boldsymbol{\hat{B}}^{(t+0.6)},\boldsymbol{A}_0)\leq\bigg\{1+2\sqrt{\frac{2s}{s^\prime}}\bigg(1+\sqrt{\frac{2s}{s^\prime}}\bigg)\bigg\}^{1/2}\text{dist}(\boldsymbol{\hat{B}}^{(t+0.3)},\boldsymbol{A}_0)+\epsilon_{privacy},
			\end{equation*}
			where
			\begin{equation}
				\label{eq:e_privacy}
				\epsilon_{privacy}=C\bigg\{\sqrt{\frac{2s}{s^\prime}}\bigg(1+\sqrt{\frac{2s}{s^{\prime}}}\bigg)s\log(p)^2\bigg\}^{1/2}\sigma_1+C\sqrt{ks^{\prime}}\sigma_2\log(p),
			\end{equation}
			and $C$ is a positive constant, the $\sigma_1,\sigma_2$ are defined as $\sigma_1=2\sigma\sqrt{3ks^{\prime}\log(2\cdot T/\delta)}/(\varepsilon/T)$ and $\sigma_2=2\sigma\sqrt{2ks^{\prime}\log(2.5\cdot T/\delta)}/(\varepsilon/T)$.
		\end{lemm}
		
		\textbf{Step 4: Mathematical Induction}
		
		Finally, we are in the position to prove the main theorem. We use mathematical induction to show the main result. Similarly to the proof of Lemma \ref{lem:grad_desc}, we define the effective support and denote $\mathcal{F}_t:=\text{supp}(\boldsymbol{\hat{B}}^{(t)})$. Then, let
		\begin{equation*}
			\mathcal{S}_t:=\mathcal{F}_t\cap\mathcal{S}\cap\mathcal{F}_{t+1}.
		\end{equation*}
		The gradient descent procedure restricted the solution to $\mathcal{S}_t$ is equivalent to
		\begin{equation*}
			\boldsymbol{\hat{B}}^{(t+0.3)}_{\mathcal{S}_t}=\boldsymbol{\hat{B}}^{(t)}_{\mathcal{S}_t}-2\eta\{-\boldsymbol{\hat{M}}_{\mathcal{S}_t\mathcal{S}_t}\boldsymbol{\hat{B}}^{(t)}_{\mathcal{S}_t}+\lambda_{penalty}\boldsymbol{\hat{\Sigma}}_{\mathcal{S}_t\mathcal{S}_t}\boldsymbol{\hat{B}}^{(t)}_{\mathcal{S}_t}(\boldsymbol{\hat{B}}^{(t)\top}_{\mathcal{S}_t}\boldsymbol{\hat{\Sigma}}_{\mathcal{S}_t\mathcal{S}_t}\boldsymbol{\hat{B}}^{(t)}_{\mathcal{S}_t}-\boldsymbol{I}_k)\}.
		\end{equation*}
		Suppose we have
		\begin{equation*}
			\text{dist}(\boldsymbol{\hat{B}}^{(t)},\boldsymbol{A}_0)\leq\xi^{t-1}\text{dist}(\boldsymbol{\hat{B}}^{(t-1)},\boldsymbol{A}_0)+\epsilon_{stat}+\epsilon_{privacy},
		\end{equation*}
		where 
		\begin{equation*}
			\epsilon_{stat}:=\frac{C}{1-\xi}\frac{\sqrt{Tks^{\prime}\log(ep/s^{\prime})/n}}{\lambda_k},
		\end{equation*}
		\begin{equation*}
			\epsilon_{privacy}:=\frac{1}{1-\xi}\epsilon_{privacy}^0,
		\end{equation*}
		$\epsilon_{privacy}^0$ is defined in \eqref{eq:e_privacy} and $\xi$ will be defined later. For $t=0$, it's the initial conditions. For $t+1$, we first check the condition in the proposition holds for $t+1$. By the Lemma \ref{lem:grad_desc},
		\begin{equation*}
			\text{dist}(\boldsymbol{\hat{B}}^{(t+0.3)}_{\mathcal{P}_{t+1}},\boldsymbol{\hat{A}}(\mathcal{S}_t))\leq\sqrt{1-\alpha\eta}\text{dist}(\boldsymbol{\hat{B}}^{(t)},\boldsymbol{\hat{A}}(\mathcal{S}_t))\leq(1-\alpha\eta/2)\text{dist}(\boldsymbol{\hat{B}}^{(t)},\boldsymbol{\hat{A}}(\mathcal{S}_t)),
		\end{equation*}
		where
		$$\alpha=\frac{\lambda_k}{4\lambda_x},\beta=C\lambda_1(1+2\lambda_x)\text{ and }\eta\leq\beta^{-1},$$
		are defined in Lemma \ref{lem:grad_desc}, and we use Taylor expansion in the previous equation. By the definition of $\boldsymbol{A}_0$ and $\boldsymbol{\hat{A}}(\mathcal{S}_t)$, we have
		\begin{equation*}
			\text{dist}(\boldsymbol{A}_0,\boldsymbol{\hat{A}}(\mathcal{S}_t))\leq\|\boldsymbol{A}_0-\boldsymbol{\hat{A}}(\mathcal{S}_t)\|_F\leq\frac{C}{\lambda_k}\sqrt{\frac{s^{\prime}\log(p/s^{\prime})}{n/T}},
		\end{equation*}
		with probability at least $1-\exp\{-\log(p)\}$. By the triangle inequality, we have
		\begin{equation*}
			\text{dist}(\boldsymbol{\hat{B}}^{(t+0.3)},\boldsymbol{A}_0)\leq(1-\alpha\eta/2)\text{dist}(\boldsymbol{\hat{B}}^{(t)},\boldsymbol{A}_0)+\frac{C}{\lambda_k}\sqrt{\frac{s^{\prime}\log(p/s^{\prime})}{n/T}}.
		\end{equation*}
		After the matrix-noisy hard thresholding procedure, by Lemma \ref{lem:peeling_effect}, we have
		\begin{align*}
			\text{dist}(\boldsymbol{\hat{B}}^{(t+0.6)},\boldsymbol{A}_0)\leq\bigg\{1+2\sqrt{\frac{2s}{s^\prime}}\bigg(1+\sqrt{\frac{2s}{s^\prime}}\bigg)\bigg\}^{1/2}\text{dist}(\boldsymbol{\hat{B}}^{(t+0.3)}_{\mathcal{P}_{t+1}},\boldsymbol{A}_0)+(1-\xi)\epsilon_{privacy}.
		\end{align*}
		Combining the previous results, we have
		\begin{align*}
			\text{dist}(\boldsymbol{\hat{B}}^{(t+1)},\boldsymbol{A}_0)&\leq\bigg\{1+2\sqrt{\frac{2s}{s^\prime}}\bigg(1+\sqrt{\frac{2s}{s^\prime}}\bigg)\bigg\}^{1/2}(1-\alpha\eta/2)\text{dist}(\boldsymbol{\hat{B}}^{(t)},\boldsymbol{A}_0)\\
			&+\bigg\{1+2\sqrt{\frac{2s}{s^\prime}}\bigg(1+\sqrt{\frac{2s}{s^\prime}}\bigg)\bigg\}^{1/2}\frac{2C}{\lambda_k}\sqrt{\frac{s^{\prime}\log(p/s^{\prime})}{n/T}}\\
			&+(1-\xi)\epsilon_{privacy},
		\end{align*}
		where we use the fact that the truncation didn't happen with probability approaching $1$ as $n,p\to\infty$. Then, define
		\begin{equation*}
			\xi=\bigg\{1+2\sqrt{\frac{s}{s^\prime}}\bigg(1+\sqrt{\frac{s}{s^\prime}}\bigg)\bigg\}^{1/2}(1-\alpha\eta/2),
		\end{equation*}
		To ensure $\xi<1$, by solving the quadratic inequality, we have
		\begin{equation*}
			\begin{split}
				\frac{2s}{s^{\prime}}\leq\frac{1}{4}\bigg\{\sqrt{\frac{2}{(1-\alpha\eta/2)^2}-1}-1\bigg\}^2,
			\end{split}
		\end{equation*}
		where we use the fact $(1-\alpha\eta/2)<1$ by the choice of $\eta$. Then, we further obtain
		\begin{equation*}
			\text{dist}(\boldsymbol{\hat{B}}^{(t+1)},\boldsymbol{A}_0)\leq\xi\text{dist}(\boldsymbol{\hat{B}}^{(t)},\boldsymbol{A}_0)+(1-\xi)(\epsilon_{stat}+\epsilon_{privacy}).
		\end{equation*}
		By iterating the previous equation $t$ times, we have
		\begin{equation*}
			\text{dist}(\boldsymbol{\hat{B}}^{(t)},\boldsymbol{A}_0)\leq\xi^{t}\text{dist}(\boldsymbol{\hat{B}}^{(0)},\boldsymbol{A}_0)+\epsilon_{stat}+\epsilon_{privacy}.
		\end{equation*}
		For $T=O\{\log(n)\}$, the first term is smaller than the third term. Thus, by the choice of $\sigma$ in Lemma \ref{lem:low_privacy}, we have
		\begin{equation*}
			\text{dist}(\boldsymbol{\hat{B}}^{(T)},\boldsymbol{A})^2\leq C^{\prime}\frac{s\log(p)\log(n)}{\lambda^2n}+C^{\prime}\frac{s^2\log^2(p)\log(1/\delta)\log^7(n)}{\lambda^2n^2\varepsilon^2},
		\end{equation*}
		for a universal constant $C^{\prime}$. Furthermore, by the proof of Corollary 4.2 in \cite{gao2021sparse}, we have for a positive constant $C$, the following holds with probability greater than $1-\exp(-C^{\prime}p)$ as $n,p\to\infty$:
		\begin{equation*}
			\text{dist}(\boldsymbol{\hat{B}}^{(T)}(\boldsymbol{\hat{B}}^{(T)\top}\boldsymbol{\Sigma}\boldsymbol{\hat{B}}^{(T)})^{-1/2},\boldsymbol{B})^2\leq \frac{C}{2\lambda_x}\frac{s\log(p)\log(n)}{\lambda^2n}+\frac{C}{2\lambda_x}\frac{s^{2}\log^2(p)\log(1/\delta)\log^7(n)}{\lambda^2n^2\varepsilon^2}.
		\end{equation*}
		We use the notation $\boldsymbol{\hat{B}}^{(n)}:=\boldsymbol{\hat{B}}^{(T)}(\boldsymbol{\hat{B}}^{(T)\top}\boldsymbol{\Sigma}\boldsymbol{\hat{B}}^{(T)})^{-1/2}$ to denote the normalized matrix. By the Hölder's inequality, we have
		\begin{equation*}
			\text{dist}(\boldsymbol{\Sigma}^{1/2}\boldsymbol{\hat{B}}^{(n)},\boldsymbol{\Sigma}^{1/2}\boldsymbol{B})^2\leq\|\boldsymbol{\Sigma}^{1/2}\|^2_{op}\times\text{dist}(\boldsymbol{\hat{B}}^{(n)},\boldsymbol{B})^2=\lambda_x\text{dist}(\boldsymbol{\hat{B}}^{(n)},\boldsymbol{B})^2.
		\end{equation*}
		By the fact $\boldsymbol{\hat{B}}^{(n)\top}\boldsymbol{\Sigma}\boldsymbol{\hat{B}}^{(n)}=\boldsymbol{I}_k$ and $\boldsymbol{B}^{\top}\boldsymbol{\Sigma}\boldsymbol{B}=\boldsymbol{I}_k$, we apply Lemma 2 in \cite{tan2020sparse},
		\begin{equation*}
			\begin{split}
				L(\boldsymbol{\hat{B}}^{(T)},\B)=&
				\|\boldsymbol{P}_{\boldsymbol{\hat{B}}^{(T)}}-\boldsymbol{P}_{\B}\|_F^2=2\times	\text{dist}(\boldsymbol{\Sigma}^{1/2}\boldsymbol{\hat{B}}^{(n)},\boldsymbol{\Sigma}^{1/2}\boldsymbol{B})^2\\
				=&C\frac{s\log(p)\log(n)}{\lambda^2n}+C\frac{s^2\log^2(p)\log(1/\delta)\log^7(n)}{\lambda^2n^2\varepsilon^2}.
			\end{split}
		\end{equation*}
		\hfill $\square$
	\end{proof}
	
	\subsection{Proof of the Lemmas}
	
	\begin{proof}[Proof of Lemma \ref{lemma:estS}]
		
		Recall that the sample estimators $\boldsymbol{\hat{M}}$ and $\boldsymbol{\hat{\Sigma}}$ of $\M$ and $\boldsymbol{\Sigma}$ are defined as,
		\begin{equation*}
			\boldsymbol{\hat{M}}=\sum_{h=1}^{H}\hat{p}_{h}[\mathbb{E}_n(\boldsymbol{x}\mid\widetilde{Y}=h)-\mathbb{E}_n\boldsymbol{x}][\mathbb{E}_n(\boldsymbol{x}\mid\widetilde{Y}=h)-\mathbb{E}_n\boldsymbol{x}]^{\top},
		\end{equation*}
		where $\hat{p}_h=\mathbb{E}(\mathbbm{1}\{\widetilde{Y}=h\})$, $\mathbb{E}_n(\boldsymbol{x}\mid\widetilde{Y}=h)=\sum_{i=1}^{n}\boldsymbol{x}_i\mathbbm{1}\{\widetilde{Y}_i=h\}/\sum_{i=1}^{n}\mathbbm{1}\{\widetilde{Y}_i=h\}$, $\widetilde{Y}$ is defined in proof of Lemma \ref{lem:low_privacy}, and
		\begin{equation*}
			\boldsymbol{\hat{\Sigma}}=\frac{1}{n}\sum_{i=1}^{n}\{\boldsymbol{x}_i-\mathbb{E}_n(\boldsymbol{x})\}\{\boldsymbol{x}_i-\mathbb{E}_n(\boldsymbol{x})\}^{\top},
		\end{equation*}
		respectively. We first consider the term $\|\boldsymbol{\hat{\Sigma}}_{\mathcal{I}\mathcal{I}}-\boldsymbol{\Sigma}_{\mathcal{I}\mathcal{I}}\|_{op}$. Note that,
		\begin{equation*}
			\begin{split}
				\|\boldsymbol{\hat{\Sigma}}_{\mathcal{I}\mathcal{I}}-\boldsymbol{\Sigma}_{\mathcal{I}\mathcal{I}}\|_{op}&=\|\boldsymbol{\Sigma}_{\mathcal{I}\mathcal{I}}^{1/2}(\boldsymbol{\Sigma}_{\mathcal{I}\mathcal{I}}^{-1/2}\boldsymbol{\hat{\Sigma}}_{\mathcal{I}\mathcal{I}}\boldsymbol{\Sigma}_{\mathcal{I}\mathcal{I}}^{-1/2}-\boldsymbol{I})\boldsymbol{\Sigma}_{\mathcal{I}\mathcal{I}}^{1/2}\|_{op}\\
				&\leq\lambda_x\|\boldsymbol{\Sigma}_{\mathcal{I}\mathcal{I}}^{-1/2}\boldsymbol{\hat{\Sigma}}_{\mathcal{I}\mathcal{I}}\boldsymbol{\Sigma}_{\mathcal{I}\mathcal{I}}^{-1/2}-\boldsymbol{I}\|_{op}=\lambda_x\|\frac{1}{n}\sum_{i=1}^{n}\boldsymbol{\hat{z}}_{i,\mathcal{I}}\boldsymbol{\hat{z}}_{i,\mathcal{I}}^T-\boldsymbol{I}\|_{op},
			\end{split}
		\end{equation*}
		where we use the notations $\boldsymbol{\hat{z}}_{i,\mathcal{I}}:=\boldsymbol{\Sigma}^{-1/2}_{\mathcal{I}\mathcal{I}}\{\boldsymbol{x}_{i,\mathcal{I}}-\mathbb{E}_n(\boldsymbol{x}_{\mathcal{I}})\}$ to represent sample normalized sub-vectors and $\boldsymbol{z}_{i,\mathcal{I}}:=\boldsymbol{\Sigma}^{-1/2}_{\mathcal{I}\mathcal{I}}\{\boldsymbol{x}_{i,\mathcal{I}}-\mathbb{E}(\boldsymbol{x}_{\mathcal{I}})\}\sim (\boldsymbol{0},\boldsymbol{I}_p)$ to represnet normalized vectors. The effect of estimating $\mathbb{E}(\boldsymbol{X})$ is
		\begin{equation*}
			\begin{split}
				&\|\mathbb{E}_n(\boldsymbol{\hat{z}}_{i,\mathcal{I}}\boldsymbol{\hat{z}}_{i,\mathcal{I}}^{\top})-\mathbb{E}_n(\boldsymbol{z}_{i,\mathcal{I}}\boldsymbol{z}_{i,\mathcal{I}}^{\top})\|_{op}\\
				=&\|\mathbb{E}_n\{(\boldsymbol{\hat{z}}_{i,\mathcal{I}}-\boldsymbol{z}_{i,\mathcal{I}})(\boldsymbol{\hat{z}}_{i,\mathcal{I}}-\boldsymbol{z}_{i,\mathcal{I}})^{\top}\}-\mathbb{E}_n\{\boldsymbol{\hat{z}}_{i,\mathcal{I}}(\boldsymbol{\hat{z}}_{i,\mathcal{I}}-\boldsymbol{z}_{i,\mathcal{I}})^{\top}\}-\mathbb{E}_n\{(\boldsymbol{\hat{z}}_{i,\mathcal{I}}-\boldsymbol{z}_{i,\mathcal{I}})\boldsymbol{\hat{z}}_{i,\mathcal{I}}^{\top}\}\|_{op}\\
				\leq&\|\mathbb{E}_n\{(\boldsymbol{\hat{z}}_{i,\mathcal{I}}-\boldsymbol{z}_{i,\mathcal{I}})(\boldsymbol{\hat{z}}_{i,\mathcal{I}}-\boldsymbol{z}_{i,\mathcal{I}})^{\top}\}\|_{op}+\|\mathbb{E}_n\{\boldsymbol{\hat{z}}_{i,\mathcal{I}}(\boldsymbol{\hat{z}}_{i,\mathcal{I}}-\boldsymbol{z}_{i,\mathcal{I}})^{\top}\}\|_{op}+\|\mathbb{E}_n\{(\boldsymbol{\hat{z}}_{i,\mathcal{I}}-\boldsymbol{z}_{i,\mathcal{I}})\boldsymbol{\hat{z}}_{i,\mathcal{I}}^{\top}\}\|_{op}\\
				=&\|\mathbb{E}_n\{(\boldsymbol{\hat{z}}_{i,\mathcal{I}}-\boldsymbol{z}_{i,\mathcal{I}})(\boldsymbol{\hat{z}}_{i,\mathcal{I}}-\boldsymbol{z}_{i,\mathcal{I}})^{\top}\}\|_{op}=\|\mathbb{E}_n(\boldsymbol{\hat{z}}_{i,\mathcal{I}}-\boldsymbol{z}_{i,\mathcal{I}})\|_2^2\leq\{4\sqrt{|\mathcal{I}|}+2\sqrt{\log(1/\delta)}\}^2/n,
			\end{split}
		\end{equation*}
		with a probability of at least $1-\delta$. As we will show later, the term due to estimating mean is relatively small. Then, we consider the term
		\begin{equation*}
			\begin{split}
				\mathbb{P}(\|\frac{1}{n}\sum_{i=1}^{n}\boldsymbol{z}_{i,\mathcal{I}}\boldsymbol{z}_{i,\mathcal{I}}^{\top}-\boldsymbol{I}\|_{op}>t)\leq\exp(C|\mathcal{I}|-Cnt^2)
			\end{split}
		\end{equation*}
		where we use Lemma \ref{lem:con} and  Lemma \ref{lem:random_matrix_eigen}. For the choice of $\delta=\exp\{-C_1|\mathcal{I}|\log(ep/|\mathcal{I}|)\}$ and $t=\sqrt{\{|\mathcal{I}|+C_1|\mathcal{I}|\log(ep/|\mathcal{I}|)\}/n}$, by applying the union bound, we have
		\begin{equation*}
			\begin{split}
				&\mathbb{P}(\sup_{\mathcal{I}}\|\mathbb{E}_n(\boldsymbol{\hat{z}}_{i,\mathcal{I}}\boldsymbol{\hat{z}}_{i,\mathcal{I}}^{\top})-\mathbb{E}_n(\boldsymbol{z}_{i,\mathcal{I}}\boldsymbol{z}_{i,\mathcal{I}}^{\top})\|_{op}\geq\{4\sqrt{|\mathcal{I}|}+2\sqrt{|\mathcal{I}|\log(ep/|\mathcal{I}|)}\}^2/n)\leq p^{|\mathcal{I}|}\delta\\
				\leq&\exp\{|\mathcal{I}|\log(p)-C_1|\mathcal{I}|\log(ep/|\mathcal{I}|)\}\leq\exp\{-C_1/2|\mathcal{I}|\log(ep/|\mathcal{I}|)\},
			\end{split}
		\end{equation*}
		and
		\begin{equation*}
			\begin{split}
				&\mathbb{P}(\sup_{\mathcal{I}}\|\frac{1}{n}\sum_{i=1}^{n}\boldsymbol{z}_{i,\mathcal{I}}\boldsymbol{z}_{i,\mathcal{I}}^{\top}-\boldsymbol{I}\|_{op}>t)\leq p^{|\mathcal{I}|}\exp(C|\mathcal{I}|-Cnt^2)\\
				=&\exp\{|\mathcal{I}|\log(p)-CC_1|\mathcal{I}|\log(ep/|\mathcal{I}|)\}\leq\exp\{-CC_1/2|\mathcal{I}|\log(ep/|\mathcal{I}|)\},
			\end{split}
		\end{equation*}
		for $C_1$ be a large enough constant, where we use the fact that the the number of combinations of choosing $|\mathcal{I}|$ items from $p$ items is bounded by $p^{|\mathcal{I}|}$. Thus, we conclude (\ref{eq:estsigma}).
		
		Then we consider the term $\|\boldsymbol{\hat{M}}_{\mathcal{I}\mathcal{I}}-\M_{\mathcal{I}\mathcal{I}}\|_{op}$. Similar to the arguments for $\|\boldsymbol{\Sigma}_{\mathcal{I}\mathcal{I}}-\boldsymbol{\hat{\Sigma}}_{\mathcal{I}\mathcal{I}}\|_{op}$, without loss of generality, we assume $\mathbb{E}_n(\boldsymbol{x})=0$ and use the notation $\M_h:=\mathbb{E}(\boldsymbol{x}\mathbbm{1}\{\widetilde{Y}=h\}),\boldsymbol{\hat{m}}_h:=\sum_{i=1}^{n}\boldsymbol{x}_i\mathbbm{1}\{\widetilde{Y}_i=h\}/n$ and $\boldsymbol{\Sigma}_h=\M_h\M_h^{\top},\boldsymbol{\hat{\Sigma}}_h=\boldsymbol{\hat{m}}_h\boldsymbol{\hat{m}}_h^{\top}$ Then, we can write $\M$ and $\boldsymbol{\hat{M}}$ as
		\begin{equation*}
			\M=\sum_{i=1}^{H}p_h^{-1}\boldsymbol{\Sigma}_h,\text{ and }\boldsymbol{\hat{M}}=\sum_{i=1}^{H}\hat{p}_h^{-1}\boldsymbol{\hat{\Sigma}}_h.
		\end{equation*}
		Note that $\hat{p}_h$ is the sample average of $n$ i.i.d. Binomial random variables. For $h=1,\dots,H$, by Bernstein's inequality, we have
		\begin{equation*}
			\mathbb{P}\{|\hat{p}_h-p_h|\geq\sqrt{p_h(1-p_h)}\sqrt{\log(p)/n}\}\leq 2\exp\bigg\{-\frac{\log(p)}{2(1+\sqrt{\log(p)/n}/3)}\bigg\}.
		\end{equation*}
		By $p_h(1-p_h)\leq 1/4$ and $\log(p)/n\to 0$, then
		\begin{equation*}
			\mathbb{P}\{|\hat{p}_h-p_h^{-1}|\geq \sqrt{\log(p)/n}/2\}\leq 2\exp\bigg\{-\frac{\log(p)}{2(1+\sqrt{\log(p)/n}/3)}\bigg\}\leq 2p^{-1/3}.
		\end{equation*}
		Furthermore,
		\begin{equation*}
			\begin{split}
				\mathbb{P}\{|\hat{p}_h^{-1}-p_h^{-1}|\geq\sqrt{\log(p)/n}/2/p_h^2\}=&\mathbb{P}\{|\hat{p}_h-p_h|\geq\sqrt{\log(p)/n}/2\hat{p}_h/p_h\}\\
				\leq& 2p^{-1/3}+\mathbb{P}\{|\hat{p}_h-p_h|\geq\sqrt{\log(p)/n}/2(1+\sqrt{\log(p)/n}/2/p_h)\}\\
				\leq& 2p^{-1/3}+\mathbb{P}\{|\hat{p}_h-p_h|\geq\sqrt{\log(p)/n}\}\leq 2(p^{-1/3}+p^{-4/3}),
			\end{split}
		\end{equation*}
		where we use Bernstein's inequality in the last inequality. Furthermore,
		\begin{align*}
			\|\boldsymbol{\hat{M}}_{\mathcal{I}\mathcal{I}}-\M_{\mathcal{I}\mathcal{I}}\|_{op} &\leq\sum_{h=1}^{H}|\hat{p}_{h}^{-1}-p_h^{-1}|\|\boldsymbol{\Sigma}_{h,\mathcal{I}\mathcal{I}}\|_{op}+\sum_{h=1}^{H}\hat{p}_{h}^{-1}\|\boldsymbol{\Sigma}_{h,\mathcal{II}}-\boldsymbol{\hat{\Sigma}}_{h,\mathcal{II}}\|_{op}\\
			&\leq C\sqrt{\frac{\log(p)}{n}}\sum_{h=1}^{H}\|\boldsymbol{\Sigma}_{h,\mathcal{I}\mathcal{I}}\|_{op}\\
			&+\sum_{h=1}^{H}\hat{p}_{h}^{-1}\|(\M_{h,\mathcal{I}\mathcal{I}}-\boldsymbol{\hat{m}}_{h,\mathcal{I}\mathcal{I}})(\M_{h,\mathcal{I}\mathcal{I}}-\boldsymbol{\hat{m}}_{h,\mathcal{I}\mathcal{I}})^{\top}\\
			&+\M_{h,\mathcal{I}\mathcal{I}}(\boldsymbol{\hat{m}}_{h,\mathcal{I}\mathcal{I}}-\M_{h,\mathcal{I}\mathcal{I}})^{\top}+(\boldsymbol{\hat{m}}_{h,\mathcal{I}\mathcal{I}}-\M_{h,\mathcal{I}\mathcal{I}})\M_{h,\mathcal{I}\mathcal{I}}^{\top}\|_{op},
		\end{align*}
		where we use the triangle inequality in the first inequality and we use the choice $C>2/p_h^2$ for the positive constant $C$. The first term in the second line of the second inequality is a cross-term, thus is sufficiently small. It is enough to consider the remaining parts. In the remaining parts, the dominant term is $\sum_{h=1}^{H}\|(\boldsymbol{\hat{m}}_{h,\mathcal{I}\mathcal{I}}-\M_{h,\mathcal{I}\mathcal{I}})\M_{h,\mathcal{I}\mathcal{I}}^{\top}\|_{op}$. Note that the
		\begin{equation*}
			\begin{split}
				\|(\boldsymbol{\hat{m}}_{h,\mathcal{I}\mathcal{I}}-\M_{h,\mathcal{I}\mathcal{I}})\M_{h,\mathcal{I}\mathcal{I}}^{\top}\|_{op}=&\sup_{\|\boldsymbol{\alpha}\|_2=1}\boldsymbol{\alpha}^{\top}(\boldsymbol{\hat{m}}_{h,\mathcal{I}\mathcal{I}}-\M_{h,\mathcal{I}\mathcal{I}})(\M_{h,\mathcal{I}\mathcal{I}}^{\top}\boldsymbol{\alpha})\\
				\leq&\sup_{\|\boldsymbol{\alpha}\|_2=1}\boldsymbol{\alpha}^{\top}(\boldsymbol{\hat{m}}_{h,\mathcal{I}\mathcal{I}}-\M_{h})\sup_{|\boldsymbol{\alpha}\|_2\leq1}(\M_{h}^{\top}\boldsymbol{\alpha})\\
				\leq&\sup_{\|\boldsymbol{\alpha}\|_2=1}\boldsymbol{\alpha}^T(\boldsymbol{\hat{m}}_{h,\mathcal{I}\mathcal{I}}-\M_{h,\mathcal{I}\mathcal{I}})\sqrt{C|\mathcal{I}|},
			\end{split}
		\end{equation*}
		where $C$ is a positive constant. By the fact $\boldsymbol{\hat{m}}_{h,\mathcal{I}\mathcal{I}}-\M_{h,\mathcal{I}\mathcal{I}}$ is an average of $n$ i.i.d. centered sub-Gaussian random vectors, and using Bernstein’s inequality,
		\begin{equation*}
			\mathbb{P}(\boldsymbol{\alpha}^{\top}(\boldsymbol{\hat{m}}_{h,\mathcal{I}\mathcal{I}}-\M_{h,\mathcal{I}\mathcal{I}})>t)\leq\exp(-Cnt^2).
		\end{equation*}
		For $t=\sqrt{C_1\log(ep/|\mathcal{I}|)/n}$. By the union bound,
		\begin{equation*}
			\begin{split}
				&\mathbb{P}(\sup_{\mathcal{I}}\|(\boldsymbol{\hat{m}}_{h,\mathcal{I}\mathcal{I}}-\M_{h,\mathcal{I}\mathcal{I}})\M_{h,\mathcal{I}\mathcal{I}}^{\top}\|_{op}\geq \sqrt{C_1C|\mathcal{I}|\log(ep/|\mathcal{I}|)/n})\\
				\leq& p^{|\mathcal{I}|}\exp\{-C_1C^2|\mathcal{I}|\log(ep/|\mathcal{I}|)\}=\exp\{|\mathcal{I}|\log(p)-C_1C^2|\mathcal{I}|\log(ep/|\mathcal{I}|)\}\\
				\leq&\exp\{-C_1C^2/2|\mathcal{I}|\log(ep/|\mathcal{I}|)\},
			\end{split}
		\end{equation*}
		for $C_1$ be a large enough constant, where we use the fact that the the number of combinations of choosing $|\mathcal{I}|$ items from $p$ items is bounded by $p^{|\mathcal{I}|}$. The conclusion (\ref{eq:estm}) holds.
		\hfill $\square$
	\end{proof}
	
	\begin{proof}[Proof of Lemma \ref{lemma:oracle}]
		
		Note that the estimator $\boldsymbol{\hat{B}}(\mathcal{I})$ is the solution of equation (\ref{eq:fixI}). It's sufficient to consider the following optimization problem,
		\begin{equation*}
			\max_{\B\in\mR^{|\mathcal{I}|\times K}}\langle\boldsymbol{\hat{M}}_{\mathcal{I}\mathcal{I}},\B\B^{\top}\rangle\text{ s.t. }\B^{\top}\boldsymbol{\hat{\Sigma}}^{-1}_{\mathcal{I}\mathcal{I}}\B=\boldsymbol{I}_k,
		\end{equation*}
		where the subscript $\mathcal{I}\mathcal{I}$ denotes the corresponding submatrix. Therefore, we restrict our attention to the subset $\mathcal{I}\subset\{1,\dots,p\}$. For the index set $\mathcal{I}$, let $\{\lambda_1,\dots,\lambda_{|\mathcal{I}|}\}$ be the ordered generalized eigenvalues of matrix pair $(\M_{\mathcal{I}\mathcal{I}},\boldsymbol{\Sigma}_{\mathcal{I}\mathcal{I}})$ and $\{\hat{\lambda}_{1},\dots,\hat{\lambda}_{|\mathcal{I}|}\}$ be the ordered generalized eigenvalues of matrix pair $(\boldsymbol{\hat{M}}_{\mathcal{I}\mathcal{I}},\boldsymbol{\hat{\Sigma}}_{\mathcal{I}\mathcal{I}})$. Notice that fo $i\geq H+1$, we have $\hat{\lambda}_i=0$. The estimated eigenvalues satisfy
		\begin{equation*}
			\begin{split}
				|\lambda_{k+1}-\hat{\lambda}_{k+1}|&\leq\|\boldsymbol{\Sigma}^{-1/2}_{\mathcal{I}\mathcal{I}}\M_{\mathcal{I}\mathcal{I}}\boldsymbol{\Sigma}^{-1/2}_{\mathcal{I}\mathcal{I}}-\boldsymbol{\hat{\Sigma}}^{-1/2}_{\mathcal{I}\mathcal{I}}\boldsymbol{\hat{M}}_{\mathcal{I}\mathcal{I}}\boldsymbol{\hat{\Sigma}}^{-1/2}_{\mathcal{I}\mathcal{I}}\|_{op}\\
				&\leq\|(\boldsymbol{\Sigma}^{-1/2}_{\mathcal{I}\mathcal{I}}-\boldsymbol{\hat{\Sigma}}^{-1/2}_{\mathcal{I}\mathcal{I}})\M_{\mathcal{I}\mathcal{I}}\boldsymbol{\Sigma}^{-1/2}_{\mathcal{I}\mathcal{I}}\|_{op}+\|\boldsymbol{\hat{\Sigma}}^{-1/2}_{\mathcal{I}\mathcal{I}}(\M_{\mathcal{I}\mathcal{I}}-\boldsymbol{\hat{M}}_{\mathcal{I}\mathcal{I}})\boldsymbol{\Sigma}^{-1/2}_{\mathcal{I}\mathcal{I}}\|_{op}\\
				&+\|\boldsymbol{\hat{\Sigma}}^{-1/2}_{\mathcal{I}\mathcal{I}}\boldsymbol{\hat{M}}_{\mathcal{I}\mathcal{I}}(\boldsymbol{\Sigma}^{-1/2}_{\mathcal{I}\mathcal{I}}-\boldsymbol{\hat{\Sigma}}^{-1/2}_{\mathcal{I}\mathcal{I}})\|_{op}\\
				&\leq\|\boldsymbol{\Sigma}^{-1/2}_{\mathcal{I}\mathcal{I}}-\boldsymbol{\hat{\Sigma}}^{-1/2}_{\mathcal{I}\mathcal{I}}\|_{op}\|\M_{\mathcal{I}\mathcal{I}}\boldsymbol{\Sigma}^{-1/2}_{\mathcal{I}\mathcal{I}}\|_{op}+\|\boldsymbol{\hat{\Sigma}}^{-1/2}_{\mathcal{I}\mathcal{I}}\|_{op}\|\M_{\mathcal{I}\mathcal{I}}-\boldsymbol{\hat{M}}_{\mathcal{I}\mathcal{I}}\|_{op}\|\boldsymbol{\Sigma}^{-1/2}_{\mathcal{I}\mathcal{I}}\|_{op}\\
				&+\|\boldsymbol{\hat{\Sigma}}^{-1/2}_{\mathcal{I}\mathcal{I}}\boldsymbol{\hat{M}}_{\mathcal{I}\mathcal{I}}\|_{op}\|\boldsymbol{\Sigma}^{-1/2}_{\mathcal{I}\mathcal{I}}-\boldsymbol{\hat{\Sigma}}^{-1/2}_{\mathcal{I}\mathcal{I}}\|_{op}\\
				&\leq C(2\lambda_1\sqrt{\lambda_x}+\lambda_x)\sqrt{\frac{|\mathcal{I}|\log(ep/|\mathcal{I}|)}{n}}\leq C\sqrt{\frac{|\mathcal{I}|\log(ep/|\mathcal{I}|)}{n}},
			\end{split}
		\end{equation*}
		where we use Lemma 2.1 in \cite{sibson1979studies} in the first inequality; we use the triangle inequality in the second inequality; we use the sub-multiplicativity of matrix norm in the third inequality; we use Lemma \ref{lemma:estS} in the last inequality. By the proof of Lemma \ref{lem:est_slices}, the generalized eigenvalues corresponding to the estimated slices satisfies $\lambda_k\geq\lambda/2$. Then, the following event holds with probability at least $1-\exp\{-C^{\prime}s^{\prime}\log(ep/s^{\prime})\}$,
		\begin{equation*}
			\hat{\lambda}_{k+1}\leq \lambda_k/2,
		\end{equation*}
		where we use the condition $\lambda_{k+1}=0$, $\sqrt{|\mathcal{I}|\log(ep/|\mathcal{I}|)/n}$ and $\sqrt{|\mathcal{I}|\log(ep/|\mathcal{I}|)/(n\lambda)}$ are small enough. Then, we have
		\begin{equation*}
			\Delta=\frac{\lambda_k-\hat{\lambda}_{k+1}}{\sqrt{(1+\lambda_1^2)(1+\hat{\lambda}_{k+1}^2)}}\geq\frac{\lambda_k}{2\sqrt{(1+\lambda_1^2)(1+\lambda_k^2/4)}}.
		\end{equation*}
		Then,
		\begin{equation*}
			c(\M_{\mathcal{I}\mathcal{I}},\boldsymbol{\Sigma}_{\mathcal{I}\mathcal{I}})=\inf_{\|\boldsymbol{x}\|_2=1}\sqrt{(\boldsymbol{x}^{\top}\M_{\mathcal{I}\mathcal{I}}\boldsymbol{x})^2+(\boldsymbol{x}^{\top}\boldsymbol{\Sigma}_{\mathcal{I}\mathcal{I}}\boldsymbol{x})^2}\geq\lambda_x^{-1},
		\end{equation*}
		\begin{equation*}
			c(\boldsymbol{\hat{M}}_{\mathcal{I}\mathcal{I}},\boldsymbol{\hat{\Sigma}}_{\mathcal{I}\mathcal{I}})=\inf_{\|\boldsymbol{x}\|_2=1}\sqrt{(\boldsymbol{x}^{\top}\boldsymbol{\hat{M}}_{\mathcal{I}\mathcal{I}}\boldsymbol{x})^2+(\boldsymbol{x}^{\top}\boldsymbol{\hat{\Sigma}}_{\mathcal{I}\mathcal{I}}\boldsymbol{x})^2}\geq\lambda_x^{-1}/2,
		\end{equation*}
		where we use the fact $\|\boldsymbol{\hat{\Sigma}}_{\mathcal{I}\mathcal{I}}-\boldsymbol{\Sigma}_{\mathcal{I}\mathcal{I}}\|_{op}\to 0$. Let $$\epsilon_{noisy}=\sqrt{\|\M_{\mathcal{I}\mathcal{I}}-\boldsymbol{\hat{M}}_{\mathcal{I}\mathcal{I}}\|_{op}^2+\|\boldsymbol{\hat{\Sigma}}_{\mathcal{I}\mathcal{I}}-\boldsymbol{\Sigma}_{\mathcal{I}\mathcal{I}}\|_{op}^2}\leq C\sqrt{|\mathcal{I}|\log(ep/|\mathcal{I}|)/n}.$$
		Then by Theorem 3.1 in \cite{sun1983perturbation} and Corollary 4.5 in \cite{stewart1979pertubation},
		\begin{align*}
			\|\mathbf{P}-\mathbf{\hat{P}}\|_F&\leq C\frac{\epsilon_{noisy}\sqrt{k}}{\Delta c(\M_{\mathcal{I}\mathcal{I}},\boldsymbol{\Sigma}_{\mathcal{I}\mathcal{I}})c(\boldsymbol{\hat{M}}_{\mathcal{I}\mathcal{I}},\boldsymbol{\hat{\Sigma}}_{\mathcal{I}\mathcal{I}})}\sqrt{\|\M_{\mathcal{I}\mathcal{I}}\|_{op}^2+\|\boldsymbol{\Sigma}_{\mathcal{I}\mathcal{I}}\|_{op}^2}\\
			&\leq C\sqrt{k}\lambda_x^3\frac{(1+\lambda_1^2)\sqrt{1+\lambda_{k}^2/4}}{\lambda_k}\sqrt{|\mathcal{I}|_0\log(ep/|\mathcal{I}|_0)/n}\leq C\frac{\sqrt{|\mathcal{I}|_0\log(ep/|\mathcal{I}|_0)/n}}{\lambda},
		\end{align*}
		where $\mathbf{P}$ and $\hat{\mathbf{P}}$ are the projection matrix on the eigen-subspace of matrix pair $(\M_{\mathcal{I}\mathcal{I}},\boldsymbol{\Sigma}_{\mathcal{I}\mathcal{I}})$ and $(\boldsymbol{\hat{M}}_{\mathcal{I}\mathcal{I}},\boldsymbol{\hat{\Sigma}}_{\mathcal{I}\mathcal{I}})$, $C$ is a positive constant.
		\hfill $\square$
	\end{proof}
	
	\begin{proof}[Proof of Lemma \ref{lem:grad_desc}]
		
		Before the proof, we first make some illustrations of notations. By the sparsity of $\boldsymbol{\hat{B}}^{(t)}$ and $\boldsymbol{\hat{B}}^{(t+1)}$, the support set $\mathcal{S}_t$ has at most $2s^{\prime}+s$ elements. Thus, by Lemma \ref{lemma:estS} and Lemma \ref{lemma:oracle}, the event $\mathcal{E}_1\cap\mathcal{E}_2$ holds with probability at least $1-\exp\{-C^{\prime}s^{\prime}\log(ep/s^{\prime})\}$. Thus, it's sufficient to consider $\boldsymbol{\hat{M}}_{\mathcal{S}_t\mathcal{S}_t}$ and $\boldsymbol{\hat{\Sigma}}_{\mathcal{S}_t\mathcal{S}_t}$, where the subscript $\mathcal{S}_t\mathcal{S}_t$ means the $|\mathcal{S}_t|\times|\mathcal{S}_t|$ submatrix. To simplify the notation, we use $\boldsymbol{\hat{M}}_{t}$ and $\boldsymbol{\hat{\Sigma}}_{t}$ to denote $\boldsymbol{\hat{M}}_{\mathcal{S}_t\mathcal{S}_t}$ and $\boldsymbol{\hat{\Sigma}}_{\mathcal{S}_t\mathcal{S}_t}$, respectively. Then, the corresponding loss function is reformulated into
		\begin{equation}
			\label{eq:newloss}
			f_t(\B):=-\langle\boldsymbol{\hat{M}}_t,\B\B^{\top}\rangle+\lambda_{penalty}\|\B^{\top}\boldsymbol{\hat{\Sigma}}_t\B-\boldsymbol{I}_k\|_F^2,
		\end{equation}
		with the corresponding gradient function
		\begin{equation}
			\label{eq:newgrad}
			\nabla f_t(\B)=-2\boldsymbol{\hat{M}}_t\B+2\lambda_{penalty}\boldsymbol{\hat{\Sigma}}_t\B(\B^{\top}\boldsymbol{\hat{\Sigma}}_t\B-\boldsymbol{I}_k),
		\end{equation}
		where we use the notation $f_t(\cdot)$ to denote the corresponding loss function at the $t$-th step. We use notations $\boldsymbol{L}_t:=\boldsymbol{\hat{B}}^{(t)}_{\mathcal{S}_t}$ and $\boldsymbol{L}_{t+1}:=\boldsymbol{\hat{B}}^{(t+0.3)}_{\mathcal{S}_t}=\boldsymbol{L}_t-\eta\nabla f_t(\boldsymbol{L}_t)$, where the subscript $\mathcal{S}_t$ indicates the restriction of support set. Recall that the support of $\boldsymbol{\hat{B}}^{(t)}$ is in the set $\mathcal{S}_t$, but  the support of $\boldsymbol{\hat{B}}^{(t+0.3)}$ doesn't need to be in the set $\mathcal{S}_t$. But considering the subvector is enough for the conclusion of the Lemma.
		
		Notice that the parameter $\B\in\mR^{|\mathcal{S}_t|\times r}$ is a matrix and the loss function (\ref{eq:newloss}) maps a matrix to a scalar, so the Hessians matrix of (\ref{eq:newloss}) will be a fourth-order tensor, which is difficult to represent. We adopt the strategy in \cite{chi2019nonconvex}, which ignores the matrix structure when calculating the Hessian matrix. For high-dimensional Z-estimators, smoothness and convexity are two commonly used conditions. A matrix value function $f(\boldsymbol{L}):\mR^{a\times b}\to\mR$ is $\beta$-smooth, if for all $\boldsymbol{Q}\in\mR^{a\times b}$,
		\begin{equation*}
			\text{vec}(\boldsymbol{Q})^{\top}\nabla^2 f(\boldsymbol{L})\text{vec}(\boldsymbol{Q})\leq\beta\|\boldsymbol{Q}\|_F^2,
		\end{equation*}
		However, the locally strong convexity condition usually doesn't hold, due to the ignoration of the matrix structure. Considering the information carried in the matrix structure carefully, one can only hope to recover the original matrix up to a global orthonormal transformation. Mathematically, a suitable error metric for $\boldsymbol{X},\boldsymbol{Y}\in\mR^{p\times r}$ is,
		\begin{equation*}
			\text{dist}(\boldsymbol{X},\boldsymbol{Y})=\min_{\boldsymbol{H}\in\mathbb{O}^{r\times r}}\|\boldsymbol{X}\boldsymbol{H}-\boldsymbol{Y}\|_F,
		\end{equation*}
		where $\mathbb{O}^{r\times r}$ is space containing all $r\times r$ orthogonal matrix. For notation convenience, we let
		\begin{equation*}
			\boldsymbol{H}_{\boldsymbol{X}}=\argmin_{\boldsymbol{H}\in\mathbb{O}^{r\times r}}\|\boldsymbol{X}\boldsymbol{H}-\boldsymbol{Y}\|_F,
		\end{equation*}
		which represents the ``best angel" between $\boldsymbol{X}$ and $\boldsymbol{Y}$. Let $\boldsymbol{L}_*$ be the optimal solution of (\ref{eq:newloss}). The following lemma is from Lemma 4 in \cite{chi2019nonconvex}.
		\begin{lemm}
			\label{lem:general_gradient_descent}
			Suppose that $f_t$ is $\beta$-smooth within a ball $\mathbb{B}_{\xi}(\boldsymbol{L}_*)=\{\boldsymbol{L}:\|\boldsymbol{L}-\boldsymbol{L}_*\|_F\leq\xi\}$, and that $\nabla f_t(\boldsymbol{L})\boldsymbol{H}=\nabla f_t(\boldsymbol{L}\boldsymbol{H})$ for any orthogonal matrix $\boldsymbol{H}\in\mathbb{O}^{r\times r}$. Assume that for any $\boldsymbol{L}_t\in\mathbb{B}_{\xi}(\boldsymbol{L}_*)$ and any $\boldsymbol{Z}$,
			\begin{equation*}
				\text{vec}(\boldsymbol{Z}\boldsymbol{H}_{\boldsymbol{Z}}-\boldsymbol{L}_*)^{\top}\nabla^2 f(\boldsymbol{L}_t)\text{vec}(\boldsymbol{Z}\boldsymbol{H}_{\boldsymbol{Z}}-\boldsymbol{L}_*)\geq\alpha\|\boldsymbol{Z}\boldsymbol{H}_{\boldsymbol{Z}}-\boldsymbol{L}_*\|^2_F,
			\end{equation*}
			where $\boldsymbol{H}_{\boldsymbol{Z}}$ is the rotation matrix with respect to $\boldsymbol{L}_*$. If $\eta_t=1/\beta$, then gradient descent update with initial value $\boldsymbol{L}_t\in\mathbb{B}_{\xi}(\boldsymbol{L}_*)$ satisfies
			\begin{equation*}
				\text{dist}^2(\boldsymbol{L}_{t+1},\boldsymbol{L}_*)\leq (1-\frac{\alpha}{\beta})\text{dist}^2(\boldsymbol{L}_t,\boldsymbol{L}_*).
			\end{equation*}
		\end{lemm}
		The previous lemma says that under some regularity conditions, the error distance decreases at an exponential rate when a good initial value is specified. By Lemma B.7 in \cite{gao2021sparse}, the condition $\boldsymbol{L}_t\in\mathbb{B}_{\xi}(\boldsymbol{L}_*)$ can be relaxed to $\text{dist}(\boldsymbol{L}_t,\boldsymbol{L}_*)\leq\xi$ without any further conditions. To apply Lemma \ref{lem:general_gradient_descent}, it remains to check the conditions in Lemma \ref{lem:general_gradient_descent} and bound three constants $(\beta,\alpha,\xi)$ one by one.
		
		\emph{Step 1}: We first check the smoothness condition and the rotation invariance condition $\nabla f_t(\boldsymbol{L})\boldsymbol{H}=\nabla f_t(\boldsymbol{L}\boldsymbol{H})$. Notice that,
		\begin{align*}
			\nabla f_t(\boldsymbol{L}\boldsymbol{H})\boldsymbol{H}^{\top}&=-2\boldsymbol{\hat{M}}_t\boldsymbol{L}\boldsymbol{H}\boldsymbol{H}^{\top}+2\lambda_{penalty}\boldsymbol{\hat{\Sigma}}_t\boldsymbol{L}\boldsymbol{H}(\boldsymbol{H}^{\top}\boldsymbol{L}^{\top}\boldsymbol{\hat{\Sigma}}_t\boldsymbol{L}\boldsymbol{H}-\boldsymbol{I}_k)\boldsymbol{H}^{\top}\\
			&=-2\boldsymbol{\hat{M}}_t\boldsymbol{L}+2\lambda_{penalty}\boldsymbol{\hat{\Sigma}}_t\boldsymbol{L}(\boldsymbol{L}^{\top}\boldsymbol{\hat{\Sigma}}_t\boldsymbol{L}-\boldsymbol{I}_k)=\nabla f_t(\boldsymbol{L}),
		\end{align*}
		where we use the fact $\boldsymbol{H}\boldsymbol{H}^{\top}=\boldsymbol{I}$. Then, it remains to check the smoothness condition. We rewrite the vectorization $\text{vec}(\boldsymbol{L})=[\boldsymbol{l_1}^{\top},\dots,\boldsymbol{l}_k^{\top}]^{\top}$ where $\boldsymbol{l}_i\in\mR^{|\mathcal{S}_t|}$ for $i=1,\dots,r$. Then, we rewrite the equation (\ref{eq:newgrad}) into vector form,
		\begin{equation*}
			\begin{split}
				&\text{vec}\nabla f_t(\boldsymbol{L})\\
				&=-2(\boldsymbol{I}_k\otimes\boldsymbol{\hat{M}}_t)\text{vec}(\boldsymbol{L})+2\lambda_{penalty}(\boldsymbol{I}_k\otimes\boldsymbol{\hat{\Sigma}}_t\boldsymbol{L}\boldsymbol{L}^{\top}\boldsymbol{\hat{\Sigma}}_t)\text{vec}(\boldsymbol{L})-2\lambda_{penalty}(\boldsymbol{I}_k\otimes\boldsymbol{\hat{\Sigma}}_t)\text{vec}(\boldsymbol{L}),
			\end{split}
		\end{equation*}
		where $\otimes$ is the kronecker product. Then, we calculate the ``Hessian" $\nabla\{\nabla\text{vec}f_t(\boldsymbol{L})\}$ term by term. The first term satisfies
		\begin{equation*}
			\nabla\{-2(\boldsymbol{I}_k\otimes\boldsymbol{\hat{M}}_t)\text{vec}(\boldsymbol{L})\}=-2\boldsymbol{I}_k\otimes\boldsymbol{\hat{M}}_t.
		\end{equation*}
		The second term satisfies
		\begin{align*}
			&\nabla\{2\lambda_{penalty}(\boldsymbol{I}_k\otimes\boldsymbol{\hat{\Sigma}}_t\boldsymbol{L}\boldsymbol{L}^{\top}\boldsymbol{\hat{\Sigma}}_t)\text{vec}(\boldsymbol{L})\}\\
			=&2\lambda_{penalty}(\boldsymbol{I}_k\otimes\boldsymbol{\hat{\Sigma}}_t\boldsymbol{L}\boldsymbol{L}^{\top}\boldsymbol{\hat{\Sigma}}_t)+2\lambda_{penalty}\nabla(\boldsymbol{I}_k\otimes\boldsymbol{\hat{\Sigma}}_t\boldsymbol{L}\boldsymbol{L}^{\top}\boldsymbol{\hat{\Sigma}}_t)\text{vec}(\boldsymbol{L})\\
			=&2\lambda_{penalty}(\boldsymbol{I}_k\otimes\boldsymbol{\hat{\Sigma}}_t\boldsymbol{L}\boldsymbol{L}^{\top}\boldsymbol{\hat{\Sigma}}_t)+2\lambda_{penalty}(\boldsymbol{I}_k\otimes\boldsymbol{\hat{\Sigma}})\text{vec}(\boldsymbol{L})\text{vec}(\boldsymbol{L})^{\top}(\boldsymbol{I}_K\otimes\boldsymbol{\hat{\Sigma}})\nonumber\\
			&+2\lambda_{penalty}\text{vec}(\boldsymbol{L})^{\top}(\boldsymbol{I}_k\otimes\boldsymbol{\hat{\Sigma}})\text{vec}(\boldsymbol{L})(\boldsymbol{I}_k\otimes\boldsymbol{\hat{\Sigma}}),
		\end{align*}
		where we use the Leibniz rule in the first and second inequality. The third term satisfies
		\begin{equation*}
			\nabla\{-2\lambda_{penalty}(\boldsymbol{I}_k\otimes\boldsymbol{\hat{\Sigma}}_t)\text{vec}(\boldsymbol{L})\}=-2\lambda_{penalty}\boldsymbol{I}_k\otimes\boldsymbol{\hat{\Sigma}}_t.
		\end{equation*}
		Then, we are in the position to calculate the smoothness condition. By calculating the quadratic form, we have
		\begin{align*}
			&\frac{1}{2}\text{vec}(\boldsymbol{Q})\nabla^2 f_t(\boldsymbol{L})\text{vec}(\boldsymbol{Q})\\
			=&-\langle\boldsymbol{\hat{M}}_t,\boldsymbol{Q}\boldsymbol{Q}^T\rangle+\lambda_{penalty}\langle\boldsymbol{\hat{\Sigma}}_t\boldsymbol{L}\boldsymbol{L}^{\top}\boldsymbol{\hat{\Sigma}}_t,\boldsymbol{Q}\boldsymbol{Q}^{\top}\rangle+\lambda_{penalty}\langle\boldsymbol{L}^{\top}\boldsymbol{\hat{\Sigma}}_t\boldsymbol{L},\boldsymbol{Q}^{\top}\boldsymbol{\hat{\Sigma}}_t\boldsymbol{Q}\rangle\nonumber\\
			&+\lambda_{penalty}\langle\boldsymbol{Q}^{\top}\boldsymbol{\hat{\Sigma}}_t\boldsymbol{L},\boldsymbol{L}^{\top}\boldsymbol{\hat{\Sigma}}_t\boldsymbol{Q}\rangle-\lambda_{penalty}\langle\boldsymbol{\hat{\Sigma}}_t,\boldsymbol{Q}\boldsymbol{Q}^{\top}\rangle.
		\end{align*}
		Then we are to bound each term in the above equation. Notice that,
		\begin{equation*}
			\langle\boldsymbol{\hat{M}}_t,\boldsymbol{Q}\boldsymbol{Q}^{\top}\rangle=\text{vec}(\boldsymbol{Q})^{\top}(\boldsymbol{I}_k\otimes\boldsymbol{\hat{M}}_t)\text{vec}(\boldsymbol{Q})\geq 0
		\end{equation*}
		and
		\begin{equation*}
			\langle\boldsymbol{\hat{\Sigma}}_t,\boldsymbol{Q}\boldsymbol{Q}^{\top}\rangle=\text{vec}(\boldsymbol{Q})^{\top}(\boldsymbol{I}_k\otimes\boldsymbol{\hat{\Sigma}}_t)\text{vec}(\boldsymbol{Q})\geq 0.
		\end{equation*}
		Then by Cauchy–Schwarz inequality and sub-multiplicity of matrix norm, we have,
		\begin{align}
			\label{eq:smoothpara}
			&\frac{1}{2}\text{vec}(\boldsymbol{Q})^{\top}\nabla^2 f_t(\boldsymbol{L})\text{vec}(\boldsymbol{Q})\nonumber\\
			&\leq\lambda_{penalty}\|\boldsymbol{L}^{\top}\boldsymbol{\hat{\Sigma}}_t\|_{op}^2\|\boldsymbol{Q}\|^2_F+\lambda_{penalty}\|\boldsymbol{L}^{\top}\boldsymbol{\hat{\Sigma}}_t\boldsymbol{L}\|_F\|\boldsymbol{Q}^{\top}\boldsymbol{\hat{\Sigma}}_t\boldsymbol{Q}\|_F\nonumber\\
			&+\lambda_{penalty}\|\boldsymbol{Q}^{\top}\boldsymbol{\hat{\Sigma}}_t\boldsymbol{L}\|_F\|\boldsymbol{L}^{\top}\boldsymbol{\hat{\Sigma}}_t\boldsymbol{Q}\|_F.
		\end{align}
		Then, we show that for $\boldsymbol{L}$ satisfying $\|\boldsymbol{\hat{\Sigma}}^{1/2}_t(\boldsymbol{L}_*-\boldsymbol{L})\|_F\leq\eta$, the smoothness parameter can be bounded. By the assumption $\|\boldsymbol{\hat{\Sigma}}^{1/2}_t(\boldsymbol{L}_*-\boldsymbol{L})\|_F\leq\eta$, we have $$\|\boldsymbol{\hat{\Sigma}}^{1/2}_t(\boldsymbol{L}_*-\boldsymbol{L})\|_{op}\leq\|\boldsymbol{\hat{\Sigma}}^{1/2}_t(\boldsymbol{L}_*-\boldsymbol{L})\|_F\leq\eta.$$
		Thus, by triangle inequality, we have
		\begin{equation*}
			\|\boldsymbol{\hat{\Sigma}}^{1/2}_t\boldsymbol{L}_*\|_{op}-\eta\leq\|\boldsymbol{\hat{\Sigma}}^{1/2}_t\boldsymbol{L}\|_{op}\leq\|\boldsymbol{\hat{\Sigma}}^{1/2}_t\boldsymbol{L}_*\|_{op}+\eta.
		\end{equation*}
		Note that the solution $\boldsymbol{L}_*$ satisfies
		\begin{equation*}
			\boldsymbol{\hat{M}}_t\boldsymbol{L}_*=\lambda_{penalty}\boldsymbol{\hat{\Sigma}}_t\boldsymbol{L}_*(\boldsymbol{L}_*^T\boldsymbol{\hat{\Sigma}}_t\boldsymbol{L}_*-\boldsymbol{I}_k).
		\end{equation*}
		The matrix $\boldsymbol{L}_*^{\top}\boldsymbol{\hat{\Sigma}}_t\boldsymbol{L}_*-\boldsymbol{I}_k$ is $k\times k$ real symmetric, so there exists an eigen-decomposition of $\boldsymbol{L}_*^{\top}\boldsymbol{\hat{\Sigma}}_t\boldsymbol{L}_*-\boldsymbol{I}_k=\boldsymbol{R}\boldsymbol{\Lambda}\boldsymbol{R}^{\top}$, where $\boldsymbol{R}$ is a $k\times k$ orthogonal matrix and $\boldsymbol{\Lambda}$ is a diagonal matrix. Then,
		\begin{equation*}
			\boldsymbol{\hat{M}}_t\boldsymbol{L}_*\boldsymbol{R}=\lambda_{penalty}\boldsymbol{\hat{\Sigma}}_t\boldsymbol{L}_*\boldsymbol{R}\boldsymbol{\Lambda}.
		\end{equation*}
		Thus, the column space of the global minimizer of (\ref{eq:newloss}) coincides with the space spanned by some generalized eigenvectors, which is not necessarily the eigenvectors corresponding to the largest eigenvalues. By the fact that $\boldsymbol{L}_*$ is non-degenerate, we assume $\boldsymbol{L}_*=\boldsymbol{\hat{L}}_t\boldsymbol{D}_t$, where $\boldsymbol{\hat{L}}_t$ is the generalized eigenvectors with eigenvalues $\{\hat{\Lambda}_{t,i}\}_{i=1}^{k}$ corresponding to the matrix pair $(\boldsymbol{\hat{M}}_t,\boldsymbol{\hat{\Sigma}}_t)$ and $\boldsymbol{D}_t$ is an inverse matrix. We use $\boldsymbol{\hat{\Lambda}}_t$ to denote a diagnoal matrix with diagonal elements $\{\hat{\Lambda}_{t,i}\}_{i=1}^{k}$. Then, by plugging into the equation, we have
		\begin{align*}
			\boldsymbol{\hat{M}}_t\boldsymbol{\hat{L}}_t\boldsymbol{D}_t&=\lambda_{penalty}\boldsymbol{\hat{\Sigma}}_t\boldsymbol{\hat{L}}_t\boldsymbol{D}_t(\boldsymbol{D}_t^{\top}\boldsymbol{D}_t-\boldsymbol{I}_k),\\
			\boldsymbol{\hat{\Sigma}}_t\boldsymbol{\hat{L}}_t\boldsymbol{\hat{\Lambda}}_t\boldsymbol{D}_t&=\lambda_{penalty}\   \boldsymbol{\hat{\Sigma}}_t\boldsymbol{\hat{L}}_t\boldsymbol{D}_t(\boldsymbol{D}_t^{\top}\boldsymbol{D}_t-\boldsymbol{I}_k).
		\end{align*}
		Thus, the $\boldsymbol{D}_t$ can be chosen as $\boldsymbol{D}_t=(\boldsymbol{I}_k+\boldsymbol{\hat{\Lambda}}_t/\lambda_{penalty})^{1/2}$. In fact, we can say more about the eigenvalues in $\boldsymbol{\hat{\Lambda}}_t$. Notice that,
		\begin{align*}
			f_t(\boldsymbol{L}_*)&=-\langle\boldsymbol{\hat{M}},\boldsymbol{\hat{L}}_t\boldsymbol{D}_t\boldsymbol{D}_t^{\top}\boldsymbol{\hat{L}}_t^{\top}\rangle+\lambda_{penalty}\|\boldsymbol{D}_t^{\top}\boldsymbol{\hat{L}}_t^{\top}\boldsymbol{\hat{\Sigma}}_t\boldsymbol{\hat{L}}_t\boldsymbol{D}_t-\boldsymbol{I}_k\|^2_F\\
			&=-\text{Tr}(\boldsymbol{D}_t^{\top}\boldsymbol{\hat{\Lambda}}_t\boldsymbol{D}_t)+\lambda_{penalty}\|\boldsymbol{D}_t^{\top}\boldsymbol{D}_t-\boldsymbol{I}_k\|^2_F\\
			&=-\text{Tr}(\boldsymbol{\hat{\Lambda}}_t)-\frac{1}{\lambda_{penalty}}\text{Tr}(\boldsymbol{\hat{\Lambda}}_t^2)+\frac{1}{\lambda_{penalty}}\text{Tr}(\boldsymbol{\hat{\Lambda}}_t^2)=-\text{Tr}(\boldsymbol{\hat{\Lambda}}_t)\\
			&=-(\sum_{i=1}^{k}\hat{\Lambda}_{t,i}),
		\end{align*}
		which is minimized when $\{\hat{\Lambda}_{t,i}\}_{i=1}^{k}$ corresponding to the $k$ largest eigenvalues of matrix pair $(\boldsymbol{\hat{M}}_t,\boldsymbol{\hat{\Sigma}}_t)$. Thus, we can let $\boldsymbol{\hat{\Lambda}}_t=\boldsymbol{\hat{\Lambda}}_k(\mathcal{S}_t)$, where $\boldsymbol{\hat{\Lambda}}_k(\mathcal{S}_t)$ is the leading $k$ generalized eigenvalues of matrix pair $(\boldsymbol{\hat{M}}_t,\boldsymbol{\hat{\Sigma}}_t)$. Recall our definition of $\boldsymbol{\hat{B}}(\mathcal{S}_t)$, which coincides with $\boldsymbol{\hat{L}}_t$. Thus, we have
		\begin{equation*}
			\boldsymbol{L}_*=\boldsymbol{\hat{B}}(\mathcal{S}_t)\bigg(\boldsymbol{I}+\frac{\boldsymbol{\Lambda}_k(\mathcal{S}_t)}{\lambda_{penalty}}\bigg)^{1/2}=\boldsymbol{\hat{A}}(\mathcal{S}_t).
		\end{equation*}
		Then, we are back to consider the equation (\ref{eq:smoothpara}).
		\begin{align*}
			&\frac{1}{2}\text{vec}(\boldsymbol{Q})\nabla^2 f_t(\boldsymbol{L})\text{vec}(\boldsymbol{Q})\nonumber\\
			&\leq\lambda_{penalty}\|\boldsymbol{L}^{\top}\boldsymbol{\hat{\Sigma}}_t^{1/2}\|_{op}^2\|\boldsymbol{Q}\|^2_F+\lambda_{penalty}\|\boldsymbol{L}^{\top}\boldsymbol{\hat{\Sigma}}_t\boldsymbol{L}\|_{op}\|\boldsymbol{Q}^{\top}\boldsymbol{\hat{\Sigma}}_t\boldsymbol{Q}\|_F\\
			&+\lambda_{penalty}\|\boldsymbol{Q}^{\top}\boldsymbol{\hat{\Sigma}}_t\boldsymbol{L}\|_F\|\boldsymbol{L}^{\top}\boldsymbol{\hat{\Sigma}}_t\boldsymbol{Q}\|_F\nonumber\\
			&\leq \lambda_{penalty}(\eta+\|\boldsymbol{\hat{A}}(\mathcal{S}_t)^{\top}\boldsymbol{\hat{\Sigma}}_t^{1/2}\|_{op})^2\|\boldsymbol{Q}\|^2_F+2\lambda_{penalty}(\eta+\|\boldsymbol{\hat{A}}(\mathcal{S}_t)^{\top}\boldsymbol{\hat{\Sigma}}_t^{1/2}\|_{op})^2\lambda_x\|\boldsymbol{Q}\|^2_F\\
			&=\lambda_{penalty}(1+2\lambda_x)(\eta+\|\boldsymbol{\hat{A}}(\mathcal{S}_t)^{\top}\boldsymbol{\hat{\Sigma}}_t^{1/2}\|_{op})^2\|\boldsymbol{Q}\|^2_F\\
			&=\lambda_{penalty}(1+2\lambda_x)\big(\eta+\sqrt{1+\hat{\lambda}_1/\lambda_{penalty}}\big)^2\|\boldsymbol{Q}\|^2_F\\
			&\leq\lambda_{penalty}(1+2\lambda_x)\big(\eta+\sqrt{1+2\lambda_1/\lambda_{penalty}}\big)^2\|\boldsymbol{Q}\|^2_F,
		\end{align*}
		where we use the initial condition of $\boldsymbol{\hat{A}}(\mathcal{S}_t)$ and the convergence of $\boldsymbol{\hat{\Sigma}}$ in the second inequality, we use the following relation in the third equality,
		\begin{equation*}
			\begin{split}
				\|\boldsymbol{\hat{A}}(\mathcal{S}_t)^T\boldsymbol{\hat{\Sigma}}_t^{1/2}\|_{op}&=\bigg\|\boldsymbol{\hat{\Sigma}}_t^{1/2}\boldsymbol{\hat{B}}(\mathcal{S}_t)\bigg(1+\frac{\boldsymbol{\Lambda}_K(\mathcal{S}_t)}{\lambda_{penalty}}\bigg)^{1/2}\bigg\|_{op}\\
				&=\sqrt{1+\frac{\hat{\lambda}_1}{\lambda_{penalty}}}\leq\sqrt{1+\frac{2\lambda_1}{\lambda_{penalty}}}.
			\end{split}
		\end{equation*}
		The last inequality follows similar arguments about convergence of $\widehat{\lambda}_1$  in the proof of Lemma \ref{lemma:oracle}. Thus, we can choose $\beta=\lambda_{penalty}(1+2\lambda_x)\big(\eta+\sqrt{1+2\lambda_1/\lambda_{penalty}}\big)^2$.
		
		\emph{Step 2}: Strong convexity parameter $\alpha$. To simplify the notation, we define $\boldsymbol{\tilde{Z}}:=\boldsymbol{Z}\boldsymbol{H}_{\boldsymbol{Z}}-\boldsymbol{\hat{A}}(\mathcal{S}_t)$. Then, we first consider the case where $\boldsymbol{L}=\boldsymbol{\hat{A}}(\mathcal{S}_t)$. By algebra, we have
		\begin{align*}
			&\frac{1}{2}\text{vec}(\boldsymbol{\tilde{Z}})^{\top}\nabla^2 f_t\{\boldsymbol{\hat{A}}(\mathcal{S}_t)\}\text{vec}(\boldsymbol{\tilde{Z}})\\
			=&-\langle\boldsymbol{\hat{M}}_t,\boldsymbol{\tilde{Z}}\boldsymbol{\tilde{Z}}^{\top}\rangle+\lambda_{penalty}\langle\boldsymbol{\hat{\Sigma}}_t\boldsymbol{\hat{A}}(\mathcal{S}_t)\boldsymbol{\hat{A}}(\mathcal{S}_t)^{\top}\boldsymbol{\hat{\Sigma}}_t,\boldsymbol{\tilde{Z}}\boldsymbol{\tilde{Z}}^{\top}\rangle+\lambda_{penalty}\langle\boldsymbol{\hat{A}}(\mathcal{S}_t)^{\top}\boldsymbol{\hat{\Sigma}}_t\boldsymbol{\hat{A}}(\mathcal{S}_t),\boldsymbol{\tilde{Z}}^{\top}\boldsymbol{\hat{\Sigma}}_t\boldsymbol{\tilde{Z}}\rangle\\
			&+\lambda_{penalty}\langle\boldsymbol{\tilde{Z}}^{\top}\boldsymbol{\hat{\Sigma}}_t\boldsymbol{\hat{A}}(\mathcal{S}_t),\boldsymbol{\hat{A}}(\mathcal{S}_t)^{\top}\boldsymbol{\hat{\Sigma}}_t\boldsymbol{\tilde{Z}}\rangle-\lambda_{penalty}\langle\boldsymbol{\hat{\Sigma}}_t,\boldsymbol{\tilde{Z}}\boldsymbol{\tilde{Z}}^{\top}\rangle\\
			=&-\langle\boldsymbol{\hat{M}}_t,\boldsymbol{\tilde{Z}}\boldsymbol{\tilde{Z}}^{\top}\rangle+\lambda_{penalty}\langle\boldsymbol{\hat{\Sigma}}_t\boldsymbol{\hat{A}}(\mathcal{S}_t)\boldsymbol{\hat{A}}(\mathcal{S}_t)^{\top}\boldsymbol{\hat{\Sigma}}_t,\boldsymbol{\tilde{Z}}\boldsymbol{\tilde{Z}}^{\top}\rangle\\
			&+\lambda_{penalty}\langle\boldsymbol{I}_k+\boldsymbol{\hat{\Lambda}}_k(\mathcal{S}_t)/\lambda_{penalty},\boldsymbol{\tilde{Z}}^{\top}\boldsymbol{\hat{\Sigma}}_t\boldsymbol{\tilde{Z}}\rangle\\
			&+\lambda_{penalty}\langle\boldsymbol{\tilde{Z}}\boldsymbol{\tilde{Z}}^{\top},\boldsymbol{\hat{\Sigma}}_t\boldsymbol{\hat{A}}(\mathcal{S}_t)\boldsymbol{\hat{A}}(\mathcal{S}_t)^{\top}\boldsymbol{\hat{\Sigma}}_t\rangle-\lambda_{penalty}\langle\boldsymbol{\hat{\Sigma}}_t,\boldsymbol{\tilde{Z}}\boldsymbol{\tilde{Z}}^{\top}\rangle\\
			=&-\langle\boldsymbol{\hat{M}}_t,\boldsymbol{\tilde{Z}}\boldsymbol{\tilde{Z}}^{\top}\rangle+\lambda_{penalty}\langle\boldsymbol{\hat{\Sigma}}_t\boldsymbol{\hat{A}}(\mathcal{S}_t)\boldsymbol{\hat{A}}(\mathcal{S}_t)^{\top}\boldsymbol{\hat{\Sigma}}_t,\boldsymbol{\tilde{Z}}\boldsymbol{\tilde{Z}}^{\top}\rangle+\langle\boldsymbol{\hat{\Lambda}}_K(\mathcal{S}_t),\boldsymbol{\tilde{Z}}^{\top}\boldsymbol{\hat{\Sigma}}_t\boldsymbol{\tilde{Z}}\rangle\\
			&+\lambda_{penalty}\langle\boldsymbol{\tilde{Z}}^{\top}\boldsymbol{\hat{\Sigma}}_t\boldsymbol{\hat{A}}(\mathcal{S}_t),\boldsymbol{\hat{A}}(\mathcal{S}_t)^{\top}\boldsymbol{\hat{\Sigma}}_t\boldsymbol{\tilde{Z}}\rangle.
		\end{align*}
		We consider two lines separately. For the first line,
		\begin{align*}
			&-\langle\boldsymbol{\hat{M}}_t,\boldsymbol{\tilde{Z}}\boldsymbol{\tilde{Z}}^{\top}\rangle+\lambda_{penalty}\langle\boldsymbol{\hat{\Sigma}}_t\boldsymbol{\hat{A}}(\mathcal{S}_t)\boldsymbol{\hat{A}}(\mathcal{S}_t)^{\top}\boldsymbol{\hat{\Sigma}}_t,\boldsymbol{\tilde{Z}}\boldsymbol{\tilde{Z}}^{\top}\rangle+\langle\boldsymbol{\hat{\Lambda}}_k(\mathcal{S}_t),\boldsymbol{\tilde{Z}}^{\top}\boldsymbol{\hat{\Sigma}}_t\boldsymbol{\tilde{Z}}\rangle\\
			=&\langle\boldsymbol{\tilde{Z}}\boldsymbol{\tilde{Z}}^{\top},\lambda_{penalty}\boldsymbol{\hat{\Sigma}}_t\boldsymbol{\hat{A}}(\mathcal{S}_t)\boldsymbol{\hat{A}}(\mathcal{S}_t)^{\top}\boldsymbol{\hat{\Sigma}}_t-\boldsymbol{\hat{M}}_t\rangle+\langle\boldsymbol{\tilde{Z}}\boldsymbol{\hat{\Lambda}}_k(\mathcal{S}_t)\boldsymbol{\tilde{Z}}^{\top},\boldsymbol{\hat{\Sigma}}_t\rangle\\
			\geq&\langle\boldsymbol{\tilde{Z}}\boldsymbol{\tilde{Z}}^{\top},\lambda_{penalty}\boldsymbol{\hat{\Sigma}}_t\boldsymbol{\hat{A}}(\mathcal{S}_t)\boldsymbol{\hat{A}}(\mathcal{S}_t)^{\top}\boldsymbol{\hat{\Sigma}}_t-\boldsymbol{\hat{M}}_t+\hat{\lambda}_k\boldsymbol{\hat{\Sigma}}_t\rangle\\
			=&\langle\boldsymbol{\tilde{Z}}\boldsymbol{\tilde{Z}}^{\top},\boldsymbol{\hat{\Sigma}}_t^{1/2}\{\lambda_{penalty}\boldsymbol{\hat{\Sigma}}_t^{1/2}\boldsymbol{\hat{A}}(\mathcal{S}_t)\boldsymbol{\hat{A}}(\mathcal{S}_t)^{\top}\boldsymbol{\hat{\Sigma}}_t^{1/2}-\boldsymbol{\hat{\Sigma}}_t^{-1/2}\boldsymbol{\hat{M}}_t\boldsymbol{\hat{\Sigma}}_t^{-1/2}+\hat{\lambda}_k\boldsymbol{I}_{|\mathcal{S}_t|}\}\boldsymbol{\hat{\Sigma}}_t^{1/2}\rangle\\
			=&\langle\boldsymbol{\tilde{Z}}\boldsymbol{\tilde{Z}}^{\top},\boldsymbol{\hat{\Sigma}}_t^{1/2}\{\lambda_{penalty}\boldsymbol{\hat{\Sigma}}_t^{1/2}\boldsymbol{\hat{B}}(\mathcal{S}_t)(\boldsymbol{I}_k+\boldsymbol{\Lambda}_k(\mathcal{S}_t)/\lambda_{penalty})\boldsymbol{\hat{B}}(\mathcal{S}_t)^{\top}\boldsymbol{\hat{\Sigma}}_t^{1/2}\\
			&-\boldsymbol{\hat{\Sigma}}_t^{-1/2}\boldsymbol{\hat{M}}_t\boldsymbol{\hat{\Sigma}}_t^{-1/2}+\hat{\lambda}_k\boldsymbol{I}_{|\mathcal{S}_t|}\}\boldsymbol{\hat{\Sigma}}_t^{1/2}\rangle:=\text{Line1},
		\end{align*}
		where we use the fact that $\widehat{\lambda}_k$ is the $k$-th largest generalized eigenvalue in the second inequality. Note that the generalized eigenvalue problem with respect to the matrix pair $(\boldsymbol{\hat{M}}_t,\boldsymbol{\hat{\Sigma}}_t)$ satisfies,
		\begin{align*}
			\boldsymbol{\hat{M}}_t\boldsymbol{\hat{B}}(\mathcal{S}_t)&=\boldsymbol{\Lambda}_k(\mathcal{S}_t)\boldsymbol{\hat{\Sigma}}_t\boldsymbol{\hat{B}}(\mathcal{S}_t)\\
			(\boldsymbol{\hat{\Sigma}}_t^{-1/2}\boldsymbol{\hat{M}}_t\boldsymbol{\hat{\Sigma}}_t^{-1/2})(\boldsymbol{\hat{\Sigma}}_t^{1/2}\boldsymbol{\hat{B}}(\mathcal{S}_t))&=\boldsymbol{\Lambda}_k(\mathcal{S}_t)(\boldsymbol{\hat{\Sigma}}_t^{1/2}\boldsymbol{\hat{B}}(\mathcal{S}_t)).
		\end{align*}
		By the fact that the matrix $\boldsymbol{\hat{\Sigma}}_t^{-1/2}\boldsymbol{\hat{M}}_t\boldsymbol{\hat{\Sigma}}_t^{-1/2}$ is real and symmetric, then there exists an eigen-decomposition. From the previous equation, we know $\{\boldsymbol{\hat{\Sigma}}_t^{1/2}\boldsymbol{\hat{B}}(\mathcal{S}_t),\boldsymbol{\Lambda}_k(\mathcal{S}_t)\}$ is part of eigenpairs of $\boldsymbol{\hat{\Sigma}}_t^{-1/2}\boldsymbol{\hat{M}}_t\boldsymbol{\hat{\Sigma}}_t^{-1/2}$. Then, we have
		\begin{equation*}
			\boldsymbol{\hat{\Sigma}}_t^{-1/2}\boldsymbol{\hat{M}}_t\boldsymbol{\hat{\Sigma}}_t^{-1/2}=(\boldsymbol{\hat{\Sigma}}_t^{1/2}\boldsymbol{\hat{B}}(\mathcal{S}_t))\boldsymbol{\Lambda}_k(\mathcal{S}_t)(\boldsymbol{\hat{B}}(\mathcal{S}_t)^{\top}\boldsymbol{\hat{\Sigma}}_t^{1/2})+\boldsymbol{\hat{\Sigma}}_t^{1/2}\boldsymbol{\hat{B}}_c\boldsymbol{\Lambda}_c\B_c^{\top}\boldsymbol{\hat{\Sigma}}_t^{1/2},
		\end{equation*}
		where $\{\boldsymbol{\hat{\Sigma}}_t^{1/2}\boldsymbol{\hat{B}}_c,\boldsymbol{\Lambda}_c\}$ is the remaining eigenpairs. Then, we have for the first line,
		\begin{align*}
			\text{Line1}&\geq\langle\boldsymbol{\tilde{Z}}\boldsymbol{\tilde{Z}}^{\top},\boldsymbol{\hat{\Sigma}}_t^{1/2}\{\lambda_{penalty}\boldsymbol{\hat{\Sigma}}_t^{1/2}\boldsymbol{\hat{B}}(\mathcal{S}_t)\boldsymbol{\hat{B}}(\mathcal{S}_t)^{\top}\boldsymbol{\hat{\Sigma}}_t^{1/2}-\boldsymbol{\hat{\Sigma}}_t^{1/2}\boldsymbol{\hat{B}}_c\boldsymbol{\Lambda}_c\B_c^{\top}\boldsymbol{\hat{\Sigma}}_t^{1/2}+\hat{\lambda}_k\boldsymbol{I}_{|\mathcal{S}_t|}\}\boldsymbol{\hat{\Sigma}}_t^{1/2}\rangle\\
			&=\langle\boldsymbol{\tilde{Z}}\boldsymbol{\tilde{Z}}^{\top},\boldsymbol{\hat{\Sigma}}_t^{1/2}\{(\lambda_{penalty}+\hat{\lambda}_k)\boldsymbol{\hat{\Sigma}}_t^{1/2}\boldsymbol{\hat{B}}(\mathcal{S}_t)\boldsymbol{\hat{B}}(\mathcal{S}_t)^{\top}\boldsymbol{\hat{\Sigma}}_t^{1/2}\\
			&+\boldsymbol{\hat{\Sigma}}_t^{1/2}\boldsymbol{\hat{B}}_c(\hat{\lambda}_k\boldsymbol{I}_{|\mathcal{S}_t|-k}-\boldsymbol{\Lambda}_c)\B_c^{\top}\boldsymbol{\hat{\Sigma}}_t^{1/2}\}\boldsymbol{\hat{\Sigma}}_t^{1/2}\rangle\\
			&=\langle\boldsymbol{\tilde{Z}}\boldsymbol{\tilde{Z}}^{\top},(\lambda_{penalty}+\hat{\lambda}_k)\boldsymbol{\hat{\Sigma}}_t\boldsymbol{\hat{B}}(\mathcal{S}_t)\boldsymbol{\hat{B}}(\mathcal{S}_t)^{\top}\boldsymbol{\hat{\Sigma}}_t+\boldsymbol{\hat{\Sigma}}_t\boldsymbol{\hat{B}}_c(\hat{\lambda}_k\boldsymbol{I}_{|\mathcal{S}_t|-k}-\boldsymbol{\Lambda}_c)\B_c^{\top}\boldsymbol{\hat{\Sigma}}_t\rangle\\
			&\geq \langle\boldsymbol{\tilde{Z}}\boldsymbol{\tilde{Z}}^{\top},(\lambda_{penalty}+\hat{\lambda}_k)\boldsymbol{\hat{\Sigma}}_t\boldsymbol{\hat{B}}(\mathcal{S}_t)\boldsymbol{\hat{B}}(\mathcal{S}_t)^{\top}\boldsymbol{\hat{\Sigma}}_t+(\hat{\lambda}_k-\hat{\lambda}_{k+1})\boldsymbol{\hat{\Sigma}}_t\boldsymbol{\hat{B}}_c\B_c^{\top}\boldsymbol{\hat{\Sigma}}_t\rangle\\
			&\geq\lambda_x^{-1}\langle\boldsymbol{\tilde{Z}}\boldsymbol{\tilde{Z}}^{\top},(\hat{\lambda}_k-\hat{\lambda}_{k+1})\boldsymbol{I}_{|\mathcal{S}|}\rangle=\lambda_x^{-1}(\hat{\lambda}_k-\hat{\lambda}_{k+1})\|\boldsymbol{\tilde{Z}}\|^2_F.
		\end{align*}
		where we use the decomposition of identity matrix in the first inequality, we use the fact that eigenvalues are ordered in the third inequality, and we use the convergence of $\boldsymbol{\hat{\Sigma}}$ in the last inequality . The second line satisfies,
		\begin{align*}
			&\lambda_{penalty}\langle\boldsymbol{\tilde{Z}}^{\top}\boldsymbol{\hat{\Sigma}}_t\boldsymbol{\hat{A}}(\mathcal{S}_t),\boldsymbol{\hat{A}}(\mathcal{S}_t)^{\top}\boldsymbol{\hat{\Sigma}}_t\boldsymbol{\tilde{Z}}\rangle=\lambda_{penalty}\text{Tr}\{\boldsymbol{\hat{A}}(\mathcal{S}_t)^{\top}\boldsymbol{\hat{\Sigma}}_t\boldsymbol{\tilde{Z}}\boldsymbol{\hat{A}}(\mathcal{S}_t)^{\top}\boldsymbol{\hat{\Sigma}}_t\boldsymbol{\tilde{Z}}\}\geq 0.
		\end{align*}
		Then, it remains to consider the term $|\frac{1}{2}\text{vec}(\boldsymbol{\tilde{Z}})^{\top}[\nabla^2 f_t\{\boldsymbol{\hat{A}}(\mathcal{S}_t)\}-\nabla^2f_t(\boldsymbol{L})]\text{vec}(\boldsymbol{\tilde{Z}})|$. Notice that,
		\begin{align*}
			&\frac{1}{2}\text{vec}(\boldsymbol{\tilde{Z}})^{\top}[\nabla^2 f_t\{\boldsymbol{\hat{A}}(\mathcal{S}_t)\}-\nabla^2f_t(\boldsymbol{L})]\text{vec}(\boldsymbol{\tilde{Z}})\\
			=&\lambda_{penalty}\langle\boldsymbol{\hat{\Sigma}}_t\{\boldsymbol{\hat{A}}(\mathcal{S}_t)\boldsymbol{\hat{A}}(\mathcal{S}_t)^{\top}-\boldsymbol{L}\boldsymbol{L}^{\top}\}\boldsymbol{\hat{\Sigma}}_t,\boldsymbol{\tilde{Z}}\boldsymbol{\tilde{Z}}^{\top}\rangle\\
			&+\lambda_{penalty}\langle\boldsymbol{\hat{A}}(\mathcal{S}_t)^{\top}\boldsymbol{\hat{\Sigma}}_t\boldsymbol{\hat{A}}(\mathcal{S}_t)-\boldsymbol{L}^{\top}\boldsymbol{\hat{\Sigma}}_t\boldsymbol{L},\boldsymbol{\tilde{Z}}^{\top}\boldsymbol{\hat{\Sigma}}_t\boldsymbol{\tilde{Z}}\rangle\\
			+&\lambda_{penalty}\langle\boldsymbol{\tilde{Z}}^{\top}\boldsymbol{\hat{\Sigma}}_t\boldsymbol{\hat{A}}(\mathcal{S}_t),\boldsymbol{\hat{A}}(\mathcal{S}_t)^{\top}\boldsymbol{\hat{\Sigma}}_t\boldsymbol{\tilde{Z}}\rangle-\lambda_{penalty}\langle\boldsymbol{\tilde{Z}}^{\top}\boldsymbol{\hat{\Sigma}}_t\boldsymbol{L},\boldsymbol{L}^{\top}\boldsymbol{\hat{\Sigma}}_t\boldsymbol{\tilde{Z}}\rangle.
		\end{align*}
		For the first term, we have
		\begin{align*}
			&|\lambda_{penalty}\langle\boldsymbol{\hat{\Sigma}}_t\{\boldsymbol{\hat{A}}(\mathcal{S}_t)\boldsymbol{\hat{A}}(\mathcal{S}_t)^{\top}-\boldsymbol{L}\boldsymbol{L}^{\top}\}\boldsymbol{\hat{\Sigma}}_t,\boldsymbol{\tilde{Z}}\boldsymbol{\tilde{Z}}^{\top}\rangle|\\
			\leq&\lambda_{penalty}\|\boldsymbol{\hat{\Sigma}}_t\{\boldsymbol{\hat{A}}(\mathcal{S}_t)\boldsymbol{\hat{A}}(\mathcal{S}_t)^{\top}-\boldsymbol{L}\boldsymbol{L}^{\top}\}\boldsymbol{\hat{\Sigma}}_t\|_{op}\|\boldsymbol{\tilde{Z}}\|_F^2\\
			\leq&\lambda_{penalty}\lambda_x\|\boldsymbol{\hat{\Sigma}}_t^{1/2}\{\boldsymbol{\hat{A}}(\mathcal{S}_t)\boldsymbol{\hat{A}}(\mathcal{S}_t)^{\top}-\boldsymbol{L}\boldsymbol{L}^{\top}\}\boldsymbol{\hat{\Sigma}}_t^{1/2}\|_{op}\|\boldsymbol{\tilde{Z}}\|_F^2\\
			\leq&\lambda_{penalty}\lambda_x\|\boldsymbol{\hat{\Sigma}}_t^{1/2}\{\boldsymbol{\hat{A}}(\mathcal{S}_t)\boldsymbol{\hat{A}}(\mathcal{S}_t)^{\top}-\boldsymbol{L}\boldsymbol{L}^{\top}\}\boldsymbol{\hat{\Sigma}}_t^{1/2}\|_{F}\|\boldsymbol{\tilde{Z}}\|_F^2\\
			\leq&\frac{9}{4}\lambda_{penalty}\lambda_x\|\boldsymbol{\hat{\Sigma}}_t^{1/2}\boldsymbol{L}\|_{op}\|\boldsymbol{\hat{\Sigma}}_t^{1/2}\boldsymbol{L}-\boldsymbol{\hat{\Sigma}}_t^{1/2}\boldsymbol{\hat{A}}(\mathcal{S}_t)\|_F\|\boldsymbol{\tilde{Z}}\|_F^2\\
			\leq&\frac{9}{4}\lambda_{penalty}\lambda_x^{3/2}(\eta+\|\boldsymbol{\hat{\Sigma}}^{1/2}\boldsymbol{L}^{*}\|_{op})\|\boldsymbol{L}-\boldsymbol{\hat{A}}(\mathcal{S}_t)\|_F\|\boldsymbol{\tilde{Z}}\|_F^2\\
			\leq&\frac{9}{4}\lambda_{penalty}\lambda_x^{3/2}(\eta+\sqrt{1+2\lambda_1/\lambda_{penalty}})\|\boldsymbol{L}-\boldsymbol{\hat{A}}(\mathcal{S}_t)\|_F\|\boldsymbol{\tilde{Z}}\|_F^2,
		\end{align*}
		where we use the Cauchy-Swartz inequality in the first, fourth, fifth inequality, we use Lemma 5.3 of \cite{tu2016low} in the third inequality. To apply Lemma 5.3, we assume for now $\|\boldsymbol{\hat{\Sigma}}^{1/2}_t\boldsymbol{L}-\boldsymbol{\hat{\Sigma}}_t^{1/2}\boldsymbol{\hat{A}}(\mathcal{S}_t)\|_F\leq\frac{1}{4}\|\boldsymbol{\hat{\Sigma}}^{1/2}\boldsymbol{L}^{*}\|_{op}$, which can be satisfied by putting the conditions for the initial values.
		
		Similarlt, for the second term, we have
		\begin{align*}
			&|\lambda_{penalty}\langle\boldsymbol{\hat{A}}(\mathcal{S}_t)^{\top}\boldsymbol{\hat{\Sigma}}_t\boldsymbol{\hat{A}}(\mathcal{S}_t)-\boldsymbol{L}^{\top}\boldsymbol{\hat{\Sigma}}_t\boldsymbol{L},\boldsymbol{\tilde{Z}}^{\top}\boldsymbol{\hat{\Sigma}}_t\boldsymbol{\tilde{Z}}\rangle|\\
			\leq&\lambda_{penalty}\|(\boldsymbol{\hat{A}}(\mathcal{S}_t)^{\top}\boldsymbol{\hat{\Sigma}}_t\boldsymbol{\hat{A}}(\mathcal{S}_t)-\boldsymbol{L}^{\top}\boldsymbol{\hat{\Sigma}}_t\boldsymbol{L})\boldsymbol{\tilde{Z}}^{\top}\|_{op}\|\boldsymbol{\hat{\Sigma}}_t\boldsymbol{\tilde{Z}}\|_F\\
			\leq&\lambda_{penalty}\|(\boldsymbol{\hat{A}}(\mathcal{S}_t)^{\top}\boldsymbol{\hat{\Sigma}}_t\boldsymbol{\hat{A}}(\mathcal{S}_t)-\boldsymbol{L}^{\top}\boldsymbol{\hat{\Sigma}}_t\boldsymbol{L})\boldsymbol{\tilde{Z}}^{\top}\|_{F}\|\boldsymbol{\hat{\Sigma}}_t\boldsymbol{\tilde{Z}}\|_F\\
			\leq&\lambda_{penalty}\lambda_x\|(\boldsymbol{\hat{A}}(\mathcal{S}_t)^{\top}\boldsymbol{\hat{\Sigma}}_t\boldsymbol{\hat{A}}(\mathcal{S}_t)-\boldsymbol{L}^{\top}\boldsymbol{\hat{\Sigma}}_t\boldsymbol{L})\boldsymbol{\tilde{Z}}^{\top}\|_{F}\|\boldsymbol{\tilde{Z}}\|_F\\
			\leq&\lambda_{penalty}\lambda_x\|\boldsymbol{\hat{A}}(\mathcal{S}_t)^{\top}\boldsymbol{\hat{\Sigma}}_t\boldsymbol{\hat{A}}(\mathcal{S}_t)-\boldsymbol{L}^{\top}\boldsymbol{\hat{\Sigma}}_t\boldsymbol{L}\|_{op}\|\boldsymbol{\tilde{Z}}\|_F^2\\
			=&\lambda_{penalty}\lambda_x\|(\boldsymbol{\hat{A}}(\mathcal{S}_t)-\boldsymbol{L})^{\top}\boldsymbol{\hat{\Sigma}}_t\boldsymbol{\hat{A}}(\mathcal{S}_t)+\boldsymbol{\hat{A}}(\mathcal{S}_t)^{\top}\boldsymbol{\hat{\Sigma}}_t(\boldsymbol{\hat{A}}(\mathcal{S}_t)-\boldsymbol{L})\\
			&-(\boldsymbol{\hat{A}}(\mathcal{S}_t)-\boldsymbol{L})^{\top}\boldsymbol{\hat{\Sigma}}_t(\boldsymbol{\hat{A}}(\mathcal{S}_t)-\boldsymbol{L})\|_{op}\|\boldsymbol{\tilde{Z}}\|_F^2\\
			\leq&\lambda_{penalty}\lambda_x\{\|(\boldsymbol{\hat{A}}(\mathcal{S}_t)-\boldsymbol{L})^{\top}\boldsymbol{\hat{\Sigma}}_t\boldsymbol{\hat{A}}(\mathcal{S}_t)\|_{op}+\|\boldsymbol{\hat{A}}(\mathcal{S}_t)^{\top}\boldsymbol{\hat{\Sigma}}_t(\boldsymbol{\hat{A}}(\mathcal{S}_t)-\boldsymbol{L})\|_{op}\\
			&+\|(\boldsymbol{\hat{A}}(\mathcal{S}_t)-\boldsymbol{L})^{\top}\boldsymbol{\hat{\Sigma}}_t(\boldsymbol{\hat{A}}(\mathcal{S}_t)-\boldsymbol{L})\|_{op}\}\|\boldsymbol{\tilde{Z}}\|_F^2\\
			=&\lambda_{penalty}\lambda_x\{2\|\boldsymbol{\hat{\Sigma}}_t^{1/2}\boldsymbol{\hat{A}}(\mathcal{S}_t)\|_{op}\|\boldsymbol{\hat{\Sigma}}_t^{1/2}(\boldsymbol{\hat{A}}(\mathcal{S}_t)-\boldsymbol{L})\|_{op}+\|\boldsymbol{\hat{\Sigma}}_t^{1/2}(\boldsymbol{\hat{A}}(\mathcal{S}_t)-\boldsymbol{L})\|_{op}^2\}\|\boldsymbol{\tilde{Z}}\|_F^2\\
			\leq&\lambda_{penalty}\lambda_x\bigg(\mu+2\sqrt{1+\frac{\hat{\lambda}_1}{\lambda_{penalty}}}\bigg)\|\boldsymbol{\hat{\Sigma}}_t^{1/2}(\boldsymbol{\hat{A}}(\mathcal{S}_t)-\boldsymbol{L})\|_{op}\|\boldsymbol{\tilde{Z}}\|_F^2\\
			\leq&\lambda_{penalty}\lambda_x^{3/2}\bigg(\mu+2\sqrt{1+\frac{2\lambda_1}{\lambda_{penalty}}}\bigg)\|\boldsymbol{\hat{A}}(\mathcal{S}_t)-\boldsymbol{L}\|_{F}\|\boldsymbol{\tilde{Z}}\|_F^2.
		\end{align*}
		where we use similar procedures as for the first term. The third term satisfies,
		\begin{align*}
			&\lambda_{penalty}|\langle\boldsymbol{\tilde{Z}}^{\top}\boldsymbol{\hat{\Sigma}}_t\boldsymbol{\hat{A}}(\mathcal{S}_t),\boldsymbol{\hat{A}}(\mathcal{S}_t)^{\top}\boldsymbol{\hat{\Sigma}}_t\boldsymbol{\tilde{Z}}\rangle-\langle\boldsymbol{\tilde{Z}}^{\top}\boldsymbol{\hat{\Sigma}}_t\boldsymbol{L},\boldsymbol{L}^{\top}\boldsymbol{\hat{\Sigma}}_t\boldsymbol{\tilde{Z}}\rangle|\\
			=&\lambda_{penalty}|\text{Tr}\{(\boldsymbol{\tilde{Z}}^{\top}\boldsymbol{\hat{\Sigma}}_t\boldsymbol{\hat{A}}(\mathcal{S}_t))^2-(\boldsymbol{\tilde{Z}}^{\top}\boldsymbol{\hat{\Sigma}}_t\boldsymbol{L})^2\}|\\
			=&\lambda_{penalty}|\text{Tr}\{\boldsymbol{\tilde{Z}}^{\top}\boldsymbol{\hat{\Sigma}}_t(\boldsymbol{\hat{A}}(\mathcal{S}_t)+\boldsymbol{L})\boldsymbol{\tilde{Z}}^{\top}\boldsymbol{\hat{\Sigma}}_t(\boldsymbol{\hat{A}}(\mathcal{S}_t)-\boldsymbol{L})\}|\\
			\leq&\lambda_{penalty}\|\boldsymbol{\tilde{Z}}^{\top}\boldsymbol{\hat{\Sigma}}_t(\boldsymbol{\hat{A}}(\mathcal{S}_t)+\boldsymbol{L})\|_F\|\boldsymbol{\tilde{Z}}^{\top}\boldsymbol{\hat{\Sigma}}_t(\boldsymbol{\hat{A}}(\mathcal{S}_t)-\boldsymbol{L})\|_F\\
			\leq&\lambda_{penalty}\lambda_x^{3/2}\|\boldsymbol{\tilde{Z}}\|_F\|\boldsymbol{L}-\boldsymbol{\hat{A}}(\mathcal{S}_t)\|_F\|\boldsymbol{\tilde{Z}}\|_F\|\boldsymbol{\hat{\Sigma}}_t^{1/2}(\boldsymbol{\hat{A}}(\mathcal{S}_t)+\boldsymbol{L})\|_{op}\\
			\leq&\lambda_{penalty}\lambda_x^{3/2}\bigg(\mu+2\sqrt{1+\frac{2\lambda_1}{\lambda_{penalty}}}\bigg)\|\boldsymbol{L}-\boldsymbol{\hat{A}}(\mathcal{S}_t)\|_F\|\boldsymbol{\tilde{Z}}\|_F^2,
		\end{align*}
		where we use similar procedures as for the first term. Thus,
		\begin{align*}
			\frac{1}{2}|\text{vec}(\boldsymbol{\tilde{Z}})^{\top}\{\nabla^2 f_t(\boldsymbol{\hat{A}}(\mathcal{S}_t))-\nabla^2f_t(\boldsymbol{L})\}\text{vec}(\boldsymbol{\tilde{Z}})|&\leq c_1\|\boldsymbol{L}-\boldsymbol{\hat{A}}(\mathcal{S}_t)\|_F\|\boldsymbol{\tilde{Z}}\|_F^2,
		\end{align*}
		where
		\begin{equation*}
			c_1=\lambda_{penalty}\lambda_x^{3/2}\bigg\{2\mu+4\sqrt{1+\frac{2\lambda_1}{\lambda_{penalty}}}+\frac{9}{4}(\eta+\sqrt{1+\frac{2\lambda_1}{\lambda_{penalty}}})\bigg\}.
		\end{equation*}
		Thus, for $\|\boldsymbol{L}-\boldsymbol{\hat{A}}(\mathcal{S}_t)\|_F\leq (\hat{\lambda}_k-\hat{\lambda}_{k+1})/(2\lambda_x c_1)$, we have
		\begin{align*}
			&\frac{1}{2}\text{vec}(\boldsymbol{\tilde{Z}})^{\top}\nabla^2f_t(\boldsymbol{L})\text{vec}(\boldsymbol{\tilde{Z}})\\ 
			\geq&\frac{1}{2}\text{vec}(\boldsymbol{\tilde{Z}})^{\top}\nabla^2f_t(\boldsymbol{\hat{A}}(\mathcal{S}_t))\text{vec}(\boldsymbol{\tilde{Z}})-\frac{1}{2}|\text{vec}(\boldsymbol{\tilde{Z}})\{\nabla^2 f_t(\boldsymbol{\hat{A}}(\mathcal{S}_t))-\nabla^2f_t(\boldsymbol{L})\}\text{vec}(\boldsymbol{\tilde{Z}})|\\
			\geq&\lambda_x^{-1}(\lambda_k-\lambda_{k+1})\|\boldsymbol{\tilde{Z}}\|_F^2-c\|\boldsymbol{L}-\boldsymbol{\hat{A}}(\mathcal{S}_t)\|_F|\boldsymbol{\tilde{Z}}|_F^2\\
			\geq&\frac{1}{2\lambda_x}(\hat{\lambda}_k-\hat{\lambda}_{k+1})\|\boldsymbol{\tilde{Z}}\|_F^2\geq\frac{1}{4\lambda_x}(\lambda_k-\lambda_{k+1})\|\boldsymbol{\tilde{Z}}\|_F^2.
		\end{align*}
		Thus, the convexity parameter $\alpha$ is
		\begin{equation*}
			\alpha=\frac{\lambda_k-\lambda_{k+1}}{4\lambda_x}=\frac{\lambda_k}{4\lambda_x}.
		\end{equation*}
		In summary, we needs the following conditions
		\begin{equation*}
			\|\boldsymbol{L}-\boldsymbol{\hat{A}}(\mathcal{S}_t)\|_F\leq\frac{\hat{\lambda}_k-\hat{\lambda}_{k+1}}{2\lambda_x c_1},
		\end{equation*}
		\begin{equation*}
			\|\boldsymbol{\hat{\Sigma}}^{1/2}_t\boldsymbol{L}-\boldsymbol{\hat{\Sigma}}_t^{1/2}\boldsymbol{\hat{A}}(\mathcal{S}_t)\|_F\leq\eta,
		\end{equation*}
		\begin{equation*}
			\|\boldsymbol{\hat{\Sigma}}^{1/2}_t\boldsymbol{L}-\boldsymbol{\hat{\Sigma}}_t^{1/2}\boldsymbol{\hat{A}}(\mathcal{S}_t)\|_F\leq\frac{1}{4}\|\boldsymbol{\hat{\Sigma}}^{1/2}\boldsymbol{L}^{*}\|_{op}=\frac{1}{4}\sqrt{1+\frac{\hat{\lambda}_1}{\lambda_{penalty}}}.
		\end{equation*}
		The above conditions lead to the choice
		\begin{equation}
			\xi=\min\bigg\{\frac{\lambda_k}{4\lambda_x c_1},\frac{\eta}{\sqrt{\lambda_x}},\frac{1}{4\sqrt{\lambda_x}}\sqrt{1+\frac{2\lambda_1}{\lambda_{penalty}}}\bigg\}.
		\end{equation}
		
		Then by Lemma \ref{lem:general_gradient_descent}, 
		\begin{equation*}
			\text{dist}(\boldsymbol{L}_{t+1},\boldsymbol{L}_*)^2\leq(1-\frac{\alpha}{\beta})\text{dist}(\boldsymbol{L}_{t},\boldsymbol{L}_*)^2,
		\end{equation*}
		where $\alpha=\frac{\lambda_k}{4\lambda_x}$ and $\beta=\lambda_{penalty}(1+2\lambda_x)\big(\eta+\sqrt{1+2\lambda_1/\lambda_{penalty}}\big)^2$, when the step size is smaller than $1/\beta$. 
		
		Then, by taking $\eta=1$ and $\lambda_{penalty}=\lambda_1/c$, we have
		\begin{equation*}
			c_1=\lambda_1\lambda_x^{3/2}\{2+4\sqrt{1+2c}+9/4(1+\sqrt{1+2c})\}/c,
		\end{equation*}
		\begin{equation*}
			\beta=\lambda_1(1+2\lambda_x)(1+\sqrt{1+2c})^2/c,
		\end{equation*}
		\begin{equation*}
			\xi=\frac{1}{\sqrt{\lambda_x}}\min\bigg\{\frac{\lambda_k}{\lambda_1\lambda_x^2c_2},1,\frac{1}{4}\sqrt{1+2c}\bigg\},
		\end{equation*}
		where $c_2=4\{2+4\sqrt{1+2c}+9/4(1+\sqrt{1+2c})\}/c.$
		\hfill $\square$
	\end{proof}
	
	\begin{proof}[Proof of Lemma \ref{lem:peeling_effect}]
		
		Before the proof of our main results, we use the following notations for support of different estimators,
		\begin{equation*}
			\text{supp}(\boldsymbol{A}_0)=\mathcal{S}\text{ and supp}(\boldsymbol{\hat{B}}^{(t+1)})=\mathcal{S}_{t+1}.
		\end{equation*}
		There are three types of sets in terms of overlapping, and then we define $\mathcal{F}_1:=\mathcal{S}\backslash\mathcal{S}_{t+1}$, $\mathcal{F}_2:=\mathcal{S}\cap\mathcal{S}_{t+1}$, $\mathcal{F}_3:=\mathcal{S}_{t+1}\backslash\mathcal{S}$, with size $k_1,k_2,k_3$, respectively. For simplicity, we use the following notations,
		\begin{equation*}
			x_1:=\|\boldsymbol{A}_{0,\mathcal{F}_1}\|_F,x_2:=\|\boldsymbol{A}_{0,\mathcal{F}_1}\|_F,
		\end{equation*}
		\begin{equation*}
			y_1:=\|\boldsymbol{\hat{B}}_{\mathcal{F}_1}^{(t+0.3)}\|_F,y_2:=\|\boldsymbol{\hat{B}}_{\mathcal{F}_2}^{(t+0.3)}\|_F, y_3:=\|\boldsymbol{\hat{B}}_{\mathcal{F}_3}^{(t+0.3)}\|_F,
		\end{equation*}
		where the subscript $\mathcal{F}_i$ indicates the subvector for $i=1,\dots,3$. Furthermore, we define the following variables:
		\begin{equation*}
			x_1^2+x_2^2=\|\boldsymbol{A}_0\|_F^2=X^2,
		\end{equation*}
		\begin{equation*}
			y_1^2+y_2^2+y_3^2=\|\boldsymbol{\hat{B}}^{(t+0.3)}_{\mathcal{S}\cup\mathcal{S}_{t+1}}\|_F^2\leq\|\boldsymbol{\hat{B}}^{(t+0.3)}\|_F^2=Y^2.
		\end{equation*}
		Notice that
		\begin{align*}
			\|\boldsymbol{\hat{B}}^{(t+0.6)}\|^2_F&=\|\boldsymbol{\hat{B}}^{(t+0.3)}_{\mathcal{P}_{t+1}}+\boldsymbol{\tilde{w}}_{t+1}\|^2_F\leq\|\boldsymbol{\hat{B}}^{(t+0.3)}_{\mathcal{P}_{t+1}}\|_F^2+\|\boldsymbol{\tilde{w}}_{t+1}\|^2_F\leq y_2^2+y_3^2+Cks^{\prime}\sigma_2^2\log(p)^2,
		\end{align*}
		with probability at least $1-\exp\{-Cs^{\prime}k\log(p)\}$ by Lemma \ref{lem:con}, where $\boldsymbol{\tilde{w}}_{t+1}$ is the noise added at the data release step defined in Line 6 of \ref{alg:vector_noisy_ht} and $\sigma_2=2\sigma\sqrt{2ks^{\prime}\log(2.5\cdot T/\delta)}/(\varepsilon/T)$ by Algorithm \ref{alg:hd_dp_sir}. Then, we consider the property of the distance $\text{dist}(\boldsymbol{U},\boldsymbol{V})$. Notice that
		\begin{align*}
			\text{dist}(\boldsymbol{U},\boldsymbol{V})&=\min_{\boldsymbol{P}\in\mathbb{O}^{d\times d}}\|\boldsymbol{U}\boldsymbol{P}-\boldsymbol{V}\|_F\\
			&=\min_{\boldsymbol{P}\in\mathbb{O}^{d\times d}}\|\boldsymbol{U}\boldsymbol{P}\|_F+\|\boldsymbol{V}\|-2\langle\boldsymbol{U}\boldsymbol{P},\boldsymbol{V}\rangle\\
			&=\min_{\boldsymbol{P}\in\mathbb{O}^{d\times d}}\|\boldsymbol{U}\|_F+\|\boldsymbol{V}\|-2\text{Tr}(\boldsymbol{P}^{\top}\boldsymbol{U}^{\top}\boldsymbol{V}).
		\end{align*}
		Assume that $\boldsymbol{A}\boldsymbol{D}\B^{\top}$ is a singular value decomposition of $\boldsymbol{U}^{\top}\boldsymbol{V}$, then
		\begin{equation*}
			\text{Tr}(\boldsymbol{P}^{\top}\boldsymbol{U}^{\top}\boldsymbol{V})=\text{Tr}(\boldsymbol{P}^{\top}\boldsymbol{A}\boldsymbol{D}\B^{\top})=\text{Tr}(\B^{\top}\boldsymbol{P}^{\top}\boldsymbol{A}\boldsymbol{D})=\text{Tr}(\boldsymbol{Z}\boldsymbol{D}),
		\end{equation*}
		where $\boldsymbol{Z}=\B^{\top}\boldsymbol{P}^{\top}\boldsymbol{A}$ is an orthogonal matrix. By the fact that $\boldsymbol{D}$ is the diagonal matrix, the trace $\text{Tr}(\boldsymbol{Z}\boldsymbol{D})=\sum_i Z_{ii}D_{ii}$ is maximized at $Z_{ii}=1$ for all $i$. Thus, we have
		\begin{align*}
			\text{dist}(\boldsymbol{U},\boldsymbol{V})=\|\boldsymbol{U}\|_F+\|\boldsymbol{V}\|_F-2\text{Tr}(\boldsymbol{D})&=\|\boldsymbol{U}\|_F+\|\boldsymbol{V}\|_F-2\text{Tr}(\sqrt{\boldsymbol{D}^{\top}\boldsymbol{D}})\\
			&=\|\boldsymbol{U}\|_F+\|\boldsymbol{V}\|_F-2\text{Tr}(|\boldsymbol{U}^{\top}\boldsymbol{V}|).
		\end{align*}
		To simplify the notation, we define $\text{Tr}(|\boldsymbol{W}|):=\text{Tr}(\sqrt{\boldsymbol{W}^{\top}\boldsymbol{W}})$. Then, it remains to consider the term $\text{Tr}(|\boldsymbol{A}_0^{\top}\boldsymbol{\hat{B}}^{(t+0.3)}|)-\text{Tr}(|\boldsymbol{A}^{\top}_0\boldsymbol{\hat{B}}^{(t+0.6)}|)$. Let $\Delta:=\text{Tr}(|\boldsymbol{A}_0^{\top}\boldsymbol{\hat{A}}^{(t+0.3)}|)$. Notice that $\mathcal{F}_1$ is part of $\mathcal{S}$ which is not in $\mathcal{S}_{t+1}$. 
		
		By Lemma 4.2 in \cite{dwork2021differentially}, we have
		\begin{equation*}
			\P\bigg(\max_{1\leq i\leq s^{\prime},1\leq j\leq p}|w_{i,j}|\leq 2\sigma_1\log(ps_1)\bigg)\geq 1-\exp\{-\log(s^{\prime}p)\}
		\end{equation*}
		where $w_{i,j}$ is defined in Line 3 of \ref{alg:vector_noisy_ht} and $\sigma_1=2\sigma\sqrt{3ks^{\prime}\log(2\cdot T/\delta)}/(\varepsilon/T)$ by Algorithm \ref{alg:hd_dp_sir}. Since the peeling mechanism is the noisy hard thresholding, we have the following relation by the selection criterion,
		\begin{equation}
			\frac{y_1^2}{k_1}\leq 2\frac{y_3^2}{k_3}+2C\sigma_1^2\log(p)^2,
		\end{equation}
		for a constant $C$ with probability at least $1-\exp\{-\log(p)\}$, where we use the inequality $(a+b)^2\leq 2a^2+2b^2$.
		
		We know $k_1\leq k_3$ by $k_1+k_2=s\leq s^{\prime}=k_2+k_3$. Then
		\begin{align*}
			\Delta^2&\leq(\|\boldsymbol{A}_{0,\mathcal{F}_1}\|_F\|\boldsymbol{\hat{B}}^{(t+0.3)}_{\mathcal{F}_1}\|_F+\|\boldsymbol{A}_{0,\mathcal{F}_2}\|_F\|\boldsymbol{\hat{B}}^{(t+0.3)}_{\mathcal{F}_2}\|_F)^2\\
			&=(x_1y_1+x_2y_2)^2\leq(y_1^2+y_2^2)X^2\leq(Y^2-y_3^2)X^2\\
			&\leq X^2Y^2-\frac{k_3}{2k_1}X^2y_1^2+2Ck_3\sigma_1^2\log^2(p),
		\end{align*}
		where we use Holder's inequality. Thus,
		\begin{align*}
			y_1^2&\leq\frac{2k_1}{k_3X^2}(X^2Y^2-\Delta^2)+\frac{k_1}{k_3X^2}Ck_3\sigma_1^2\log^2(p)\\
			&\leq 2\frac{k_1+k_2}{k_3+k_2}(Y^2-\Delta^2/X^2)+C\frac{k_1}{X^2}\sigma_1^2\log^2(p)\\
			&=\frac{2s}{s^{\prime}}(Y^2-\Delta^2/X^2)+C\frac{k_1}{X^2}\sigma_1^2\log^2(p),
		\end{align*}
		where we use the fact $k_1\leq k_3$. Furthermore, by definition,
		\begin{equation*}
			x_1y_1+\sqrt{Y^2-y_1^2}\sqrt{X^2-x_1^2}\geq x_1y_1+x_2y_2\geq\Delta.
		\end{equation*}
		The left-hand side of the equation is the sum of a linear function and a left-bottom circle, thus being convex. Thus, by solving the quadratic inequality, we have
		\begin{equation*}
			x_1\leq\frac{\Delta y_1+\sqrt{(X^2Y^2-\Delta^2)(Y^2-y_1^2)}}{Y^2}.
		\end{equation*}
		Furthermore, by $x_1\leq X$ and $\Delta\leq XY$,
		\begin{align*}
			x_1&\leq\min\bigg\{X,\frac{Xy_1+\sqrt{X^2Y^2-\Delta^2}}{Y}\bigg\}\\
			&\leq\min\bigg\{X,\frac{\sqrt{\frac{2s}{s^{\prime}}(Y^2X^2-\Delta^2)+Ck_1\sigma_1^2\log^2(p)}+\sqrt{X^2Y^2-\Delta^2}}{Y}\bigg\}.
		\end{align*}
		Thus,
		\begin{align*}
			x_1y_1&\leq\min\bigg\{\sqrt{\frac{2s}{s^{\prime}}(Y^2X^2-\Delta^2)+Ck_1\sigma_1^2\log^2(p)},\\
			&,\frac{\frac{2s}{s^{\prime}}+\sqrt{\frac{2s}{s^{\prime}}}}{XY}\bigg((X^2Y^2-\Delta^2)+Ck_1\sigma_1^2\log^2(p)\bigg)\bigg\},
		\end{align*}
		where the right hand side is denoted by $\Delta_{max}$. Then, combining all the previous results, by the relation,
		\begin{equation*}
			\text{Tr}(|\boldsymbol{A}_0^{\top}\boldsymbol{\hat{B}}^{(t+0.3)}|)-\text{Tr}(|\boldsymbol{A}_0^{\top}\boldsymbol{\hat{B}}^{(t+0.3)}_{\mathcal{P}_{t+1}}|)\leq\text{Tr}\{|\boldsymbol{A}_0^{\top}(\boldsymbol{\hat{B}}^{(t+0.3)}_{\mathcal{P}_{t+1}}-\boldsymbol{\hat{B}}^{(t+0.0.3})|\}\leq x_1y_1\leq\Delta_{max},
		\end{equation*}
		we have
		\begin{align*}
			\text{dist}^2(\boldsymbol{\hat{B}}^{(t+0.3)}_{\mathcal{P}_{t+1}},\boldsymbol{A}_0)&=\|\boldsymbol{A}_0\|^2_F+\|\boldsymbol{\hat{B}}^{(t+0.3)}_{\mathcal{P}_{t+1}}\|^2_F-2\text{Tr}(|\boldsymbol{A}^{\top}_0\boldsymbol{\hat{B}}^{(t+0.3)}_{\mathcal{P}_{t+1}}|)\\
			&\leq \|\boldsymbol{A}_0\|^2_F+\|\boldsymbol{\hat{B}}^{(t+0.3)}\|^2_F-2\text{Tr}(|\boldsymbol{A}^{\top}_0\boldsymbol{\hat{B}}^{(t+0.3)}|)+\Delta_{max}\\
			&\leq\text{dist}^2(\boldsymbol{\hat{B}}^{(t+0.3)},\boldsymbol{A}_0)+2\sqrt{\frac{2s}{s^\prime}}\frac{1+\sqrt{\frac{2s}{s^{\prime}}}}{XY}\bigg(X^2Y^2-\Delta^2+Ck_1\sigma_1^2\log^2(p)\bigg).
		\end{align*}
		Notice that,
		\begin{align*}
			X^2Y^2-\Delta^2&=(XY+\Delta)(XY-\Delta)\leq 2XY(XY-\Delta)\\
			&\leq XY(X^2+Y^2-2\Delta)=XY(\|\boldsymbol{A}_0\|^2_F+\|\boldsymbol{\hat{B}}^{(t+0.3)}\|^2_F-2\text{Tr}(|\boldsymbol{A}_0^{\top}\boldsymbol{\hat{B}}^{(t+0.3)}|))\\
			&=XY\text{dist}^2(\boldsymbol{\hat{B}}^{(t+0.3)},\boldsymbol{A}_0).
		\end{align*}
		Thus, we have
		\begin{equation*}
			\text{dist}^2(\boldsymbol{\hat{B}}^{(t+0.3)}_{\mathcal{P}_{t+1}},\boldsymbol{A}_0)\leq\bigg\{1+2\sqrt{\frac{2s}{s^\prime}}\bigg(1+\sqrt{\frac{2s}{s^\prime}}\bigg)\bigg\}\text{dist}^2(\boldsymbol{\hat{B}}_{(t+0.3)},\boldsymbol{A}_0)+\epsilon_1^2,
		\end{equation*}
		where
		\begin{equation*}
			\epsilon^2_1=2\sqrt{\frac{s}{2s^\prime}}\frac{1+\sqrt{\frac{2s}{s^{\prime}}}}{XY}Cs\sigma_1^2\log^2(p).
		\end{equation*}
		Notice that by definition
		\begin{equation*}
			X=\|\boldsymbol{A}_0\|_F\geq\|\B_0\|_F\sqrt{1+\frac{\lambda_k}{2\lambda_{penalty}}},
		\end{equation*}
		and
		\begin{equation*}
			Y=\|\boldsymbol{\hat{B}}^{(t+0.3)}\|_F\geq\|\boldsymbol{\hat{B}}^{(t+0.3)}_{\mathcal{S}_t}\|_F\geq C\|\B_0\|_F\sqrt{1+\frac{\lambda_k}{2\lambda_{penalty}}},
		\end{equation*}
		for some constant $C$ under the conditions of Lemma \ref{lem:grad_desc}. We can write $\epsilon_1^2=C\sqrt{\frac{2s}{s^{\prime}}}(1+\sqrt{\frac{2s}{s^{\prime}}})s\sigma_1^2\log^2(p)$. Then, we have
		\begin{equation*}
			\text{dist}(\boldsymbol{\hat{B}}^{(t+0.3)}_{\mathcal{P}_{t+1}},\boldsymbol{A}_0)\leq\bigg\{1+2\sqrt{\frac{2s}{s^\prime}}\bigg(1+\sqrt{\frac{2s}{s^\prime}}\bigg)\bigg\}^{1/2}\text{dist}(\boldsymbol{\hat{B}}_{(t+0.3)},\boldsymbol{A}_0)+\epsilon_1,
		\end{equation*}
		where we use the inequality $\sqrt{a^2+b^2}\leq a+b$ for $a,b\geq 0$. Then, the effect due to additional noise is
		\begin{equation}
			\text{dist}(\boldsymbol{\hat{B}}^{(t+0.3)}_{\mathcal{P}_{t+1}},\boldsymbol{\hat{B}}_{(t+0.6)})\leq\|\boldsymbol{\tilde{w}}\|_F\leq \sqrt{C\sigma_2^2ks^{\prime}\log^2(p)}.
		\end{equation}
		The final result is followed by the triangle inequality.
	\end{proof}

	\subsection{Proof of the Initial Estimator}
	
	We first show that the initial estimator of Algorithm \ref{alg:hd_ini} is $(2\varepsilon,\delta)$-DP.
	
	\begin{proof}[Proof of Lemma \ref{lem:hd_ini}]
		
		By Lemma \ref{lem:est_slices}, estimation intervals $\{\widehat{I}_h\}_{h=1}^{H}$ by DP-histogram is $(\varepsilon,0)$-DP. By the proof of Lemma \ref{lem:low_privacy_ini}, the $\ell_{\infty}$ sensitivity of $\boldsymbol{\hat{M}}$ is bounded by $7c_x^2/n$. By Theorem 3 in \cite{dwork2021differentially}, returning the index set $\widehat{\mathcal{P}}$ is $(\varepsilon/2,\delta/2)$-DP. By the composition theorem and Lemma \ref{lem:low_privacy_ini}, we conclude that the output of Algorithm \ref{alg:hd_ini} is $(2\varepsilon,\delta)$-DP.
	\end{proof}
	
	We then show the initial estimator is consistent. 
	
	\begin{proof}[Proof of Theorem \ref{thm:hd_ini}]
		
		Under Assumption \ref{assum:sparse_cov}, the support set of the kernel matrix $\M$ is denoted by $\mathcal{P}$. For $j\in\mathcal{P}$, there exists $\boldsymbol{v}\in\text{span}(\M)$ such that the $j$th component of $\boldsymbol{v}$ is nonzero. Due to the fact that $\boldsymbol{\Sigma}\cdot\text{span}(\B)=\text{span}(\M)$ by the conditions of sliced inverse regression, there exists $\boldsymbol{\beta}\in\text{span}(\B)$ such that $\boldsymbol{v}=\boldsymbol{\Sigma}\boldsymbol{\beta}$, which implies $j\in\text{supp}(\boldsymbol{\Sigma}\boldsymbol{\beta})$. By the sparsity of $\B$ and $\boldsymbol{\Sigma}$, we have $|\mathcal{P}|\leq s\max_{1\leq i\leq p}|\text{supp}(\boldsymbol{\Sigma})_{*,i}|=O(s)$. Thus, the support set $\mathcal{P}$ is sparse. For $Ls\leq s^{\prime}\leq Cs$, by Lemma 4.2 in \cite{dwork2021differentially}, we have
		\begin{equation*}
			\P\bigg(\max_{1\leq i\leq s_1,1\leq j\leq p}|w_{i,j}|\leq\frac{7c_x^2}{n}\frac{\sqrt{3s_1\log(2/\delta)}}{\varepsilon/2}\log(ps^{\prime}/\alpha)\bigg)\geq 1-\alpha,
		\end{equation*}
		for $0<\alpha<1$. By choosing $\alpha=1/p$ and using the fact $s^{\prime}\leq p$, we have
		\begin{equation*}
			\max_{1\leq i\leq s_1,1\leq j\leq p}|w_{i,j}|\leq\frac{21c_x^2}{n}\frac{\sqrt{3s^{\prime}\log(2/\delta)}}{\varepsilon/2}\log(p),
		\end{equation*}
		with probability at least $1-p^{-1}$. Furthermore, by triangle inequality, we have
		\begin{equation*}
			\begin{split}
				\widehat{M}_{j,j}+w_{i,j}&\geq M_{j,j}+w_{i,j}-|\widehat{M}_{j,j}-M_{j,j}|\geq M_{j,j}-\max_{i,j}|w_{i,j}|-|\widehat{M}_{j,j}-M_{j,j}|\\
				&\geq c-\sqrt{\frac{C\log(p)}{n}}-\frac{21c_x^2}{n}\frac{\sqrt{3s_1\log(2/\delta)}}{\varepsilon/2}\log(p),
			\end{split}
		\end{equation*}
		for $j\in\mathcal{P}$, where we use Lemma \ref{lemma:estS} in the last inequality. Under the conditions in Theorem \ref{thm:hd_ini}, we have $	\widehat{M}_{j,j}+w_{i,j}\geq 2/3\cdot c$. Similarly, for $j\notin\mathcal{P}$, by triangle inequality,
		\begin{equation*}
			\begin{split}
				\widehat{M}_{j,j}+w_{i,j}&\leq M_{j,j}+w_{i,j}+|\widehat{M}_{j,j}-M_{j,j}|\leq M_{j,j}+\max_{i,j}|w_{i,j}|-|\widehat{M}_{j,j}-M_{j,j}|\\
				&\leq 0+\sqrt{\frac{C\log(p)}{n}}+\frac{21c_x^2}{n}\frac{\sqrt{3s_1\log(2/\delta)}}{\varepsilon/2}\log(p)\leq 1/3\cdot c.
			\end{split}
		\end{equation*}
		Thus, we have
		\begin{equation*}
			\widehat{M}_{j,j}+w_{i,j}\geq\widehat{M}_{j^{\prime},j^{\prime}}+w_{i,j^{\prime}}
		\end{equation*}
		for $j\in\mathcal{P}$ and $j^{\prime}\notin\mathcal{P}$. Then, we have $\mathcal{P}\subset\widehat{\mathcal{P}}$ with probability at least $1-\exp(-C\log(p))$. Then, by applying Lemma \ref{thm:dpsir_errorbound_ini}, we have
		\begin{equation*}
			L(\boldsymbol{\hat{B}}^{(0)},\B)\to 0\text{ and }|\tilde{\lambda}_j-\lambda_j|\to 0\text{ for }j=1,\dots,H,
		\end{equation*}
		with probability converging to one as $n,p\to\infty$ .
		
	\end{proof}
	
	\section{Technical Lemmas}\label{sec: lemmas}
	
	\begin{lemm}[Concentration of the norm]
		\label{lem:con}
		Let $\boldsymbol{x}=(X_1,\dots,X_n)\in\mR^n$ be a random vector with independent sub-Gaussian coordinates $X_i$ that satisfy $\mathbb{E}(X_i^2)=1$ for $i=1,\dots,n$. Then
		\begin{equation*}
			\|\boldsymbol{x}\|_{2}\leq 4\sqrt{n}+2C\sqrt{\log\bigg(\frac{1}{\delta}\bigg)},
		\end{equation*}
		with probability at least $1-\delta$, where the constant $C$ only depends on the maximal sub-Gaussian norm $\max_{i}\|X_i\|_{\psi_2}$. 
	\end{lemm}
	\begin{proof}
		This result is Theorem 3.1.1 in \cite{vershynin2018high}. 
		\hfill $\square$
	\end{proof}
	
	\begin{lemm}
		\label{lem:random_matrix_eigen}
		Let $\boldsymbol{A}$ be an $n\times k$ random matrix whose columns $\boldsymbol{a}_i$ are independent isotropic random vectors in $\mR^k$ with $\|\boldsymbol{a}_i\|_2=\sqrt{k}$ almost surely. For any $t>0$, we have
		\begin{equation*}
			\mathbb{P}\bigg\{\big\|\frac{1}{n}\boldsymbol{A}^{\top}\boldsymbol{A}-\boldsymbol{I}_k\big\|_{op}\leq C\sqrt{\frac{k}{n}}+t\bigg\}\geq 1-2e^{-nt^2/2},
		\end{equation*}
		with probability at least $1-2\exp(-cnt^2)$, where the constant $C$ and $c$ only depend on the maximal sub-Gaussian norm $\max_{i}\|\boldsymbol{a}_i\|_{\psi_2}$. 
	\end{lemm}
	\begin{proof}
		Lemma \ref{lem:random_matrix_eigen} generalizes Proposition D.1. in \cite{ma2013sparse} from i.i.d normal entries to independent sub-Gaussian columns. Let $\boldsymbol{W}_{nk}=\boldsymbol{A}^{\top}\boldsymbol{A}/n$ and $\delta_k(t)=C\sqrt{k/n}+t$. By the fact that $\lambda_i(\boldsymbol{W}_{nk})=n^{-1/2}\lambda_i(\boldsymbol{A})$ for $i=1,\dots,\max(n,k)$ where $\lambda_i(\cdot)$ is the $i$th eigenvalue for a symmetric matrix and $i$th singular-value for an asymmetric matrix, respectively. Thus, $\{\lambda_1(\boldsymbol{W}_{nk})>1+\delta_k(t)\}=\{\lambda_1(\boldsymbol{A})>\sqrt{n}+C\sqrt{k}+\sqrt{n}t\}$ and $\{\lambda_k(\boldsymbol{W}_{nk})<1-\delta_k(t)\}=\{\lambda_k(\boldsymbol{A})<\sqrt{n}-C\sqrt{k}-\sqrt{n}t\}$. By the fact that $\|\boldsymbol{W}_{nk}-\boldsymbol{I}_k\|_{op}=\max\{\lambda_1(\boldsymbol{W}_{nk})-1,1-\lambda_k(\boldsymbol{W}_{nk})\}$, we obtain
		\begin{equation*}
			\begin{split}
				\mathbb{P}\{\|\boldsymbol{W}_{nk}-\boldsymbol{I}_p\|_{op}\geq\delta_k(t)\}\leq&\mathbb{P}\{\lambda_1(\boldsymbol{W}_{nk})>1+\delta_k(t)\}+\mathbb{P}\{\lambda_k(\boldsymbol{W}_{nk})<1-\delta_k(t)\}\\
				\leq&\mathbb{P}\{\lambda_1(\boldsymbol{A})>\sqrt{n}\\
				&+C\sqrt{k}+\sqrt{n}t\}+\mathbb{P}\{\lambda_k(\boldsymbol{A})>\sqrt{n}-C\sqrt{k}-\sqrt{n}t\}.
			\end{split}
		\end{equation*}
		By Theorem 5.58 in \cite{vershynin2010introduction},
		\begin{equation*}
			\sqrt{n}-C\sqrt{k}-\sqrt{n}t\leq\lambda_k(\boldsymbol{A})\leq\lambda_1(\boldsymbol{A})\leq\sqrt{n}+C\sqrt{k}+\sqrt{n}t,
		\end{equation*}
		with probability at least $1-2\exp(-cnt^2)$, where the constant $C$ and $c$ only depend on the maximal sub-Gaussian norm $\max_{i}\|\boldsymbol{a}_i\|_{\psi_2}$. Then, the conclusion of lemma \ref{lem:random_matrix_eigen} holds. A similar conclusion is discussed by Remark 5.59 by \cite{vershynin2010introduction}.
		\hfill $\square$
	\end{proof}
	
	We then provide two important lemmas on DP. To ensure privacy, it is necessary to add random noises to the algorithms. Next, we introduce two well-known techniques: the Laplace mechanism that achieves $(\varepsilon,0)$-DP, and the Gaussian mechanism that achieves $(\varepsilon,\delta)$-DP.
	
	\begin{lemm}[\cite{dwork2014algorithmic}]
		\label{lemma:laplace_gaussian_mechanism}
		$\ $
		\begin{enumerate}
			\item (Laplace mechanism): For a vector-valued deterministic algorithm $\mathcal{T}(\cdot)$ with $l_1$ sensitivity $\Delta_1(\mathcal{T})$, the randomized algorithm $\mathcal{M}(\cdot):=\mathcal{T}(\cdot)+\boldsymbol{\xi}$ achieves $(\varepsilon,0)$-DP, where $\boldsymbol{\xi}=(\xi_1,\dots,\xi_m)^{\top}$ follows i.i.d. Laplace distribution with scale parameter $\Delta_1(\mathcal{T})/\varepsilon$.
			\item (Gaussian mechanism): For a vector-valued deterministic algorithm $\mathcal{T}(\cdot)$ with $l_2$ sensitivity $\Delta_2(\mathcal{T})$, the randomized algorithm $\mathcal{M}(\cdot):=\mathcal{T}(\cdot)+\boldsymbol{\xi}$ achieves $(\varepsilon,\delta)$-DP, where $\boldsymbol{\xi}=(\xi_1,\dots,\xi_m)^{\top}$ follows i.i.d. Gaussian distribution with mean $0$ and standard deviation $\sqrt{2\log(1.25/\delta)}\Delta_2(\mathcal{T})/\varepsilon$.
		\end{enumerate}
	\end{lemm}
	
	Lemma \ref{lemma:laplace_gaussian_mechanism} demonstrates the feasibility of converting deterministic algorithms into $(\varepsilon,\delta)$-DP algorithms. However, accurately calculating the sensitivity of an algorithm can be a highly intricate and challenging task. To overcome this difficulty, the post-processing and composition properties of DP can be employed, enabling the construction of complex algorithms from simpler ones in a convenient manner. The subsequent lemmas offer valuable insights and guidance in this aspect.
	
	\begin{lemm}
		\label{lem:post_combin_dp}
		DP algorithms have the following properties \citep{dwork2006calibrating, dwork2010boosting}:
		\begin{enumerate}
			\item Post-processing: Let $\mathcal{M}(\cdot)$ be an $(\varepsilon,\delta)$-DP algorithm and $f(\cdot)$ be a deterministic function that maps $\mathcal{M}(\cdot)$ to real Euclidean space, then $f\{\mathcal{M}(\cdot)\}$ is also an $(\varepsilon,\delta)$-DP algorithm.
			\item Composition: Let $\mathcal{M}_1(\cdot)$ and $\mathcal{M}_2(\cdot)$ be $(\varepsilon_1,\delta_1)$ and $(\varepsilon_2,\delta_2)$ DP, then the composition $\mathcal{M}_1\circ\mathcal{M}_2$ is $(\varepsilon_1+\varepsilon_2,\delta_1+\delta_2)$-DP.
			\item Advanced Composition: Let $\mathcal{M}(\cdot)$ be $(\varepsilon,0)$-DP and $0<\delta^{\prime}<1$, then $k$-fold adaptive composition of $\mathcal{M}(\cdot)$ is $(\varepsilon^{\prime},\delta^{\prime})$-DP for $\varepsilon^{\prime}=k\varepsilon(e^{\varepsilon}-1)+\varepsilon\sqrt{2k\log(1/\delta^{\prime})}$.
		\end{enumerate}
	\end{lemm}

	\section{The Algorithms of the DP Rayleigh Flow}\label{sec: hu2023privacy}
	
	For the readers' convenience, we provide the algorithms proposed by \cite{hu2023privacy} in this section. 
	\begin{algorithm}
		\caption{Differential Private Rayleigh Flow}
		\begin{algorithmic}[1]
			\Require $\{\boldsymbol{x}_{i},\tilde{Y}_{i}\}_{i=1,\dots,n}$, noise scale $\sigma^2_1$ and $\sigma^2_2$, initial vector $\boldsymbol{v}_0\in\mR^p$ with $\|\boldsymbol{v}_0\|=1$, step size $\eta$, iteration numbers $T$.
			\State Estimate $\boldsymbol{\hat{\Sigma}}$ and $\boldsymbol{\hat{M}}$ using data $\{\boldsymbol{x}_{i},\widetilde{Y}_{i}\}_{i=1,\dots,n}$;
			\For{$t=1,\dots,T$}
			\State Let $\boldsymbol{E}_1^{(t)}\in\mR^{p\times p}$ and $\boldsymbol{E}_2^{(t)}\in\mR^{p\times p}$ be two symmetric matrices where the upper triangle (including the diagonal) elements are i.i.d. samples generated from $N(0,\sigma^2_1)$ and $N(0,\sigma^2_2)$, respectively;
			\State Let $\boldsymbol{\widetilde{\Sigma}}^{(t)}:=\boldsymbol{\hat{\Sigma}}+\boldsymbol{E}_1^{(t)}$ and $\boldsymbol{\widetilde{M}}^{(t)}:=\boldsymbol{\hat{M}}+\boldsymbol{E}_2^{(t)}$;
			\State Calculate: $\rho_t=(\boldsymbol{v}_{t-1}^{\top}\boldsymbol{\widetilde{M}}^{(t)}\boldsymbol{v}_{t-1})/(\boldsymbol{v}_{t-1}^{\top}\boldsymbol{\widetilde{\Sigma}}^{(t)}\boldsymbol{v}_{t-1})$ and 
			$$\boldsymbol{C}^{(t)}=\boldsymbol{I}_p+\frac{\eta}{\rho_t}(\boldsymbol{\widetilde{M}}^{(t)}-\rho_t\boldsymbol{\widetilde{\Sigma}}^{(t)});$$
			\State Update $\boldsymbol{v}_t=\boldsymbol{C}^{(t)}\boldsymbol{v}_{t-1}/\|\boldsymbol{C}^{(t)}\boldsymbol{v}_{t-1}\|_2$;
			\EndFor
			\Ensure $\boldsymbol{v}_T$.
		\end{algorithmic}
	\end{algorithm}
	\begin{algorithm}
		\caption{Differential Private Truncated Rayleigh Flow}
		\begin{algorithmic}[1]
			\Require $\{\boldsymbol{x}_{i},\widetilde{Y}_{i}\}_{i=1,\dots,n}$, sparsity $s$, noise scale $\sigma^2_1$ and $\sigma^2_2$, initial vector $\boldsymbol{v}_0\in\mR^p$ with $\|\boldsymbol{v}_0\|=1$, step size $\eta$, iteration numbers $T$.
			\State Estimate $\boldsymbol{\hat{\Sigma}}$ and $\boldsymbol{\hat{M}}$ using data $\{\boldsymbol{x}_{i},\widetilde{Y}_{i}\}_{i=1,\dots,n}$;
			\For{$t=1,\dots,T$}
			\State Let $\boldsymbol{E}_1^{(t)}\in\mR^{p\times p}$ and $\boldsymbol{E}_2^{(t)}\in\mR^{p\times p}$ be two symmetric matrices where the upper triangle (including the diagonal) elements are i.i.d. samples generated from $N(0,\sigma^2_1)$ and $N(0,\sigma^2_2)$, respectively;
			\State Let $\boldsymbol{\widetilde{\Sigma}}^{(t)}:=\boldsymbol{\hat{\Sigma}}+\boldsymbol{E}_1^{(t)}$ and $\boldsymbol{\widetilde{M}}^{(t)}:=\boldsymbol{\hat{M}}+\boldsymbol{E}_2^{(t)}$;
			\State Calculate: $\rho_t=(\boldsymbol{v}_{t-1}^{\top}\boldsymbol{\widetilde{M}}^{(t)}\boldsymbol{v}_{t-1})/(\boldsymbol{v}_{t-1}^{\top}\boldsymbol{\widetilde{\Sigma}}^{(t)}\boldsymbol{v}_{t-1})$ and 
			$$\boldsymbol{C}^{(t)}=\boldsymbol{I}_p+\frac{\eta}{\rho_t}(\boldsymbol{\widetilde{M}}^{(t)}-\rho_t\boldsymbol{\widetilde{\Sigma}}^{(t)});$$
			\State Update $\boldsymbol{v}_t^{\prime}=\boldsymbol{C}^{(t)}\boldsymbol{v}_{t-1}/\|\boldsymbol{C}^{(t)}\boldsymbol{v}_{t-1}\|_2$;
			\State Let $F^{(t)}$ be the set of indices of $\boldsymbol{v}_{t}^{\prime}$ with the largest $s$ absolute values.
			\State Let $\boldsymbol{\hat{v}}_t$ be the truncated vector of $\boldsymbol{v}_t^{\prime}$ by setting the $i$th coordinate $(\boldsymbol{v}_t^{\prime})_i=0$ if $i\notin F^{(t)}$.
			\State Update $\boldsymbol{v}_t=\boldsymbol{\hat{v}}_{t}/\|\boldsymbol{\hat{v}}_{t}\|_2$;
			\EndFor
			\Ensure $\boldsymbol{v}_T$.
		\end{algorithmic}
	\end{algorithm}
	
	\newpage
	
	\bibliographystyle{refstyle.bst}
	\bibliography{ref}

\begin{thebibliography}{51}
\newcommand{\enquote}[1]{``#1''}
\providecommand{\natexlab}[1]{#1}
\expandafter\ifx\csname urlstyle\endcsname\relax
  \providecommand{\doi}[1]{doi:\discretionary{}{}{}#1}\else
  \providecommand{\doi}{doi:\discretionary{}{}{}\begingroup
  \urlstyle{rm}\Url}\fi

\bibitem[{Avella-Medina et~al.(2023)Avella-Medina, Bradshaw, and
  Loh}]{avella2021differentially}
Avella-Medina, M., Bradshaw, C., and Loh, P.L. (2023).
\newblock \enquote{Differentially private inference via noisy optimization.}
\newblock \emph{The Annals of Statistics}, \textbf{51(5)}, 2067--2092.

\bibitem[{Cai et~al.(2021)Cai, Wang, and Zhang}]{cai2021cost}
Cai, T.T., Wang, Y., and Zhang, L. (2021).
\newblock \enquote{The cost of privacy: Optimal rates of convergence for
  parameter estimation with differential privacy.}
\newblock \emph{The Annals of Statistics}, \textbf{49(5)}, 2825--2850.

\bibitem[{Cai et~al.(2023{\natexlab{a}})Cai, Wang, and Zhang}]{cai2023score}
Cai, T.T., Wang, Y., and Zhang, L. (2023{\natexlab{a}}).
\newblock \enquote{Score attack: A lower bound technique for optimal
  differentially private learning.}
\newblock \emph{arXiv preprint arXiv:2303.07152}.

\bibitem[{Cai et~al.(2022{\natexlab{a}})Cai, Lei, and Roeder}]{cai2022model}
Cai, Z., Lei, J., and Roeder, K. (2022{\natexlab{a}}).
\newblock \enquote{Model-free prediction test with application to genomics
  data.}
\newblock \emph{Proceedings of the National Academy of Sciences},
  \textbf{119(34)}, e2205518119.

\bibitem[{Cai et~al.(2024)Cai, Lei, and Roeder}]{cai2024asymptotic}
Cai, Z., Lei, J., and Roeder, K. (2024).
\newblock \enquote{Asymptotic distribution-free independence test for
  high-dimension data.}
\newblock \emph{Journal of the American Statistical Association},
  \textbf{119(547)}, 1794--1804.

\bibitem[{Cai et~al.(2022{\natexlab{b}})Cai, Li, and
  Zhang}]{cai2022distribution}
Cai, Z., Li, R., and Zhang, Y. (2022{\natexlab{b}}).
\newblock \enquote{A distribution free conditional independence test with
  applications to causal discovery.}
\newblock \emph{Journal of Machine Learning Research}, \textbf{23(85)}, 1--41.

\bibitem[{Cai et~al.(2020)Cai, Li, and Zhu}]{cai2020online}
Cai, Z., Li, R., and Zhu, L. (2020).
\newblock \enquote{Online sufficient dimension reduction through sliced inverse
  regression.}
\newblock \emph{Journal of Machine Learning Research}, \textbf{21(10)}, 1--25.

\bibitem[{Cai et~al.(2023{\natexlab{b}})Cai, Li, Xia, and
  Zhang}]{cai2023private}
Cai, Z., Li, S., Xia, X., and Zhang, L. (2023{\natexlab{b}}).
\newblock \enquote{Private estimation and inference in high-dimensional
  regression with {FDR} control.}
\newblock \emph{arXiv preprint arXiv:2310.16260}.

\bibitem[{Cai et~al.(2022{\natexlab{c}})Cai, Xi, Zhu, and Li}]{cai2022causal}
Cai, Z., Xi, D., Zhu, X., and Li, R. (2022{\natexlab{c}}).
\newblock \enquote{Causal discoveries for high dimensional mixed data.}
\newblock \emph{Statistics in Medicine}, \textbf{41(24)}, 4924--4940.

\bibitem[{Chaudhuri et~al.(2012)Chaudhuri, Sarwate, and
  Sinha}]{chaudhuri2012near}
Chaudhuri, K., Sarwate, A., and Sinha, K. (2012).
\newblock \enquote{Near-optimal differentially private principal components.}
\newblock \emph{Advances in neural information processing systems},
  \textbf{25}.

\bibitem[{Chen et~al.(2018)Chen, Fan, and Li}]{chen2018error}
Chen, Z., Fan, J., and Li, R. (2018).
\newblock \enquote{Error variance estimation in ultrahigh-dimensional additive
  models.}
\newblock \emph{Journal of the American Statistical Association},
  \textbf{113(521)}, 315--327.

\bibitem[{Chi et~al.(2019)Chi, Lu, and Chen}]{chi2019nonconvex}
Chi, Y., Lu, Y.M., and Chen, Y. (2019).
\newblock \enquote{Nonconvex optimization meets low-rank matrix factorization:
  An overview.}
\newblock \emph{IEEE Transactions on Signal Processing}, \textbf{67(20)},
  5239--5269.

\bibitem[{Cook(2004)}]{cook2004testing}
Cook, R.D. (2004).
\newblock \enquote{Testing predictor contributions in sufficient dimension
  reduction.}
\newblock \emph{Annals of Statistics}, \textbf{32(3)}, 1062--1092.

\bibitem[{Cook and Weisberg(1991)}]{cook1991sliced}
Cook, R.D. and Weisberg, S. (1991).
\newblock \enquote{Sliced inverse regression for dimension reduction: Comment.}
\newblock \emph{Journal of the American Statistical Association},
  \textbf{86(414)}, 328--332.

\bibitem[{Du et~al.(2022)Du, Cai, and Roeder}]{du2022robust}
Du, J.H., Cai, Z., and Roeder, K. (2022).
\newblock \enquote{Robust probabilistic modeling for single-cell multimodal
  mosaic integration and imputation via scvaeit.}
\newblock \emph{Proceedings of the National Academy of Sciences},
  \textbf{119(49)}, e2214414119.

\bibitem[{Dua and Graff(2017)}]{Dua:2019}
Dua, D. and Graff, C. (2017).
\newblock \enquote{{UCI} machine learning repository.}

\bibitem[{Dwork et~al.(2006)Dwork, McSherry, Nissim, and
  Smith}]{dwork2006calibrating}
Dwork, C., McSherry, F., Nissim, K., and Smith, A. (2006).
\newblock \enquote{Calibrating noise to sensitivity in private data analysis.}
\newblock In \enquote{Theory of Cryptography: Third Theory of Cryptography
  Conference, TCC 2006, New York, NY, USA, March 4-7, 2006. Proceedings 3,}
  pages 265--284. Springer.

\bibitem[{Dwork et~al.(2014{\natexlab{a}})Dwork, Roth
  et~al.}]{dwork2014algorithmic}
Dwork, C., Roth, A., et~al. (2014{\natexlab{a}}).
\newblock \enquote{The algorithmic foundations of differential privacy.}
\newblock \emph{Foundations and Trends{\textregistered} in Theoretical Computer
  Science}, \textbf{9(3--4)}, 211--407.

\bibitem[{Dwork et~al.(2010)Dwork, Rothblum, and Vadhan}]{dwork2010boosting}
Dwork, C., Rothblum, G.N., and Vadhan, S. (2010).
\newblock \enquote{Boosting and differential privacy.}
\newblock In \enquote{2010 IEEE 51st Annual Symposium on Foundations of
  Computer Science,} pages 51--60. IEEE.

\bibitem[{Dwork et~al.(2021)Dwork, Su, and Zhang}]{dwork2021differentially}
Dwork, C., Su, W., and Zhang, L. (2021).
\newblock \enquote{Differentially private false discovery rate control.}
\newblock \emph{Journal of Privacy and Confidentiality}, \textbf{11(2)}.

\bibitem[{Dwork et~al.(2014{\natexlab{b}})Dwork, Talwar, Thakurta, and
  Zhang}]{dwork2014analyze}
Dwork, C., Talwar, K., Thakurta, A., and Zhang, L. (2014{\natexlab{b}}).
\newblock \enquote{Analyze gauss: optimal bounds for privacy-preserving
  principal component analysis.}
\newblock In \enquote{Proceedings of the forty-sixth annual ACM symposium on
  Theory of computing,} pages 11--20.

\bibitem[{Erd{\H{o}}s et~al.(2012)Erd{\H{o}}s, Yau, and
  Yin}]{erdHos2012rigidity}
Erd{\H{o}}s, L., Yau, H.T., and Yin, J. (2012).
\newblock \enquote{Rigidity of eigenvalues of generalized wigner matrices.}
\newblock \emph{Advances in Mathematics}, \textbf{229(3)}, 1435--1515.

\bibitem[{Fan et~al.(2020)Fan, Li, Zhang, and Zou}]{fan2020statistical}
Fan, J., Li, R., Zhang, C.H., and Zou, H. (2020).
\newblock \emph{Statistical foundations of data science}.
\newblock CRC press.

\bibitem[{Gao and Ma(2023)}]{gao2021sparse}
Gao, S. and Ma, Z. (2023).
\newblock \enquote{Sparse {gca} and thresholded gradient descent.}
\newblock \emph{Journal of Machine Learning Research}, \textbf{24(135)}, 1--61.

\bibitem[{Guyon(2003)}]{guyon2003design}
Guyon, I. (2003).
\newblock \enquote{Design of experiments of the nips 2003 variable selection
  benchmark.}
\newblock In \enquote{NIPS 2003 workshop on feature extraction and feature
  selection,} volume 253, page~40.

\bibitem[{Hall and Li(1993)}]{hall1993almost}
Hall, P. and Li, K.C. (1993).
\newblock \enquote{On almost linearity of low dimensional projections from high
  dimensional data.}
\newblock \emph{Annals of Statistics}, pages 867--889.

\bibitem[{He et~al.(2023)He, Zhang, and Chen}]{he2023differentially}
He, S., Zhang, J., and Chen, X. (2023).
\newblock \enquote{Differentially private sliced inverse regression in the
  federated paradigm.}
\newblock \emph{arXiv preprint arXiv:2306.06324}.

\bibitem[{Hu et~al.(2023)Hu, Xiang, Liu, and Wang}]{hu2023privacy}
Hu, L., Xiang, Z., Liu, J., and Wang, D. (2023).
\newblock \enquote{Privacy-preserving sparse generalized eigenvalue problem.}
\newblock In \enquote{International Conference on Artificial Intelligence and
  Statistics,} pages 5052--5062. PMLR.

\bibitem[{Li(2018)}]{li2018sufficient}
Li, B. (2018).
\newblock \emph{Sufficient dimension reduction: Methods and applications with
  R}.
\newblock CRC Press.

\bibitem[{Li et~al.(2005)Li, Zha, and Chiaromonte}]{li2005contour}
Li, B., Zha, H., and Chiaromonte, F. (2005).
\newblock \enquote{Contour regression: A general approach to dimension
  reduction.}
\newblock \emph{The Annals of Statistics}, \textbf{33(4)}, 1580--1616.

\bibitem[{Li(1991)}]{li1991sliced}
Li, K.C. (1991).
\newblock \enquote{Sliced inverse regression for dimension reduction.}
\newblock \emph{Journal of the American Statistical Association},
  \textbf{86(414)}, 316--327.

\bibitem[{Lin et~al.(2021)Lin, Li, Huang, and Liu}]{lin2021optimal}
Lin, Q., Li, X., Huang, D., and Liu, J.S. (2021).
\newblock \enquote{{On the optimality of sliced inverse regression in high
  dimensions}.}
\newblock \emph{The Annals of Statistics}, \textbf{49(1)}, 1 -- 20.
\newblock \doi{10.1214/19-AOS1813}.

\bibitem[{Lin et~al.(2018)Lin, Zhao, and Liu}]{lin2018consistency}
Lin, Q., Zhao, Z., and Liu, J.S. (2018).
\newblock \enquote{On consistency and sparsity for sliced inverse regression in
  high dimensions.}
\newblock \emph{The Annals of Statistics}, \textbf{46(2)}, 580--610.

\bibitem[{Lin et~al.(2019)Lin, Zhao, and Liu}]{lin2019sparse}
Lin, Q., Zhao, Z., and Liu, J.S. (2019).
\newblock \enquote{Sparse sliced inverse regression via lasso.}
\newblock \emph{Journal of the American Statistical Association},
  \textbf{114(528)}, 1726--1739.

\bibitem[{Liu et~al.(2022{\natexlab{a}})Liu, Ke, Liu, and Li}]{liu2022model}
Liu, W., Ke, Y., Liu, J., and Li, R. (2022{\natexlab{a}}).
\newblock \enquote{Model-free feature screening and {FDR} control with knockoff
  features.}
\newblock \emph{Journal of the American Statistical Association},
  \textbf{117(537)}, 428--443.

\bibitem[{Liu et~al.(2022{\natexlab{b}})Liu, Kong, Jain, and Oh}]{liu2022dp}
Liu, X., Kong, W., Jain, P., and Oh, S. (2022{\natexlab{b}}).
\newblock \enquote{Dp-pca: Statistically optimal and differentially private
  pca.}
\newblock In S.~Koyejo, S.~Mohamed, A.~Agarwal, D.~Belgrave, K.~Cho, and A.~Oh,
  editors, \enquote{Advances in Neural Information Processing Systems,}
  volume~35, pages 29929--29943. Curran Associates, Inc.

\bibitem[{Ma(2013)}]{ma2013sparse}
Ma, Z. (2013).
\newblock \enquote{{Sparse principal component analysis and iterative
  thresholding}.}
\newblock \emph{The Annals of Statistics}, \textbf{41(2)}, 772 -- 801.
\newblock \doi{10.1214/13-AOS1097}.

\bibitem[{Qiu et~al.(2023)Qiu, Lei, and Roeder}]{qiu2023gradient}
Qiu, Y., Lei, J., and Roeder, K. (2023).
\newblock \enquote{Gradient-based sparse principal component analysis with
  extensions to online learning.}
\newblock \emph{Biometrika}, \textbf{110(2)}, 339--360.

\bibitem[{Sibson(1979)}]{sibson1979studies}
Sibson, R. (1979).
\newblock \enquote{Studies in the robustness of multidimensional scaling:
  Perturbational analysis of classical scaling.}
\newblock \emph{Journal of the Royal Statistical Society: Series B
  (Methodological)}, \textbf{41(2)}, 217--229.

\bibitem[{Stewart(1979)}]{stewart1979pertubation}
Stewart, G.W. (1979).
\newblock \enquote{Pertubation bounds for the definite generalized eigenvalue
  problem.}
\newblock \emph{Linear algebra and its applications}, \textbf{23}, 69--85.

\bibitem[{Sun(1983)}]{sun1983perturbation}
Sun, J.g. (1983).
\newblock \enquote{The perturbation bounds for eigenspaces of a definite
  matrix-pair.}
\newblock \emph{Numerische Mathematik}, \textbf{41(3)}, 321--343.

\bibitem[{Tan et~al.(2020)Tan, Shi, and Yu}]{tan2020sparse}
Tan, K., Shi, L., and Yu, Z. (2020).
\newblock \enquote{Sparse {SIR}: Optimal rates and adaptive estimation.}
\newblock \emph{The Annals of Statistics}, \textbf{48(1)}, 64 -- 85.
\newblock \doi{10.1214/18-AOS1791}.

\bibitem[{Tu et~al.(2016)Tu, Boczar, Simchowitz, Soltanolkotabi, and
  Recht}]{tu2016low}
Tu, S., Boczar, R., Simchowitz, M., Soltanolkotabi, M., and Recht, B. (2016).
\newblock \enquote{Low-rank solutions of linear matrix equations via procrustes
  flow.}
\newblock In \enquote{International Conference on Machine Learning,} pages
  964--973. PMLR.

\bibitem[{Vershynin(2010)}]{vershynin2010introduction}
Vershynin, R. (2010).
\newblock \enquote{Introduction to the non-asymptotic analysis of random
  matrices.}
\newblock \emph{arXiv preprint arXiv:1011.3027}.

\bibitem[{Vershynin(2018)}]{vershynin2018high}
Vershynin, R. (2018).
\newblock \emph{High-dimensional probability: An introduction with applications
  in data science}, volume~47.
\newblock Cambridge university press.

\bibitem[{Wainwright(2019)}]{wainwright2019high}
Wainwright, M.J. (2019).
\newblock \emph{High-dimensional statistics: A non-asymptotic viewpoint},
  volume~48.
\newblock Cambridge university press.

\bibitem[{Wasserman and Zhou(2010)}]{wasserman2010statistical}
Wasserman, L. and Zhou, S. (2010).
\newblock \enquote{A statistical framework for differential privacy.}
\newblock \emph{Journal of the American Statistical Association},
  \textbf{105(489)}, 375--389.

\bibitem[{Xia and Cai(2023)}]{xia2023adaptive}
Xia, X. and Cai, Z. (2023).
\newblock \enquote{Adaptive false discovery rate control with privacy
  guarantee.}
\newblock \emph{Journal of Machine Learning Research}, \textbf{24}, 1--35.

\bibitem[{Yin and Hilafu(2015)}]{yin2015sequential}
Yin, X. and Hilafu, H. (2015).
\newblock \enquote{Sequential sufficient dimension reduction for large p, small
  n problems.}
\newblock \emph{Journal of the Royal Statistical Society: Series B: Statistical
  Methodology}, pages 879--892.

\bibitem[{Zhu et~al.(2006)Zhu, Miao, and Peng}]{zhu2006sliced}
Zhu, L., Miao, B., and Peng, H. (2006).
\newblock \enquote{On sliced inverse regression with high-dimensional
  covariates.}
\newblock \emph{Journal of the American Statistical Association},
  \textbf{101(474)}, 630--643.

\bibitem[{Zou et~al.(2006)Zou, Hastie, and Tibshirani}]{zou2006sparse}
Zou, H., Hastie, T., and Tibshirani, R. (2006).
\newblock \enquote{Sparse principal component analysis.}
\newblock \emph{Journal of computational and graphical statistics},
  \textbf{15(2)}, 265--286.

\end{thebibliography}

\end{document}